\documentclass[sigconf]{acmart}

\newif\ifarxiv
\arxivtrue % Set to false for Camera-ready

\usepackage{mathtools}

\usepackage{multirow}
\usepackage{makecell}
\usepackage{tabularx}   % Auto-adjust column width and line breaks
\usepackage{subcaption}
\usepackage{graphicx}

\usepackage{algorithm}
\usepackage{algorithmic}

\usepackage{soul}
\usepackage{xspace}
\usepackage{pifont}

\usepackage{balance}

\usepackage{colortbl}

\definecolor{LightCyan}{gray}{0.9}

\definecolor{dodgerblue}{rgb}{0.12,0.565,1}
\definecolor{antiquefuchsia}{rgb}{0.57, 0.36, 0.51}
\definecolor{RosePink}{RGB}{235,122,119}
\definecolor{DarkRed}{RGB}{195, 0, 0}
\definecolor{DarkOrange}{RGB}{255,140,0}
\definecolor{myPurple}{RGB}{128,0,128}

\newcommand{\revised}[1]{\textcolor{black}{#1}}

\newenvironment{revise}{\color{black}}{}

\newcommand{\prob}{{\sc WSAD}\xspace}
\newcommand{\workname}{WSADBench\xspace}
\newcommand{\nalgorithms}{36\xspace}

\newcommand{\rla}{\gamma_{la}}
\newcommand{\nla}{N_{la}}
\newcommand{\numu}{N_{u}}  % \nu is already defined

\newcommand{\cmark}{\ding{51}}%

\newcommand{\arxivlink}{http://arxiv.org/abs/2605.26068}
\newcommand{\arxivurl}{\url{\arxivlink}}

\copyrightyear{2026}
\acmYear{2026}
\setcopyright{cc}
\setcctype{by}
\acmConference[KDD '26]{Proceedings of the 32nd ACM SIGKDD Conference on Knowledge Discovery and Data Mining V.2}{August 09--13, 2026}{Jeju Island, Republic of Korea}
\acmBooktitle{Proceedings of the 32nd ACM SIGKDD Conference on Knowledge Discovery and Data Mining V.2 (KDD '26), August 09--13, 2026, Jeju Island, Republic of Korea}
\acmDOI{10.1145/3770855.3817536}
\acmISBN{979-8-4007-2259-2/2026/08}

\begin{document}

%%
%% Title
%%
\title{Rethinking Weak Supervision in Anomaly Detection: A Comprehensive Benchmark}

%%
%% Authors and Affiliations
%% KDD D&B Track is single-blind, so author info is visible
%% Each author has their own affiliation block for cleaner display
%%

% Author 1 - Xu Yao (Equal Contribution)
\author{Xu Yao}
\authornote{Equal contribution.}
\email{yaoxu@stu.sufe.edu.cn}
\affiliation{%
  \institution{Shanghai University of Finance and Economics}
  \city{Shanghai}
  \country{China}
}

% Author 2 - Siyuan Zhou (Equal Contribution)
\author{Siyuan Zhou}
\authornotemark[1]
\email{zhousiyuan@stu.sufe.edu.cn}
\affiliation{%
  \institution{Shanghai University of Finance and Economics}
  \city{Shanghai}
  \country{China}
}

% Author 3 - Zhenbo Wu (Equal Contribution)
\author{Zhenbo Wu}
\authornotemark[1]
\email{zhenbowu@stu.sufe.edu.cn}
\affiliation{%
  \institution{Shanghai University of Finance and Economics}
  \city{Shanghai}
  \country{China}
}

% Author 4 - Chaochuan Hou (Equal Contribution)
\author{Chaochuan Hou}
\authornotemark[1]
\email{houchaochuan@foxmail.com}
\affiliation{%
  \institution{Shanghai University of Finance and Economics}
  \city{Shanghai}
  \country{China}
}

% Author 5 - Shuang Liang (Equal Contribution)
\author{Shuang Liang}
\authornotemark[1]
\email{liangs1104@stu.sufe.edu.cn}
\affiliation{%
  \institution{Shanghai University of Finance and Economics}
  \city{Shanghai}
  \country{China}
}

% Author 6 - Shiping Wang
\author{Shiping Wang}
\authornotemark[1]
\email{shiping.wsp@antgroup.com}
\affiliation{
  \institution{Ant Group}
  \city{Shanghai}
  \country{China}
}

% Author 7 - Hailiang Huang
\author{Hailiang Huang}
\authornotemark[2]
\email{hlhuang@shufe.edu.cn}
\affiliation{%
  \department{Key Laboratory of Interdisciplinary Research of Computation and Economics}
  \institution{Shanghai University of Finance and Economics}
  \city{Shanghai}
  \country{China}
}

% Author 8 - Songqiao Han
\author{Songqiao Han}
\authornotemark[2]
\email{han.songqiao@shufe.edu.cn}
\affiliation{%
  \department{Key Laboratory of Interdisciplinary Research of Computation and Economics}
  \institution{Shanghai University of Finance and Economics}
  \city{Shanghai}
  \country{China}
}

% Author 9 - Minqi Jiang (Corresponding)
\author{Minqi Jiang}
\authornote{Corresponding author.}
\email{jiangmq95@163.com}
\affiliation{%
  \department{Key Laboratory of Interdisciplinary Research of Computation and Economics}
  \institution{Shanghai University of Finance and Economics}
  \city{Shanghai}
  \country{China}
}

%% Short author list for headers
\renewcommand{\shortauthors}{Xu Yao et al.}

%%
%% Abstract
%%
\begin{abstract}
  % Weakly supervised anomaly detection (WSAD) has evolved along three isolated directions: incomplete, inexact, and inaccurate supervision.
% However, the lack of a unified evaluation framework prevents distinguishing whether these paradigms address unique challenges or share fundamental mechanics.
% This paper introduces WSADBench, the first systematic benchmark designed to evaluate the independence and comparability of WSAD methods.
% By unifying experimental protocols across 4 data modalities and 36 algorithms, we control variations in label quantity, granularity, and quality to map the performance boundaries of anomaly detection.
% Scenario Interconnectivity:
% The Specialization Paradox:
% Unlabeled Data Variance:
% Asymmetric Sensitivity:
Weakly supervised anomaly detection (WSAD) has developed in three primary directions: incomplete, inexact, and inaccurate supervision.
However, these directions remain isolated, lacking a unified framework to assess whether they address unique challenges or share fundamental mechanisms.
This paper introduces WSADBench, the first benchmark that unifies evaluation across distinct weakly supervised scenarios, benchmarking diverse approaches from specialized WSAD methods to advanced tabular foundation models.
WSADBench establishes standardized protocols to evaluate 36 algorithms across 4 modalities by systematically varying label quantity, granularity, and quality, revealing the performance boundaries of various methods.
Based on over 700K experiments, WSADBench reveals four critical insights:
(i) Strong intrinsic correlations exist between these weak supervision scenarios, challenging the isolation of current research directions.
(ii) Specialized WSAD algorithms excel only in extreme label-scarcity regimes but are quickly dominated by tabular foundation models and general classification methods as supervision increases or in OOD scenarios.
(iii) Unlabeled data shows inconsistent utility across settings, with marginal gains compared to label refinement.
(iv) Models exhibit asymmetric sensitivity to different types of label noise.
We release WSADBench as an open-source benchmark with code and datasets to facilitate future WSAD research: \url{https://github.com/SUFE-AILAB/WSADBench}.

\vspace{-0.5em}

\end{abstract}

%%
%% CCS Concepts
%% Generated from: https://dl.acm.org/ccs
%%

% \ccsdesc[500]{Computing methodologies~Semi-supervised learning settings}

%%
%% Keywords
%%

%%
%% Make title
%%
\maketitle

\newcommand\kddavailabilityurl{https://github.com/SUFE-AILAB/WSADBench}
\newcommand\kddzenododoi{https://doi.org/10.5281/zenodo.20389325}
\ifdefempty{\kddavailabilityurl}{}{
  \begingroup\small\noindent\raggedright\textbf{Resource Availability:}\\
  The code and datasets are permanently archived at \url{\kddzenododoi}, with active updates maintained at \url{\kddavailabilityurl}.
  \endgroup
}

%%
%% Body
%%
\section{Introduction}

% Comparison table with existing AD benchmarks
\begin{table*}[t]
  \centering
  \caption{Comparison of \workname with existing anomaly detection benchmarks.
  }
  \label{tab:comparison}
  \vspace{-8pt}
  \resizebox{1\textwidth}{!}{
    \begin{tabular}{lccccccccccc}
      \toprule
      \multirow{2}{*}{\textbf{Benchmark}} & \multirow{2}{*}{\textbf{Year}} & \multirow{2}{*}{\textbf{Modality}} & \multirow{2}{*}{\textbf{Scope}} & \multirow{2}{*}{\textbf{\# Dataset}} & \multirow{2}{*}{\textbf{\# Model}} & \multicolumn{2}{c}{\textbf{Task Level}} & \multirow{2}{*}{\textbf{OOD}} & \multicolumn{2}{c}{\textbf{Supervision}} & \multirow{2}{*}{\textbf{TFM}} \\
      \cmidrule(lr){7-8} \cmidrule(lr){10-11}
      & & & & & & \textbf{Inst.} & \textbf{Bag} & & \textbf{Unsup.} & \textbf{Weak.} & \\
      \midrule
      ADBench~\cite{han2022adbench} & 2022 & Tab., Img., Txt. & ML/DL & 57 & 30 & \cmark & & & \cmark & \cmark & \\
      AnoShift~\cite{dragoi2022anoshift} & 2022 & Tab. & ML/DL & 1 & 11 & \cmark & & \cmark & \cmark & & \\
      UBnormal~\cite{acsintoae2022ubnormal} & 2022 & Vid. & DL & 10 & 3 & \cmark & \cmark & \cmark & \cmark & \cmark & \\
      % OpenOOD~\cite{yang2022openood} & 2022 & Img. & DL & 9 & 35 & & \cmark & & \cmark & \cmark & \cmark \\
      BMAD~\cite{bao2024bmad} & 2024 & Img. & DL & 6 & 15 & \cmark & & & \cmark & & \\
      NLP-ADBench~\cite{li2024nlp} & 2024 & Txt. & ML/DL & 8 & 19 & \cmark & & \cmark & \cmark & & \cmark \\
      MMAD~\cite{jiang2024mmad} & 2025 & Img., Txt. & DL & 4 & 12 & \cmark & & \cmark & \cmark & & \cmark \\
      Text-ADBench~\cite{xiao2025text} & 2025 & Txt. & ML/DL & 12 & 12 & \cmark & & & \cmark & & \cmark \\
      % Abdalla et al.~\cite{abdalla2025video} & 2025 & Vid. & DL & 62 & 50 & \cmark & \cmark & \cmark & \cmark & \cmark & \cmark \\
      \midrule
      \rowcolor{gray!20} \textbf{\workname (Ours)} & \textbf{2026} & \textbf{Tab., Img., Txt., Vid.} & \textbf{ML/DL} & \textbf{61} & \textbf{\nalgorithms} & \textbf{\cmark} & \textbf{\cmark} & \textbf{\cmark} & \textbf{\cmark} & \textbf{\cmark} & \textbf{\cmark} \\
      \bottomrule
    \end{tabular}
  }
\end{table*}

Anomaly detection (AD) plays a critical role in risk-sensitive tasks \cite{pang2021deep, Aggarwal2013a, zhao2019pyod}, ranging from fraud prevention to medical diagnosis.
Yet, practical systems must typically learn under imperfect supervision: labels that are scarce, coarse-grained, or corrupted by noise.
This reality has given rise to \textit{Weakly Supervised Anomaly Detection} (\prob),
where algorithms compensate for label deficiencies by leveraging unlabeled data, exploiting aggregate supervision signals, or explicitly modeling annotation noise.
These include positive-unlabeled (PU) learning methods that treat anomaly detection as learning from labeled anomalies and unlabeled samples~\cite{pang2023deep,ruff2019deep}, Multiple Instance Learning (MIL)
approaches for coarse bag-level labels~\cite{sultani2018real,tian2021weakly},
and noise-robust techniques that explicitly model annotation errors~\cite{zhao2022admoe,dong2021isp}.

However, this parallel evolution of isolated research directions raises fundamental questions that remain unanswered.
(\textit{i}) Are these three supervision deficiencies truly independent research problems?
For instance, bag-level labels in inexact supervision can be broadcast to all instances within each bag, converting the problem into inaccurate supervision with noisy instance-level labels.
(\textit{ii}) Do \prob algorithms necessarily require modality-specific designs and specialized architectures?
Recent tabular foundation models like TabPFN~\cite{grinsztajn2025tabpfn} and LimiX~\cite{zhang2509limix} achieve competitive performance on tabular tasks without anomaly-specific inductive biases, questioning whether specialized \prob techniques remain necessary.
(\textit{iii}) Are existing evaluations even comparable across studies?
Different works employ inconsistent feature extractors (e.g., I3D\cite{carreira2017quo} vs. C3D\cite{tran2015learning} for video), incompatible ground-truth label definitions, and varied experimental protocols, preventing rigorous cross-method comparison.
These three questions motivate a systematic rethinking of the \prob landscape.

We present \textbf{\workname}, a comprehensive benchmark designed to address these fundamental questions through three key design principles.
\textit{First}, to examine cross-scenario connections, we evaluate whether techniques designed for one supervision setting can transfer to others.
\textit{Second}, to assess algorithm necessity, we benchmark diverse approaches ranging from specialized \prob techniques to general-purpose classifiers and tabular foundation models, determining when domain-specific designs provide value.
\textit{Third}, to enable fair comparison, we establish unified protocols with standardized feature extractors, consistent ground-truth alignments, and controlled experimental settings across diverse data modalities.

\vspace{5pt}
\noindent\textbf{Contributions.} Driven by comprehensive evaluation spanning over \textbf{700k experiments}, this work systematizes the \prob domain and delivers the following contributions:
\begin{list}{$\bullet$}{\leftmargin=1em\itemsep=3pt\parsep=0pt\topsep=1pt\partopsep=0pt}
\item \textbf{Unified framework for cross-scenario evaluation.}
  We establish the first benchmark that systematically evaluates \prob methods across different supervision deficiencies,
  revealing connections and transfer potential between traditionally isolated research directions.

  \begin{revise}
  \item \textbf{Novel analytical perspectives and findings.}
    Beyond standard leaderboard comparisons, we introduce analytical
    perspectives such as progressive OOD distribution shift, decoupled
    asymmetric noise injection, inclusion of tabular foundation models
    in \prob method comparison, and joint labeled--unlabeled data
    utility mapping.
    These perspectives reveal new insights into algorithmic robustness,
    noise sensitivity, and the effective boundaries of specialized
    \prob designs.
  \end{revise}

\item \textbf{Standardized evaluation protocols.}
  Through systematic experiments varying label quantity, granularity, and quality, we standardize experimental settings, enabling rigorous method comparisons that were previously hindered by the inconsistent protocols prevalent in existing WSAD research.

\item \textbf{Comprehensive coverage and open-source platform.}
  We benchmark \nalgorithms diverse algorithms across 4 modalities, including specialized \prob methods, deep learning, and tabular foundation models.
  To foster reproducibility, we publicly release all code, datasets and standardized preprocessing pipelines at \url{https://github.com/SUFE-AILAB/WSADBench}.

\end{list}

\vspace{-8pt}
\section{Related Work}

\subsection{Anomaly Detection Methods}
\label{sec:related_methods}

% Anomaly detection has been extensively studied across various domains and modalities.
% The majority of existing methods focus on unsupervised settings, assuming training data consists entirely of normal samples.
% Semi-supervised AD methods utilize a small set of labeled anomalies to improve detection performance~\cite{han2022adbench}.
% However, real-world applications often involve more complex supervision scenarios, collectively referred to as WSAD~\cite{pang2023deep}.
% Following the established taxonomy of weakly supervised learning~\cite{zhou2018brief}, we organize WSAD methods based on the type of label imperfection: incomplete, inexact, and inaccurate labels.

Traditional unsupervised methods operate on unlabeled data containing both normal and anomalous instances, distinguishing outliers based on distributional assumptions such as low-density regions (e.g., LOF~\cite{breunig2000lof}) or distance-based isolation (e.g., kNN~\cite{angiulli2002fast}).
Semi-supervised approaches assume access to clean normal samples for one-class learning~\cite{ruff2018deep, goyal2020drocc}.
However, in real-world scenarios, only scarce label information is available, which is neither purely normal nor fully reliable.
Exploiting such limited supervision has been shown to substantially improve detection performance~\cite{han2022adbench}, motivating the study of Weakly-Supervised Anomaly Detection (WSAD)~\cite{pang2023deep}.
Drawing upon the established taxonomy of weakly supervised learning~\cite{zhou2018brief}, we introduce this framework to the AD domain and propose \workname, the first systematic WSAD benchmark that categorizes methods by three types of label imperfection: \textit{incomplete}, \textit{inexact}, and \textit{inaccurate} supervision.
\sloppy{\textit{Incomplete Supervision} refers to scenarios where only a subset of anomalies are labeled.
  Methods in this category employ techniques such as anomaly score learning in DevNet~\cite{pang2019deepdevnet} and PReNet~\cite{pang2023deep}, feature-guided strategies in FEAWAD~\cite{zhou2021feature}, or data augmentation in RoSAS~\cite{xu2023rosas} to estimate class priors or re-weight unlabeled samples, mitigating bias from label scarcity.
In contrast, \textit{Inexact Supervision} deals with coarse-grained bag-level labels rather than instance-level annotations.}
Particularly prominent in video anomaly detection~\cite{abdalla2025video}, temporal MIL methods~\cite{sultani2018real} leverage ranking losses, feature magnitude constraints introduced by RTFM~\cite{tian2021weakly}, or center-based regularization employed in AR-Net~\cite{wan2020weakly} to localize anomalies within sequences.
Complementing these settings, \textit{Inaccurate Supervision} addresses the challenge of label noise, where annotations themselves may be erroneous.
Although robust learning has been extensively studied in classification~\cite{song2022learning, zhang2021understanding}, its application to AD remains less systematic, with emerging work on noise-tolerant losses and confidence-based filtering~\cite{sattarov2025diffusion}.

\vspace{-8pt}
\subsection{Anomaly Detection Benchmarks}
\label{sec:related_benchmarks}

\begin{figure*}[t]
  \centering
  \includegraphics[width=0.99\textwidth]{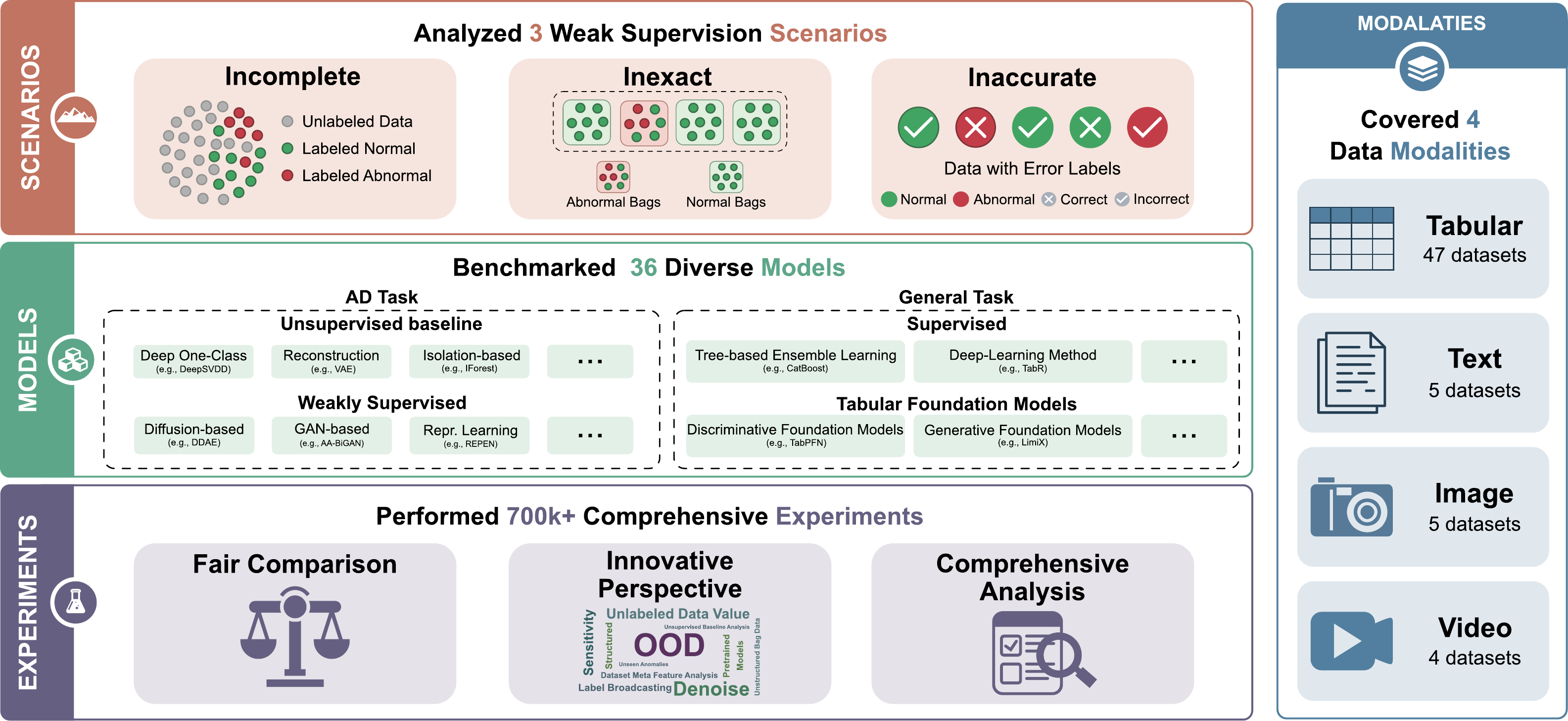}
  \vspace{-8pt}
  \caption{Overview of the \workname. It integrates datasets spanning diverse modalities and varied supervision scenarios into a comprehensive evaluation pipeline. The analysis uncovers the limits and trade-offs of existing WSAD algorithms.
  }
  \label{fig:overview}
  \vspace{-8pt}
\end{figure*}

Several benchmarks have been proposed for anomaly detection.
% However, existing benchmarks exhibit key limitations that prevent comprehensive evaluation of WSAD methods.
Most benchmarks focus on unsupervised or semi-supervised settings, treating weak supervision as a minor extension rather than a systematic research direction.
For instance, AnoShift~\cite{dragoi2022anoshift} and our prior work ADBench~\cite{han2022adbench} cover tabular, image, and text modalities but primarily evaluate unsupervised methods with limited weak supervision scenarios.
Similarly, modality-specific benchmarks such as BMAD~\cite{bao2024bmad} for medical images, UBnormal~\cite{acsintoae2022ubnormal} for video AD, NLP-ADBench~\cite{li2024nlp} and Text-ADBench~\cite{xiao2025text} for text, along with video AD surveys~\cite{abdalla2025video}, provide valuable domain-focused evaluations but lack unified cross-modal assessment.
Beyond standard AD, OpenOOD~\cite{yang2022openood} addresses out-of-distribution detection, while MMAD~\cite{jiang2024mmad} explores multimodal foundation models but remains confined to unsupervised settings.
Furthermore, WRENCH~\cite{zhang2021wrench}, though designed for weak supervision, targets general classification rather than AD-specific challenges.
We compare \workname with existing benchmarks, as summarized in Table~\ref{tab:comparison}.
Additionally, current benchmarks lack systematic evaluation of tabular foundation models under weak supervision---while recent works~\cite{jiang2024mmad, xiao2025text, bao2024bmad} explore zero-shot capabilities, their robustness across different weak supervision scenarios (incomplete, inexact, inaccurate) and OOD settings remains unexplored.

\workname addresses these gaps by providing the \textit{first} comprehensive benchmark dedicated to weakly-supervised anomaly detection.
It unifies evaluation across three supervision deficiency types, spans four modalities, and systematically assesses both specialized WSAD algorithms and tabular foundation models under standardized protocols, yielding critical insights into the effective boundaries and robustness of different approaches.
\vspace{-5pt}
% \vspace{-5pt}
\section{\workname}
\label{sec:taxonomy}

We design \workname (Figure~\ref{fig:overview}) to address the fragmentation and inconsistency in current \prob research. Our rationale comprises three core principles:
(1) \textbf{Cross-Scenario Transferability}—we break scenario isolation by enabling methods specialized for one deficiency to be evaluated across all scenarios, testing whether these supervisions are independent problems or convertible formulations.
(2) \textbf{Broad Baseline Inclusion}—we require specialized \prob algorithms and general-purpose tabular foundation models to compete on the same stage, evaluating the necessity of model specialization.
(3) \textbf{Standardized Fair Evaluation}—we enforce unified feature extractors, standardized metrics, and consistent evaluation protocols to resolve the inconsistent practices found in existing studies.

\subsection{Unified Benchmarking Framework.}
While these three regimes have traditionally been studied in isolation, \workname\ unifies them by establishing explicit connections between supervision scenarios:
\textit{(i) Incomplete-to-Inexact Transferability}. Methods designed for incomplete supervision (e.g., weakly-supervised AD algorithms) are evaluated on inexact supervision tasks (video MIL), revealing whether designs optimized for label scarcity can generalize to label granularity challenges.
\textit{(ii) Inexact-to-Incomplete Transferability}. Specialized inexact methods (e.g., MIL models) are tested on incomplete supervision tasks with sparse instance-level labels, examining whether temporal or structural inductive biases transfer to traditional tabular settings.
\textit{(iii) Robustness under Inaccurate Supervision}. Methods from different supervision scenarios are evaluated under label noise and corruption, examining whether designs optimized for specific supervision types maintain robustness when label quality degrades.
By systematically implementing these evaluations across diverse modalities and different models, we enable rigorous assessment of how tabular foundation models and specialized WSAD algorithms perform when supervision is imperfect in quantity, granularity, or quality—challenging the assumption that methods should only be tested within their original problem settings.

\begin{table}[tb!]
  \centering
  \vspace{-10pt}
  \caption{Classification of benchmarked AD methods.}
  \label{tab:model_classification}
  \vspace{-8pt}
  \resizebox{\columnwidth}{!}{
    \begin{tabular}{ll | ll}
      \toprule
      \textbf{Method} & \textbf{Category} & \textbf{Method} & \textbf{Category} \\
      \midrule
      \multicolumn{2}{c|}{\cellcolor{gray!15}\textbf{Weakly-supervised (Instance)}} & \multicolumn{2}{c}{\cellcolor{gray!15}\textbf{Unsupervised (Instance)}} \\
      DevNet~\cite{pang2019deepdevnet} & Score Learning & IForest~\cite{liu2008isolation} & Isolation-based \\
      DeepSAD~\cite{ruff2019deep} & Score Learning & AutoEncoder~\cite{zong2018deep} & Reconstruction \\
      PReNet~\cite{pang2023deep} & Score Learning & VAE~\cite{kingma2013auto} & Reconstruction \\
      REPEN~\cite{pang2018learning} & Repr. Learning & PCA~\cite{shyu2003novel} & Reconstruction \\
      XGBOD~\cite{zhao2018xgbod} & Repr. Learning & DeepSVDD~\cite{ruff2018deep} & Deep One-class \\
      RoSAS~\cite{xu2023rosas} & Data Aug. & ECOD~\cite{li2022ecod} & Probabilistic \\
      Dual-MGAN~\cite{li2022dual} & Data Aug. & CBLOF~\cite{he2003discovering} & Cluster-based \\
      FEAWAD~\cite{zhou2021feature} & Reconstruction & LOF~\cite{breunig2000lof} & Density-based \\
      DDAE~\cite{sattarov2025diffusion} & Diffusion DAE   & LUNAR~\cite{goodge2022lunar} & GNN-based \\
      SOEL-NTL~\cite{pang2019deepdevnet} & Pseudo-Labeling & & \\
      AA-BiGAN~\cite{tian2022anomaly} & GAN-based & \multicolumn{2}{c}{\cellcolor{gray!15}\textbf{Supervised (Instance)}} \\
      GANomaly~\cite{akcay2018ganomaly} & GAN-based & XGBoost~\cite{chen2016xgboost} & GBDT  \\
      \multicolumn{2}{c|}{\cellcolor{gray!15}\textbf{Weakly-supervised (Bag)}}  & CatBoost~\cite{prokhorenkova2018catboost} & GBDT \\
      Sultani~\cite{sultani2018real} & Vanilla MIL & FTTransformer~\cite{gorishniy2021revisiting} & Deep (Sup.) \\
      RTFM~\cite{tian2021weakly} & Magnitude MIL & TabM~\cite{gorishniy2024tabm} & Deep (Sup.) \\
      MGFN~\cite{chen2023mgfn} & Magnitude MIL & TabR-S~\cite{gorishniy2023tabr} & Deep (Sup.) \\
      AR-Net~\cite{wan2020weakly} & Dynamic MIL &  &  \\
      VadCLIP~\cite{wu2024vadclip} & Language-Guided MIL & \multicolumn{2}{c}{\cellcolor{gray!15}\textbf{Foundation Models (Instance)}} \\
      UR-DMU~\cite{zhou2023dual} & Uncertainty-Aware MIL & TabPFN~\cite{grinsztajn2025tabpfn} & Found. Model \\
      GCN-Anomaly~\cite{zhong2019graph} & Label Denoising & LimiX~\cite{zhang2509limix} & Found. Model \\
      PUMA~\cite{perini2023learning} & PU MIL & & \\

      \bottomrule
    \end{tabular}
  }
  \vspace{-15pt}
\end{table}

\subsection{Datasets and Models}
\noindent\textbf{Datasets.}
To comprehensively verify algorithm detection performance across diverse data types, \workname incorporates \textbf{61} datasets covering four distinct modalities: Tabular, Computer Vision (CV), Natural Language Processing (NLP), and Video.
The datasets for Tabular, CV, and NLP are adopted from ADBench~\cite{han2022adbench} (e.g., Pima, CIFAR-10, Agnews).
For Video modality, we re-collected and reprocessed real-world surveillance datasets, including UCF-Crime\cite{sultani2018real}, XD-Violence\cite{wu2020not}, TAD\cite{xu2024tad}, and ShanghaiTech\cite{liu2018future}.
We strictly follow the ``Standard / One-vs-Rest'' protocol for converting multi-class datasets into AD tasks.
Detailed statistics for all datasets are provided in Appendix~\ref{appendix:datasets}.

\noindent\textbf{Anomaly Detection Methods.}
We re-evaluate the landscape of effective anomaly detection by expanding the comparison beyond specialized WSAD algorithms to include \textit{Generalist Baselines}, such as Tabular Foundation Models and Generic Tree-based/Deep-learning-based Classifiers.
This broadens the competitive landscape to determine whether specialized AD designs are truly necessary or if general-purpose learners suffice.
We benchmark a comprehensive collection of \nalgorithms algorithms.
Table~\ref{tab:model_classification} presents the complete list of algorithms evaluated in our experiments.
Note that this table serves as an inventory of benchmarked methods rather than a strict taxonomy.
Methods are organized by supervision granularity: \textit{instance-level} supervision
and \textit{bag-level} supervision (most originally proposed for video anomaly detection).

\subsection{Experimental Protocols}
Inconsistent evaluation protocols often hinder fair comparison. We therefore enforce strict experimental standardization covering the following aspects:
(i) \textit{Unified Feature Representation}. Pretrained features alter method rankings (detailed in \ifarxiv Appendix~\ref{appendix:pretrained_model}\else the full version\fi). We ensure all methods in a modality use the same features to isolate algorithmic impact.
(ii) \textit{Standardized Ground Truth Alignment}. Subtle differences in ground truth definitions (e.g., frame-level vs. clip-level labels) can lead to contradictory conclusions (see \ifarxiv Appendix~\ref{sec:ground_truth_alignment}\else the full version\fi). We unify these definitions to prevent annotation-induced variance.
(iii) \textit{Reproducible Configurations and Controlled Variation}. We evaluate methods with default hyperparameters and fixed data splits, while systematically mapping performance boundaries through controlled variation of supervision parameters (e.g., label ratios, noise rates).
(iv) \textit{Rigorous Metrics and Statistical Validation}. We assess performance using both AUC-ROC (ranking quality) and AUC-PR (imbalanced robustness), reporting mean and standard deviation across multiple seeds with statistical hypothesis testing to ensure significance.
\vspace{-8pt}
\section{Experiments and Analysis}
\label{sec:analysis}
We evaluate WSAD methods across three dimensions: label \textit{Scarcity}, \textit{Granularity}, and \textit{Quality}.
% For each dimension, we identify when specialized techniques outperform general alternatives.
We compare specialized WSAD methods against baselines under varying supervision conditions, then investigate their generalization, robustness, and limitations.
Finally, we provide a global performance summary over all evaluated scenarios.

\subsection{Basic WSAD Experiments}
\label{sec:wsad_core}
% ====== Instance-Level Supervision ======
\begin{table*}[tbp]
  \centering
  \scriptsize
  \caption{AUCPR comparison of different models under varying label ratios.}
  % REMOVED: "(Mean(Rank))" format already shown in table cells, ratio list redundant (already in column headers)
  \label{tab:merged_aucpr_final_newnlp}
  \vspace{-8pt}
  \renewcommand{\arraystretch}{1.05}
  \begin{tabular}{c l | cccc | cccc | cccc}
    \toprule
    \multirow{2}{*}{} & \multirow{2}{*}{\textbf{Model}} & \multicolumn{4}{c|}{\textbf{Tabular}} & \multicolumn{4}{c|}{\textbf{Image}} & \multicolumn{4}{c}{\textbf{Text}} \\
    \cmidrule(lr){3-6} \cmidrule(lr){7-10} \cmidrule(lr){11-14}
    & & $\gamma_{la}=1\%$ & $\gamma_{la}=10\%$ & $\gamma_{la}=50\%$ & $\gamma_{la}=100\%$ & $\gamma_{la}=1\%$ & $\gamma_{la}=10\%$ & $\gamma_{la}=50\%$ & $\gamma_{la}=100\%$ & $\gamma_{la}=1\%$ & $\gamma_{la}=10\%$ & $\gamma_{la}=50\%$ & $\gamma_{la}=100\%$  \\
    \midrule
    \multirow{12}{*}{\rotatebox{90}{\textbf{Weakly Sup.}}} & XGBOD & 0.481(6) & 0.688(3) & 0.816(2) & 0.864(2) & 0.388(7) & 0.633(7) & 0.759(7) & 0.798(8) & 0.166(7) & 0.366(6) & 0.571(7) & 0.669(9) \\
    & DeepSAD & 0.379(10) & 0.568(12) & 0.752(8) & 0.813(7) & 0.268(16) & 0.474(12) & 0.722(11) & \textbf{0.826(1)} & 0.105(14) & 0.225(12) & 0.503(11) & 0.726(2) \\
    & REPEN & 0.408(8) & 0.602(9) & 0.739(9) & 0.779(9) & 0.288(11) & 0.592(9) & 0.773(4) & 0.816(3) & 0.134(11) & 0.344(9) & 0.618(5) & \textbf{0.741(1)} \\
    & AA-BiGAN & 0.308(15) & 0.310(16) & 0.321(16) & 0.351(16) & 0.332(8) & 0.320(14) & 0.330(15) & 0.366(15) & 0.067(16) & 0.078(16) & 0.071(16) & 0.073(16) \\
    & GANomaly & 0.314(14) & 0.316(15) & 0.335(15) & 0.382(15) & 0.277(15) & 0.279(16) & 0.285(16) & 0.311(16) & 0.110(13) & 0.107(14) & 0.114(15) & 0.136(15) \\
    & DevNet & \textbf{0.555(1)} & 0.666(6) & 0.713(11) & 0.722(11) & 0.459(4) & \textbf{0.694(1)} & 0.782(2) & 0.798(7) & 0.272(2) & 0.512(2) & 0.661(2) & 0.705(4) \\
    & FEAWAD & 0.524(3) & 0.691(2) & 0.772(5) & 0.786(8) & 0.449(5) & 0.666(4) & \textbf{0.790(1)} & 0.812(4) & 0.221(5) & 0.461(5) & 0.646(4) & 0.717(3) \\
    & PReNet & 0.547(2) & 0.676(4) & 0.721(10) & 0.729(10) & 0.468(2) & 0.688(3) & 0.781(3) & 0.795(9) & 0.261(3) & \textbf{0.513(1)} & \textbf{0.668(1)} & 0.702(6) \\
    & RoSAS & 0.378(12) & 0.570(11) & 0.762(7) & 0.826(6) & 0.285(13) & 0.548(10) & 0.747(8) & 0.820(2) & 0.164(8) & 0.356(8) & 0.582(6) & 0.703(5) \\
    & Dual-MGAN & 0.518(5) & 0.603(8) & 0.647(12) & 0.671(12) & \textbf{0.520(1)} & 0.689(2) & 0.764(6) & 0.776(11) & 0.253(4) & 0.498(3) & 0.647(3) & 0.685(8) \\
    & SOEL-NTL & 0.378(11) & 0.415(13) & 0.416(13) & 0.489(14) & 0.292(10) & 0.292(15) & 0.383(14) & 0.449(14) & 0.087(15) & 0.104(15) & 0.135(14) & 0.144(14) \\
    & DDAE & 0.320(13) & 0.331(14) & 0.386(14) & 0.517(13) & 0.321(9) & 0.332(13) & 0.404(13) & 0.592(13) & 0.172(6) & 0.176(13) & 0.194(13) & 0.224(13) \\
    \midrule % [1em]
    \multirow{4}{*}{\rotatebox{90}{ \textbf{Sup.}}} & XGBoost & 0.241(16) & 0.622(7) & 0.795(3) & 0.858(3) & 0.283(14) & 0.596(8) & 0.744(9) & 0.795(10) & 0.151(9) & 0.361(7) & 0.557(9) & 0.667(10) \\
    & CatBoost & 0.521(4) & \textbf{0.715(1)} & \textbf{0.842(1)} & \textbf{0.883(1)} & 0.448(6) & 0.635(6) & 0.764(5) & 0.802(6) & 0.143(10) & 0.317(10) & 0.568(8) & 0.686(7) \\
    & TabM & 0.468(7) & 0.675(5) & 0.787(4) & 0.830(5) & 0.461(3) & 0.660(5) & 0.741(10) & 0.762(12) & \textbf{0.278(1)} & 0.478(4) & 0.520(10) & 0.552(12) \\
    & TabR-S & 0.405(9) & 0.596(10) & 0.768(6) & 0.851(4) & 0.287(12) & 0.488(11) & 0.696(12) & 0.808(5) & 0.130(12) & 0.249(11) & 0.483(12) & 0.660(11) \\
    \midrule
    \multicolumn{2}{l|}{\textit{Median Unsup.}} & \multicolumn{4}{c|}{0.365} & \multicolumn{4}{c|}{0.360} & \multicolumn{4}{c}{0.093} \\
    \bottomrule
  \end{tabular}
\end{table*}

\subsubsection{Instance-Level Supervision}
\label{sec:wsad_instance}

We evaluate instance-level weak supervision on Tabular, CV, and NLP benchmarks under two labeling configurations: (1) Ratio-labeled Anomaly (RLA, $\rla \in \{1\%, 5\%, 10\%\}$), where a fixed percentage of anomalies are labeled; and (2) Number-labeled Anomaly (NLA, $\nla \in \{1, 5, 10\}$), where a fixed count of anomalies are labeled regardless of dataset size.
Table~\ref{tab:merged_aucpr_final_newnlp} presents RLA results; NLA results are provided in \ifarxiv Appendix~\ref{appendix:incomplete_detailed}\else the full version\fi, with both settings exhibiting consistent patterns.

% \paragraph{The Narrow Effective Scope of Weak Supervision.}
\noindent \textbf{The narrow effective scope of weak supervision}.
Our results demonstrate that specialized WSAD methods outperform supervised baselines only under extreme label scarcity. For example, at $\gamma_{la}=1\%$, DevNet achieves AUCPR 0.555 compared to 0.521 for the best supervised baseline CatBoost.
However, this advantage diminishes rapidly as labeled data increases.
This pattern also occurs in CV tasks, where Dual-MGAN leads at $\gamma_{la}=1\%$ (AUCPR 0.520 vs. TabM 0.461) but falls behind as labels increase.
These findings reveal that WSAD methods' specialized inductive biases, while valuable for sparse supervision, become redundant once sufficient labeled data enables general supervised approaches to learn effective decision boundaries.

% ====== Bag-Level Supervision ======

\subsubsection{Bag-Level Supervision}
\label{sec:vad_simple_comparison}
Video anomaly detection exemplifies bag-level supervision, where video-level labels (bags) are used to train models that predict frame-level anomalies (instances). Table~\ref{tab:vad_simple_comparison} shows our evaluation of representative MIL-based VAD methods across multiple pretrained feature extractors to assess their robustness to different backbones, addressing inconsistent backbone usage in prior comparative studies.
Detailed experimental settings and results are provided in \ifarxiv Appendix~\ref{appendix:inexact_video_results}\else the full version\fi.

% \paragraph{Center Loss Dominates MIL Ranking.}\focus{Robust feature aggregation mechanisms (e.g., Center Loss) prove more effective than complex temporal modeling for weakly-supervised VAD.}
% ARNet consistently achieves rank 1 across all pretrain-segmentation combinations, with AUCPR ranging from 0.430 to 0.445.
% This stability stems from its center loss mechanism, which enforces intra-class compactness and suppresses label noise inherent in bag-level supervision.
% Sultani and RTFM alternate between ranks 2-3, while MGFN consistently underperforms (rank 6-7).
% These findings suggest that robust feature aggregation—rather than complex temporal modeling—determines success in weakly-supervised VAD.
% MGFN

% VadClip
% ）

% URDMU

% ARNet

% ZhongGCNAD

% Sultani

\begin{table}[ht]
  \centering
  % \vspace{-3pt}
  \caption{AUCPR comparison of VAD models across various pretraining methods and segmentations.}
  \label{tab:vad_simple_comparison}
  \vspace{-8pt}
  \resizebox{0.45\textwidth}{!}{%
    \begin{tabular}{lcccccc}
      \toprule
      \textbf{Model} & \textbf{i3d} & \textbf{x3d} & \textbf{sf50} & \textbf{mvit} & \textbf{sf} & \textbf{Mean} \\
      \midrule
      \multicolumn{7}{l}{\textit{32 Segments}} \\
      \midrule
      AR-Net & \textbf{0.430 (1)} & \textbf{0.434 (1)} & \textbf{0.436 (1)} & \textbf{0.441 (1)} & \textbf{0.445 (1)} & \textbf{0.437 (1)} \\
      Sultani & 0.392 (2) & 0.377 (3) & 0.403 (3) & 0.399 (2) & 0.412 (3) & 0.397 (2) \\
      MGFN & 0.258 (7) & 0.279 (7) & 0.337 (6) & 0.308 (7) & 0.337 (7) & 0.304 (7) \\
      RTFM & 0.386 (3) & 0.397 (2) & 0.418 (2) & 0.394 (3) & 0.427 (2) & 0.404 (3) \\
      UR-DMU & 0.340 (5) & 0.339 (6) & 0.377 (4) & 0.363 (4) & 0.376 (4) & 0.359 (5) \\
      VadClip & 0.344 (4) & 0.343 (5) & 0.357 (5) & 0.354 (5) & 0.350 (6) & 0.350 (6) \\
      GCN-Anomaly & 0.304 (6) & 0.364 (4) & 0.307 (7) & 0.351 (6) & 0.354 (5) & 0.336 (4) \\
      \midrule
      \multicolumn{7}{l}{\textit{200 Segments}} \\
      \midrule
      AR-Net & \textbf{0.429 (1)} & \textbf{0.433 (1)} & \textbf{0.436 (1)} & \textbf{0.442 (1)} & \textbf{0.445 (1)} & \textbf{0.437 (1)} \\
      Sultani & 0.377 (3) & 0.398 (2) & 0.400 (3) & 0.397 (2) & 0.412 (3) & 0.397 (3) \\
      MGFN & 0.297 (7) & 0.291 (7) & 0.308 (7) & 0.333 (7) & 0.346 (7) & 0.315 (7) \\
      RTFM & 0.380 (2) & 0.379 (4) & 0.408 (2) & 0.389 (3) & 0.423 (2) & 0.396 (2) \\
      UR-DMU & 0.353 (4) & 0.350 (6) & 0.387 (4) & 0.371 (6) & 0.374 (4) & 0.367 (4) \\
      VadClip & 0.348 (5) & 0.351 (5) & 0.359 (5) & 0.372 (5) & 0.350 (6) & 0.356 (5) \\
      GCN-Anomaly & 0.303 (6) & 0.381 (3) & 0.349 (6) & 0.375 (4) & 0.373 (5) & 0.356 (6) \\
      \midrule
      \textbf{Overall Mean} & 0.352 & 0.361 & 0.374 & 0.377 & \textbf{0.387} & 0.370 \\
      \bottomrule
    \end{tabular}%
    % \vspace{-50pt}
  }
\end{table}

\noindent \textbf{Simpler methods prevail under standardized evaluation}.
% \paragraph{Simpler Methods Prevail Under Standardized Evaluation.}
A notable finding is that AR-Net~\cite{wan2020weakly}, one of the earliest MIL-based methods, ranks first across \emph{all} backbone--segmentation combinations, with mean AUCPR 0.437.
This strong performance stems from a remarkably simple but effective design: adding the center loss to the foundational margin-ranking framework of Sultani~\cite{sultani2018real}, which encourages normal-snippet features to cluster tightly.
In contrast, more recent and complex methods---such as MGFN (mean AUCPR 0.309, rank 7) and UR-DMU (0.363, rank 4)---perform notably worse and exhibit higher variance across backbones.
We attribute this partly to evaluation protocol: most prior methods report the best epoch selected on the test set, whereas \workname{} adopts a strict last-epoch protocol without peeking at test labels.
This change particularly penalizes methods with unstable convergence, as they can no longer identify and select the best-performing checkpoints.

\subsection{Further Experiments and Analysis}

\subsubsection{Tabular Foundation Models: Challenging the Necessity of Specialized WSAD Methods}
% We evaluate tabular foundation models across varying supervision levels, modalities, and dataset characteristics to assess whether specialized WSAD methods are necessary or general pre-trained models achieve comparable performance.
\revised{We evaluate tabular foundation models across varying supervision levels, modalities, and dataset characteristics to assess the necessity of specialized \prob designs.
  To adapt these models, training data formatting follows a Positive-Unlabeled (PU) learning proxy.
  This setup constructs the Context Set \((X_{ctx}, Y_{ctx})\) by assigning \(y=1\) to labeled anomalies and pseudo-labeling unlabeled instances as \(y=0\).
During inference, in-context learning predicts anomaly probabilities for test instances \(X_{query}\): \(\hat{Y}_{query} = f(X_{query} \mid X_{ctx}, Y_{ctx})\).}

% \vspace{4pt}

\begin{figure}[h]
  \centering
  % \vspace{-0.5em}
  \includegraphics[width=1.0\linewidth]{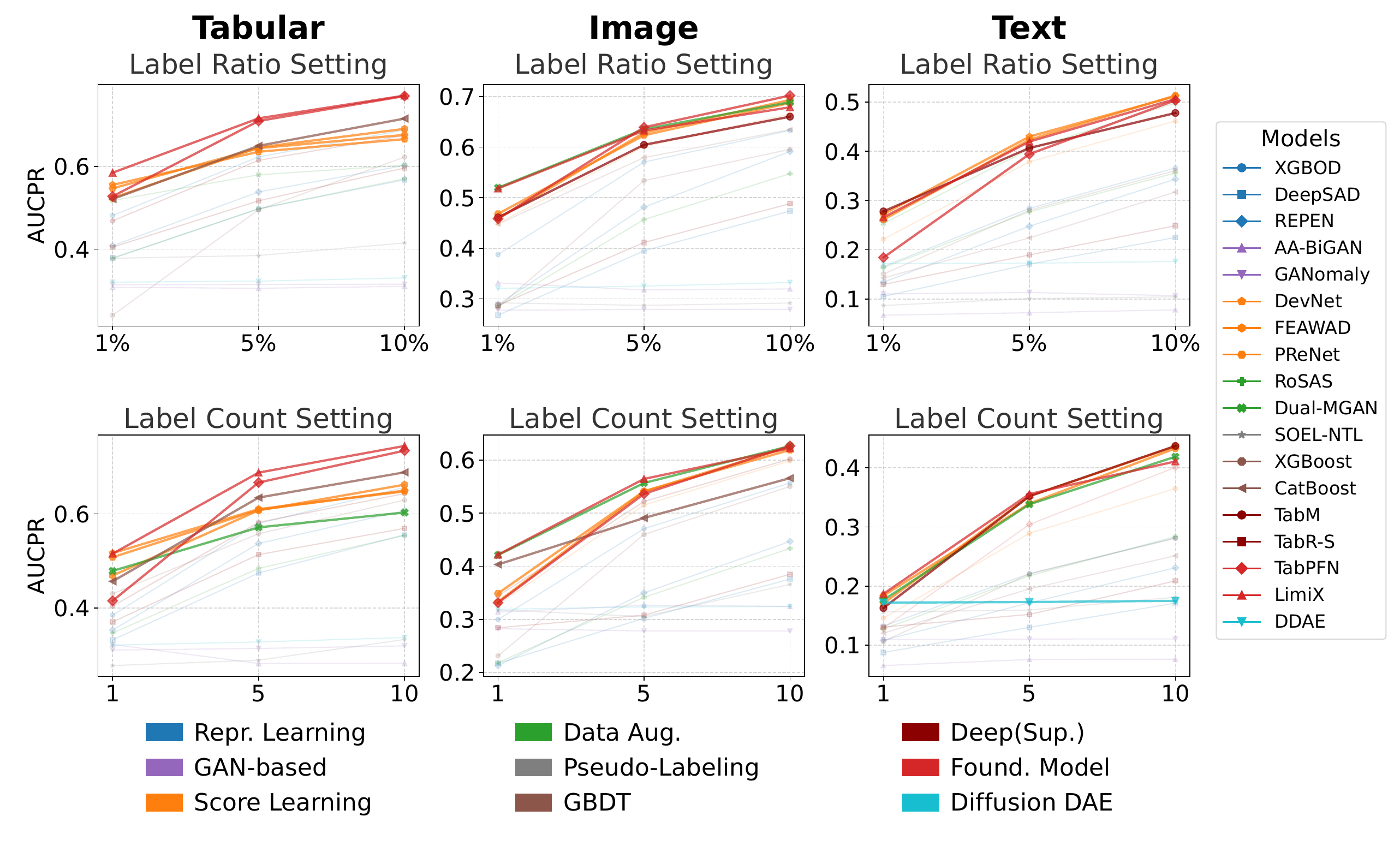}
  \vspace{-20pt}
  \caption{Tabular Foundation model performance (AUCPR) for different modalities under varying labeled anomaly ratios (upper) and counts (lower).}
  \label{fig:global_incomplete_heatmap}
  % \vspace{-12pt}
\end{figure}

\noindent \textbf{Tabular Foundation models dominate specialized WSAD across supervision levels and modalities}.
As shown in Figure~\ref{fig:global_incomplete_heatmap}, tabular foundation models (e.g., LimiX, TabPFN) consistently match or exceed specialized WSAD methods across the majority of supervision settings.
Specialized WSAD methods remain competitive only when labels are extremely scarce ($N_{la}=1$).
Once supervision exceeds this minimal threshold, tabular foundation models achieve substantially higher performance (e.g., LimiX 0.641 vs. PReNet 0.578 w.r.t. $N_{la}=5$ \revised{on tabular data}).
This finding challenges the traditional view that specialized anomaly detection designs are essential for weak supervision. In contrast, general-purpose pre-trained representations often prove sufficient or even superior.

\noindent \textbf{Task characteristics define the boundaries of tabular foundation model superiority.}
Given the dominance of tabular foundation models on tabular data, we investigate model performance across modalities through a Spearman correlation analysis between meta-features and the performance gap, as detailed in Table~\ref{tab:multi_target_corr_main}. The explanation of all meta-features can be found in \ifarxiv Table~\ref{tab:meta_feature_appendix}\else the full version\fi.
First, \textbf{Dimensionality} shows a significant negative correlation with the performance advantage of tabular foundation models over specialized methods ($\rho = -0.44$). This suggests that while tabular foundation models excel on lower-dimensional tabular data, their advantage diminishes in high-dimensional feature spaces (e.g., embeddings from ViT or RoBERTa used in CV and NLP).
Second, \textbf{Linear Separability (Fisher's Ratio)} also negatively correlates with the tabular foundation model advantage. Specialized methods like DevNet are more efficient when boundaries are distinct, whereas tabular foundation models excel at modeling complex, non-linear boundaries where traditional methods struggle.
These factors—dimensionality and separability—jointly characterize the conditions where tabular foundation models dominate versus where specialized methods remain competitive.

\begin{table}[h!]
  \centering
  \caption{Spearman correlation ($\rho$) between meta-features and pairwise AUCPR gaps across all datasets ($\gamma_{la}=100\%$).}
  % REMOVED: "Figure X:" error (should be Table), column explanation obvious from header, Appendix reference moved to text
  \label{tab:multi_target_corr_main}
  % \vspace{-8pt}
  \resizebox{\columnwidth}{!}{
    \begin{tabular}{lccc}
      \toprule
      \textbf{Meta-Feature} & \textbf{LimiX - DevNet} & \textbf{TabPFN - DevNet} & \textbf{LimiX - TabPFN} \\
      \midrule
      Dimensionality           & $-0.44^{***}$ & $-0.37^{***}$ & $-0.48^{***}$ \\
      Sparsity Ratio                 & $0.31^{***}$ & $0.24^{**}$  & $0.44^{***}$ \\
      Avg.Feat.Sim.           & $0.12$       & $0.02$       & $0.33^{***}$ \\
      Eff. Rank            & $-0.28^{**}$ & $-0.10$      & $-0.73^{***}$ \\
      Eff. Rank Ratio           & $0.44^{***}$ & $0.43^{***}$ & $0.24^{**}$  \\
      Fisher's Ratio (F1)           & $-0.49^{***}$ & $-0.61^{***}$ & $0.36^{***}$ \\
      Borderline Pts. (N1)           & $0.40^{***}$ & $0.48^{***}$ & $-0.12$      \\
      Intra/Inter Ratio (N2)           & $-0.01$      & $0.18$       & $-0.62^{***}$ \\
      RF's F1               & $-0.31^{***}$ & $-0.50^{***}$ & $0.62^{***}$ \\
      Local/Global Outlier Ratio      & $0.30^{***}$ & $0.41^{***}$ & $-0.03$      \\
      \bottomrule
    \end{tabular}
  }
  \smallskip\noindent
  {\raggedright\fontsize{5pt}{6pt}\selectfont \textit{Note: $^{*}:p<0.05, ^{**}:p<0.01, ^{***}:p<0.001$. Eff.:Effective, Feat:Feature, Sim:Similarity, Pts: Points.}\par}
  % \vspace{-10pt}
\end{table}
\subsubsection{The Value of Unlabeled Data}
\label{sec:unlabeled_value}
We investigate the value of unlabeled data across varying amounts of labeled anomalies, revealing its dependency on both supervision levels and model architecture.

\vspace{4pt}

\begin{figure}[ht]
  \centering
  % First subfigure (Left)
  % Increased width from 0.4 to 0.48 to reduce gap and enlarge figure
  \begin{subfigure}[b]{0.48\linewidth}
    \centering
    % Use trim and clip to remove whitespace. Adjust values {left bottom right top} as needed.
    % Example: trim={1cm 0.5cm 1cm 0.5cm}, clip
    \includegraphics[width=\linewidth, trim={0.3cm 0cm 0cm 0.2cm}, clip]{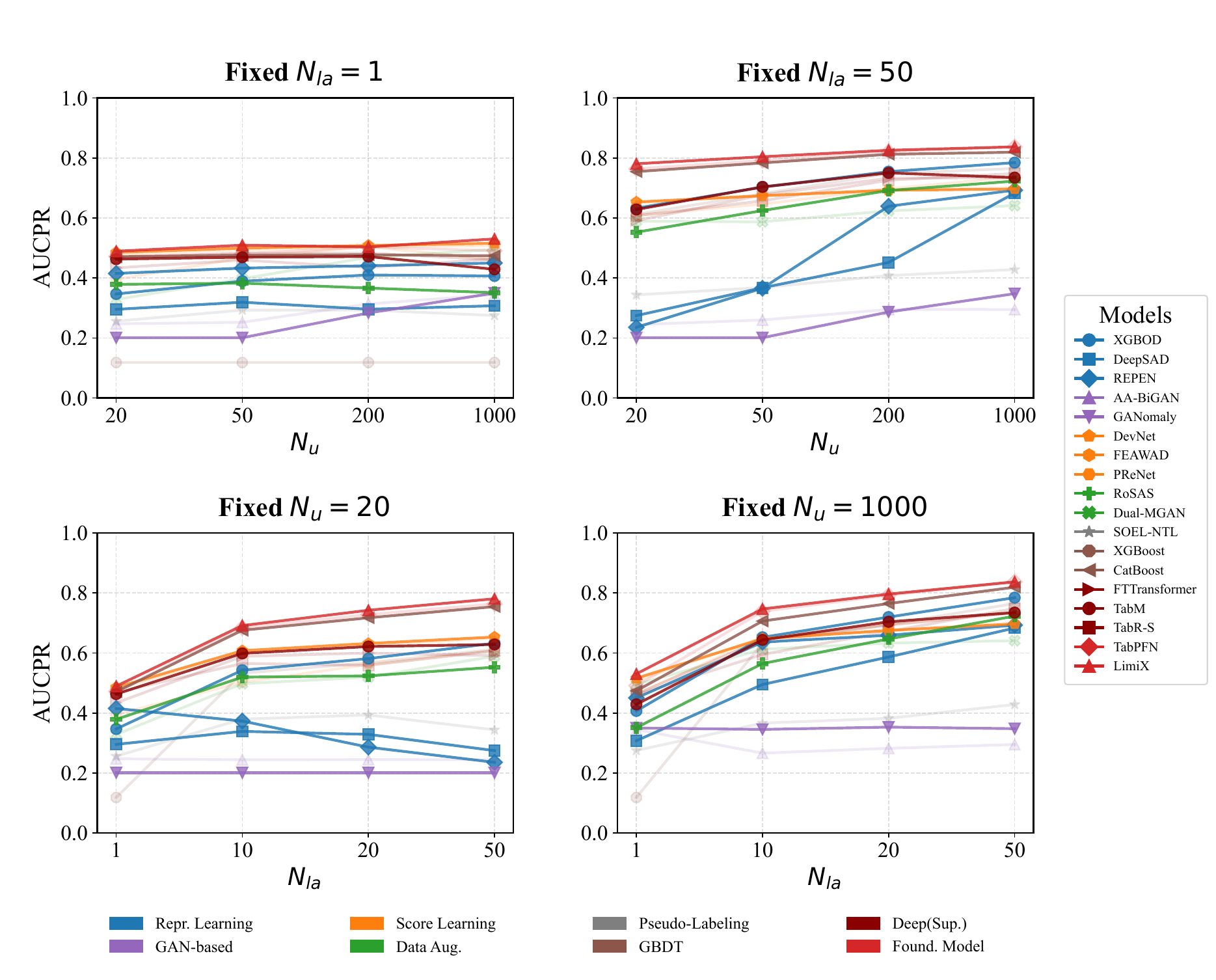}
    \caption{Marginal effect}
    \label{fig:performance_curve}
  \end{subfigure}
  \hfill % Fills space between subfigures
  % Second subfigure (Right)
  \begin{subfigure}[b]{0.48\linewidth}
    \centering
    % Apply trim and clip here as well. Adjust values based on the actual figure's margins.
    \includegraphics[width=\linewidth, trim={0.15cm 0.2cm 0cm 0.2cm}, clip]{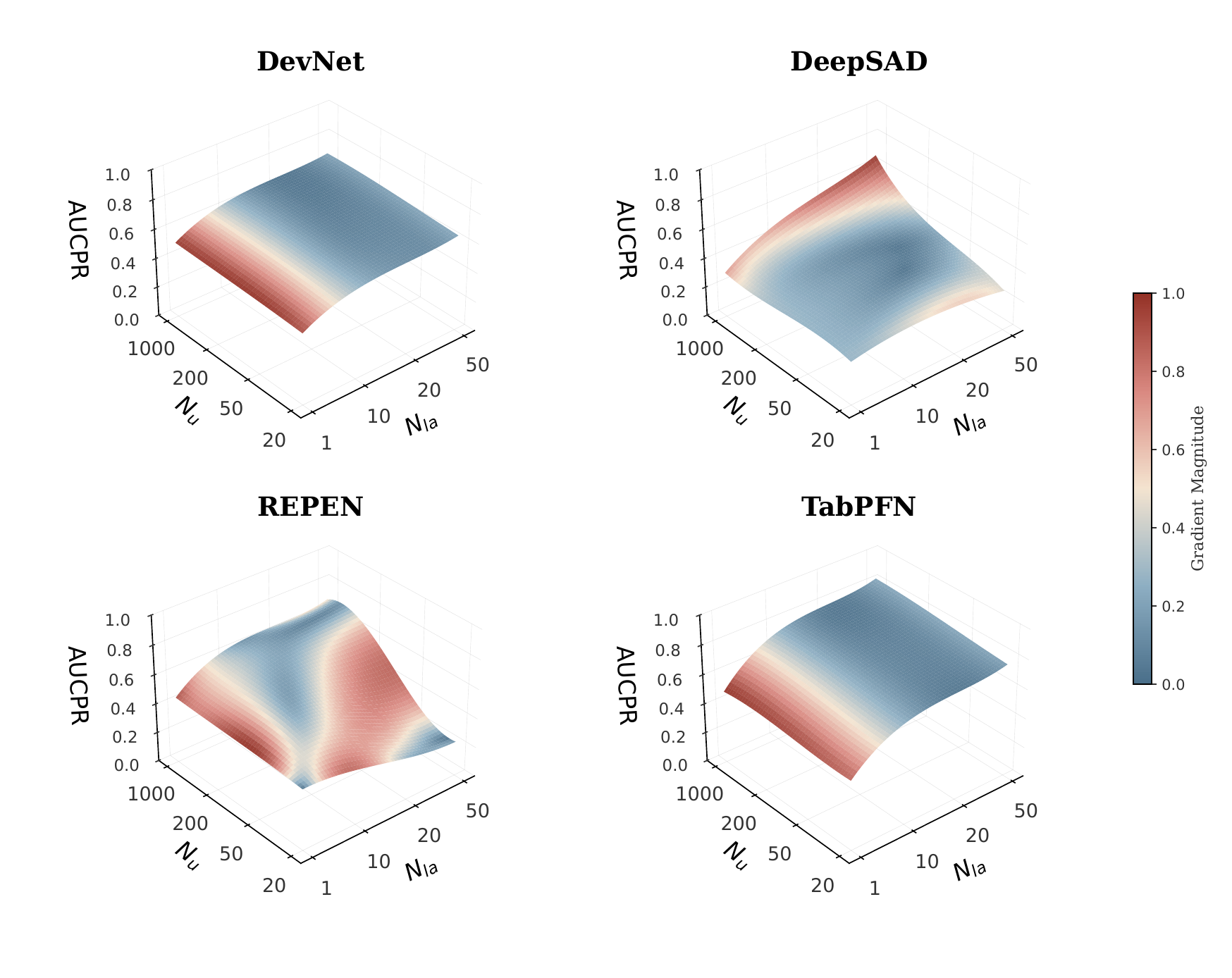}
    \caption{Joint effect}
    \label{fig:gradient_surface}
  \end{subfigure}
  % Main Caption
  \caption{AUCPR results on tabular datasets under varying labeled ($\nla$) and unlabeled ($\numu$) data amounts.}
  \label{fig:merged_classical_analysis}
\end{figure}

\noindent \textbf{Unlabeled data utility depends on label availability}.
Figure~\ref{fig:performance_curve} demonstrates that supervision levels shape the effectiveness of unlabeled data.
Under single-shot supervision (\(\nla=1\)), incorporating massive unlabeled samples yields minimal performance gains.
With only one labeled anomaly, the supervision signal is likely insufficient to guide model learning.
Consequently, models may struggle to exploit the distribution of unlabeled normal samples.
Conversely, when labeled anomalies are sufficient (\(\nla=50\)), scaling \(\numu\) to 1000 amplifies the performance gains.
Specifically, the maximum AUCPR improvement exceeds 0.40 in certain models.
These results reveal a synergistic relationship between labeled anomalies and unlabeled data.

\noindent\textbf{Method-specific unlabeled data dependencies}.
Different methods exhibit distinct dependencies on unlabeled data. Representation learning approaches (e.g., DeepSAD, REPEN) require a minimum threshold of unlabeled samples to define normal patterns; insufficient unlabeled data can cause performance degradation when labeled anomalies are added, due to instability of learned features or overfitting to sparse anomalies.
Surprisingly, despite pre-training on vast external corpora, tabular foundation models (e.g., LimiX, TabPFN) display scaling behaviors similar to specialized WSAD methods with respect to unlabeled data (Figure~\ref{fig:gradient_surface}): performance gains plateau as labeled data increases, while unlabeled data remains valuable when labeled examples are extremely scarce. This confirms that tabular foundation models inherently possess high label efficiency, enabling them to effectively maximize the utility of minimal supervision signals comparable to specialized methods.
\subsubsection{Sensitivity to Label Noise}
\label{sec:inaccurate_supervision}
% \noindent\textbf{Theme: Noise Tolerance limits and the Unsupervised Floor.}
We investigate how label noise affects model performance \revised{on tabular datasets}, revealing that noise impact varies systematically by corruption type and model architecture, and evaluate whether noise cleaning methods can effectively mitigate these effects.
% \noindent \textbf{Experiment Setup}.
Label noise is simulated by injecting errors into the training set: \textbf{Flip-Normal} (Normal $\to$ Anomaly), \textbf{Flip-Abnormal} (Anomaly $\to$ Normal), and \textbf{Double-Noise} (Both).
We introduce label noise by setting specific error ratios, as real-world annotation errors affect both classes at similar rates.

\begin{figure}[ht]
  \centering
  \begin{subfigure}[b]{0.48\linewidth}
    \includegraphics[width=\linewidth,trim={0.5cm 2.5cm 0cm 0.5cm},clip]{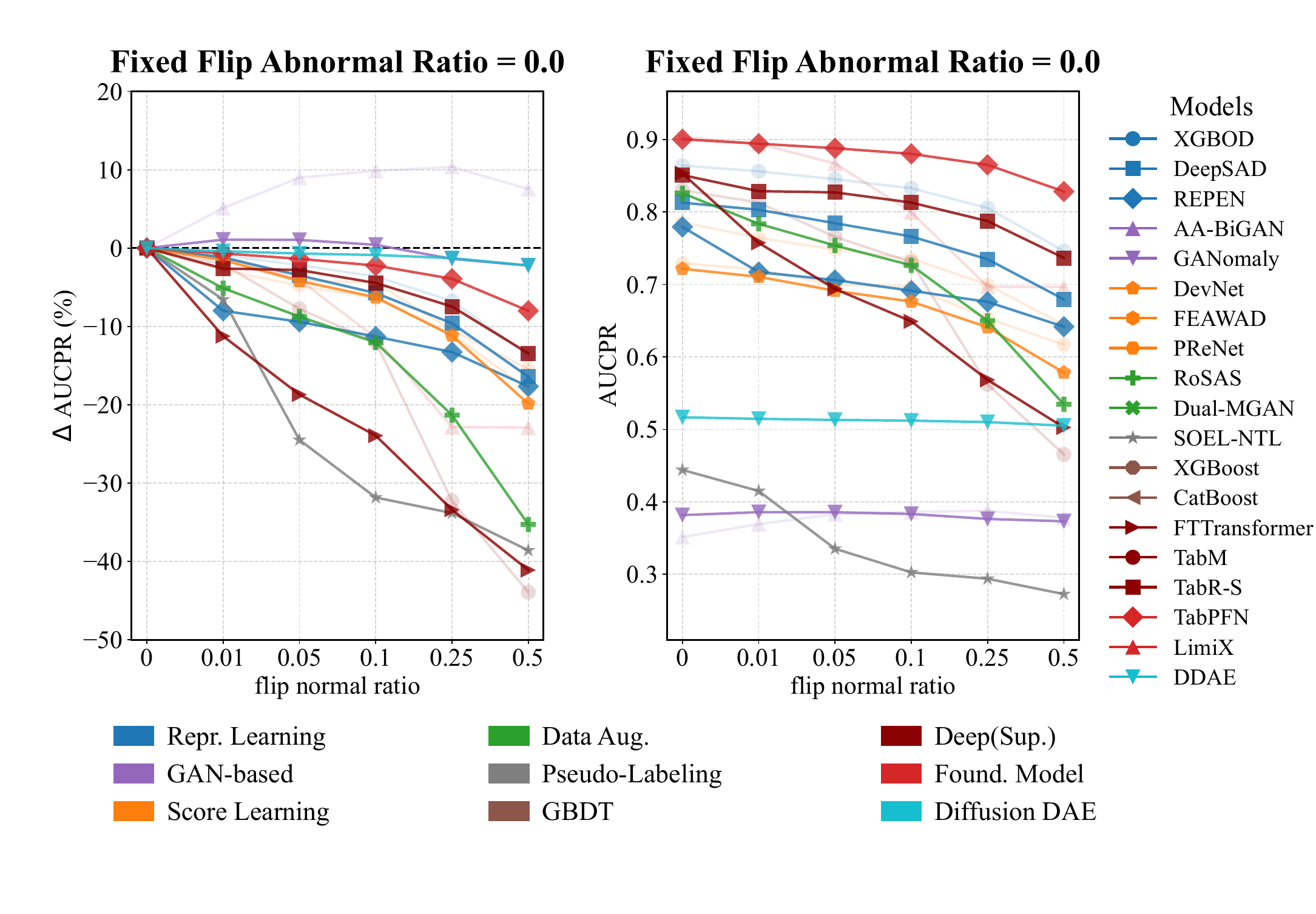}
    \vspace{-10pt}
    \caption{Flip normal}
    \label{fig:noise_flip_normal}
  \end{subfigure}
  \hfill
  \begin{subfigure}[b]{0.48\linewidth}
    \includegraphics[width=\linewidth,trim={0.5cm 2.5cm 0cm 0.5cm},clip]{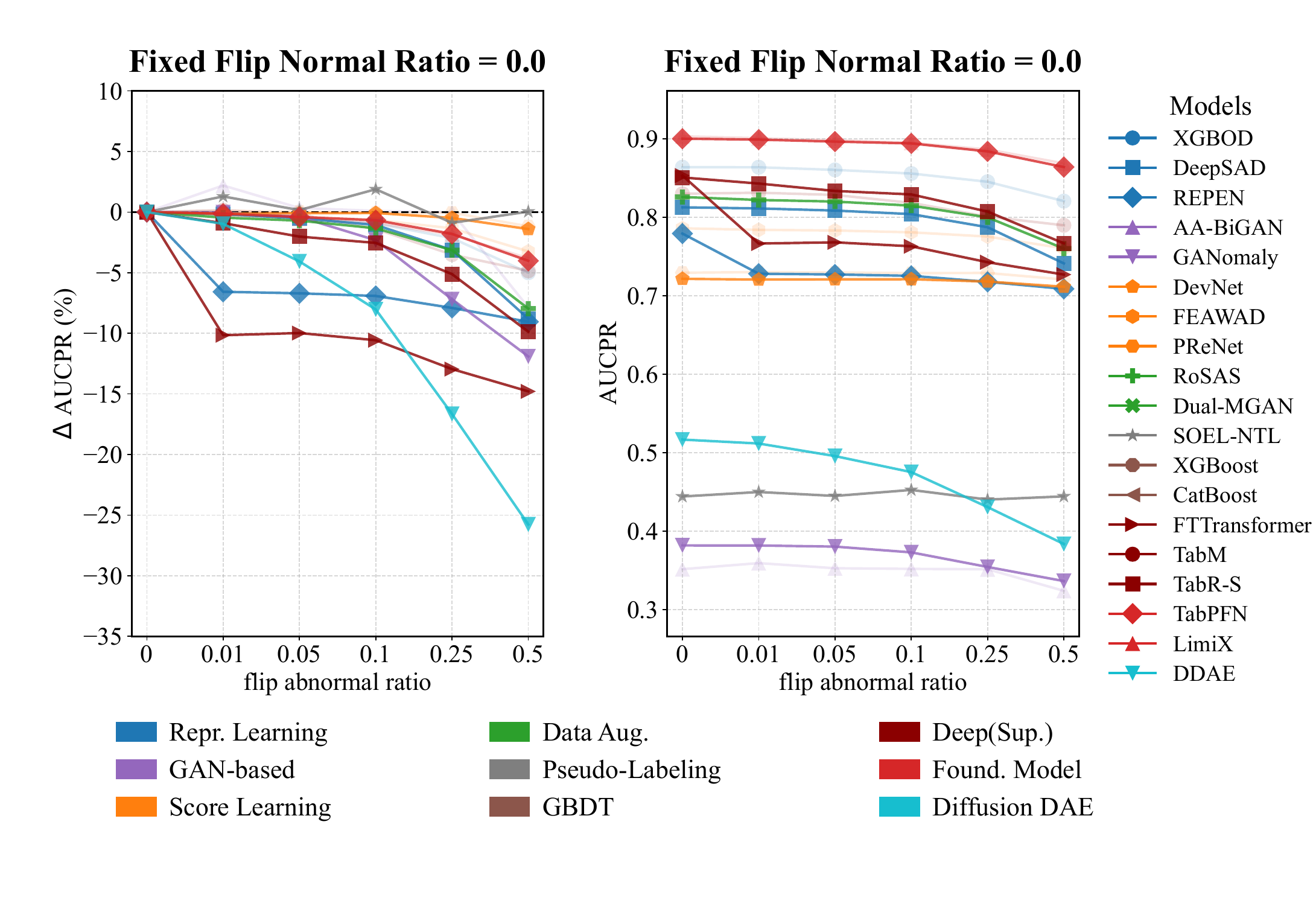}
    % \vspace{-10pt}
    \caption{Flip abnormal}
    \label{fig:noise_flip_abnormal}
  \end{subfigure}
  % vspace removed to save space
  \begin{subfigure}[b]{1.0\linewidth}
    \includegraphics[width=\linewidth,trim={0.3cm 2.3cm 0.3cm 1cm},clip]{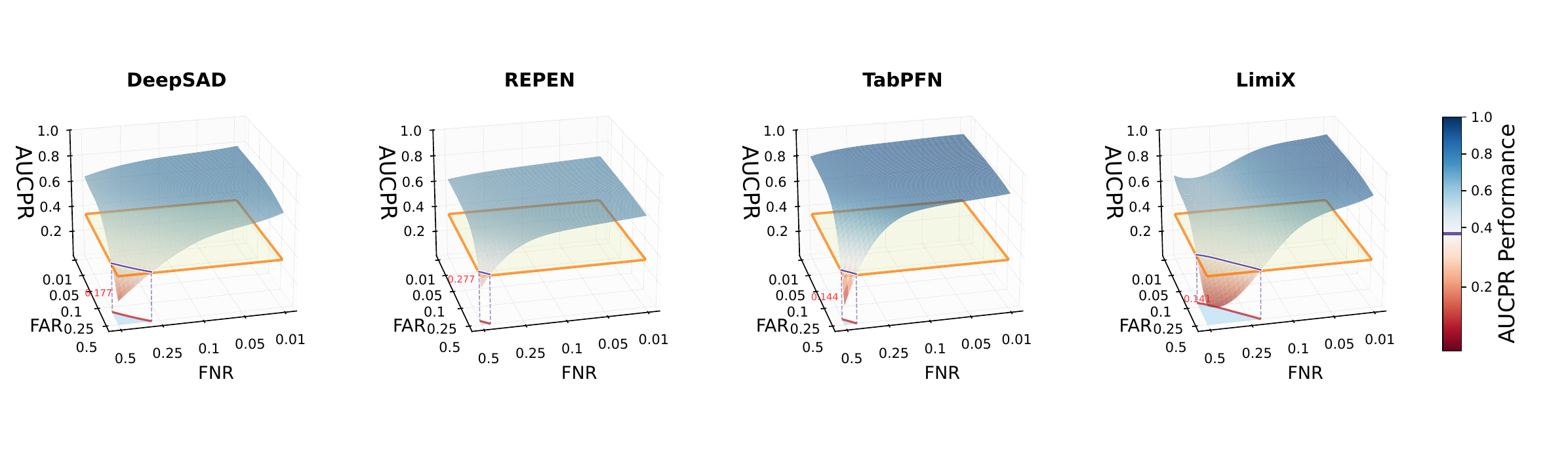}
    % \vspace{-10pt}
    \caption{Stability landscape}
    \label{fig:noise_stability_landscape}
  \end{subfigure}
  \caption{Model sensitivity to label noise: AUCPR degradation under asymmetric noise types.}
  \label{fig:noise_sensitivity_combined}
\end{figure}

\noindent \textbf{Asymmetric sensitivity to label noise types}.
Normal label noise is more destructive than abnormal label noise.
As illustrated in Figure.~\ref{fig:noise_sensitivity_combined}a and \ref{fig:noise_sensitivity_combined}b, models generally exhibit a steeper performance decline under Flip-Normal noise (FNR) compared to Flip-Abnormal noise (FAR).
This asymmetry stems from the anomaly-aware nature of WSAD methods.
FNR (Normal $\to$ Anomaly) introduces false positive anomalies that directly corrupt the anomaly pattern WSAD models explicitly explore during training, severely distorting learned abnormal representations.
In contrast, FAR (Anomaly $\to$ Normal) merely reduces labeled anomalies, which WSAD methods can tolerate due to their inherent robustness to unlabeled data noise, treating mislabeled anomalies as part of the unlabeled pool.

\noindent \textbf{Inconsistent robustness across models}.
As illustrated in Figure~\ref{fig:noise_sensitivity_combined}c, tabular foundation models demonstrate contrasting noise tolerance behaviors.
TabPFN forms a high, flat plateau indicating broad noise tolerance with ``cliff effect'' only under extreme dual-noise conditions, while LimiX exhibits sharp surface decline at lower noise thresholds despite superior clean-data performance.
Specialized WSAD methods (DeepSAD and REPEN) exhibit moderate noise robustness, degrading more gradually than LimiX but less robustly than TabPFN.

\noindent \textbf{Effectiveness of data cleaning method}.
We further evaluate whether automated data cleaning tools like Cleanlab~\cite{cleanlab} can mitigate label noise impact. As detailed in Table~\ref{tab:cleanlab_change}, external filtering may significantly benefit models lacking inherent robustness such as SOEL-NTL, TabM, and FTTransformer, which are the most sensitive to flip-normal noise (as shown in Figure~\ref{fig:noise_flip_normal}). Specifically, SOEL-NTL and TabM achieve substantial performance recovery under flip-normal noise, with AUCPR improvements of 9.61\% and 8.17\%, respectively. Conversely, this preprocessing proves detrimental to intrinsically robust architectures, as TabPFN and DevNet suffer performance degradations of 3.16\% and 1.71\% under the same conditions. This discrepancy suggests that cleaning helps sensitive models by correcting mislabeled samples but harms robust models by removing valuable difficult samples needed for generalization.

% Figure \ref{fig:cleanlab_efficacy} illustrates the relative performance variation ($\Delta \text{AUCPR}$, where positive values indicate improvement and negative values indicate degradation) induced by automated label cleaning (via Cleanlab) across different noise scenarios. In the Flip-Normal setting, where normal samples are mislabeled as anomalies, we observe substantial performance recovery for specific architectures. Notably, models lacking inherent robustness mechanisms, such as TabMCls and SOEL-NTL, exhibit remarkable gains,(e.g., both exceeding 30\% improvement at flip-normal ratio = 25\%). This indicates that external filtering effectively mitigates the impact of severe class imbalance caused by noise for sensitive models.

% In contrast, the interaction between Cleanlab and models designed with intrinsic robustness (e.g., TabPFN, DevNet) appears more complex. While cleaning exerts negligible influence at low noise levels (flip normal ratio < 10\%), it becomes increasingly detrimental to these robust models as the noise ratio rises. A striking example is TabPFN, which suffers significant performance degradation at flip-normal ratio=50\%, dropping to the lowest rank among evaluated models. This phenomenon suggests that using external cleaning tools at high noise levels may inadvertently discard informative hard samples, thereby undermining the generalization capability of robust algorithms that rely on these samples to define decision boundaries.

\begin{table}[htb]
  \centering
  \caption{Average absolute performance improvement ($\Delta$ AUCPR \%) after Cleanlab denoising.}
  \label{tab:cleanlab_change}
  % \vspace{-8pt}
  \resizebox{\linewidth}{!}{%
    \begin{tabular}{lccc}
      \toprule
      \textbf{Model} & \textbf{Flip-Normal} & \textbf{Flip-Abnormal} & \textbf{Average} \\
      \midrule
      SOEL-NTL & \textbf{9.61\%} & 0.87\% & \textbf{5.24\%} \\
      TabM & \textbf{8.17\%} & 0.80\% & \textbf{4.49\%} \\
      FTTransformer & \textbf{2.13\%} & 0.41\% & \textbf{1.27\%} \\
      DDAE & 0.29\% & 0.67\% & 0.48\% \\
      GANomaly & 0.49\% & 0.25\% & 0.37\% \\
      DeepSAD & -0.15\% & 0.08\% & -0.04\% \\
      RoSAS & -0.24\% & 0.15\% & -0.04\% \\
      REPEN & -0.68\% & -0.08\% & -0.38\% \\
      XGBOD & -1.00\% & -0.32\% & -0.66\% \\
      DevNet & -1.71\% & 0.02\% & -0.84\% \\
      CatBoost & -1.56\% & -0.46\% & -1.01\% \\
      TabPFN & -3.16\% & -0.43\% & -1.80\% \\
      \bottomrule
    \end{tabular}%
  }
\end{table}
\subsubsection{Out-of-Distribution Generalization Analysis}
\label{sec:ood_main_text}
\begin{figure}[htbp]
  \centering
  \begin{subfigure}[b]{\linewidth}
    \centering
    \includegraphics[width=\linewidth]{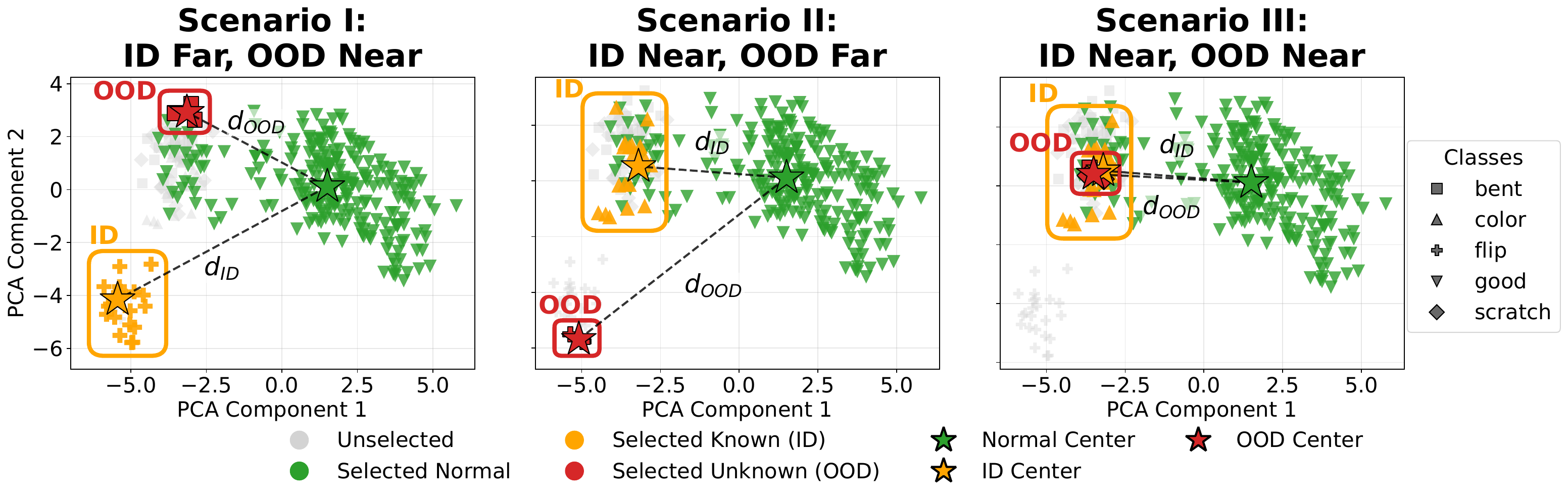}
    \vspace{-16pt}
    \caption{PCA visualization of feature distributions.}
    \label{fig:ood_pca}
  \end{subfigure}

  \begin{subfigure}[b]{\linewidth}
    \hspace*{0.02\linewidth}
    \includegraphics[width=0.97\linewidth]{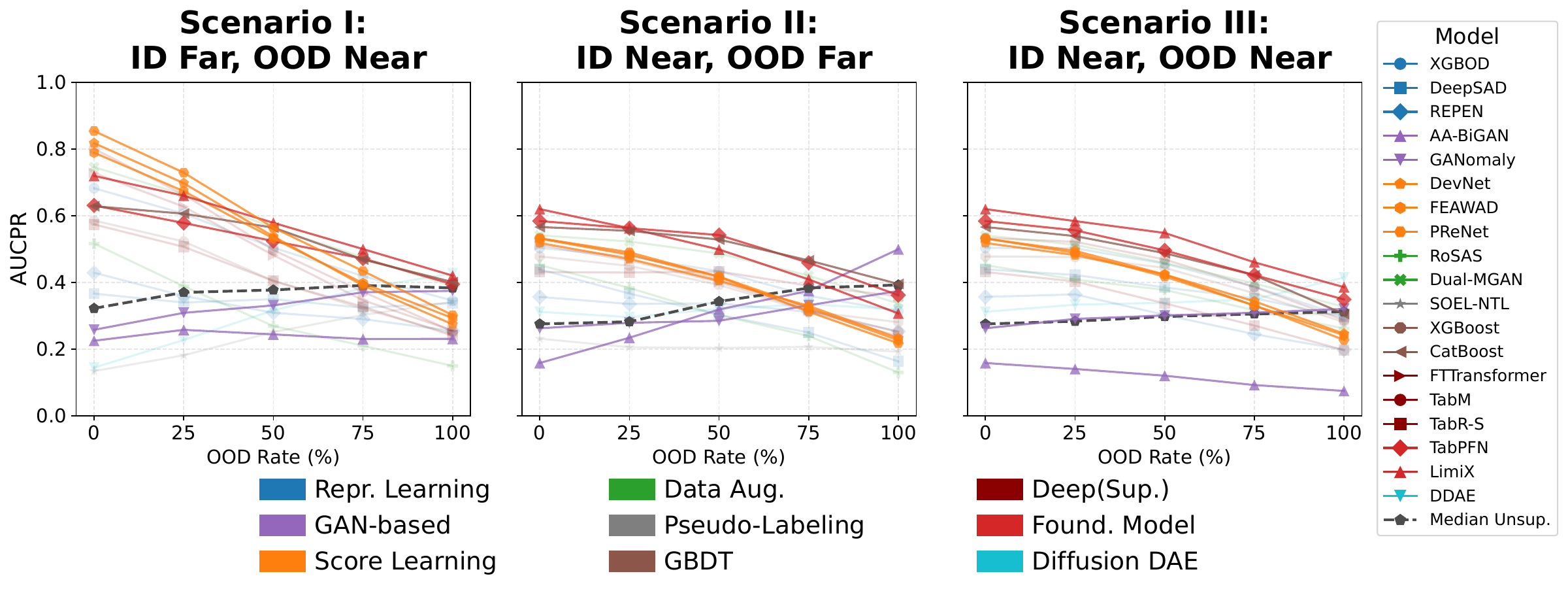}
    \vspace{-16pt}
    \centering
    \caption{AUCPR performance ($\gamma_{la}=10\%$).}
    \vspace{-1pt}
    \label{fig:ood_pr_scenarios}
  \end{subfigure}

  % \vspace{-8pt}
  \caption{Analysis of incomplete OOD across 3 Settings.}
  \label{fig:ood_combined_analysis}
  \vspace{-5pt}
\end{figure}

We evaluate robustness under distribution shift \revised{on tabular datasets} by controlling the distance between training (ID) and testing (OOD) anomalies.
Using PCA-derived feature-space distances from the normal centroid (Figure~\ref{fig:ood_pca}), we partition anomaly classes into Known ($\mathcal{A}_{known}$, used for training) and Unknown ($\mathcal{A}_{unknown}$, reserved for OOD testing), where Known anomalies are sampled at label ratio $r_{la}$.
This yields three distinct settings: \textbf{Setting I} (ID Far, OOD Near) assesses whether models trained on obvious anomalies can generalize to detect subtle threats; \textbf{Setting II} (ID Near, OOD Far) examines if fitting hard anomalies sacrifices the performance on distinct, easy outliers; and \textbf{Setting III} (ID Near, OOD Near) evaluates the boundary precision when all anomalies are near-normal.

\noindent\textbf{WSAD methods struggle to generalize beyond known anomaly types}.
In Scenario~I of Figure~\ref{fig:ood_pr_scenarios}, where ID anomalies are far from normal, WSAD methods (DevNet, PReNet, FEAWAD) achieve the highest AUCPR at OOD rate = 0\%, surpassing even tabular foundation models.
However, this advantage is narrow: performance drops sharply as the OOD rate increases, falling below tabular foundation models and GBDTs at moderate OOD rates.
This performance reversal reveals an ID-specific optimization that collapses when novel anomaly types appear.
In contrast, Settings~II and~III show no such reversal: When ID anomalies are near-normal, harder training conditions prevent WSAD methods from surpassing tabular foundation models, leaving them consistently below even at OOD rate = 0\%.

\noindent\textbf{Non-WSAD methods exhibit varied OOD tolerance}.
Tabular foundation models (e.g., LimiX, TabPFN) and gradient boosting (e.g., CatBoost) degrade less severely than WSAD methods as OOD rates increase.
They remain competitive even when unknown anomalies dominate (OOD rate $\to$ 100\%).
This tolerance likely stems from pre-trained or ensemble strategies that generalize beyond specific types.
In contrast, reconstruction methods (e.g., GANomaly and AA-BiGAN) maintain stable performance across OOD rates by modeling normal patterns, but their absolute performance remains inferior, typically at or below unsupervised baselines.
An exception occurs in Scenario~II: when OOD anomalies are far from normal, GANomaly's AUCPR rises with OOD rate, as the normal-centric objective naturally separates distant anomalies.
\ifarxiv Additional results for all datasets are shown in Figure~\ref{fig:ood_dist_combined} (Appendix~\ref{sec:appendix_ood_extend}).\else Additional results for all datasets are provided in the full version.\fi

\begin{figure}[htbp]
  \centering
  \includegraphics[width=\linewidth]{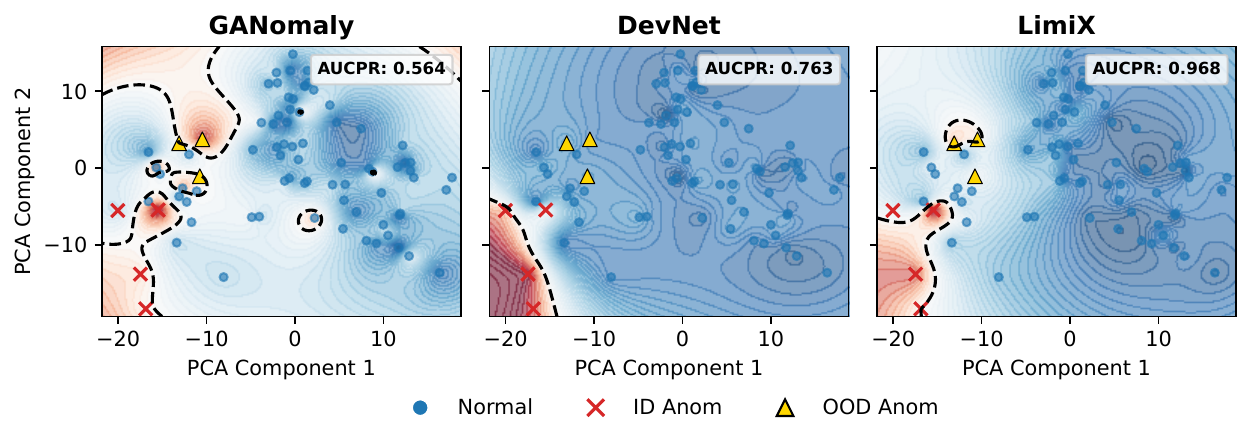}
  \vspace{-20pt}
  \caption{Anomaly score decision boundaries on Metal\_nut (Scenario I, $\gamma_{la}=10\%$, OOD rate=50\%). Redder regions indicate higher anomaly scores.}
  \label{fig:ood_decision_boundary}
\end{figure}

Figure~\ref{fig:ood_decision_boundary} visualizes the output anomaly score for three representative models on Metal\_nut (Scenario I).
GANomaly detects both ID and OOD anomalies but assigns uniformly low scores with excessive false positives, as normal-only reconstruction cannot leverage supervision effectively.
DevNet assigns high scores to ID anomalies but near-zero scores to OOD samples.
Its decision regions narrowly surround known anomaly types, failing to generalize to nearby OOD samples.
Conversely, LimiX constructs compact boundaries that cover both ID and OOD anomalies, achieving superior generalization without fragmenting the decision space.
This reveals that WSAD methods memorize specific instances rather than learning generalizable patterns—a form of ID overfitting that prevents generalization even when OOD anomalies are nearby.

\subsubsection{Can Methods Transfer Across Supervision Types?}
\label{sec:inexact_interop}
We evaluate whether methods designed for one supervision type transfer to another by testing incomplete supervision methods on the inexact scenario (video MIL tasks) and vice versa, examining how domain-specific inductive biases affect generalization between scenarios.

\begin{table}[htbp!]
  \centering
  % \vspace{-8pt}
  \caption{Supervision type transferability evaluation.}
  \label{tab:comparison_results}
  \vspace{-8pt}
  \resizebox{\columnwidth}{!}{%
    \begin{tabular}{lllcc}
      \toprule
      \textbf{Paradigm} & \textbf{Model} & \textbf{Category} & \textbf{Video MIL} & \textbf{Tabular MIL} \\
      \midrule
      \multirow{7}{*}{MIL}
      & AR-Net & Dynamic MIL & 0.441 (3) & - \\
      & Sultani & Vanilla MIL & 0.398 (7) & 0.252 (2) \\
      & MGFN & Magnitude MIL & 0.308 (12) & - \\
      & RTFM & Magnitude MIL & 0.395 (8) & - \\
      & UR-DMU & Uncertainty-Aware MIL & 0.363 (9) & - \\
      & VadCLIP & Language-Guided MIL & 0.354 (10) & - \\
      & GCN-Anomaly & Label Denoising & 0.351 (11) & 0.127 (8) \\
      \midrule
      \multirow{6}{*}{Broadcast}
      & IForest & Isolation-based & 0.143 (13) & 0.174 (7) \\
      & CatBoost & GBDT & 0.428 (5) & 0.230 (6) \\
      & DevNet & Score Learning & 0.437 (4) & 0.235 (5) \\

      & FEAWAD & Reconstruction & 0.452 (2) & 0.241 (3) \\
      & DeepSAD & Score Learning & \textbf{0.453 (1)} & 0.240 (4) \\

      & TabPFN & Found. Model & 0.408 (6) & \textbf{0.260 (1)} \\
      \bottomrule
    \end{tabular}%
  }
  % \vspace{-8pt}
\end{table}

\noindent\textbf{Transferability across supervision types reveals domain-specific constraints}.
% Propagating bag-level labels to instance-level in inexact supervision—when video structure is ignored and treated as i.i.d. tabular rows—transforms the problem into handling instance-level label noise, the core challenge of inaccurate supervision.
Label broadcasting transforms inexact supervision into an inaccurate supervision problem. By treating video bags as i.i.d. tabular instances, bag-level labels propagate to instance-level, introducing label noise as the primary challenge.
Surprisingly, Table~\ref{tab:comparison_results} illustrates that methods designed for incomplete supervision—despite lacking temporal modeling or MIL pooling—even outperform specialized inexact approaches on video MIL tasks: DeepSAD achieves mean AUCPR 0.453, surpassing the specialized MIL model AR-Net's AUCPR 0.441.
Conversely, specialized inexact methods show mixed transferability to tabular MIL. Sultani transfers relatively well and ranks second, but it is still outperformed by TabPFN, while GCN-Anomaly remains less effective in both settings.
% These results suggest that a method's original supervision type does not strictly determine where it performs best. This highlights the importance of evaluating WSAD methods through a unified taxonomy.
These results demonstrate the transferability of methods across different supervision types in \prob. This compatibility suggests that future algorithm designs can draw mutual inspiration from different weak supervision settings.

\begin{figure}[t!]
  \centering
  \includegraphics[width=\linewidth, trim=0cm 0.9cm 0cm 0cm, clip]{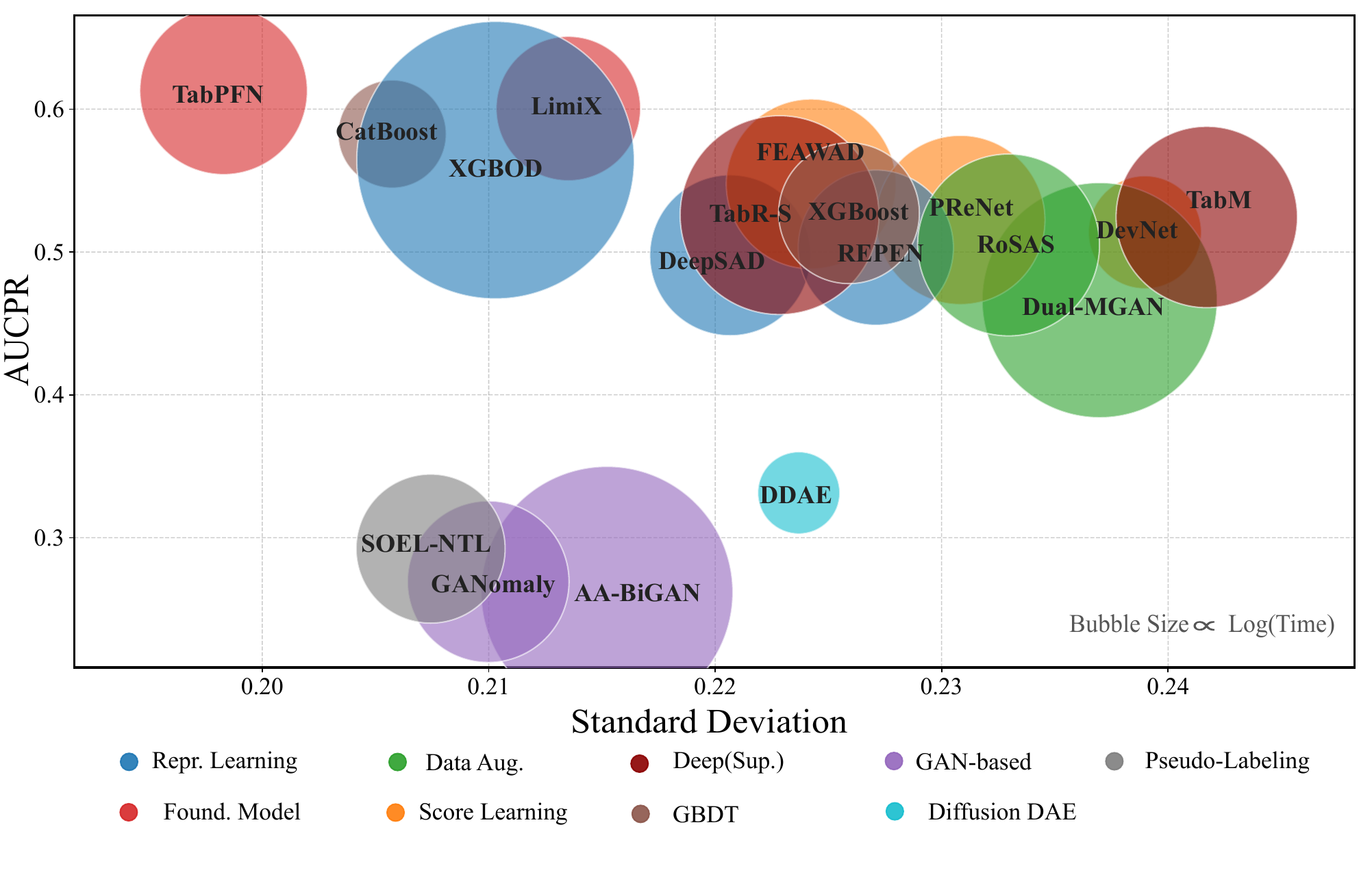}
  \vspace{-20pt}
  \caption{Performance across all scenarios.}
  \label{fig:global_perf_time}
  \vspace{-10pt}
\end{figure}

\begin{figure}[t!]
  \centering
  \includegraphics[width=\linewidth]{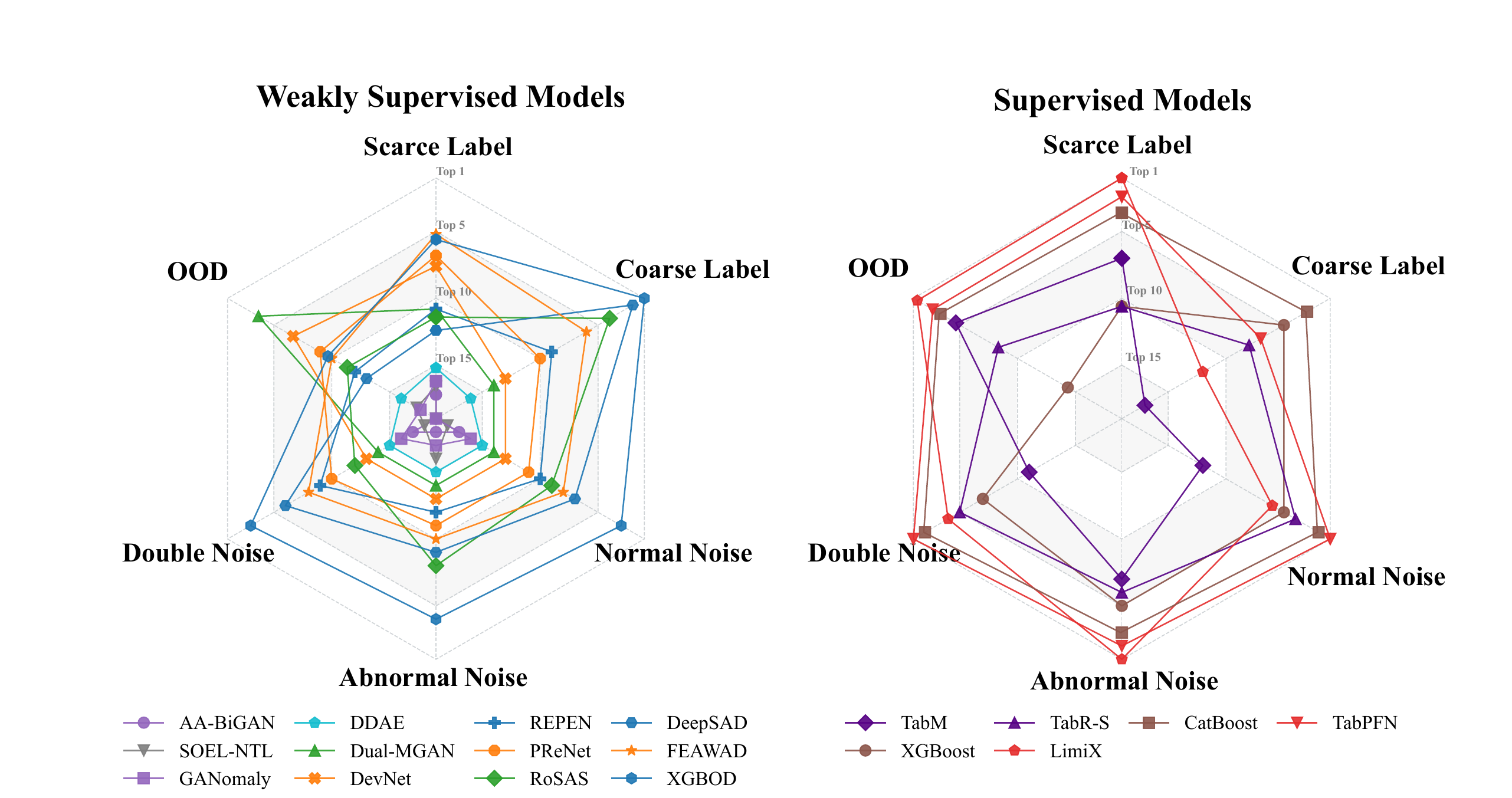}
  \vspace{-20pt}
  \caption{Model ranking radar chart.}
  \label{fig:global_radar}
  \vspace{-10pt}
\end{figure}

\subsection{Performance Summary}
\label{sec:global_performance}

To conclude our analysis, we present a global overview of algorithmic performance across all settings.
Figure~\ref{fig:global_perf_time} visualizes the trade-off among performance, stability, and efficiency, where models in the top-left region with smaller bubbles achieve the best balance.
TabPFN dominates this space with the highest mean AUCPR (\(\sim\)0.65), the lowest variability, and the smallest training cost.
Most \prob{} methods, regardless of category, cluster tightly in a narrow band, suggesting that different algorithmic designs yield only marginal differentiation under aggregated evaluation.
GAN-based approaches fall to the bottom of the chart despite incurring the largest computational overhead.

Figure~\ref{fig:global_radar} presents model rankings (based on mean AUCPR) across eight distinct scenarios spanning varying labeled anomaly ratios (100\%, 50\%, 1\%), noise conditions (FNR, FAR, Double Noise), and three data modalities (Tabular, Inexact, VAD).
\revised{Crucially, the ranking for each scenario-modality combination is calculated independently within its corresponding modality, without any cross-modality ranking aggregation.}
This cross-scenario comparison confirms that specialized WSAD algorithms are inferior to tabular foundation models in most scenarios, with no single method dominating all settings, underscoring the need for detailed per-scenario analysis and scenario-specific algorithm adaptation/selection.

\vspace{-8pt}
\section{Conclusion}
\label{sec:conclusion}

This work presents \workname, the first unified evaluation framework for \prob that enables fair comparison across diverse supervision settings and algorithms.
We evaluate 36 algorithms on 61 datasets spanning tabular, image, text, and video modalities, conducting over 700K experiments.
Results show that specialized WSAD algorithms excel only under extreme label scarcity; as labeled data increases or anomaly types shift, tabular foundation models and general classification methods consistently dominate.
We also reveal strong correlations across supervision settings and asymmetric model sensitivity to different noise types.

Our findings suggest two directions for future research.
First, WSAD can benefit through either leveraging \textbf{general-purpose foundation models} to improve representation quality in downstream AD-specific models, or developing \textbf{AD foundation models} that capture universal abnormal patterns across domains.
Second, algorithmic challenges such as poor OOD generalization and asymmetric noise sensitivity highlight the need for algorithms that learn robust, generalizable abnormal patterns. Going forward,
We plan to expand \workname with additional modalities and continuously update the benchmark with emerging algorithms.

\begin{acks}
  \label{sec.acknowledgments}
  \noindent This work was supported by the National Natural Science Foundation of China (Nos. 72271151, 72342009, 72442024, 72172085) and Ant Group.
\end{acks}

% \clearpage

\bibliographystyle{ACM-Reference-Format}
\balance
\bibliography{wsadbench}

% \clearpage
%%
%% Appendix
%%
\appendix
\section{Benchmark Details}

\subsection{Dataset Summaries and Processing Specs}
\label{appendix:datasets}
\begin{revise}
  We detail the diverse collection of datasets evaluated in \workname{}, spanning tabular, image, text and video modalities.
  These benchmarks are selected to ensure broad representativeness across varying data structures and AD scenarios.

  \subsubsection{Statistical Summaries of Benchmarks}
  \label{appendix:datasets_statistical_summaries}

  We evaluate our benchmark on a diverse suite of datasets covering tabular, image, and text modalities.
  The tabular benchmarks span domains including network intrusion, medical diagnosis, and industrial fault detection.
  The visual and natural language benchmarks cover critical defect detection and text classification tasks.
  Table~\ref{tab:all_datasets_summary} summarizes the comprehensive statistics and properties for these three modalities.

  We also incorporate 4 standard video anomaly detection (VAD) benchmarks, UCF-Crime, ShanghaiTech, XD-Violence, and TAD, covering diverse anomaly types in surveillance and action settings.
  Table~\ref{tab:dataset_diff} summarizes their detailed characteristics, including video counts and clip distributions.

  \subsubsection{Out-of-Distribution Datasets}
  \label{appendix:datasets_ood}

  We incorporate diverse anomaly datasets featuring various anomaly types for out-of-distribution generalization research.
  Table~\ref{tab:ood_description} summarizes the detailed statistics and properties of these OOD datasets.

  \begin{table}[htb]
  \caption{Description of OOD datasets}
  \vspace{-10pt}
  \label{tab:ood_description}
  \resizebox{\columnwidth}{!}{%
    \begin{tabular}{l l c c c c c}
      \toprule
      \textbf{Type} & \textbf{Dataset} & \textbf{\shortstack{Train\\(Normal)}} & \textbf{\shortstack{Test\\(Normal)}} & \textbf{\shortstack{Test\\(Ano)}} & \textbf{Ano Classes} & \textbf{Source} \\
      \midrule
      \multirow{3}{*}{Texture}
      & AITEX & 1,692 & 564 & 183 & 12 & \href{https://tinyurl.com/aitex-defect}{Link} \\
      & Carpet & 280 & 28 & 89 & 5 & \href{https://www.mvtec.com/company/research/datasets/mvtec-ad}{Link} \\ % https://www.mvtec.com/company/research/datasets/mvtec-ad
      & ELPV & 1,131 & 377 & 715 & 2 & \href{https://github.com/zae-bayern/elpv-dataset}{Link} \\% https://github.com/zae-bayern/elpv-dataset
      \midrule
      \multirow{2}{*}{Object}
      & Mastcam & 9,302 & 426 & 881 & 11 & \href{https://zenodo.org/records/3732485}{Link} \\  % https://zenodo.org/records/3732485
      & Metal Nut & 220 & 22 & 93 & 4 & \href{https://www.mvtec.com/company/research/datasets/mvtec-ad}{Link} \\ % https://www.mvtec.com/company/research/datasets/mvtec-ad
      \midrule
      Medical
      & HyperKvasir & 2,021 & 674 & 757 & 4 & \href{https://datasets.simula.no/hyper-kvasir/}{Link} \\% https://datasets.simula.no/hyper-kvasir/
      \bottomrule
    \end{tabular}%
  }
\end{table}

  \subsubsection{Video Feature Specifications and Standardization}
  \label{appendix:datasets_video_specifications}

  To unify VAD with other modalities under MIL, we decouple feature extraction from weakly supervised reasoning.
  This design isolates algorithm evaluation from upstream video perception.
  Extracting features using pre-trained models before applying WSAD algorithms is standard in VAD \cite{sultani2018real,tian2021weakly,zhou2023dual}.
  Under this setting, pre-trained feature extractors capture temporal dynamics and spatial patterns, preserving the complexity of VAD.

  Additionally, historical VAD evaluations often suffered from inconsistencies due to different pre-trained backbones or checkpoints.
  This mismatch makes it difficult to determine whether performance gains stem from the algorithm or superior feature representations.
  To ensure fair comparisons, we constructed a standardized feature extraction pipeline from raw videos.
  Our pipeline reads video frames, applying RGB normalization and TenCrop spatial augmentations.
  We extract dense, non-overlapping clips using sliding windows and group them into segments.
  A producer-consumer architecture feeds these segments into identical frozen pre-trained 3D models, such as I3D \cite{carreira2017quo} and SlowFast \cite{feichtenhofer2019slowfast}.
  This process generated over 2.4TB of standardized, memory-mapped embedding files.
  We open-source these features to establish a fair benchmark for evaluating pure weakly supervised reasoning.

  \begin{table*}[t]
  \centering
  \caption{\revised{Summary of tabular, image, and text datasets evaluated in \workname{}}}
  \vspace{-10pt}
  \label{tab:all_datasets_summary}
  \resizebox{\textwidth}{!}{%
    \begin{tabular}{l c c c c @{\hskip 0.6cm} l c c c c @{\hskip 0.6cm} l c c c c}
      \toprule
      \textbf{Dataset} & \textbf{\#Samples} & \textbf{\#Features} & \textbf{Anomaly \%} & \textbf{Domain} &
      \textbf{Dataset} & \textbf{\#Samples} & \textbf{\#Features} & \textbf{Anomaly \%} & \textbf{Domain} &
      \textbf{Dataset} & \textbf{\#Samples} & \textbf{\#Features} & \textbf{Anomaly \%} & \textbf{Domain} \\
      \midrule
      \multicolumn{5}{c}{\textbf{Tabular Datasets (Part 1)}} &
      \multicolumn{5}{c}{\textbf{Tabular Datasets (Part 2)}} &
      \multicolumn{5}{c}{\textbf{Tabular Datasets (Part 3)}} \\
      \cmidrule(lr){1-5} \cmidrule(lr){6-10} \cmidrule(lr){11-15}
      ALOI & 49,534 & 27 & 3.04 & Image & Lymphography & 148 & 18 & 4.05 & Healthcare & Waveform & 3,443 & 21 & 2.90 & Physics \\
      annthyroid & 7,200 & 6 & 7.42 & Healthcare & magic.gamma & 19,020 & 10 & 35.16 & Physical & WBC & 223 & 9 & 4.48 & Healthcare \\
      backdoor & 95,329 & 196 & 2.44 & Network & mammography & 11,183 & 6 & 2.32 & Healthcare & WDBC & 367 & 30 & 2.72 & Healthcare \\
      breastw & 683 & 9 & 34.99 & Healthcare & mnist & 7,603 & 100 & 9.21 & Image & Wilt & 4,819 & 5 & 5.33 & Botany \\
      campaign & 41,188 & 62 & 11.27 & Finance & musk & 3,062 & 166 & 3.17 & Chemistry & wine & 129 & 13 & 7.75 & Chemistry \\
      cardio & 1,831 & 21 & 9.61 & Healthcare & optdigits & 5,216 & 64 & 2.88 & Image & WPBC & 198 & 33 & 23.74 & Healthcare \\
      Cardiotocography & 2,114 & 21 & 22.04 & Healthcare & PageBlocks & 5,393 & 10 & 9.46 & Document & yeast & 1,484 & 8 & 34.16 & Biology \\
      \cmidrule(lr){11-15}
      celeba & 202,599 & 39 & 2.24 & Image & pendigits & 6,870 & 16 & 2.27 & Image & \multicolumn{5}{c}{\textbf{Image Datasets}} \\
      \cmidrule(lr){11-15}
      census & 299,285 & 500 & 6.20 & Sociology & Pima & 768 & 8 & 34.90 & Healthcare & CIFAR10 & 5,263 & 512 & 5.00 & Image \\
      cover & 286,048 & 10 & 0.96 & Botany & satellite & 6,435 & 36 & 31.64 & Astronautics & FashionMNIST & 6,315 & 512 & 5.00 & Image \\
      donors & 619,326 & 10 & 5.93 & Sociology & satimage-2 & 5,803 & 36 & 1.22 & Astronautics & MNIST-C & 10,000 & 512 & 5.00 & Image \\
      fault & 1,941 & 27 & 34.67 & Physical & shuttle & 49,097 & 9 & 7.15 & Astronautics & MVTec-AD & 5,354 & 512 & 23.50 & Image \\
      fraud & 284,807 & 29 & 0.17 & Finance & skin & 245,057 & 3 & 20.75 & Image & SVHN & 5,208 & 512 & 5.00 & Image \\
      \cmidrule(lr){11-15}
      glass & 214 & 7 & 4.21 & Forensic & smtp & 95,156 & 3 & 0.03 & Web & \multicolumn{5}{c}{\textbf{Text Datasets}} \\
      \cmidrule(lr){11-15}
      Hepatitis & 80 & 19 & 16.25 & Healthcare & SpamBase & 4,207 & 57 & 39.91 & Document & Agnews & 10,000 & 768 & 5.00 & NLP \\
      http & 567,498 & 3 & 0.39 & Web & speech & 3,686 & 400 & 1.65 & Linguistics & Amazon & 10,000 & 768 & 5.00 & NLP \\
      InternetAds & 1,966 & 1,555 & 18.72 & Image & Stamps & 340 & 9 & 9.12 & Document & Imdb & 10,000 & 768 & 5.00 & NLP \\
      Ionosphere & 351 & 33 & 35.90 & Physical & thyroid & 3,772 & 6 & 2.47 & Healthcare & Yelp & 10,000 & 768 & 5.00 & NLP \\
      landsat & 6,435 & 36 & 20.71 & Astronautics & vertebral & 240 & 6 & 12.50 & Biology & 20newsgroups & 11,905 & 768 & 4.96 & NLP \\
      letter & 1,600 & 32 & 6.25 & Image & vowels & 1,456 & 12 & 3.43 & Linguistics & & & & & \\
      \bottomrule
    \end{tabular}%
  }
\end{table*}

  \begin{table}[t]
  \centering
  \caption{Statistics of video anomaly detection Benchmarks.}
  \vspace{-10pt}
  \label{tab:dataset_diff}
  \resizebox{\columnwidth}{!}{%
    \begin{tabular}{lccccccccccc}
      \toprule
      \multirow{2}{*}{\textbf{Dataset}} & \multirow{2}{*}{\textbf{Video Num}} & \multicolumn{5}{c}{\textbf{Clip Number Statistics}} & \multicolumn{5}{c}{\textbf{Clip Range Distribution (\%)}} \\
      \cmidrule(lr){3-7} \cmidrule(lr){8-12}

      & & \textbf{Max} & \textbf{Min} & \textbf{Mean} & \textbf{Median} & \textbf{Std} & \textbf{$\leq$50} & \textbf{51--100} & \textbf{101--200} & \textbf{201--500} & \textbf{$>$500} \\
      \midrule

      TAD          & 500  & 938   & 2  & 67.62  & 21  & 118.27 & 74.60\% & 7.80\%  & 7.00\%  & 9.00\%  & 1.60\% \\
      UCF-Crime   & 1900 & 61032 & 7  & 453.32 & 133 & 2055.3 & 13.30\% & 23.60\% & 27.70\% & 21.80\% & 13.50\% \\
      XD-Violence  & 4751 & 16225 & 3  & 246.56 & 158 & 478.77 & 11.20\% & 19.80\% & 29.80\% & 30.00\% & 9.20\% \\
      ShanghaiTech & 437  & 261   & 13 & 45.66  & 46  & 24.85  & 67.00\% & 31.10\% & 1.40\%  & 0.50\%  & 0.00\% \\

      \bottomrule
    \end{tabular}
  }

\end{table}
\end{revise}
\subsection{Details of Tabular MIL Bags Generation}
\label{appendix:mil_bag_generation}

We formulate the bag generation process as a sampling procedure from decoupled distributions of normal ($\mathcal{X}_n$) and abnormal ($\mathcal{X}_a$) data.
The goal is to generate a dataset of bags $\mathcal{B} = \{B_1, B_2, \dots, B_{K}\}$, where each bag $B_i$ contains $L$ instances ($L=N_{samples}$).
For each bag $B_i$, a label $Y_i \in \{0, 1\}$ is determined by a random variable $p \sim \text{Uniform}(0, 1)$ and a threshold $P$:

\noindent\textbf{Normal Bags} ($Y_i = 0$): Condition: $p < P$. The bag is composed entirely of normal instances:
$$B_i = \{x_j\}_{j=1}^{L}, \quad \text{where } x_j \sim \mathcal{X}_n$$

\noindent\textbf{Abnormal Bags} ($Y_i = 1$): Condition: $p \ge P$. The bag is constructed by injecting a random number of anomalies.
Let $m$ be the number of abnormal instances, sampled uniformly such that $m \sim U\{1, \dots, \lfloor L/2 \rfloor - 1\}$.
The bag is composed of:
$$B_i = \{x_j\}_{j=1}^{m} \cup \{x_k\}_{k=1}^{L-m}, \quad \text{where } x_j \sim \mathcal{X}_a, x_k \sim \mathcal{X}_n$$

This protocol ensures that normal bags are pure, while abnormal bags contain a minority of anomalies mixed with background normal samples, mimicking real-world MIL scenarios.
Following the experimental setup in~\cite{perini2023learning}, we set the number of instances per bag to 10.
Our experiments were conducted on the Tabular datasets, which were preprocessed into bag-structured formats suitable for Multi-Instance Learning.

\subsection{Experiment Setting}

\label{appendix:hyperparams}

This section provides additional details on experiment setting.

\paragraph{Hyperparameter Configurations.}
For fair comparison and reproducibility, \textbf{all baseline methods are evaluated using their official default hyperparameters without additional tuning}.
Classical unsupervised baselines (e.g., LOF, IForest) follow scikit-learn defaults, while deep models adopt consistent optimization strategies (Adam optimizer, learning rate 1e-3 unless otherwise noted). A complete list of parameters—including embedding dimensions, hidden layer widths, loss coefficients, and training epochs—for each model–dataset pair is provided in the supplementary material (see repository link in Appendix~\ref{appendix:github}).

\paragraph{Computational Resources.}
Classical anomaly detection models are run on an Intel(R) Xeon(R) Platinum 8358 CPU @ 2.60GHz, 1TB RAM, 128-core server. For deep learning models, experiments are conducted on a server equipped with eight NVIDIA A800-SXM4-80GB GPUs.
\paragraph{Evaluation Metrics.} We employ two standard metrics: Area Under the Receiver Operating Characteristic curve (AUCROC) and Area Under the Precision-Recall curve (AUCPR). AUCROC evaluates the global ranking quality by plotting the True Positive Rate (TPR) against the False Positive Rate (FPR) across all thresholds. It is robust to decision thresholds but can be overly optimistic in highly imbalanced scenarios. AUCPR  plots Precision against Recall, focusing specifically on the minority class (anomalies). Given the severe class imbalance in anomaly detection, AUCPR is often considered a more informative metric for practical deployment performance.

\section{Calibration Analysis}
\label{sec:appendix_calibration}

\begin{revise}

  Anomaly detection benchmarks typically evaluate performance via ranking metrics such as AUC-PR and AUC-ROC.
  These metrics measure the model's ability to separate anomalies from normal instances but do not assess probability reliability.
  Expected Calibration Error (ECE)~\cite{naeini2015obtaining,guo2017calibration} is a complementary metric that quantifies whether predicted probabilities match empirical anomaly frequencies.
  We report ECE to provide a reliability perspective beyond ranking quality, evaluating how well each model's output scores can serve as probability estimates.

  \paragraph{Model Selection.}
  We evaluate five models spanning three algorithm families:
  specialized WSAD methods (DevNet, XGBOD), general supervised learning (CatBoost),
  and tabular foundation models (TabPFN, LimiX).
  These models are competitive on the 47 tabular datasets and cover the main supervision paradigms in \workname{}.

  \paragraph{Calibration.}
  To obtain comparable probability estimates, we reserve a validation split for calibration.
  Specifically, we partition the training data into training and validation subsets, resulting in a 60\%/10\%/30\% train/validation/test split.
  Under extreme label scarcity, we ensure the validation subset contains at least one anomalous sample to guarantee calibrator convergence.
  Let \( f(x) \in \mathbb{R} \) be the mapping function outputting the raw anomaly score for sample \( x \).
  We fit a Platt scaling calibrator on the validation subset to map raw scores to the binary label space \( Y \in \{0, 1\} \).
  The calibrated probability is formulated as:
  \[
    \hat{p}(y=1 \mid x) = \sigma(A \cdot f(x) + B) = \frac{1}{1 + \exp(A \cdot f(x) + B)},
  \]
  where parameters \( A, B \) are optimized by minimizing the negative log-likelihood loss \( \mathcal{L} \) on the validation subset.
  Finally, ECE is computed on the calibrated test probabilities.
  We partition the predicted probability range \( [0, 1] \) into \( M = 10 \) equally spaced bins.
  The ECE aggregates the absolute difference between average confidence and empirical accuracy across all bins, weighted by sample frequency.

  \begin{table}[tbp]
  \centering
  \caption{Expected Calibration Error (ECE) across different supervision settings.}
  \vspace{-10pt}
  \label{tab:ece_results}
  \resizebox{\linewidth}{!}{
    \begin{tabular}{lcccccccc}
      \toprule
      Model & $\rla=1\%$ & $\rla=10\%$ & $\rla=50\%$ & $\rla=100\%$ & $\nla=1$ & $\nla=5$ & $\nla=10$ & $\nla=50$ \\
      \midrule
      DevNet & 0.0265(5) & 0.0248(1) & 0.0241(1) & 0.0235(1) & 0.0285(5) & 0.0246(5) & 0.0297(5) & 0.0251(1) \\
      CatBoost & 0.0014(1) & 0.0315(2) & 0.0599(3) & 0.0378(2) & 0.0011(2) & 0.0017(2) & 0.0044(1) & 0.0497(2) \\
      XGBOD & 0.0014(1) & 0.0329(3) & 0.0587(2) & 0.0381(3) & 0.0011(3) & 0.0015(1) & 0.0044(2) & 0.0530(3) \\
      TabPFN & 0.0023(3) & 0.0407(4) & 0.0645(4) & 0.0388(4) & 0.0021(4) & 0.0019(3) & 0.0057(3) & 0.0663(4) \\
      LimiX & 0.0026(4) & 0.0435(5) & 0.0670(5) & 0.0388(5) & 0.0011(1) & 0.0032(4) & 0.0069(4) & 0.0664(5) \\
      \bottomrule
    \end{tabular}
  }
\end{table}
  \begin{table}[t]
  \centering
  \caption{AUCPR performance of additional tabular foundation models across different supervision settings.}
  \vspace{-8pt}
  \label{tab:tfm_results}
  \resizebox{\linewidth}{!}{
    \begin{tabular}{lcccccccc}

      \toprule
      Model & $\rla=1\%$ & $\rla=10\%$ & $\rla=50\%$ & $\rla=100\%$ & $\nla=1$ & $\nla=5$ & $\nla=10$ & $\nla=50$ \\
      \midrule
      XGBOD & 0.481(8) & 0.688(7) & 0.816(7) & 0.864(6) & 0.385(8) & 0.580(8) & 0.657(7) & 0.791(7) \\
      DevNet & 0.555(3) & 0.666(8) & 0.713(8) & 0.722(8) & 0.517(1) & 0.610(7) & 0.648(8) & 0.697(8) \\
      CatBoost & 0.521(5) & 0.715(6) & 0.842(5) & 0.883(4) & 0.457(4) & 0.635(5) & 0.689(5) & 0.809(5) \\
      TabPFN & 0.528(4) & 0.770(2) & 0.864(2) & 0.900(2) & 0.415(6) & 0.667(3) & 0.735(2) & 0.842(2) \\
      TabICLv2 & 0.570(1) & 0.773(1) & 0.872(1) & 0.902(1) & 0.482(3) & 0.710(1) & 0.761(1) & 0.859(1) \\
      Orion-MSP & 0.562(2) & 0.745(3) & 0.844(4) & 0.872(5) & 0.492(2) & 0.685(2) & 0.733(3) & 0.835(3) \\
      TabDPT & 0.505(6) & 0.743(4) & 0.850(3) & 0.885(3) & 0.397(7) & 0.649(4) & 0.709(4) & 0.823(4) \\
      Mitra & 0.485(7) & 0.719(5) & 0.831(6) & 0.863(7) & 0.429(5) & 0.623(6) & 0.688(6) & 0.800(6) \\
      \bottomrule

    \end{tabular}
  }
\end{table}

  \paragraph{Evaluation and Analysis.}
  We evaluate ECE across both RLA and NLA settings, covering
  \(\rla \in \{1\%, 10\%, 50\%, 100\%\}\) and \(\nla \in \{1, 5, 10, 50\}\),
  on all 47 tabular datasets.
  The computed ECE values are averaged across all 47 tabular datasets, with results reported in Table~\ref{tab:ece_results}.
  Overall, the ECE results complement the ranking-based evaluation.
  The best-calibrated method varies across supervision regimes.
  CatBoost and XGBOD perform well under extreme label scarcity, whereas DevNet improves when more labeled anomalies are available.
  TabPFN and LimiX do not show a consistent ECE advantage in this experiment.

\end{revise}

\section{Further Experimental Results on Tabular Foundation Models}
\label{sec:appendix_tfm}

% To address the concern regarding the representativeness of tabular foundation models, we revise our terminology to ``Tabular Foundation Models (TFMs)''. This choice aligns with our benchmark's core objective: to evaluate an algorithm's pure learning and reasoning capability by decoupling feature extraction from anomaly scoring. In our framework, data from all modalities are first transformed into high-dimensional embeddings using frozen, pre-trained extractors. Within this unified feature space, TFMs serve as the direct modern counterparts to traditional WSAD methods. They excel at in-context learning from structured vectors to overcome limited supervision. By contrast, models such as Outformer \cite{ding2026from} target zero-shot unsupervised outlier detection without labeled anomalies, while Vision Foundation Models (VFMs) or Large Language Models (LLMs) tightly couple representation learning with reasoning. Both therefore fall outside the scope of the WSAD setting studied here.
\begin{revise}
  To solidify our findings on the dominance of tabular foundation models, we expanded our evaluation to include four additional state-of-the-art tabular foundation models representing diverse design strategies: TabICLv2~\cite{qu2026tabiclv2} and Orion-MSP~\cite{bouadi2025orion} extend the in-context learning (ICL) paradigm with improved attention mechanisms for better scalability; TabDPT~\cite{ma2025tabdpt} incorporates self-supervised learning on real data to scale with larger corpora; and Mitra~\cite{zhang2025mitra} focuses on synthetic prior design for improved generalization. Together, these models cover the major development directions in current tabular foundation model research.

  As shown in Table~\ref{tab:tfm_results}, the newly evaluated tabular foundation models generally outperform the WSAD baselines (DevNet and XGBOD) and supervised baseline (CatBoost) across supervision levels, reinforcing our conclusion that modern tabular foundation models hold a strong advantage in WSAD tasks. Among these models, TabICLv2 achieves the best overall performance under ratio-based supervision (\(\rla\)), while Orion-MSP demonstrates a relative advantage under the extremely scarce one-shot setting (\(\nla=1\)). The performance gap between tabular foundation models and non-foundation baselines widens as supervision increases, suggesting that tabular foundation models benefit more from additional labeled anomalies due to stronger in-context learning capacity.
\end{revise} 

\section{Open-Source Repository}
\label{appendix:github}

To facilitate reproducibility and future research, we release the full implementation of the \workname{} benchmark at:

\begin{center}
  \textbf{\url{https://github.com/SUFE-AILAB/WSADBench}}
\end{center}

\ifarxiv
We encourage the research community to use and extend \workname{} to explore new directions in WSAD.
\else
In addition to the repository, the full version at \arxivurl{} provides detailed experimental results, algorithm lists, and dataset meta-feature correlation analysis.
\fi

\clearpage

\ifarxiv

\section{Extended Results for WSAD Performance}

\subsection{Incomplete Supervision}
\label{appendix:incomplete_detailed}

\begin{figure*}[t]
  \centering

  \begin{subfigure}{0.32\textwidth}
    \centering
    \includegraphics[width=\linewidth]{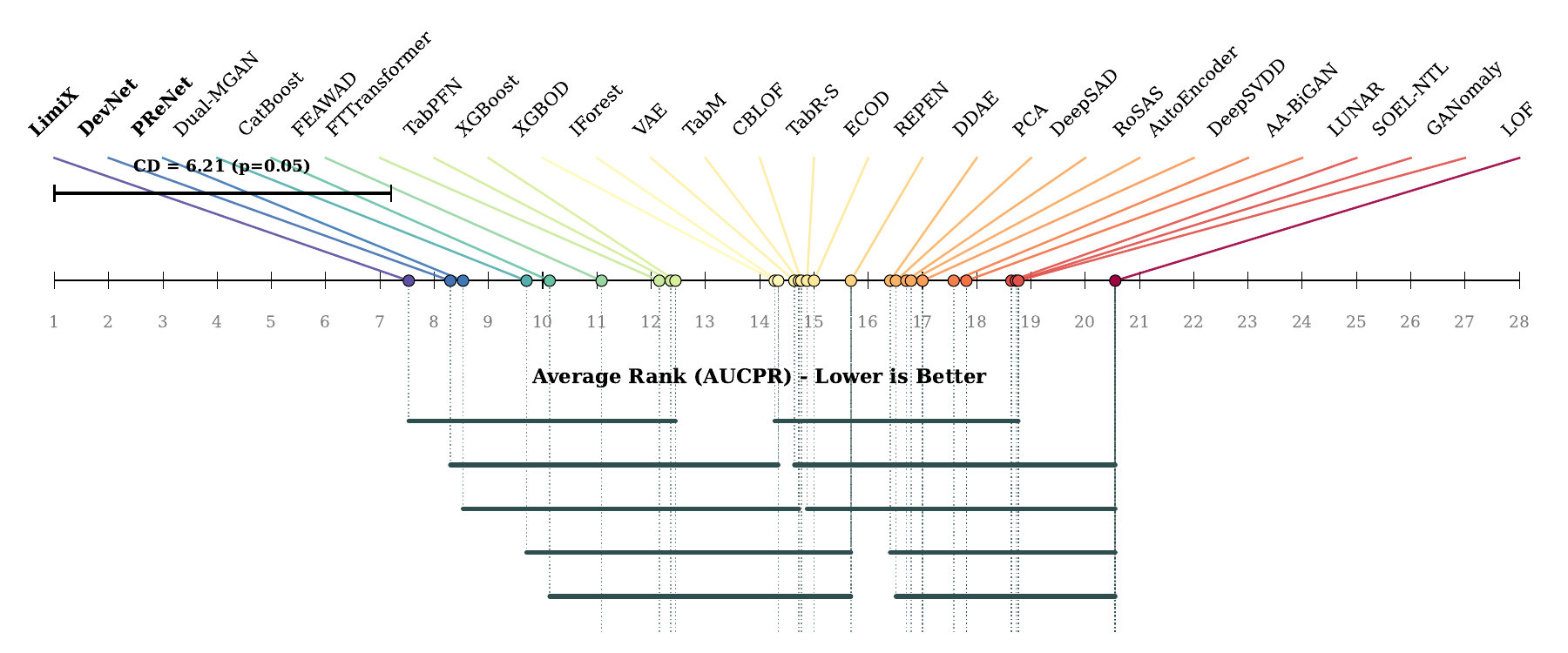}
    \caption{Tabular Datasets ($N_{la}=1$)}
    \label{fig:cd_classical}
  \end{subfigure}
  \hfill
  \begin{subfigure}{0.32\textwidth}
    \centering
    \includegraphics[width=\linewidth]{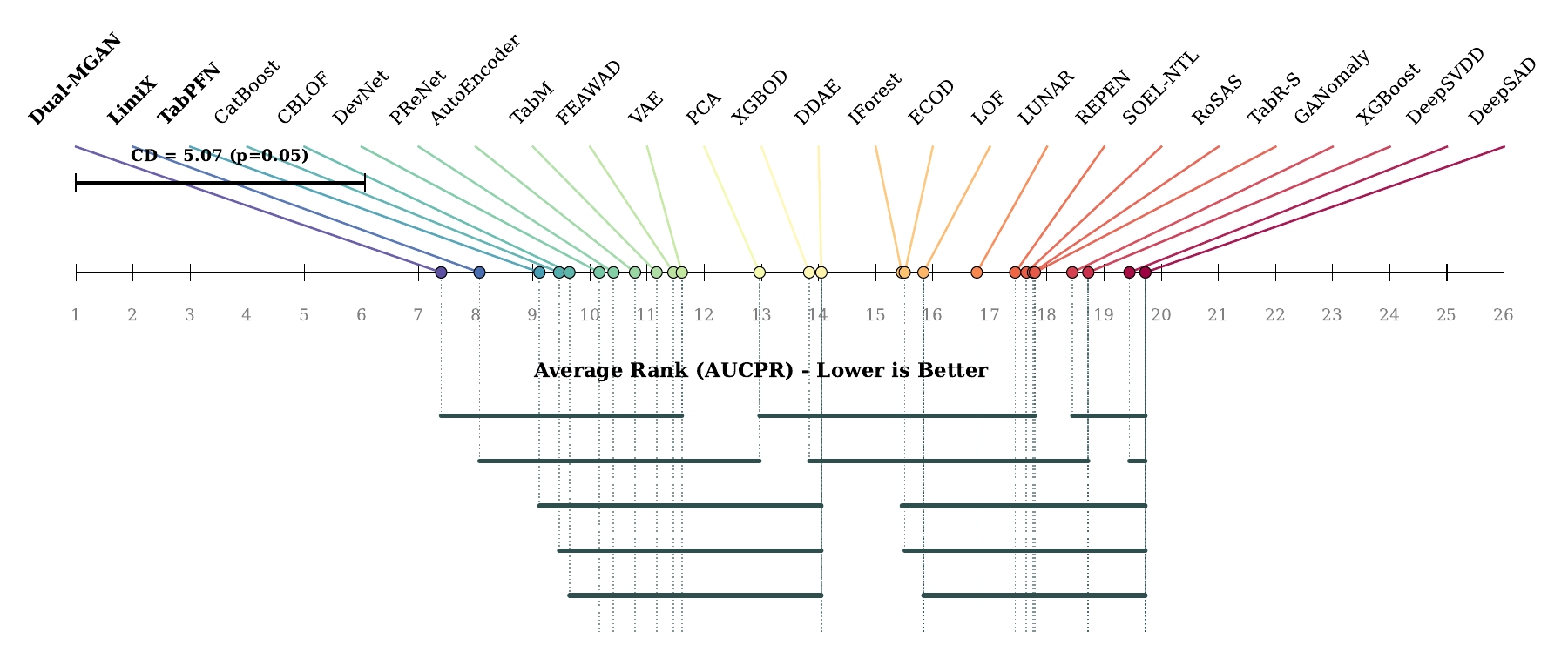}
    \caption{Image Datasets ($\rla=1\%$)}
    \label{fig:cd_cv}
  \end{subfigure}
  \hfill
  \begin{subfigure}{0.32\textwidth}
    \centering
    \includegraphics[width=\linewidth]{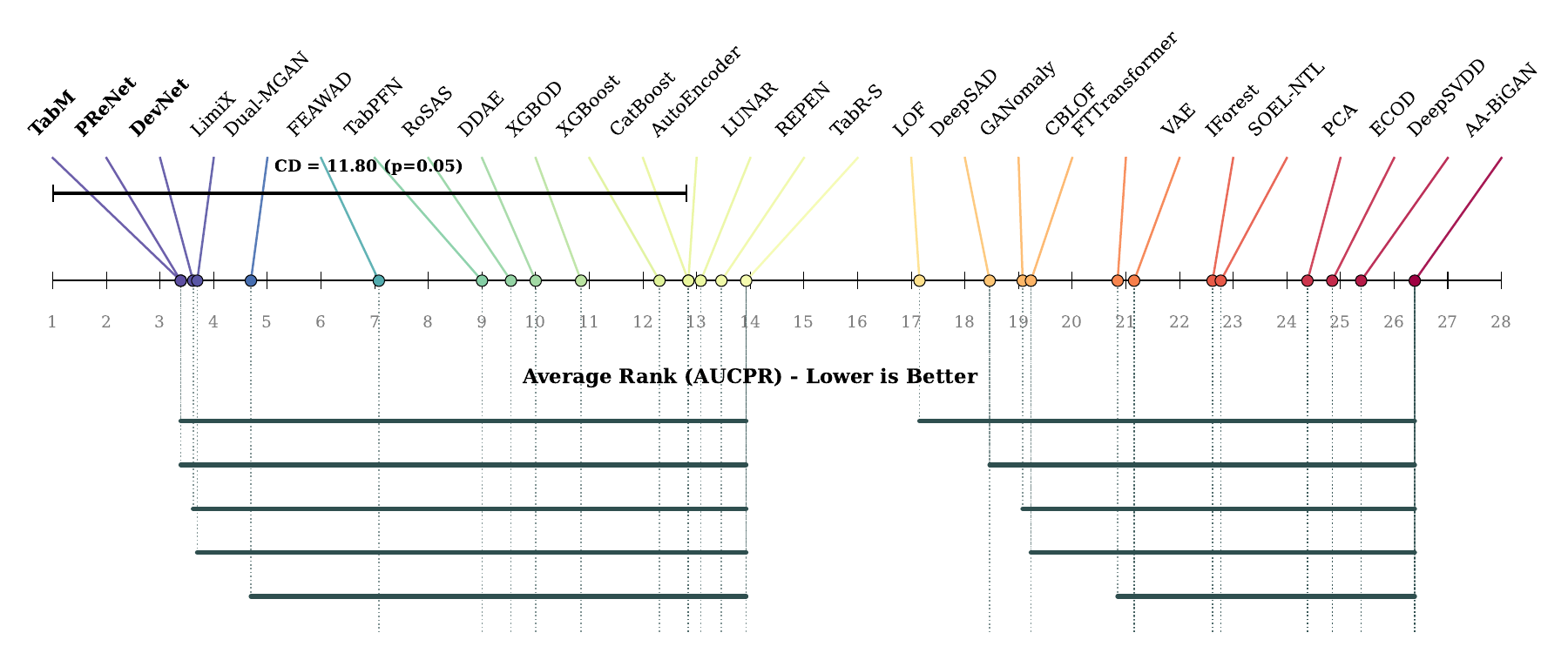}
    \caption{Text Datasets ($\rla=1\%$)}
    \label{fig:cd_cv_v2}
  \end{subfigure}

  \begin{subfigure}{0.32\textwidth}
    \centering
    \includegraphics[width=\linewidth]{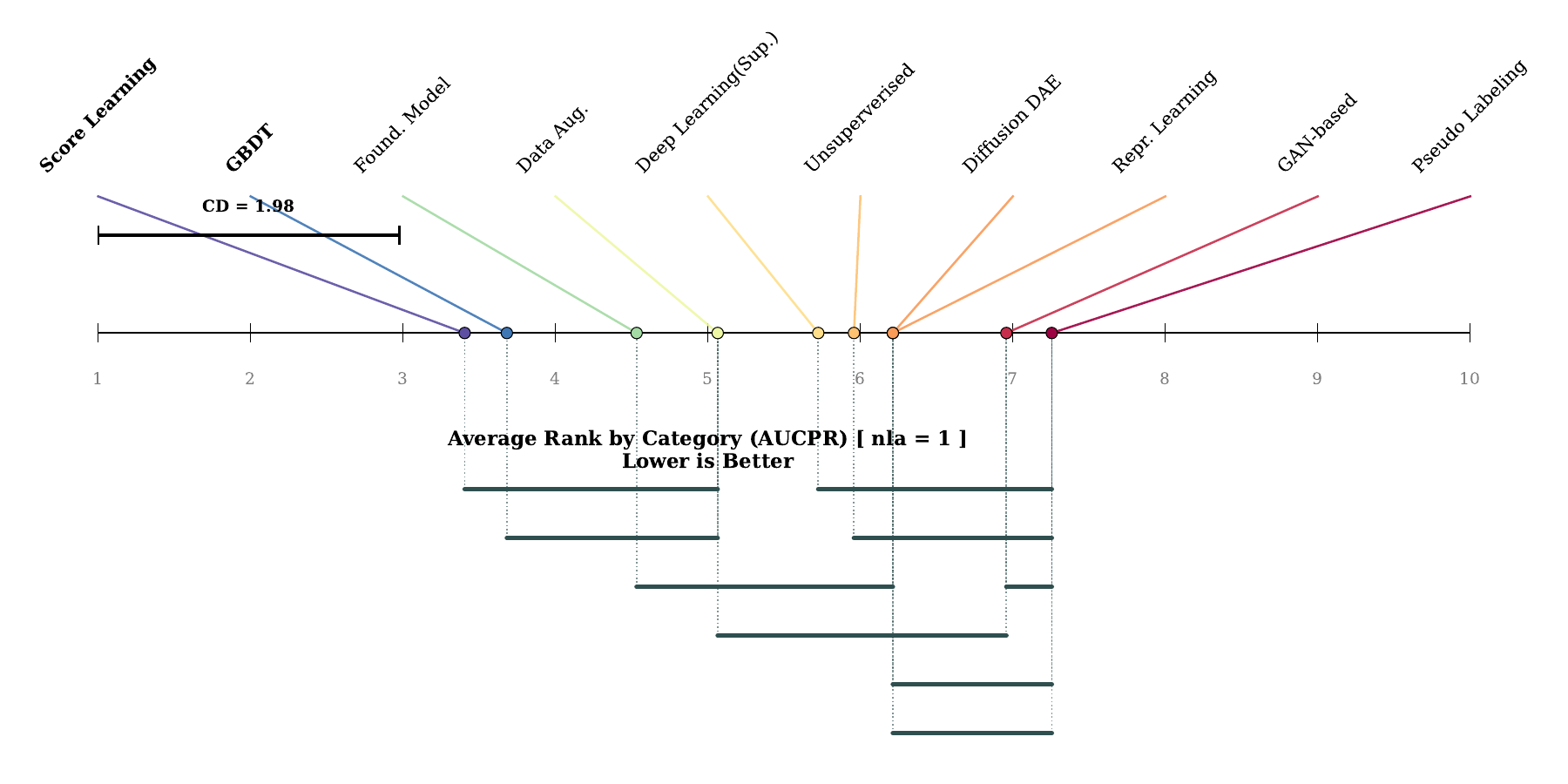}
    \caption{Tabular Datasets ($N_{la}=1$)}
    \label{fig:type_cd_tabular}
  \end{subfigure}
  \hfill
  \begin{subfigure}{0.32\textwidth}
    \centering
    \includegraphics[width=\linewidth]{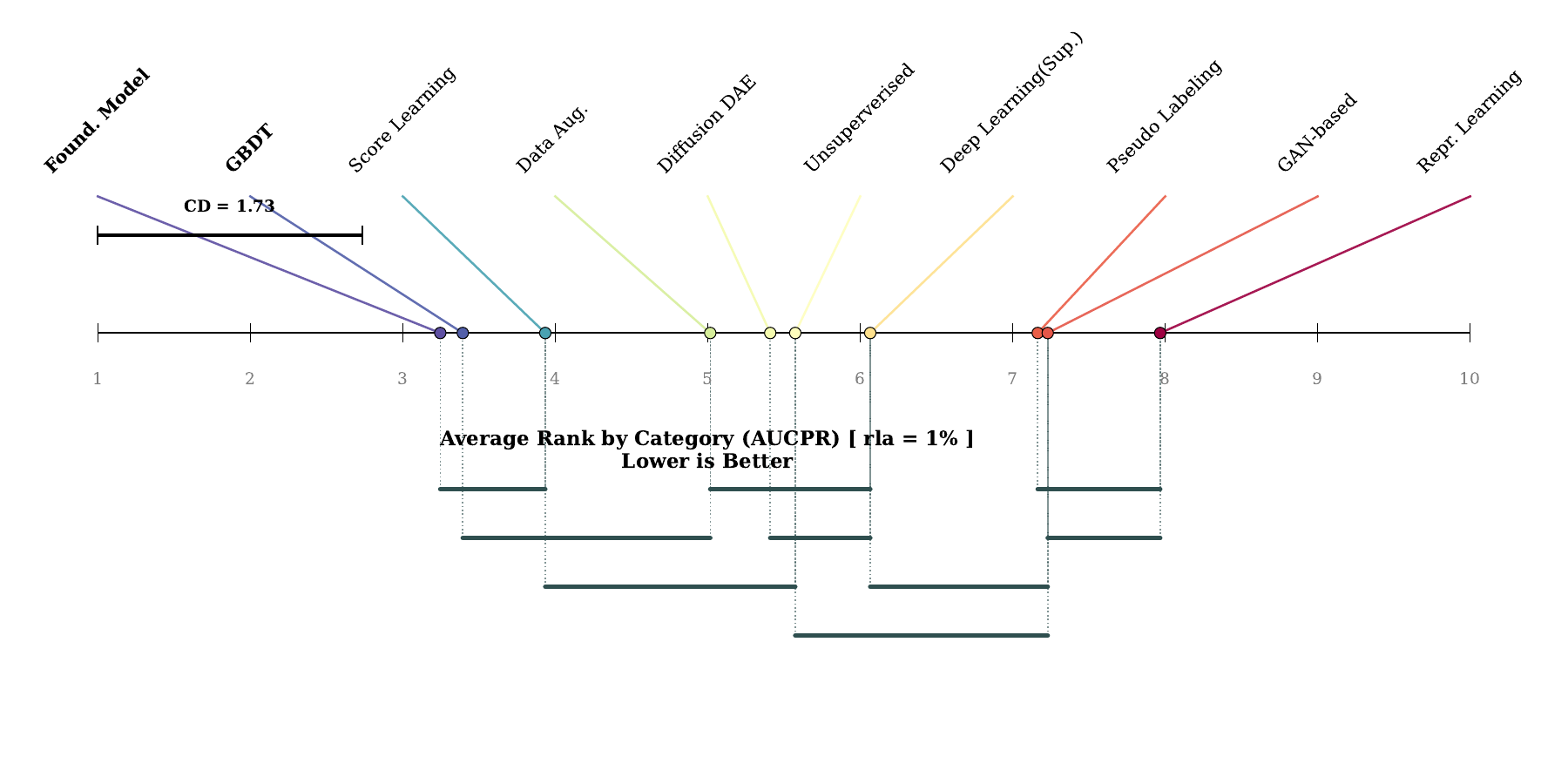}
    \caption{Image Datasets ($\rla=1\%$)}
    \label{fig:type_cd_image}
  \end{subfigure}
  \hfill
  \begin{subfigure}{0.32\textwidth}
    \centering
    \includegraphics[width=\linewidth]{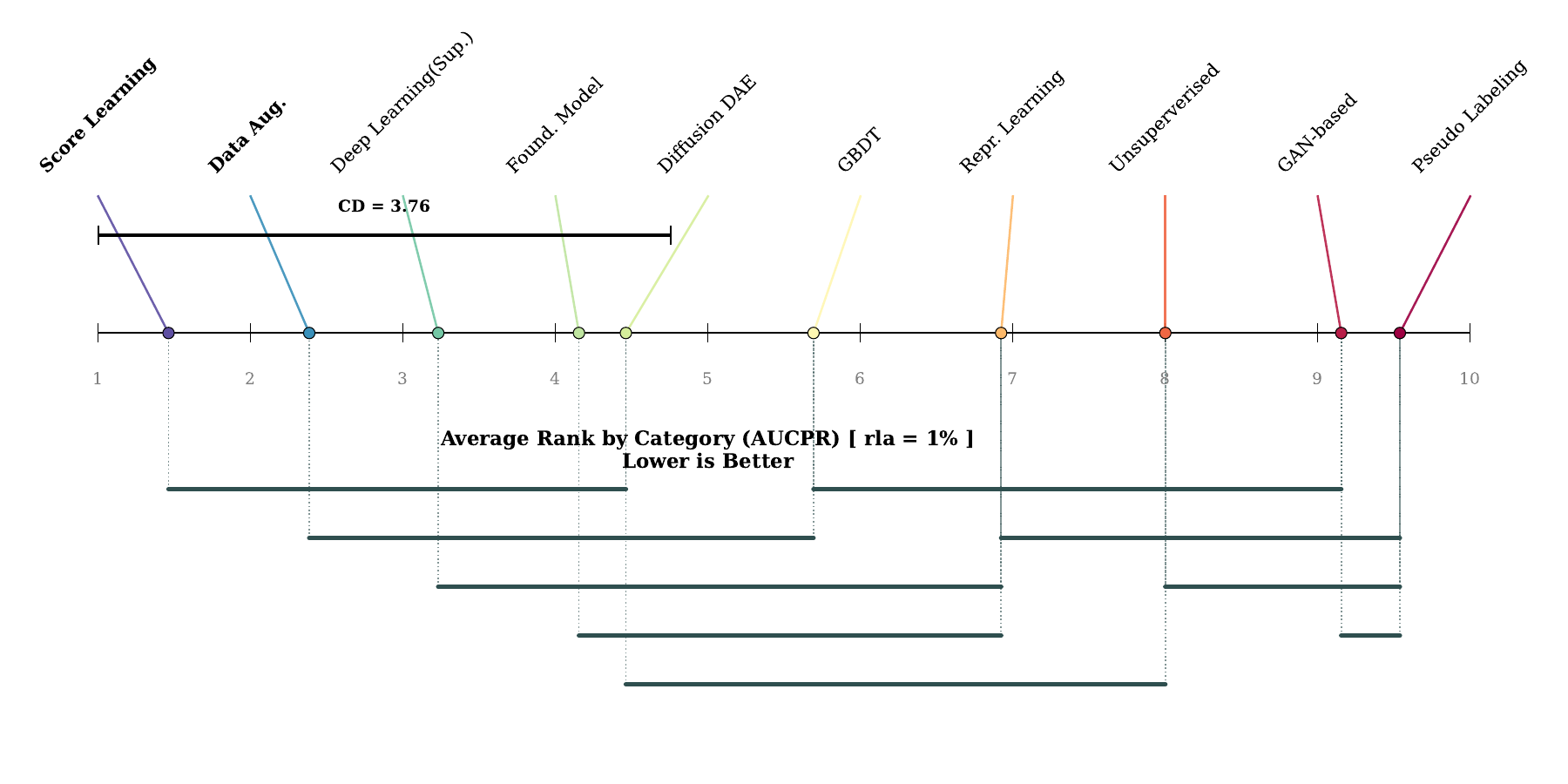}
    \caption{Text Datasets ($\rla=1\%$)}
    \label{fig:type_cd_text}
  \end{subfigure}

  \vspace{-5pt}

  % \caption{Critical Difference (CD) diagrams comparing performance across different modalities under extremely limited supervision. The top row (a-c) compares the performance of \textbf{individual models}, while the bottom row (d-f) compares different \textbf{types of algorithms} to further strengthen the generalizability of the conclusions. The average rank is calculated based on AUCPR (lower rank indicates better performance). Horizontal bars connect groups of models or algorithm types that are not statistically significantly different according to the Nemenyi post-hoc test at significance level $\alpha=0.05$.}
  \caption{Critical Difference (CD) diagrams under extremely limited supervision for each modality. The top row (a-c) compares \textbf{individual models}, whereas the bottom row (d-f) compares algorithm \textbf{categories}. Average ranks are based on AUCPR (lower rank is better). Horizontal bars indicate groups without statistically significant differences (Nemenyi test, $\alpha=0.05$).}
  \label{fig:cd_combined}
\end{figure*}

This section presents the detailed experimental results corresponding to the incomplete supervision analysis in the main paper.
% The comprehensive performance summaries are provided in Tables~\ref{tab:supp_incomp_aucpr_rla_cls}, \ref{tab:supp_incomp_aucpr_nla_cls}, and~\ref{tab:supp_incomp_aucroc_nla_cls} (Tabular Benchmark), Tables~\ref{tab:supp_incomp_aucpr_rla_cv}, \ref{tab:supp_incomp_aucpr_nla_cv}, \ref{tab:supp_incomp_aucroc_rla_cv}, and~\ref{tab:supp_incomp_aucroc_nla_cv} (CV Datasets), and Tables~\ref{tab:supp_incomp_aucpr_rla_nlp}, \ref{tab:supp_incomp_aucpr_nla_nlp}, \ref{tab:supp_incomp_aucroc_rla_nlp}, and~\ref{tab:supp_incomp_aucroc_nla_nlp} (NLP Datasets). Additionally, we provide complete benchmarking results on multimodality datasets in Tables~\ref{tab:supp_incomp_alldata_rla_1}, \ref{tab:rank_aucpr_1}, \ref{tab:supp_incomp_alldata_rla_5}, \ref{tab:rank_aucpr_0.1}, \ref{tab:rank_aucpr_0.25}, \ref{tab:rank_aucpr_0.5}, and~\ref{tab:supp_incomp_alldata_rla_100}  (AUCPR), as well as Tables~\ref{tab:rank_aucroc_0.01}, \ref{tab:rank_aucroc_1}, \ref{tab:rank_aucroc_0.05}, \ref{tab:rank_aucroc_0.1}, \ref{tab:rank_aucroc_0.25}, \ref{tab:rank_aucroc_0.5}, and~\ref{tab:rank_aucroc_1.0} (AUCROC). These tables offer detailed rankings, average scores, and standard deviations for each algorithm under varying supervision ratios ($\rla$) and counts ($\nla$), serving as an extended reference for the benchmark evaluation.
\revised{Tables~\ref{tab:supp_incomp_aucpr_rla_cls}, \ref{tab:supp_incomp_aucpr_nla_cls}, \ref{tab:classical_aucroc_refined}, and~\ref{tab:supp_incomp_aucroc_nla_cls} present the Tabular Benchmark results.
  Tables~\ref{tab:supp_incomp_aucpr_rla_cv}, \ref{tab:supp_incomp_aucpr_nla_cv}, \ref{tab:supp_incomp_aucroc_rla_cv}, and~\ref{tab:supp_incomp_aucroc_nla_cv} summarize the CV Datasets.
  Similarly, Tables~\ref{tab:supp_incomp_aucpr_rla_nlp}, \ref{tab:supp_incomp_aucpr_nla_nlp}, \ref{tab:supp_incomp_aucroc_rla_nlp}, and~\ref{tab:supp_incomp_aucroc_nla_nlp} provide the NLP Datasets performance.
  For multimodality datasets, Tables~\ref{tab:supp_incomp_alldata_rla_1}--\ref{tab:rank_aucroc_1.0} exhibit the Ratio-labeled Anomaly (\(\rla\)) results.
  Tables~\ref{tab:rank_aucpr_1}--\ref{tab:rank_aucroc_10} provide the corresponding Number-labeled Anomaly (\(\nla\)) performance.
  These tables offer detailed rankings, average scores, and standard deviations for each algorithm.
They serve as an extended reference for the benchmark evaluation.}

To rigorously compare the performance differences among algorithms under extremely limited supervision settings, we employ the Nemenyi post-hoc test to generate Critical Difference (CD) diagrams, as shown in Figure~\ref{fig:cd_combined}.
Figure~\ref{fig:cd_classical} visualizes the statistical ranking on the Tabular Benchmark with only one labeled anomaly ($N_{la}=1$).
Similarly, Figure~\ref{fig:cd_cv} and Figure~\ref{fig:cd_cv_v2} illustrate the model comparisons on image and text datasets under the 1\% labeled anomaly ratio ($\gamma_{la}=1\%$), respectively. These diagrams provide a clear perspective on the relative effectiveness of different methods in data-scarce scenarios.

\begin{revise}
  Furthermore, Figure~\ref{fig:cd_combined} bottom row aggregates performance by algorithm categories to align with Section~\ref{sec:wsad_instance}.
  Across all modalities under extreme scarcity (\( \nla = 1 \) or \( \rla = 1\% \)), tabular foundation models exhibit no statistically significant superiority.
  Specifically, they share critical difference bars with other top-performing categories, including score learning, GBDTs, and data augmentation.
  These statistical ties indicate that no single algorithm category, including tabular foundation models, achieves absolute dominance under such scarce supervision.
\end{revise}

To further demonstrate the effectiveness of using limited labeled anomalies, we compare the weakly supervised performance against the best-performing unsupervised baselines (PCA for Tabular, CBLOF for CV, and AutoEncoder for NLP).
Table~\ref{tab:supp_incomp_baseline_cls}, Table~\ref{tab:supp_incomp_baseline_cv}, and Table~\ref{tab:supp_incomp_baseline_nlp} report the AUCPR performance under varying $\gamma_{la}$ and $n_{la}$ settings compared to these unsupervised benchmarks. These comparisons highlight the performance gains achieved by introducing varying degrees of incomplete supervision.

% We provide the comprehensive performance data for incomplete supervision scenarios, spanning both Ratio-labeled Anomaly (RLA, $\rla$) and Number-labeled Anomaly (NLA, $\nla$) settings across all modalities.  % \label{tab:classical_aucpr_checked} \label{tab:rank_aucpr_0.01}
% Tables~\ref{tab:incomplete_overall} to~\ref{tab:incom_meta_analysis} detail the mean AUCPR and AUCROC values for each model under varying label densities, Complementing the summary in Section~\ref{sec:wsad_instance}.

% specialized deviation-based methods (e.g., DevNet) excel in the extreme few-shot regime (e.g., $\nla=1$), whereas general-purpose supervised baselines (e.g., CatBoost) rapidly regain dominance as the number of labeled anomalies increases towards $\nla=10$ or $\rla=1\%$.

\subsection{Inexact Supervision}

\label{appendix:inexact_video_results}

% vad_pre_train

% This section presents the complete performance comparison across different pretrain models for video anomaly detection. Table~\ref{tab:inexact_vad_auc} and Table~\ref{tab:inexact_vad_pr} show the detailed AUCROC and AUCPR results for each combination of pretrain model (i3d, x3d, sf50, mvit, sf) and detection algorithm.

This section provides three supplementary analyses to extend the main text (Section~\ref{sec:vad_simple_comparison}), detailing experimental configurations, comprehensive performance across different feature backbones, and additional findings on model sensitivity.

Specifically, Table~\ref{tab:inexcat_vad_tab_pretrain} includes the comprehensive performance of general-purpose tabular classifiers applied to video anomaly detection tasks, providing the detailed data supporting the interoperability discussion in Section~\ref{sec:inexact_interop}.

\subsubsection{Benchmark Results on Video AD}
\begin{table}[h]
  \centering
  \caption{Specifications and average performance of pre-trained video backbones. Inputs denote (Frames \(\times\) Sample Rate), and metrics are averaged over downstream algorithms.}
  \vspace{-5pt}
  \label{tab:backbone_configs_and_performance}
  \resizebox{\columnwidth}{!}{%
    \begin{tabular}{lcccccc}
      \toprule
      \textbf{Backbone} & \textbf{Abbr.} & \textbf{Input (\(T \times \tau\))} & \textbf{Params (M)} & \textbf{Feat. Dim.} & \textbf{AUCROC} & \textbf{AUCPR} \\
      \midrule
      I3D\_R50        & i3d  & 8\(\times\)8   & 28.04 & 2048 & 0.8199 (5)          & 0.3528 (5) \\
      X3D\_M          & x3d  & 16\(\times\)5  & 3.79  & 2048 & 0.8278 (3)          & 0.3654 (4) \\
      MViT\_B         & mvit & 32\(\times\)3  & 36.61 & 768  & 0.8274 (4)          & 0.3778 (2) \\
      SlowFast\_R50   & sf50 & 8\(\times\)8   & 34.57 & 2304 & 0.8393 (2)          & 0.3772 (3) \\
      SlowFast\_R101  & sf   & 8\(\times\)8   & 62.83 & 2304 & \textbf{0.8489 (1)} & \textbf{0.3875 (1)} \\
      \bottomrule
    \end{tabular}%
  }
\end{table}
\begin{table}[htbp]
  \centering
  \caption{Performance comparison between 32-seg and 200-seg settings (AUCPR).}
  \vspace{-5pt}
  \label{tab:seg_comparison}
  \resizebox{\columnwidth}{!}{
    \begin{tabular}{l c c c}
      \toprule
      \textbf{Model} & \textbf{32-seg} & \textbf{200-seg} & \textbf{Improvement (\%)} \\
      \midrule
      AR-Net      & 0.430 {\tiny $\pm$ 0.189} (1) & 0.429 {\tiny $\pm$ 0.188} (1) & -0.13 \\
      Sultani    & 0.392 {\tiny $\pm$ 0.153} (2)          & 0.377 {\tiny $\pm$ 0.149} (3)          & -3.89 \\
      MGFN       & 0.258 {\tiny $\pm$ 0.124} (7)          & 0.297 {\tiny $\pm$ 0.129} (7)          & 15.03 \\
      RTFM       & 0.386 {\tiny $\pm$ 0.162} (3)          & 0.380 {\tiny $\pm$ 0.162} (2)          & -1.56 \\
      UR-DMU      & 0.340 {\tiny $\pm$ 0.169} (5)          & 0.353 {\tiny $\pm$ 0.169} (4)          & 3.56 \\
      VadCLIP    & 0.344 {\tiny $\pm$ 0.140} (4)          & 0.348 {\tiny $\pm$ 0.142} (5)          & 1.20 \\
      GCN-Anomaly & 0.304 {\tiny $\pm$ 0.189} (6)          & 0.303 {\tiny $\pm$ 0.203} (6)          & -0.31 \\
      \bottomrule
    \end{tabular}
  }
\end{table}

% \textbf{Experimental Settings.}
Following standard MIL-based VAD protocols (e.g., Sultani), we extract clip-level features (16-frame non-overlapping clips) using pre-trained backbones. To handle variable video lengths, we employ temporal pooling to standardize each video into a fixed sequence of $n$ segments (default $n=32$, with $n=200$ for granularity analysis).
% During training, we apply minority class oversampling to balance normal and anomalous videos, and train using a Multiple Instance Learning (MIL) strategy with random batch sampling.
Evaluation is performed on the test set using the final epoch model, with frame-level scores aligned to calculate AUCROC and AUCPR. The performance is summarized in Table~\ref{tab:inexact_vad_auc} and Table~\ref{tab:inexact_vad_pr}.

\paragraph{Selection of Pre-trained Models Significantly Impacts Average VAD Performance.}
Our experiments reveal distinct performance disparities driven solely by the choice of feature extractor. As summarized in Table~\ref{tab:backbone_configs_and_performance}, the top-performing backbone, SlowFast-R101, achieves an average AUC-ROC of 0.8489 across all downstream algorithms, whereas the older I3D-R50 backbone reaches only 0.8199. This substantial gap highlights that upgrading the upstream feature representation is as critical as designing sophisticated anomaly detection heads.

% The SlowFast architecture's ability to simultaneously capture spatial context (Slow path) and fine-grained motion (Fast path) aligns well with VAD requirements, surpassing single-stream counterparts.

\paragraph{Feature Quality Weighs More Than Parameter Count.}
Counter-intuitively, larger models do not guarantee better features for VAD. As detailed in Table~\ref{tab:backbone_configs_and_performance}, X3D-M, despite having only 3.79M parameters, outperforms the much larger I3D-R50 (28.04M). Similarly, MViT-B achieves competitive results with a compact 768-dimensional feature space, outperforming 2048-dimensional features from ResNet-based models. This suggests that the semantic discriminability of features is more critical than raw model capacity or embedding dimensionality.

\paragraph{Temporal Granularity Differentially Impacts Model Architectures.}
Table~\ref{tab:seg_comparison} reveals that finer temporal segmentation ($n=200$ vs $n=32$) benefits complex models while hindering simpler ones. Attention-based models like MGFN gain significantly (15\% AUCPR increase) from higher resolution, leveraging their "focus" modules to pinpoint anomalies. Conversely, simpler MIL models like Sultani suffer a performance drop (3.89\% AUCPR decrease), likely because finer granularity introduces noise that overwhelms their coarse ranking constraints. Models with robust feature aggregation, such as AR-Net and RTFM, remain stable across granularities.

\subsubsection{Relative Contribution of Backbone and Model}
\label{appendix:pretrained_model}

\begin{table}[tbp]
  \centering
  \caption{Multi-factor ANOVA results quantifying the contribution of factors to AUCROC and AUCPR variance.}
  \vspace{-5pt}
  \label{tab:anova_combined}
  \resizebox{\linewidth}{!}{%
    \begin{tabular}{lcccccc}
      \toprule
      \multirow{2}{*}{\textbf{Factor (Interaction)}} & \multicolumn{3}{c}{\textbf{AUCROC}} & \multicolumn{3}{c}{\textbf{AUCPR}} \\
      \cmidrule(lr){2-4} \cmidrule(lr){5-7}
      & \textbf{Sum Sq.} & \textbf{$P$-Value} & \textbf{Con.(\%)} & \textbf{Sum Sq.} & \textbf{$P$-Value} & \textbf{Con.(\%)} \\
      \midrule
      Dataset & 0.7251 & $3.00 \times 10^{-104}$ & \textbf{31.82} & 16.5964 & $< 10^{-300}$ & \textbf{82.24} \\
      Dataset $\times$ Classifier & 0.3141 & $2.08 \times 10^{-44}$ & \textbf{13.78} & 1.0222 & $1.44 \times 10^{-92}$ & \textbf{5.07} \\
      Classifier & 0.2572 & $6.47 \times 10^{-44}$ & \textbf{11.29} & 1.2465 & $8.26 \times 10^{-117}$ & \textbf{6.18} \\
      Dataset $\times$ Pretraining & 0.2037 & $1.19 \times 10^{-31}$ & 8.94 & 0.1916 & $7.23 \times 10^{-21}$ & 0.95 \\
      Pretraining & 0.0774 & $5.02 \times 10^{-15}$ & 3.40 & 0.1042 & $1.88 \times 10^{-14}$ & 0.52 \\
      Classifier $\times$ Pretraining & 0.0699 & $5.37 \times 10^{-6}$ & 3.07 & 0.1266 & $6.88 \times 10^{-9}$ & 0.63 \\
      Seed & 0.0036 & 0.467 & 0.16 & 0.0130 & 0.056 & 0.06 \\
      \textit{Residual} & 0.6279 & - & 27.55 & 0.8800 & - & 4.36 \\
      \bottomrule
    \end{tabular}%
  }
\end{table}

\begin{table}[tbp]
  \centering
  \caption{Comparison of Weakly Supervised Video Anomaly Detection at clip and frame levels.}
  \vspace{-5pt}
  \label{tab:gt_performance}
  \resizebox{\linewidth}{!}{
    \begin{tabular}{lcccc}
      \toprule
      \textbf{Model} & \textbf{AUC (Clip)} & \textbf{AUC (Frame)} & \textbf{PR (Clip)} & \textbf{PR (Frame)} \\
      \midrule
      AR-Net      & 0.783$\pm$0.008 (2) & \textbf{0.806$\pm$0.001 (1)} & 0.196$\pm$0.006 (3) & 0.207$\pm$0.000 (3) \\
      MGFN       & \textbf{0.798$\pm$0.006 (1)} & 0.795$\pm$0.006 (3) & 0.172$\pm$0.004 (6) & 0.171$\pm$0.003 (6) \\
      RTFM       & 0.770$\pm$0.015 (4) & 0.779$\pm$0.014 (5) & \textbf{0.203$\pm$0.021 (1)} & 0.209$\pm$0.024 (2) \\
      Sultani    & 0.778$\pm$0.002 (3) & \textbf{0.806$\pm$0.001 (1)} & 0.199$\pm$0.003 (2) & \textbf{0.216$\pm$0.002 (1)} \\
      UR-DMU      & 0.769$\pm$0.005 (5) & 0.760$\pm$0.014 (6) & 0.178$\pm$0.011 (4) & 0.185$\pm$0.017 (5) \\
      VadCLIP    & 0.753$\pm$0.011 (7) & 0.755$\pm$0.010 (7) & 0.159$\pm$0.013 (7) & 0.162$\pm$0.013 (7) \\
      GCN-Anomaly & 0.767$\pm$0.006 (6) & 0.793$\pm$0.002 (4) & 0.178$\pm$0.006 (5) & 0.188$\pm$0.003 (4) \\
      \bottomrule
    \end{tabular}
  }
\end{table}

% \begin{table*}[h!]
% \centering
% \caption{Ground Truth Alignment Methods in Different WSAD Approaches}
% \label{tab:gt}
% \begin{tabular}{lcccccccc}
% \hline
% GT Alignment & WSAD & MGFN & Sultani & VAD-CLIP & ZhongGCN & ARNet & RTFM & URDMU \\
% \hline
% Aligned with frames      & \checkmark &          & \checkmark &          & \checkmark &          &          &         \\
% Aligned with test clips  &            & \checkmark &          & \checkmark &          &  & \checkmark &  \checkmark         \\
% Unclear &          &          &          &          &          &    \checkmark     &          &  \\
% \hline
% \end{tabular}
% \end{table*}

\begin{table}[tbp]
  \centering
  \caption{Ground truth alignment methods in different WSAD approaches}
  \vspace{-5pt}
  \label{tab:gt_transposed}
  \resizebox{1.0\columnwidth}{!}{
    \begin{tabular}{lccc}
      \toprule
      \textbf{Method} & \textbf{Aligned with frames} & \textbf{Aligned with test clips} & \textbf{Unclear} \\
      \midrule
      MGFN      &            & \cmark &            \\
      Sultani   & \cmark &            &            \\
      VadCLIP  &            & \cmark &            \\
      GCN-Anomaly  & \cmark &            &            \\
      AR-Net     &            &            & \cmark \\
      RTFM      &            & \cmark &            \\
      UR-DMU     &            & \cmark &            \\
      \midrule
      \textbf{WSADBench(Ours)}      & \cmark &            &            \\
      \bottomrule
    \end{tabular}
  }
\end{table}
% cls

We examine the relative contributions of feature representation (pre-trained backbone) and inductive bias (classifier design) to VAD performance, as well as their potential synergistic effects. By analyzing these factors, we aim to uncover whether performance gains are driven primarily by better features or superior detection algorithms.
To decouple the impact of feature representation and inductive bias, we conduct an N-way Analysis of Variance (ANOVA) across 4 datasets, 5 backbones, and 7 algorithms.
We model performance ($y$) as a linear combination of main effects (Dataset $D$, Pre-training $P$, Classifier $C$, Seed $S$) and their interactions:
\begin{equation}
  y = \mu + \alpha_{D} + \beta_{P} + \gamma_{C} + \delta_{S} + (\alpha\beta)_{DP} + (\alpha\gamma)_{DC} + (\beta\gamma)_{PC} + \epsilon
\end{equation}
Contribution rates, derived from the Sum of Squares, quantify the variance explained by each factor.
Statistical significance is assessed at $P < 0.05$.
Detailed results for AUCROC and AUCPR are provided in Table~\ref{tab:anova_combined}.

The ANOVA results reveal two pivotal findings.
First, across both metrics the classifier consistently explains more variance than the backbone (AUCROC: 11.29\% vs.\ 3.40\%; AUCPR: 6.18\% vs.\ 0.52\%), confirming that inductive bias outweighs feature representation.
Second, the Dataset x Classifier interaction under AUCROC (13.78\%) exceeds the Classifier main effect (11.29\%), indicating that the optimal algorithm varies across datasets---no single classifier universally dominates.

% First, Inductive Bias Dominates Feature Representation: The choice of classifier contributes significantly more to performance variance (6.18\%) than the pre-trained backbone (0.52\%), highlighting that the downstream algorithm's capacity to model the decision boundary is the primary driver of detection success.
% Second, Synergy is Secondary: Although the interaction between backbone and classifier is statistically significant ($P < 10^{-20}$), its contribution (0.63\%) is minimal compared to the main effects. This implies that the relationship is largely additive: superior classifiers consistently perform well across diverse backbones without requiring highly specific pairing optimization.

\subsubsection{Ground Truth Alignment}
\label{sec:ground_truth_alignment}

During baseline reproduction, we observed inconsistencies in how different methods align predicted anomaly scores with ground truth (GT) labels. Table \ref{tab:gt_transposed} summarizes these divergent protocols, which we hypothesize introduce systematic bias and potentially alter model rankings. We compare two prevalent alignment strategies:

\noindent \textbf{Protocol 1: Frame-canonical Alignment.} This approach prioritizes the physical duration of the raw video. Spatially, scores from multiple crops are averaged to a single value per clip. Temporally, clip-level scores are interpolated to restore frame-level resolution and then truncated to match the exact length of the original video ($L_{raw}$). This method strictly evaluates valid frames, ignoring padding artifacts introduced by feature extraction.

\noindent \textbf{Protocol 2: Feature-grid Alignment.} This approach operates on the feature grid, preserving all crops and padding. Spatially, crops are treated as independent instances (flattened). Temporally, the ground truth is padded with "normal" labels to match the feature grid size ($L_{grid} > L_{raw}$) and tiled to align with each crop. This artificially inflates the evaluation set with non-existent padding frames and treats spatially disjoint crops as independent samples.

We empirically quantify the impact of these strategies using the UCF-Crime dataset with I3D features, standardized to 32 temporal segments.

\noindent \textbf{Analysis.}
Quantitative results (Table~\ref{tab:gt_performance}) demonstrate that alignment strategies significantly impact model rankings.
Protocol 1 (Frame-canonical) yields higher performance for spatially sensitive models like AR-Net and Sultani (0.806 AUC), whereas Protocol 2 (Feature-grid) results in a performance drop (e.g., Sultani falls to 0.778), shifting the lead to MGFN (0.798 AUC).
This shift can be explained by a structural conflict we term the "Spatial Sparsity Penalty."
Protocol 2 mechanically broadcasts the frame-level anomaly label to all spatial crops.
Consequently, in scenarios where the anomaly is localized (e.g., a specific object) against a normal background, Protocol 2 treats the entire grid as anomalous.
Models like Sultani, designed to suppress background scores via sparsity constraints, are thus penalized for correctly predicting low scores on these background regions (False Negatives under Protocol 2).
In contrast, Protocol 1 averages spatial scores to evaluate the frame as a single unit. This allows localized anomalies to sufficiently elevate the frame score, avoiding penalties for validly suppressed background regions.
Therefore, we adopt Protocol 1 as the standard for VAD benchmarking to ensure fairness for spatially discriminative methods.

% \textbf{Key Insights.}

% \begin{itemize}
%   \item \textbf{Alignment Rationality}:  Our analysis of ground truth alignment strategies confirms that strict frame-level correspondence is essential for a rational and precise evaluation, avoiding the ambiguities of looser video-level protocols.

% \end{itemize}

% \par

% \subsection{Inaccurate Supervision}
% \label{appendix:inaccurate_results}

% % This section presents the complete sensitivity analysis for label noise experiments, covering both Flip-Normal and Flip-Abnormal scenarios.
% This section presents the detailed experimental results corresponding to the inaccurate supervision (label noise) analysis in the main paper. The comprehensive summaries of the sensitivity tests are provided in Table~\ref{tab:supp_inacc_flip_normal_nlp} and Table~\ref{tab:supp_inacc_flip_abnormal_nlp} (NLP Datasets), and Table~\ref{tab:supp_inacc_flip_normal_cls} and Table~\ref{tab:supp_inacc_flip_abnormal_cls} (Classical Datasets). These tables offer detailed AUCPR sensitivity metrics for different models under varying flipping ratios of normal and abnormal labels, serving as an extended reference for assessing algorithmic robustness against label noise.

\subsection{OOD Generalization}
\label{sec:appendix_ood_extend}

\begin{figure}[h]
  \centering
  \includegraphics[width=\linewidth]{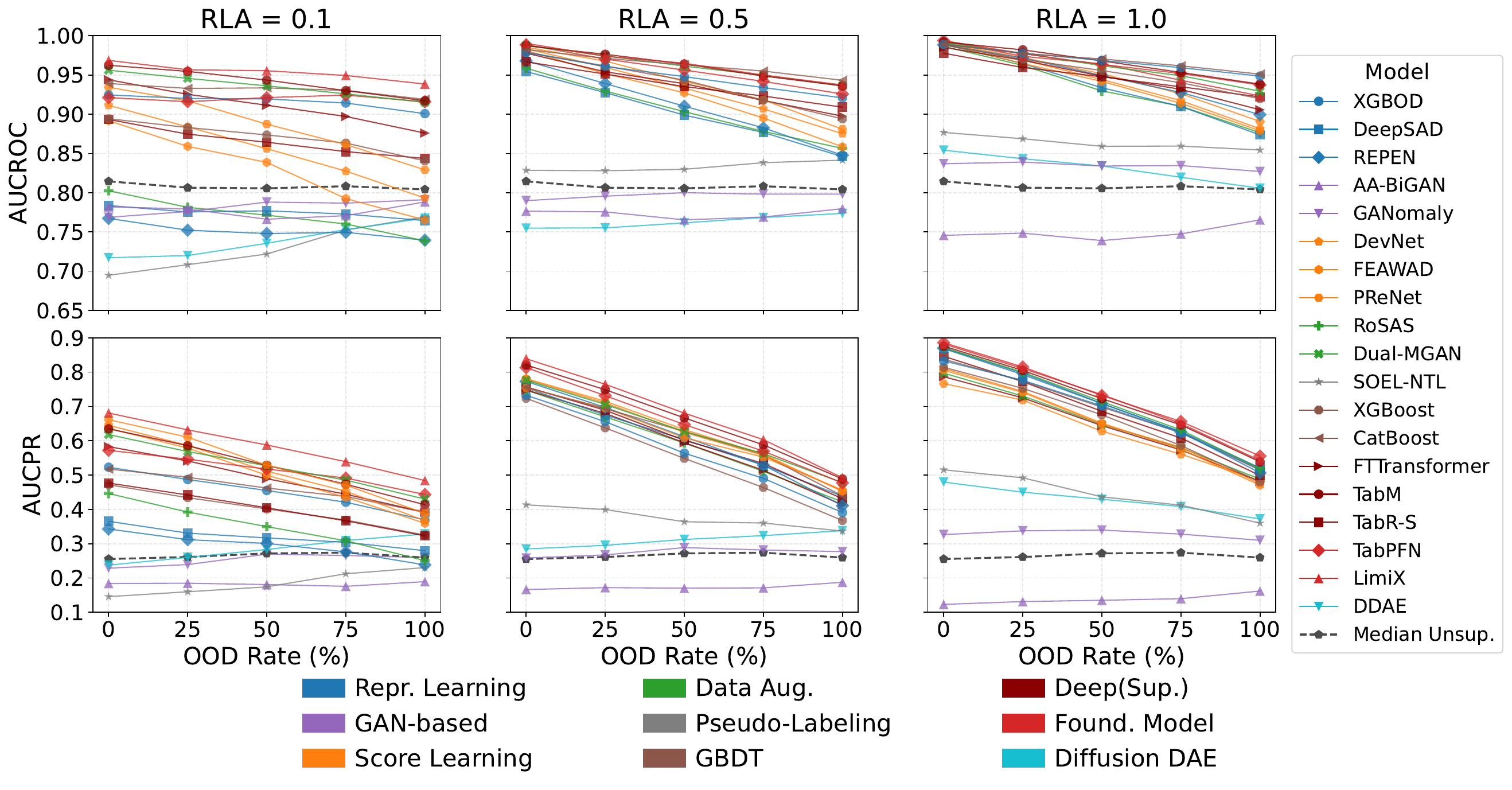}

  % \end{subfigure}

  \caption{Visualization of incomplete OOD results. Row 1 shows the AUCROC performance, and Row 2 shows the AUCPR performance under OOD conditions.}
  \label{fig:ood_general}
\end{figure}

\begin{figure*}[t]
  \centering
  \includegraphics[width=\linewidth]{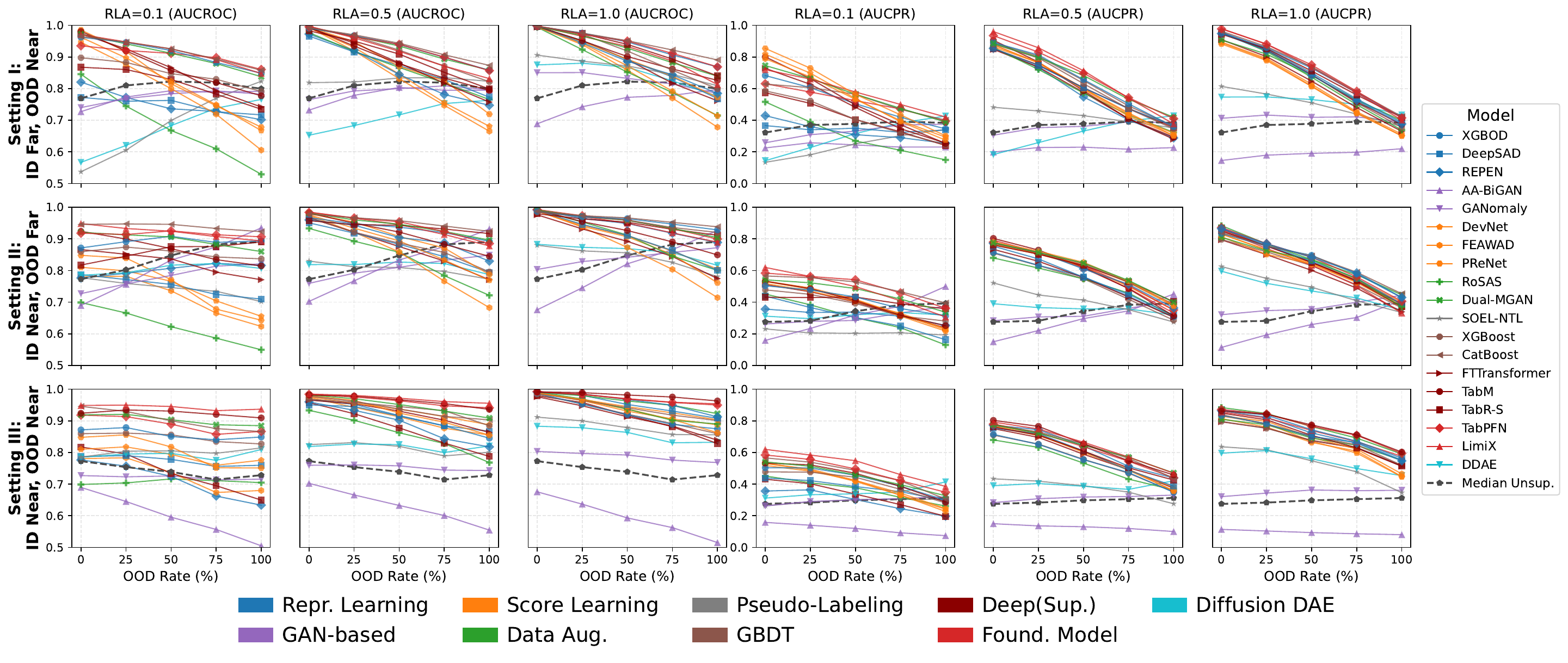}
  \caption{Visualization of AUCROC and AUCPR distributions under three different incomplete OOD settings.}
  \label{fig:ood_dist_combined}
\end{figure*}

% Following \cite{ding2022catching}, the OOD protocol evaluates generalization by splitting anomalies based on distinct classes: one class is designated as Known ($\mathcal{A}_{known}$) and all others as Unknown ($\mathcal{A}_{unknown}$). To ensure rigorous evaluation, we select datasets containing at least two anomaly types and minimum 15 samples, employing a 7:3 stratified train-test split. Training is restricted to $\mathcal{A}_{known}$ under limited supervision, where only a fraction ($\rla$) of samples is labeled, and the rest are treated as unlabeled background to simulate realistic label noise. The test set evaluates robustness by mixing ID ($\mathcal{A}_{known}$) and OOD ($\mathcal{A}_{unknown}$) anomalies, with the proportion of unknown classes controlled by the OOD rate.\par

Unlike the controlled OOD sampling based on feature distance presented in the main text, this section evaluates generalization under a more generalized setting without explicit distance constraints. We follow the protocol proposed by Ding et al.~\cite{ding2022catching}, where anomalies are split solely based on distinct semantic classes: one class is designated as Known ($\mathcal{A}_{known}$) and all others as Unknown ($\mathcal{A}_{unknown}$). To ensure rigorous evaluation, we select datasets containing at least two anomaly types and a minimum of 15 samples, employing a 7:3 stratified train-test split. Training is restricted to $\mathcal{A}_{known}$ under limited supervision, where only a fraction ($\rla$) of samples is labeled, and the rest are treated as unlabeled background to simulate realistic label noise. The test set evaluates robustness by mixing ID ($\mathcal{A}_{known}$) and OOD ($\mathcal{A}_{unknown}$) anomalies, with the proportion of unknown classes explicitly controlled by the OOD rate.\par
Figure \ref{fig:ood_general} presents the complete performance landscape across OOD datasets. These detailed visualizations serve as supplementary empirical evidence reinforcing the robustness analysis discussed in Section~\ref{sec:ood_main_text}, demonstrating that the observed trends are consistent across varying evaluation metrics (AUCROC and AUCPR).

\subsection{Tabular MIL Bag Construction Sensitivity}

\begin{revise}
  To verify that benchmark conclusions are not sensitive to bag construction choices, we conduct a sensitivity analysis on tabular Multiple Instance Learning (MIL).
  We vary two key hyperparameters independently: bag size and abnormal-bag ratio.
  For the bag size analysis, we set the abnormal-bag probability to \(0.3\) and label ratio to \(100\%\), while varying the bag size in \(\{10, 20, 30\}\).
  For the abnormal-bag ratio analysis, we fix the bag size to \(10\) and label ratio to \(100\%\), varying the abnormal-bag probability in \(\{0.1, 0.2, 0.3\}\).
  We evaluate five representative models: DeepSAD~\cite{ruff2019deep}, REPEN~\cite{pang2018learning}, DevNet~\cite{pang2019deepdevnet}, TabPFN~\cite{grinsztajn2025tabpfn}, and Sultani~\cite{sultani2018real}.

  \begin{table}[ht]
  \centering
  \caption{%
    AUCPR sensitivity of models to Tabular MIL bag construction hyperparameters.
    Parentheses denote rank within each column.
  }
  \vspace{-10pt}
  \label{tab:mil_sensitivity}
  \resizebox{\linewidth}{!}{%
    \begin{tabular}{l ccc ccc}
      \toprule
      & \multicolumn{3}{c}{$\text{prob}=0.3$, $\rla=100\%$}
      & \multicolumn{3}{c}{$\text{bag}=10$, $\rla=100\%$} \\
      \cmidrule(lr){2-4} \cmidrule(lr){5-7}
      Model & bag=10 & bag=20 & bag=30 & prob=0.1 & prob=0.2 & prob=0.3 \\
      \midrule
      DeepSAD & \textbf{0.302(1)} & \textbf{0.328(1)} & \textbf{0.322(1)} & \textbf{0.286(1)} & \textbf{0.285(1)} & \textbf{0.302(1)} \\
      REPEN   & 0.276(3)          & 0.303(2)          & 0.309(2)          & 0.255(3)          & 0.258(3)          & 0.276(3)          \\
      DevNet  & 0.265(5)          & 0.302(3)          & 0.309(2)          & 0.239(5)          & 0.245(5)          & 0.265(5)          \\
      TabPFN  & 0.284(2)          & 0.286(5)          & 0.274(5)          & 0.277(2)          & 0.266(2)          & 0.284(2)          \\
      Sultani & 0.272(4)          & 0.296(4)          & 0.303(4)          & 0.251(4)          & 0.252(4)          & 0.272(4)          \\
      \bottomrule
    \end{tabular}%
  }
\end{table}

  Table~\ref{tab:mil_sensitivity} reports the average AUCPR values and corresponding model ranks under these configurations.
  The experimental results demonstrate that relative model rankings remain highly consistent across the evaluated settings.
  For instance, DeepSAD consistently dominates the comparison by achieving the top rank in all test cases.
  Similarly, Sultani consistently occupies the fourth rank under every hyperparameter setting.
  Although minor rank shifts exist between REPEN and DevNet under larger bag sizes, their overall performance differences remain small.
  These findings verify that the relative performance of anomaly detection methods is robust to the specific choices of MIL dataset construction.
\end{revise}

\section{Algorithms List}
\label{appendix:algorithms}

\begin{table*}[ht]
  \footnotesize
  \centering
  \caption{\textbf{Overview of evaluated algorithms}}
  \vspace{-8pt}
  \label{tab:ad_methods}
  \resizebox{0.95\textwidth}{!}{%
    \begin{tabular}{llccccc}
      \toprule
      \textbf{Supervision Type} & \textbf{Method} & \textbf{Category} & \textbf{Venue} & \textbf{Backbone} & \textbf{Modality} & \textbf{Code} \\
      \midrule

      % Weakly-supervised (Instance) Group
      \multirow{11}{*}{{Weakly-supervised (Instance)}}
      & XGBOD & Repr. Learning & IJCNN'18 & Tree-based & Tabular & \href{https://github.com/yzhao062/XGBOD}{Link} \\
      & DeepSAD & Score Learning & ICLR'20 & MLP & Tabular & \href{https://github.com/lukasruff/Deep-SAD-PyTorch}{Link} \\
      & REPEN & Repr. Learning & KDD'18 & MLP & Tabular & \href{https://github.com/GuansongPang/deep-outlier-detection}{Link} \\
      & AA-BiGAN & GAN-based & IJCAI'22 & GAN & Tabular & \href{https://github.com/tbw162/AA-BiGAN}{Link} \\
      & Dual-MGAN & Data Aug. & TKDD'22 & GAN & Tabular & \href{https://github.com/leibinghe/Dual-MGAN}{Link} \\
      & DevNet & Score Learning & KDD'19 & MLP & Tabular & \href{https://github.com/GuansongPang/deviation-network}{Link} \\
      & FEAWAD & Reconstruction & TNNLS'21 & AE & Tabular & \href{https://github.com/yj-zhou/Feature_Encoding_with_AutoEncoders_for_Weakly-supervised_Anomaly_Detection}{Link} \\
      & PReNet & Score Learning & KDD'23 & MLP & Tabular & \href{https://github.com/mala-lab/PReNet}{Link} \\
      & RoSAS & Data Aug. & IP\&M'23 & MLP & Tabular & \href{https://github.com/xuhongzuo/rosas}{Link} \\
      & SOEL-NTL & Pseudo-Labeling & ICML'23 & MLP & Tabular & \href{https://github.com/aodongli/Active-SOEL}{Link} \\
      & DDAE & Diffusion DAE & KDD'25 & Diffusion AE & Tabular & \href{https://github.com/sattarov/AnoDDAE}{Link} \\
      & GANomaly & GAN-based & ACCV'18 & GAN & Tabular & \href{https://github.com/samet-akcay/ganomaly}{Link} \\
      \midrule

      % Weakly-supervised (Bag) Group
      \multirow{7}{*}{Weakly-supervised (Bag)}
      & Sultani & Vanilla MIL & CVPR'18 & MLP & Video & \href{https://github.com/seominseok0429/Real-world-Anomaly-Detection-in-Surveillance-Videos-pytorch}{Link} \\
      & VadCLIP & Lang-Guided MIL & AAAI'24 & Attention & Video & \href{https://github.com/nwpu-zxr/VadCLIP}{Link} \\
      & MGFN & Magnitude MIL & AAAI'23 & CNN+Attn & Video & \href{https://github.com/carolchenyx/MGFN.}{Link} \\
      & RTFM & Magnitude MIL & ICCV'21 & CNN+Attn & Video & \href{https://github.com/tianyu0207/RTFM}{Link} \\
      & AR-Net & Dynamic MIL & ICME'20 & MLP & Video & \href{https://github.com/wanboyang/Anomaly_AR_Net_ICME_2020}{Link} \\
      & UR-DMU & Uncertainty MIL & AAAI'23 & Attention & Video & \href{https://github.com/henrryzh1/UR-DMU}{Link} \\
      & GCN-Anomaly & Label Denoising & CVPR'19 & GCN & Video & \href{https://github.com/jx-zhong-for-academicpurpose/GCN-Anomaly-Detection}{Link} \\
      & PUMA & PU MIL & KDD'23 & AE & Tabular & \href{https://github.com/Lorenzo-Perini/PU-MIL-AD}{Link} \\
      \midrule

      % Unsupervised (Instance) Group
      \multirow{9}{*}{Unsupervised (Instance)}
      & IForest & Isolation-based & ICDM'08 & Tree-based & Tabular & \href{https://github.com/yzhao062/pyod}{Link} \\
      & AutoEncoder & Reconstruction & ICLR'18 & MLP & Tabular & \href{https://github.com/yzhao062/pyod}{Link} \\
      & DeepSVDD & Deep One-class & ICML'18 & MLP & Tabular & \href{https://github.com/lukasruff/Deep-SAD-PyTorch}{Link} \\
      & VAE & Reconstruction & ICLR'14 & MLP & Tabular & \href{https://github.com/yzhao062/pyod}{Link} \\
      & PCA & Reconstruction & ICDM'03 workshop & Linear & Tabular & \href{https://github.com/yzhao062/pyod}{Link} \\
      & ECOD & Probabilistic & TKDE'22 & Probabilistic & Tabular & \href{https://github.com/yzhao062/pyod}{Link} \\
      & CBLOF & Cluster-based & PRL'03 & Cluster-based & Tabular & \href{https://github.com/yzhao062/pyod}{Link} \\
      & LOF & Density-based & SIGMOD'00 & Density-based & Tabular & \href{https://github.com/yzhao062/pyod}{Link} \\
      & LUNAR & GNN-based & AAAI'22 & GNN & Tabular & \href{https://github.com/agoodge/LUNAR}{Link} \\
      \midrule

      % Supervised (Instance) Group
      \multirow{5}{*}{Supervised (Instance)}
      & XGBoost & GBDT & KDD'16 & GBDT & Tabular & \href{https://github.com/dmlc/xgboost}{Link} \\
      & CatBoost & GBDT & NeurIPS'18 & GBDT & Tabular & \href{https://github.com/catboost/catboost}{Link} \\
      & FTTransformer & Deep (Sup.) & NeurIPS'21 & Transformer & Tabular & \href{https://github.com/Yura52/rtdl}{Link} \\
      & TabM & Deep (Sup.) & ICLR'25 & MLP & Tabular & \href{https://github.com/yandex-research/TabM}{Link} \\
      & TabR-S & Deep (Sup.) & ICLR'24 & MLP & Tabular & \href{https://github.com/yandex-research/tabular-dl-tabr}{Link} \\
      \midrule

      % Foundation Models Group
      \multirow{2}{*}{Foundation Models (Instance)}
      & TabPFN & Found. Model & ICLR'23 & Transformer & Tabular & \href{https://github.com/PriorLabs/TabPFN}{Link} \\
      & LimiX & Found. Model & ArXiv'25 & Transformer & Tabular & \href{https://github.com/limix-ldm/LimiX}{Link} \\

      \bottomrule
    \end{tabular}%
  }
\end{table*}

We provide a comprehensive overview of the \nalgorithms{} algorithms evaluated in \workname{}, spanning diverse supervision paradigms including weakly-supervised (instance/bag), unsupervised, supervised, and tabular foundation models. Detailed attributes for each method, such as publication venue, backbone architecture, and code availability, are summarized in Table~\ref{tab:ad_methods}.

\subsection{Weakly-supervised (Instance)}
% REPEN DevNet DeepSAD FEAWAD PReNet RoSAS XGBOD AABiGAN DualMGAN GANomaly AnoDDAE
\begin{itemize}
  \item \textbf{REPEN} \cite{pang2018learning}: A representation learning method that leverages limited labeled anomalies to learn low-dimensional embeddings where anomalies are significantly distant from normal data, optimized via a triplet ranking loss.
    % End-to-end score learning）。
  \item \textbf{DevNet} \cite{pang2019deepdevnet}: An end-to-end framework that learns anomaly scores by enforcing a Gaussian prior on normal data and maximizing the Z-score deviation of anomalies from the normal distribution.
  \item \textbf{DeepSAD} \cite{ruff2019deep}: A deep semi-supervised extension of DeepSVDD that minimizes the distance of normal samples to the hypersphere center while penalizing the inverse distance of labeled anomalies to push them away.
    % (Feature Encoding with AutoEncoder
  \item \textbf{FEAWAD} \cite{zhou2021feature}: A reconstructive framework that utilizes an autoencoder to map instances into a latent space where anomalies are poorly reconstructed, while simultaneously training a classifier to separate encoded features of labeled anomalies from normal data.
  \item \textbf{PReNet} \cite{pang2023deep}: A pairwise relation prediction network that learns anomaly scores by contrasting ordinal pairs of labeled anomalies and unlabeled samples, enforcing a large margin between known anomalies and the unlabeled distribution.
  \item \textbf{RoSAS} \cite{xu2023rosas}: A robust semi-supervised framework that employs a purity-aware label smoothing strategy to mitigate noise in unlabeled data, combined with a deviation loss to learn discriminative anomaly scores.

    % Feature Construction + Ensemble
  \item \textbf{XGBOD} \cite{zhao2018xgbod}: An extreme gradient boosting-based outlier detection framework that enhances unsupervised outlier scores with new features generated from various extraction methods, optimized under a semi-supervised setting.
    % Bi-directional GAN + Class Imbalance
  \item \textbf{AA-BiGAN} \cite{tian2022anomaly}: An augmented adversarial bi-directional GAN that addresses class imbalance by augmenting labeled anomalies and enforces cycle-consistency to map normal data to a latent distribution distinct from anomalies.
    % Dual Memory + Mode Collapse
  \item \textbf{Dual-MGAN} \cite{li2022dual}: A dual-memory GAN framework that employs separate memory modules for normal and abnormal patterns to mitigate mode collapse, learning discriminative features through adversarial training.

    % PU Learning + Multi-Instance
  \item \textbf{PUMA} \cite{perini2023learning}: A framework that formulates anomaly detection as a Positive-Unlabeled (PU) learning problem under multi-instance constraints, estimating the class prior to robustly learn from bags with unknown instance labels.

    % Encoder-Decoder-Encoder + Reconstruction Error
  \item \textbf{GANomaly} \cite{akcay2018ganomaly}: A semi-supervised adversarial framework that employs an encoder-decoder-encoder architecture to minimize the reconstruction error of normal samples in both image and latent spaces.
    % Diffusion DAE + Noise as Score
  \item \textbf{DDAE} \cite{sattarov2025diffusion}: A denoising diffusion autoencoder that utilizes diffusion noise to corrupt input data and trains a denoiser to reconstruct the original signal, using the reconstruction error as the anomaly score.

\end{itemize}

\subsection{Unsupervised (Instance)} % DeepSVDD", "CBLOF", "LOF", "AutoEncoder", "VAE", "ECOD", "LUNAR", "IForest", "PCA"

\begin{itemize}
    % One-class Classification + Hypersphere
  \item \textbf{DeepSVDD} \cite{ruff2018deep}: A deep one-class classification method that maps normal data into a minimum-volume hypersphere in feature space.
    % Cluster-based + Size/Distance
  \item \textbf{CBLOF} \cite{he2003discovering}: A clustering-based method that assigns anomaly scores based on the size of the cluster a point belongs to and its distance to the cluster center.
    % Density-based + Local Neighborhood
  \item \textbf{LOF} \cite{breunig2000lof}: A density-based algorithm that measures the local density deviation of a given data point with respect to its neighbors.
    % Reconstruction + Bottleneck
  \item \textbf{AutoEncoder} \cite{zong2018deep}: A deep neural network trained to reconstruct input data, identifying anomalies through high reconstruction errors.
    % Probabilistic + Reconstruction
  \item \textbf{VAE} \cite{kingma2013auto}: A probabilistic generative model that learns the underlying data distribution, detecting anomalies by their low likelihood or high reconstruction loss.
    % Empirical Distribution + Tail Probability
  \item \textbf{ECOD} \cite{li2022ecod}: An unsupervised method that estimates the underlying distribution of data using empirical cumulative distribution functions.
    % Graph Neural Network + Nearest Neighbors
  \item \textbf{LUNAR} \cite{goodge2022lunar}: A graph neural network-based approach that learns to classify nodes as normal or anomalous based on their local neighborhood structure.
    % Tree-based + Path Length
  \item \textbf{IForest} \cite{liu2008isolation}: An ensemble method that isolates anomalies using random trees, exploiting the property that anomalies are susceptible to isolation (shorter path lengths).
    % Linear Projection + Variance
  \item \textbf{PCA} \cite{shyu2003novel}: A linear dimensionality reduction technique that detects anomalies by projecting data onto principal components and measuring reconstruction error.
\end{itemize}

\begin{table*}[ht]
  \footnotesize
  \centering
  \caption{Comprehensive description and interpretation of dataset meta-features.}
  \vspace{-8pt}
  \label{tab:meta_feature_appendix}
  \resizebox{0.95\textwidth}{!}{%
    \begin{tabular}{llcp{4.9cm}p{7.3cm}}
      \toprule
      \textbf{Category} & \textbf{Meta-Feature} & \textbf{Sym.} & \textbf{Definition} & \textbf{Interpretation} \\
      \midrule

      % --- Category 1: Basic Statistics ---
      \multirow{3}{*}{\textbf{Basic Stats}}
      & Sample Size & $N$ & Total number of instances. & Larger $N$ improves model stability but increases training cost. \\
      & Dimensionality & $D$ & Number of features. & High $D$ with low $N$ increases risk of overfitting (curse of dimensionality). \\
      & Sparsity Ratio & $\rho_{sp}$ & Fraction of zero elements. & High values indicate sparse data (e.g., text), suitable for specialized sparse solvers. \\
      \midrule

      % --- Category 2: Dimensionality & Correlation ---
      \multirow{5}{*}{\shortstack[l]{\textbf{Intrinsic}\\ \textbf{Dim. \&}\\ \textbf{Corr.}}}
      & Avg. Feat. Sim. & $\bar{\rho}_{f}$ & Mean pairwise correlation. & High values imply high redundancy and potential for feature selection. \\
      & Effective Rank & $ER$ & Number of significant principal components. & Indicates the true linear dimensionality of the data. \\
      & Eff. Rank Ratio & $R_{ER}$ & $ER$ normalized by $D$. & Low values imply features reside in a low-dimensional linear subspace. \\
      & Intrinsic Dim. & $ID$ & MLE of manifold dimension. & High values indicate complex non-linear structures. \\
      & ID Ratio & $R_{ID}$ & $ID$ normalized by $D$. & Measures compactness of the data manifold relative to feature space. \\
      \midrule

      % --- Category 3: Complexity Measures ---
      \multirow{4}{*}{\textbf{Complexity}}
      & Fisher's Ratio & $F1$ & Max Fisher discriminant ratio. & High values indicate high linear separability between classes. \\
      & Borderline Pts. & $N1$ & Fraction of points in MST crossing class boundary. & High values imply complex, overlapping decision boundaries. \\
      & Intra/Inter Ratio & $N2$ & Ratio of within-class to between-class distance. & Low values indicate distinct, compact clusters (easier classification). \\
      & RF's F1 & $ - $ & F1-score of Random Forest. & Serves as a proxy for overall task difficulty (High = Easy). \\
      \midrule

      % --- Category 4: Anomaly Topology ---
      \multirow{2}{*}{\shortstack[l]{\textbf{Anomaly}\\ \textbf{Topology}}}
      & Anom. Cluster. degree & $Clus$ & Clustering coefficient of anomalies. & High values mean anomalies form dense micro-clusters rather than being scattered. \\
      & Local/Global Outlier Ratio & $LGR$ & Ratio of local to global outliers. & High values indicate anomalies are context-dependent rather than extreme values. \\

      \bottomrule
    \end{tabular}%
  }
  \par\vspace{3pt}
  \begin{minipage}{0.95\textwidth}
    \footnotesize \textbf{Note:} Complexity measures (\(F1, N1, N2\)) are adopted from Ho \& Basu \cite{ho2002complexity}; \(ID\) uses the MLE estimator \cite{levina2004maximum}; \(ER\) is based on \cite{roy2007effective}.
  \end{minipage}
\end{table*}

\subsection{Weakly-supervised (Bag)}
\label{appendix:weakly_supervised_bag}

\begin{itemize}
    % Dynamic MIL + Center Loss (Compactness)
  \item \textbf{AR-Net} \cite{wan2020weakly}: An MIL framework incorporating Center Loss to enforce intra-class compactness of normal features and a dynamic instance selection strategy to filter noisy labels.
    % Vanilla MIL + Ranking Loss
  \item \textbf{Sultani} \cite{sultani2018real}: The seminal MIL-based method that introduces a ranking loss to maximize the separability between the highest-scoring instance in abnormal bags and that in normal bags.
    % Magnitude Contrast + Attention
  \item \textbf{MGFN} \cite{chen2023mgfn}: An attention-based network that integrates feature magnitude learning with a glance-and-focus mechanism to capture global context and refine local anomaly localization.
    % Text-Guided + Prompt Learning
  \item \textbf{VadCLIP} \cite{wu2024vadclip}: A vision-language framework that adapts CLIP for VAD using dual-branch prompt learning to align video segments with anomaly-related textual descriptions.
    % Uncertainty Learning + Dual Memory
  \item \textbf{UR-DMU} \cite{zhou2023dual}: An uncertainty-regulated dual-memory unit that uses separate memory banks for normal/abnormal patterns and an uncertainty module to weigh feature reliability and filter noise.
    % Graph Denoising + Pseudo-labeling
  \item \textbf{GCN-Anomaly} \cite{zhong2019graph}: A graph-based framework that treats anomaly detection as a label noise cleaning problem, using GCNs to propagate pseudo-labels based on feature similarity and temporal consistency.
    % Feature Magnitude + Smoothness/Sparsity
  \item \textbf{RTFM} \cite{tian2021weakly}: A robust temporal feature magnitude learning method that enforces separability in the feature magnitude domain with additional temporal smoothness and sparsity constraints.
\end{itemize}

\subsection{Supervised (Instance)}
\begin{itemize}
    % Gradient Boosting + Tree Ensemble
  \item \textbf{XGBoost} \cite{chen2016xgboost}: A scalable tree boosting system that implements gradient-boosted decision trees with regularization to control overfitting and optimize computational efficiency.
    % Categorical Features + Gradient Boosting
  \item \textbf{CatBoost} \cite{prokhorenkova2018catboost}: A gradient boosting framework specialized for categorical features, using ordered boosting to mitigate prediction shift and handle categorical variables without extensive preprocessing.
    % Transformer for Tabular + Feature Tokenization
  \item \textbf{FTTransformer} \cite{gorishniy2021revisiting}: A Transformer-based architecture adapted for tabular data, treating features as tokens and applying self-attention to capture complex feature interactions.
    % Multiple Class Ideas + Deep Tabular (Note: TabMCIs is likely a specialized variant or typo, assuming generic Deep Tabular or specific paper if known. Based on context, likely a deep tabular baseline. If citation exists, use it.)
    % Assuming TabMCIs refers to "Model with Class-Imbalance strategies" or similar. Since exact mechanism is niche, generalized description:
  \item \textbf{TabM} \cite{gorishniy2024tabm}: A modern deep tabular learning framework that leverages model ensembling and advanced regularization techniques to improve robustness on heterogeneous tabular datasets.
    % Retrieval-augmented + Tabular
  \item \textbf{TabR-S} \cite{gorishniy2023tabr}: A retrieval-augmented tabular deep learning model that enhances prediction by retrieving and aggregating information from similar training examples (neighbors) during inference.
\end{itemize}

\subsection{Tabular Foundation Models (Instance)}
\begin{itemize}
    % Prior-Data Fitting + In-Context Learning
  \item \textbf{TabPFN} \cite{grinsztajn2025tabpfn}: A Prior-Data Fitted Network that learns a Bayesian inference algorithm on synthetic datasets, enabling efficient in-context learning for tabular classification without gradient-based training.
    % Large Scale Pretraining + Transfer Learning
  \item \textbf{LimiX} \cite{zhang2509limix}: A general-purpose large model for structured data modeling that leverages massive cross-table pretraining to transfer knowledge to downstream tasks, exhibiting strong zero-shot generalization capabilities.
\end{itemize}

\section{Dataset Meta-features and Correlation Analysis}

\begin{figure*}[t]
  \centering
  \begin{subfigure}[b]{0.32\textwidth}
    \includegraphics[width=\linewidth]{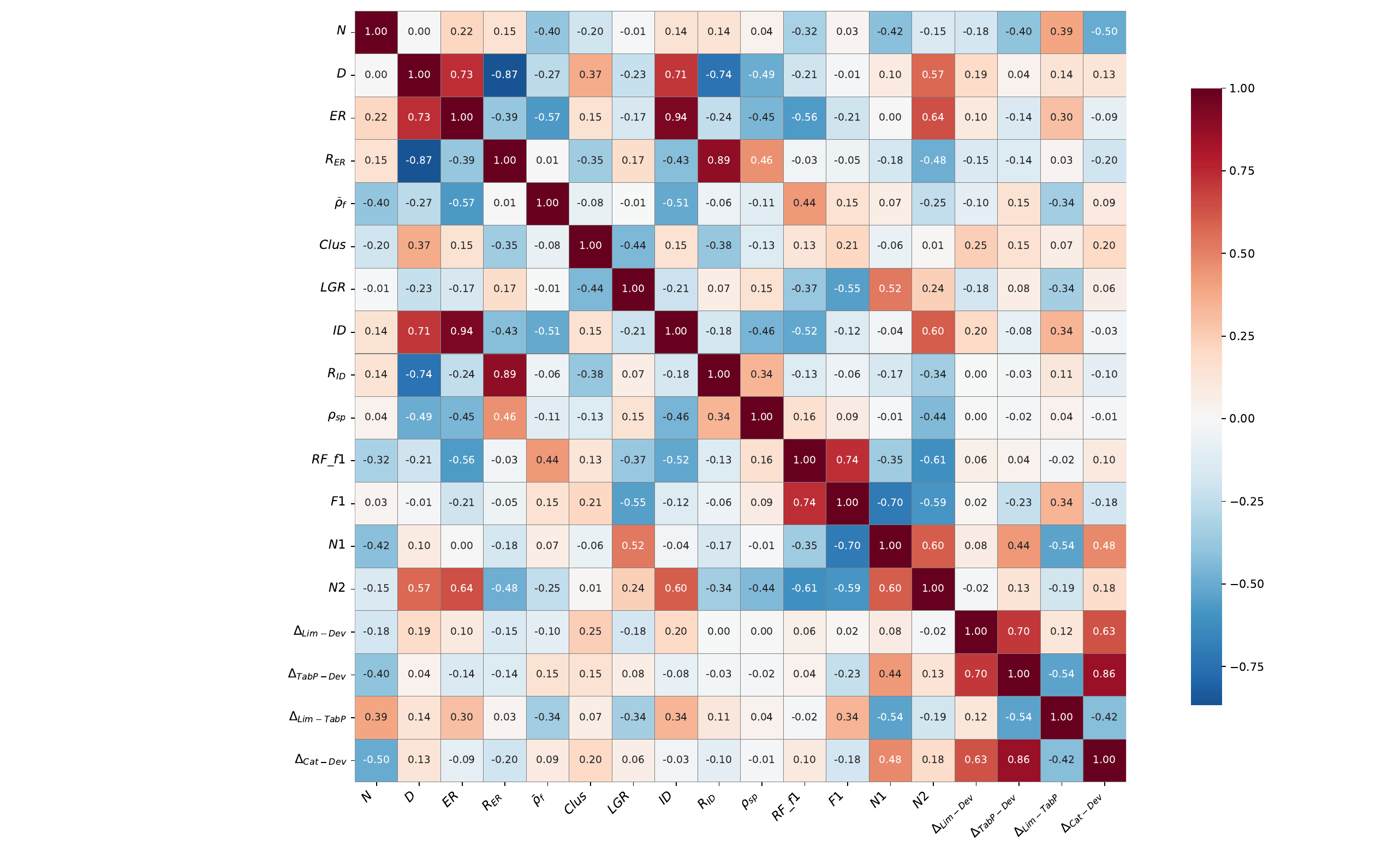}
    \caption{$\rla=1\%$}
    \label{fig:meta_corr_rla001}
  \end{subfigure}
  \hfill
  \begin{subfigure}[b]{0.32\textwidth}
    \includegraphics[width=\linewidth]{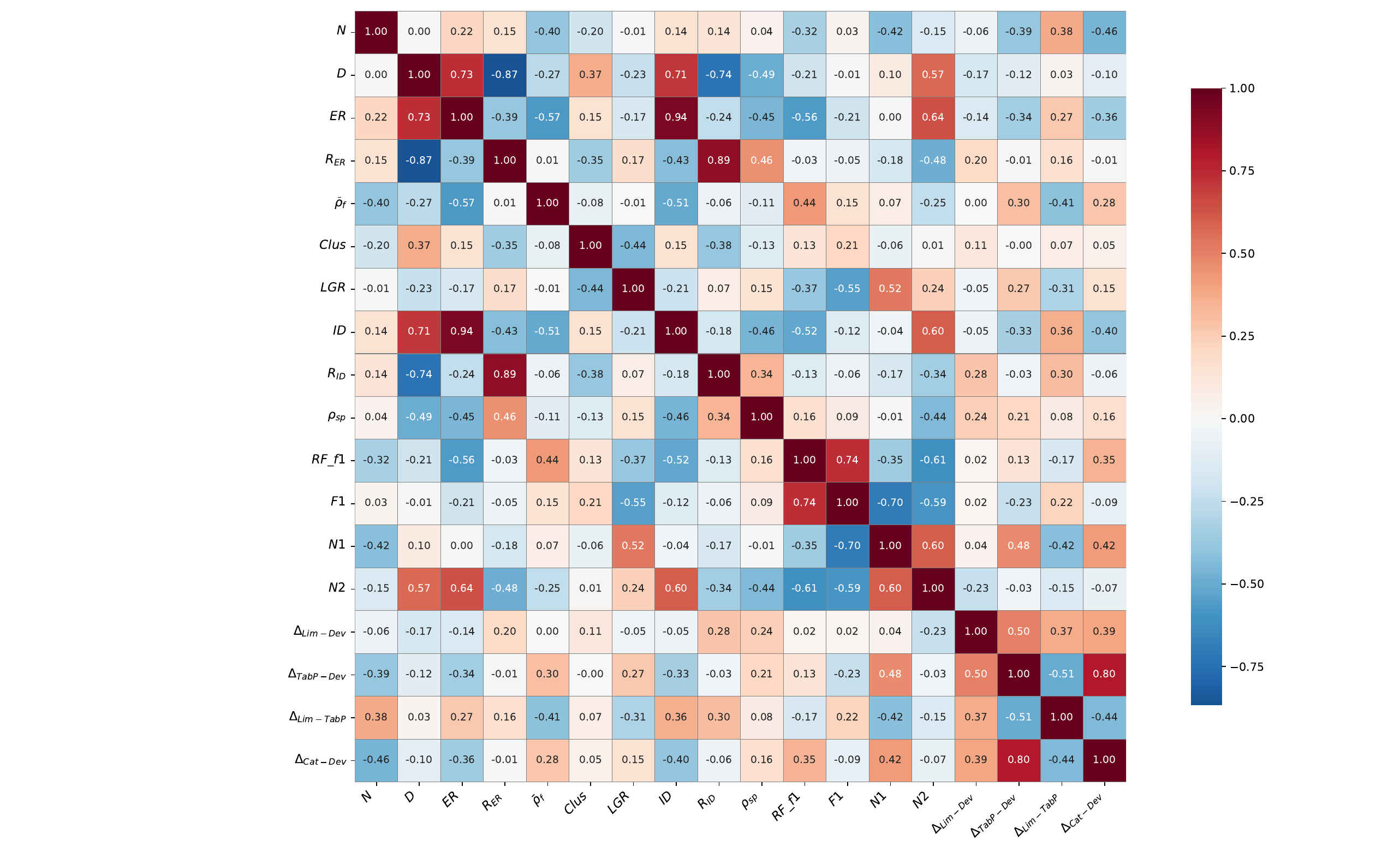}
    \caption{$\rla=5\%$}
    \label{fig:meta_corr_rla005}
  \end{subfigure}
  \hfill
  \begin{subfigure}[b]{0.32\textwidth}
    \includegraphics[width=\linewidth]{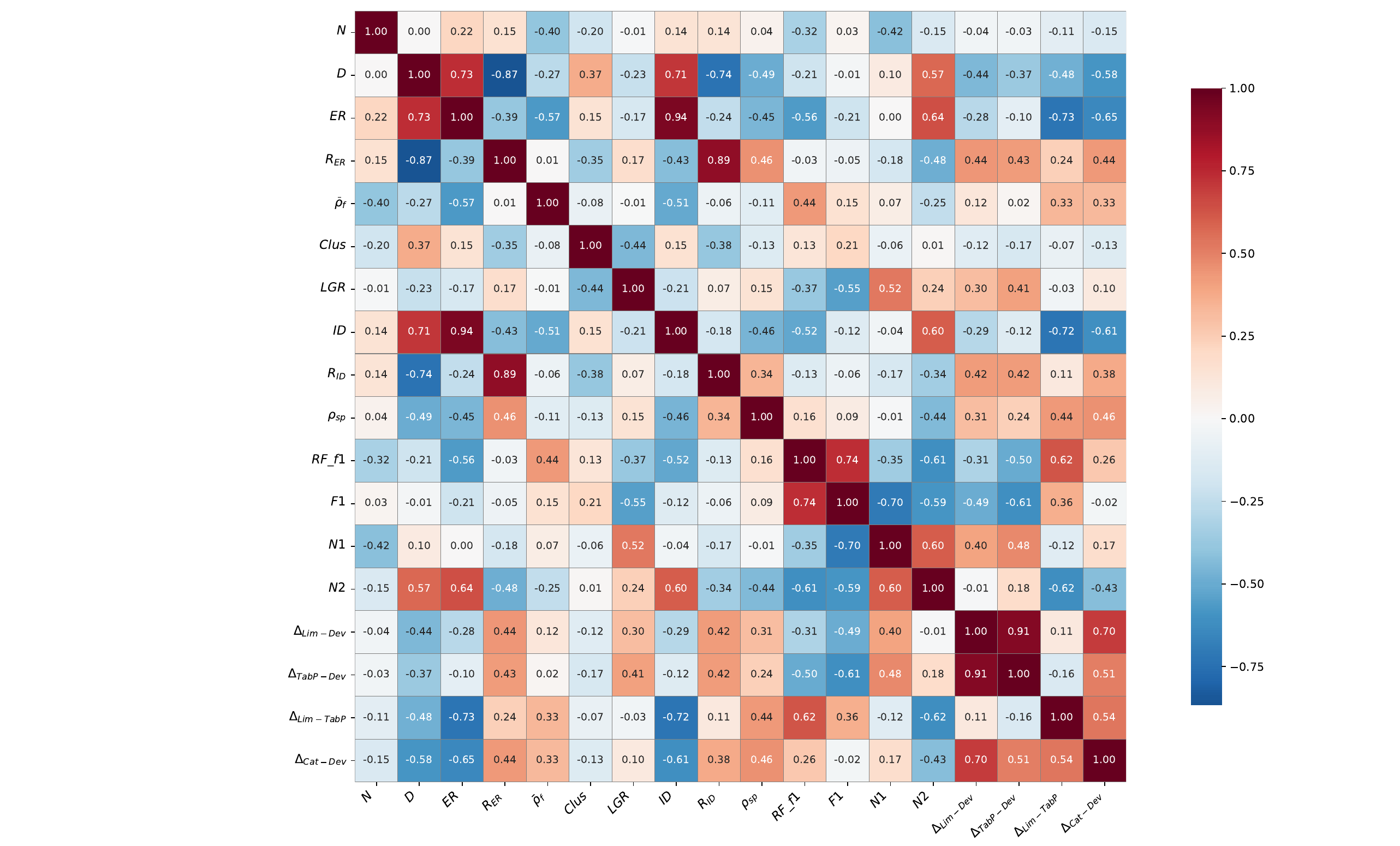}
    \caption{$\rla=100\%$}
    \label{fig:meta_corr_rla100}
  \end{subfigure}
  \vspace{-8pt}
  \caption{Heatmap of spearman correlations between dataset meta-features (denoted by symbols) and model performance under varying supervision ratios. Red (high) indicates positive correlation, while blue (low) indicates negative. $\Delta_{A-B}$ denotes the performance gap between model A and model B.}
  \label{fig:meta_corr_all}
\end{figure*}

\begin{table*}[t]
  \begin{minipage}[t]{0.49\textwidth}
    \centering
    \caption{Spearman correlation ($\rho$) of meta-features across different model comparisons on $\rla=1\%$ of all tabular datasets.}
    \vspace{-10pt}
    \label{tab:multi_target_corr}
    \resizebox{\textwidth}{!}{
      \begin{tabular}{lcccc}
        \toprule
        \textbf{Meta-Feature} & \textbf{LimiX - DevNet} & \textbf{TabPFN - DevNet} & \textbf{LimiX - TabPFN} & \textbf{CatB - DevNet} \\
        \midrule
        Sample Size               & $-0.18^{*}$  & $-0.40^{***}$ & $0.39^{***}$ & $-0.50^{***}$ \\
        Dimensionality            & $0.19^{*}$   & $0.04$       & $0.14$       & $0.13$       \\
        Avg. Feat. Sim.           & $-0.10$      & $0.15$       & $-0.34^{***}$ & $0.09$       \\
        Effective Rank            & $0.10$       & $-0.14$      & $0.30^{***}$ & $-0.09$      \\
        Eff. Rank Ratio           & $-0.15$      & $-0.14$      & $0.03$       & $-0.20^{*}$  \\
        Intrinsic Dim.            & $0.20^{*}$   & $-0.08$      & $0.34^{***}$ & $-0.03$      \\
        Fisher's Ratio            & $0.02$       & $-0.23^{**}$ & $0.34^{***}$ & $-0.18^{*}$  \\
        Borderline Pts.           & $0.08$       & $0.44^{***}$ & $-0.54^{***}$ & $0.48^{***}$ \\
        Intra/Inter Ratio         & $-0.02$      & $0.13$       & $-0.19^{*}$  & $0.18^{*}$   \\
        Anom. Cluster. degree     & $0.25^{**}$  & $0.15$       & $0.07$       & $0.20^{*}$   \\
        Local/Global Outlier Ratio & $-0.18^{*}$  & $0.08$       & $-0.34^{***}$ & $0.06$       \\
        \bottomrule
      \end{tabular}
    }
    \vspace{1mm}
    \begin{flushleft}
      \scriptsize \textit{Note: $^{*}p<0.05, ^{**}p<0.01, ^{***}p<0.001$. Non-significant entries are unstarred.}
    \end{flushleft}
  \end{minipage}
  \hfill
  \begin{minipage}[t]{0.49\textwidth}
    \centering
    \caption{Spearman correlation ($\rho$) of meta-features across different model comparisons on $\rla=5\%$ of all tabular datasets.}
    \vspace{-10pt}
    \label{tab:multi_target_corr_v2}
    \resizebox{\textwidth}{!}{
      \begin{tabular}{lcccc}
        \toprule
        \textbf{Meta-Feature} & \textbf{LimiX - DevNet} & \textbf{TabPFN - DevNet} & \textbf{LimiX - TabPFN} & \textbf{CatB - DevNet} \\
        \midrule
        Sample Size               & $-0.06$      & $-0.39^{***}$ & $0.38^{***}$ & $-0.46^{***}$ \\
        Sparsity Ratio            & $0.24^{**}$  & $0.21^{*}$   & $0.08$       & $0.16$       \\
        Avg. Feat. Sim.           & $0.00$       & $0.30^{**}$  & $-0.41^{***}$ & $0.28^{**}$  \\
        Effective Rank            & $-0.14$      & $-0.34^{***}$ & $0.27^{**}$  & $-0.36^{***}$ \\
        Eff. Rank Ratio           & $0.20^{*}$   & $-0.01$      & $0.16$       & $-0.01$      \\
        Intrinsic Dim.            & $-0.05$      & $-0.33^{***}$ & $0.36^{***}$ & $-0.40^{***}$ \\
        ID Ratio                  & $0.28^{**}$  & $-0.03$      & $0.30^{***}$ & $-0.06$      \\
        Fisher's Ratio            & $0.02$       & $-0.23^{*}$  & $0.22^{*}$   & $-0.09$      \\
        Borderline Pts.           & $0.04$       & $0.48^{***}$ & $-0.42^{***}$ & $0.42^{***}$ \\
        Intra/Inter Ratio         & $-0.23^{*}$  & $-0.03$      & $-0.15$      & $-0.07$      \\
        RF's F1                   & $0.02$       & $0.13$       & $-0.17$      & $0.35^{***}$ \\
        Local/Global Outlier Ratio & $-0.05$      & $0.27^{**}$  & $-0.31^{***}$ & $0.15$       \\
        \bottomrule
      \end{tabular}
    }
    \vspace{1mm}
    \begin{flushleft}
      \scriptsize \textit{Note: $^{*}p<0.05, ^{**}p<0.01, ^{***}p<0.001$. Non-significant entries are unstarred.}
    \end{flushleft}
  \end{minipage}
\end{table*}

\subsection{Meta-feature Definitions}
\label{appendix:meta_feature_definitions}

To systematically characterize the intrinsic properties of datasets and investigate their impact on anomaly detection performance, we extracted a comprehensive set of meta-features. The definitions and mathematical formulations of some key meta-features are detailed below. Let $\mathcal{D} = \{(x_i, y_i)\}_{i=1}^N$ denote a dataset with $N$ samples and $D$ features, where $y_i \in \{0, 1\}$ represents the label (0 for normal, 1 for abnormal).

Average Feature Similarity (Avg. Feat. Sim., $\bar{\rho}_f$): This metric quantifies feature redundancy by averaging the absolute Pearson correlation coefficients between all pairs of features.$$\bar{\rho}_f = \frac{2}{D(D-1)} \sum_{i=1}^{D-1} \sum_{j=i+1}^{D} |\text{corr}(f_i, f_j)|$$where $f_i$ denotes the $i$-th feature vector. A higher $\bar{\rho}_f$ indicates strong linear dependencies among features.

Intrinsic Dimensionality ($ID$): We estimate the non-linear manifold dimension of the data using the Maximum Likelihood Estimation (MLE) method based on nearest neighbor distances, which can explain 95\% variance. For each sample $x_i$, let $r_k(x_i)$ be the distance to its $k$-th nearest neighbor. The $ID$ is estimated as:$$ID = \left[ \frac{1}{N} \sum_{i=1}^{N} \frac{1}{k-1} \sum_{j=1}^{k-1} \ln \frac{r_k(x_i)}{r_j(x_i)} \right]^{-1}$$This value represents the minimum number of variables required to describe the data without significant information loss.

ID Ratio ($R_{ID}$): Defined as the ratio of the intrinsic dimensionality to the original feature space dimension ($D$), this metric measures the compactness of the data manifold relative to the embedding space.$$R_{ID} = \frac{ID}{D}$$A lower $R_{ID}$ suggests that the data resides on a low-dimensional manifold within the high-dimensional feature space.

Fisher’s Discriminant Ratio ($F1$): To assess linear separability, we compute the maximum Fisher’s discriminant ratio across all individual feature dimensions.$$F1 = \max_{j=1 \dots D} \frac{(\mu_{0, j} - \mu_{1, j})^2}{\sigma_{0, j}^2 + \sigma_{1, j}^2}$$where $\mu_{c, j}$ and $\sigma_{c, j}^2$ are the mean and variance of the $j$-th feature for class $c$. A higher $F1$ indicates that at least one feature provides strong discriminative power between normal and abnormal classes.

Fraction of Borderline Points ($N1$): This complexity measure quantifies the intricacy of the decision boundary. We construct a Minimum Spanning Tree (MST) over the entire dataset and calculate the fraction of edges connecting samples from different classes.$$N1 = \frac{1}{N} \sum_{(x_i, x_j) \in \text{MST}} \mathbb{I}(y_i \neq y_j)$$High values of $N1$ imply that normal and abnormal samples are interleaved, indicating a complex classification boundary.

Ratio of Intra/Inter Class Nearest Neighbor Distances ($N2$): This metric evaluates the clustering structure by comparing within-class compactness to between-class separation.$$N2 = \frac{\sum_{i=1}^N d(x_i, NN_{\text{same}}(x_i))}{\sum_{i=1}^N d(x_i, NN_{\text{diff}}(x_i))}$$where $d(\cdot)$ is the Euclidean distance, and $NN_{\text{same}}$ ($NN_{\text{diff}}$) denotes the nearest neighbor from the same (different) class. Lower values indicate distinct, compact clusters.

Anomaly Clustering Degree ($Clus$): This feature measures the tendency of anomalies to form dense micro-clusters rather than being randomly scattered. It is calculated as the average local clustering coefficient of the subgraph induced by abnormal samples in the $k$-nearest neighbor graph.$$Clus = \frac{1}{N_{anom}} \sum_{x \in \mathcal{S}_{anom}} C(x)$$where $C(x)$ is the local clustering coefficient of node $x$. High values suggest that anomalies exhibit local structural patterns.

\subsection{Correlation with Model Performance}
To visualize the relationship between these meta-features and model performance, we present the Spearman correlation heatmaps across varying supervision levels in Figure~\ref{fig:meta_corr_all} ($\rla=1\%$ in \subref{fig:meta_corr_rla001}, $\rla=5\%$ in \subref{fig:meta_corr_rla005}, and $\rla=100\%$ in \subref{fig:meta_corr_rla100}). These visualizations highlight how the importance of specific dataset characteristics shifts as supervision signals become more abundant. While the main text focuses on the meta-feature analysis under full supervision ($\gamma_{la}=100\%$), we also investigate how these correlations evolve under limited supervision. Table \ref{tab:multi_target_corr} and Table \ref{tab:multi_target_corr_v2} present the Spearman correlations between meta-features and model performance gaps at $\gamma_{la}=1\%$ and $\gamma_{la}=5\%$, respectively. These results highlight the consistency of certain meta-features (e.g., dimensionality) in influencing model superiority even when labeled data is scarce.

\begin{figure*}[htbp]
  \centering

  % 第一张子图 (a)
  \begin{subfigure}[b]{0.48\textwidth}
    \centering
    \includegraphics[width=\textwidth, trim=0pt 0pt 30pt 0pt, clip]{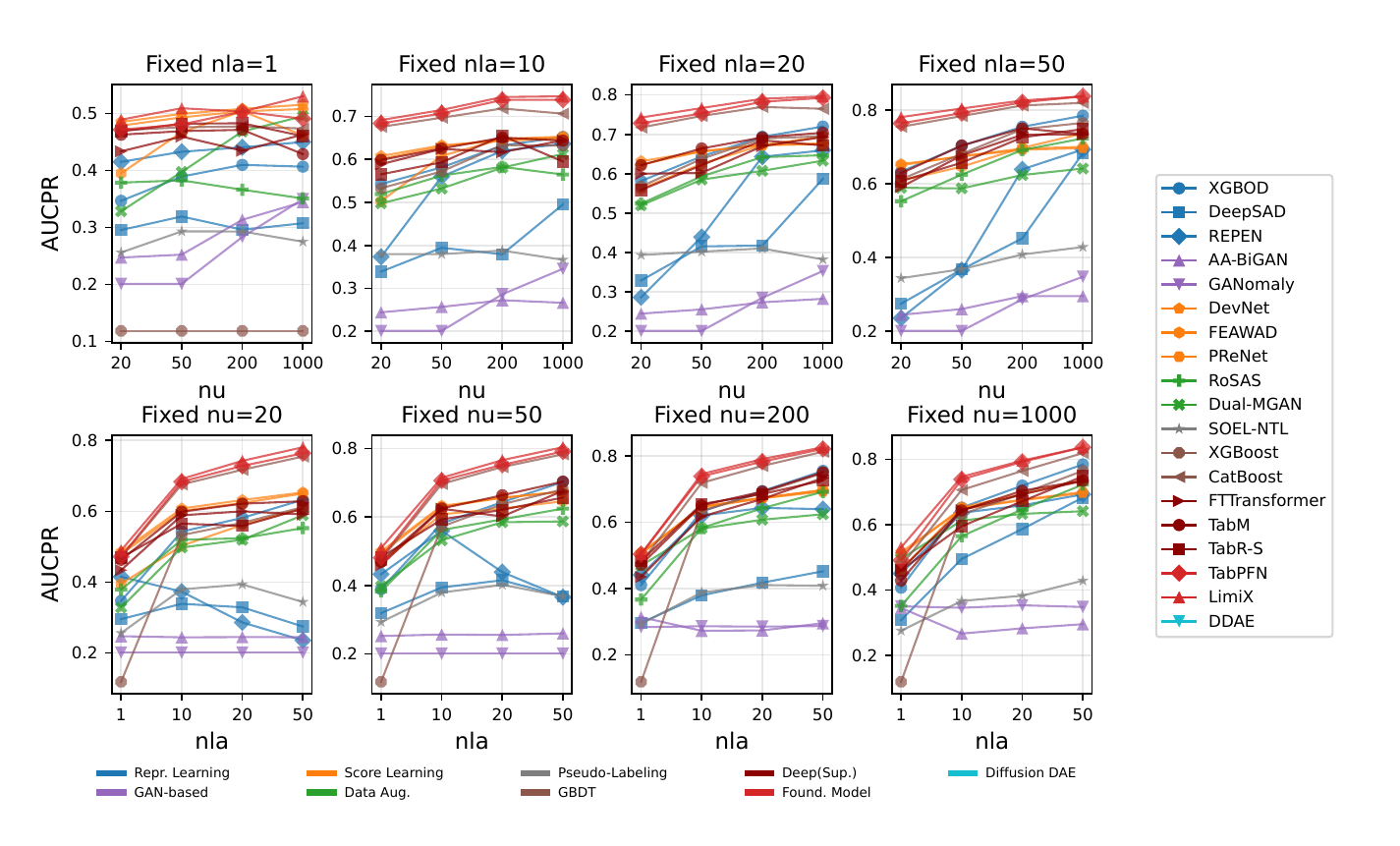}
    \caption{AUCPR Marginal effect}
    \label{fig:unlabel_aucpr}
  \end{subfigure}
  \hfill % 使用 hfill 自动填充两张图之间的空白
  % 第二张子图 (b)
  \begin{subfigure}[b]{0.48\textwidth}
    \centering
    \includegraphics[width=\textwidth, trim=0pt 0pt 30pt 0pt, clip]{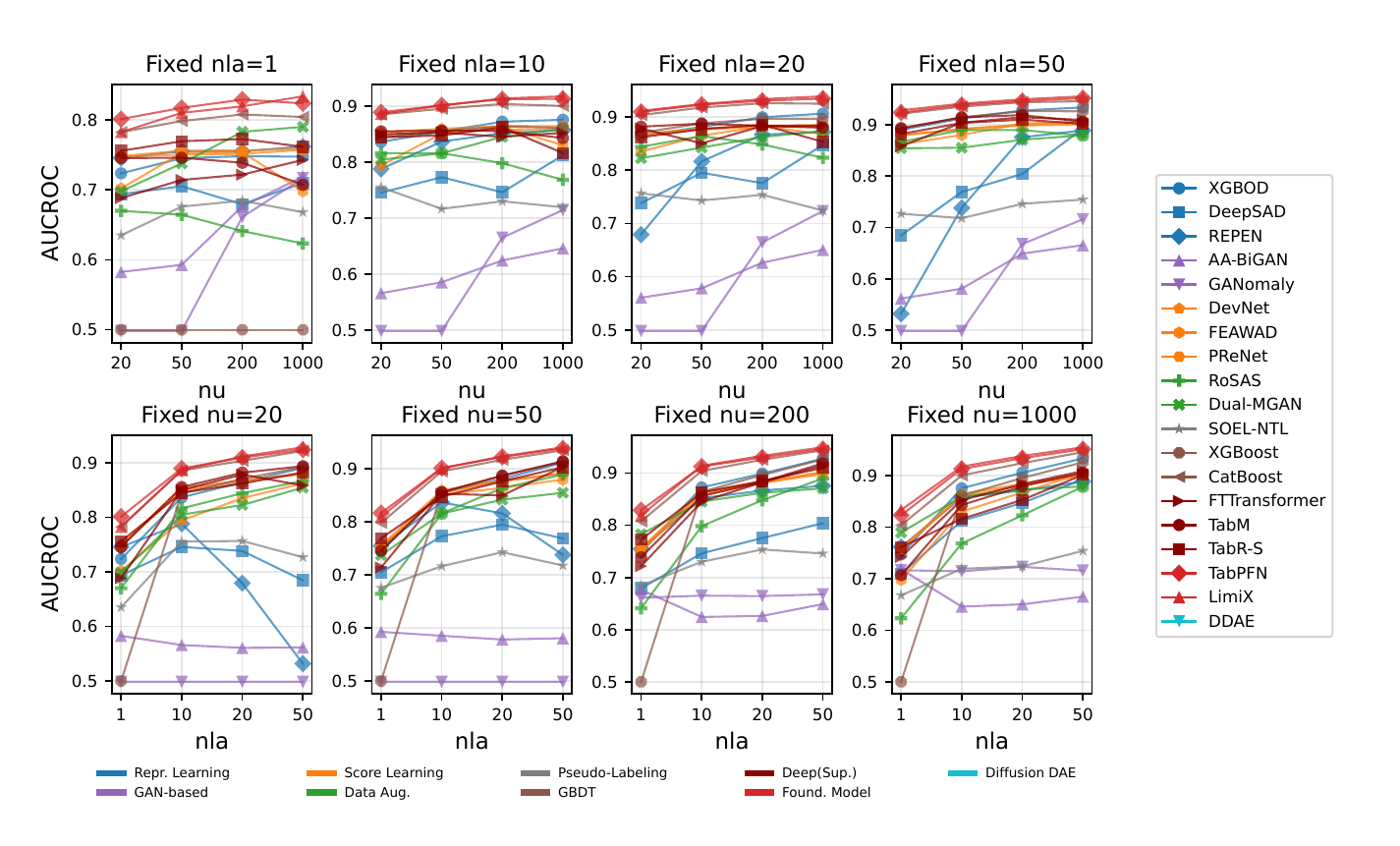}
    \caption{AUCROC Marginal effect}
    \label{fig:unlabel_aucroc}
  \end{subfigure}

  \vspace{-5pt}
  \caption{AUCPR (a) and AUCROC (b) results on Tabular datasets under varying labeled ($\nla$) and unlabeled ($\numu$) data amounts.}
  \label{fig:unlabel_overall}
\end{figure*}

%--------------------------------------------------------

\begin{figure*}[htbp]
  \centering
  \begin{subfigure}[b]{0.48\textwidth}
    \centering
    % trim = <left> <bottom> <right> <top> (左 下 右 上)
    \includegraphics[width=\textwidth, trim=0pt 0pt 30pt 0pt, clip]{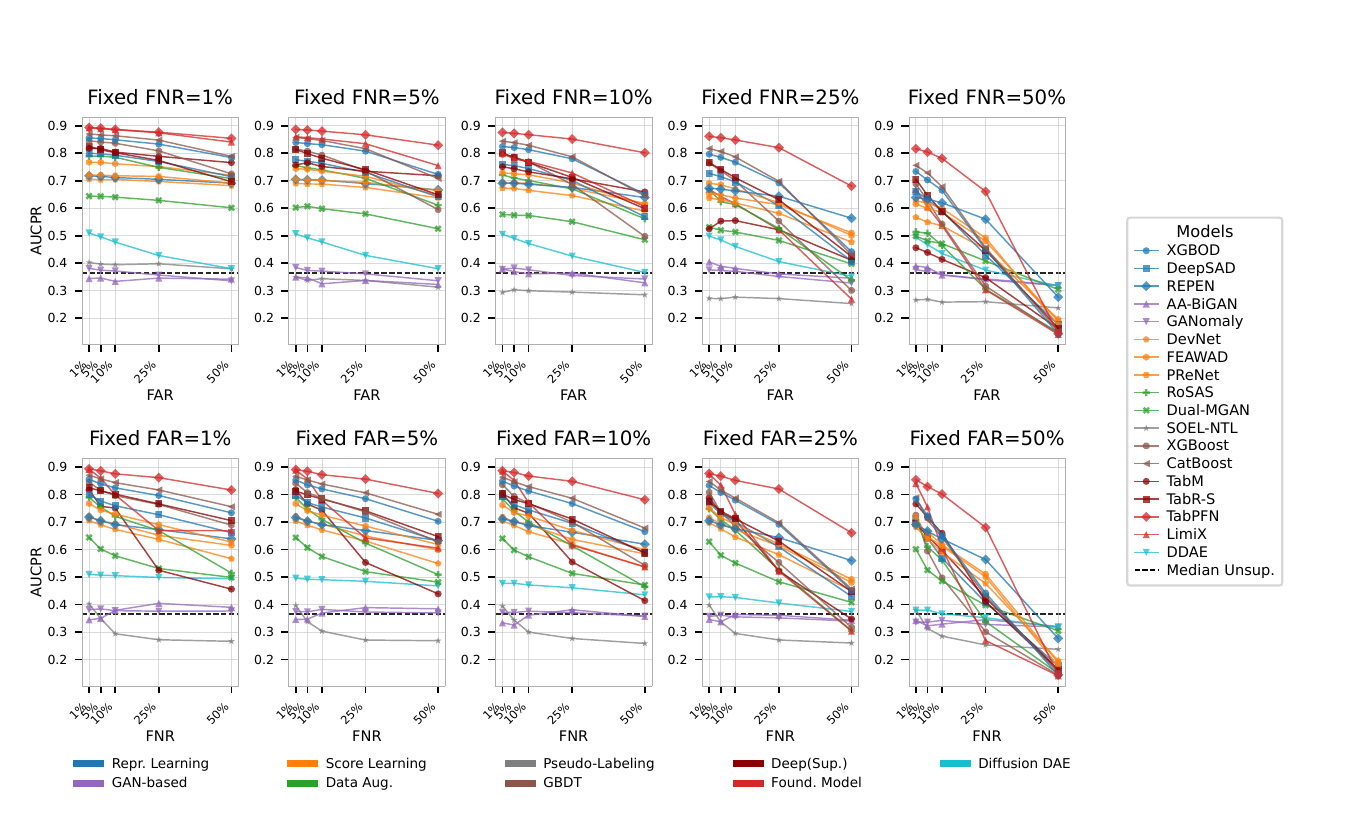}
    \caption{AUCPR performance}
    \label{fig:inacc_aucpr}
  \end{subfigure}
  \hfill
  \begin{subfigure}[b]{0.48\textwidth}
    \centering
    % trim = <left> <bottom> <right> <top> (左 下 右 上)
    \includegraphics[width=\textwidth, trim=0pt 0pt 30pt 0pt, clip]{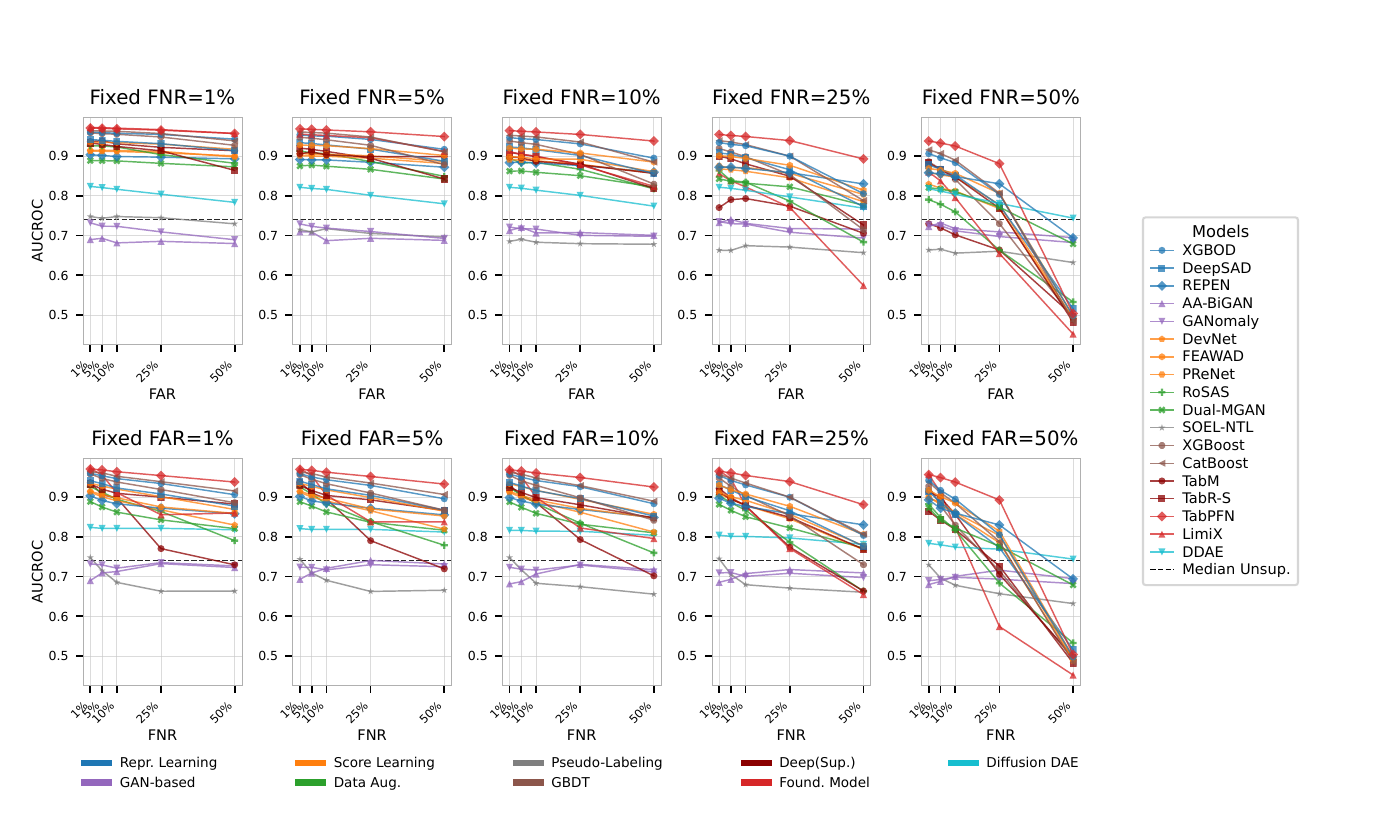}
    \caption{AUCROC performance}
    \label{fig:inacc_aucroc}
  \end{subfigure}

  \vspace{-5pt}
  \caption{Performance comparison under inaccurate conditions (label noise) on tabular datasets.}
  \label{fig:inacc_overall}
\end{figure*}

\section{Supplement to Unlabeled Data Utilization Experiments}
\label{appendix:unlabeled_data}

This section extends the unlabeled data utilization analysis by evaluating a broader range of model architectures.
Specifically, Figure~\ref{fig:unlabel_overall} presents the 2D marginal effects of labeled anomalies and unlabeled data across all evaluated methods.
These curves confirm that the trends observed for representative models in Section~\ref{sec:unlabeled_value} generalize globally.
For instance, the performance of representation-based methods (e.g., DeepSAD, REPEN) depends heavily on the volume of unlabeled data.
Conversely, supervised methods and tabular foundation models consistently maintain high label efficiency.

Furthermore, Figure~\ref{fig:3d_full_unlabe} visualizes the joint effects of labeled and unlabeled data using 3D surfaces for five representative models.
These models span different categories, including tabular foundation models (TabPFN, LimiX) and representation-based methods (DeepSAD).
The 3D visualizations verify that the synergy between labeled and unlabeled samples is a common characteristic.
Specifically, for all evaluated models, scaling unlabeled data yields the most significant gains when anomalies are scarce.
This extended comparison demonstrates that the dependencies discussed in the main text are not model-specific but architectural traits.

\section{Supplement to Inaccurate Scenario Experiments}
\label{appendix:inaccurate_scenarios}

We extend the label noise sensitivity analysis to a wider range of algorithms to verify the generalizability of our findings.
Figure~\ref{fig:inacc_overall} reports the AUCPR and AUCROC performance of all evaluated models under varying noise ratios.
These detailed curves confirm the asymmetric sensitivity to noise type as a universal phenomenon across architectures.
Specifically, every model degrades sharper under Flip-Normal noise (FNR) compared to Flip-Abnormal noise (FAR).
This consistent trend validates that FNR is inherently more destructive to weakly-supervised anomaly detection.

Additionally, Figure~\ref{fig:3d_inacc_overall} displays the 3D performance landscapes of five representative models under double-noise conditions.
These surfaces highlight key differences in noise tolerance dynamics.
Tabular foundation models (TabPFN, LimiX) maintain broad, flat plateaus under low-to-moderate noise, indicating superior initial tolerance.
However, their performance collapses abruptly once the noise exceeds a critical threshold, dropping below other methods.
In contrast, representation-based models like DeepSAD degrade earlier and more steadily under mild noise, lacking initial tolerance plateaus.
Consequently, under extreme contamination, some representation-based models like REPEN eventually exhibit higher remaining performance than foundation models.

\clearpage

\begin{figure*}[p]
  \centering

  % --- Row 1 ---
  \begin{subfigure}[b]{0.16\textwidth}
    \centering
    \includegraphics[width=\textwidth, height=0.94\linewidth]{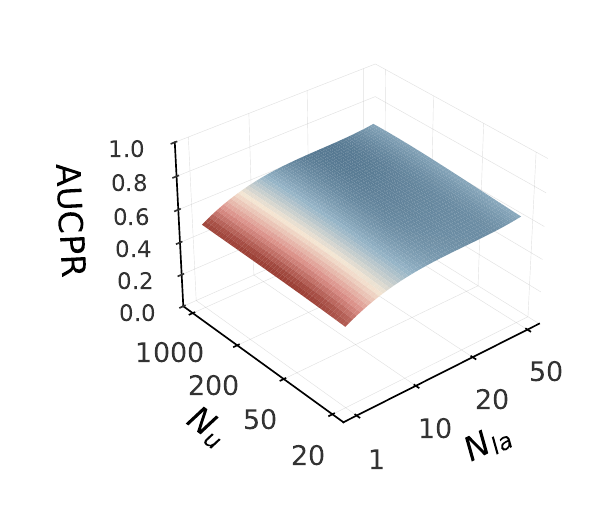}
    \caption{DevNet}
    \label{fig:3d_unlabe_DevNet}
  \end{subfigure}
  \hfill
  \begin{subfigure}[b]{0.16\textwidth}
    \centering
    \includegraphics[width=\textwidth, height=0.94\linewidth]{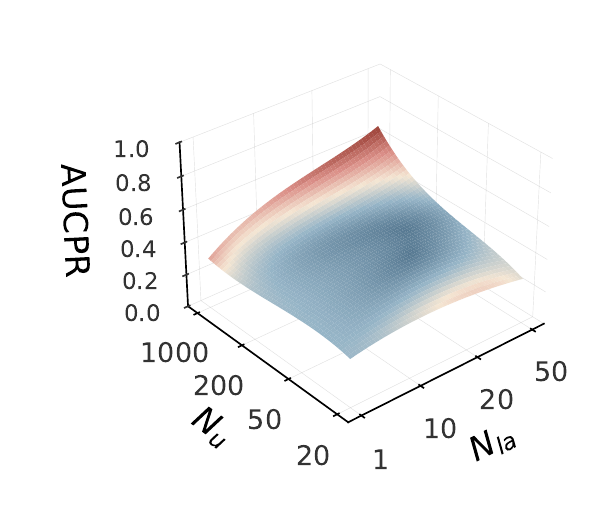}
    \caption{DeepSAD}
    \label{fig:3d_unlabe_DeepSAD}
  \end{subfigure}
  \hfill
  \begin{subfigure}[b]{0.16\textwidth}
    \centering
    \includegraphics[width=\textwidth, height=0.94\linewidth]{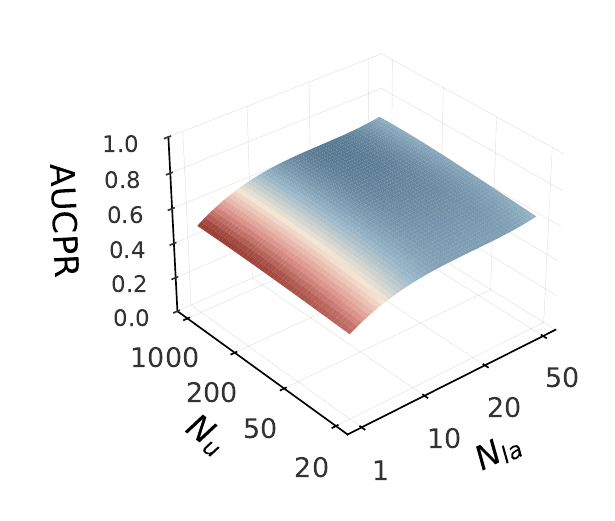}
    \caption{PReNet}
    \label{fig:3d_unlabe_PReNet}
  \end{subfigure}
  \hfill
  \begin{subfigure}[b]{0.16\textwidth}
    \centering
    \includegraphics[width=\textwidth, height=0.94\linewidth]{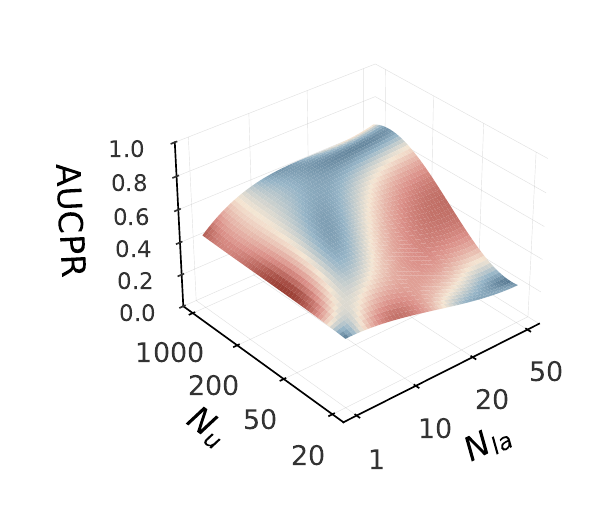}
    \caption{REPEN}
    \label{fig:3d_unlabe_REPEN}
  \end{subfigure}
  \hfill
  \begin{subfigure}[b]{0.16\textwidth}
    \centering
    \includegraphics[width=\textwidth, height=0.94\linewidth]{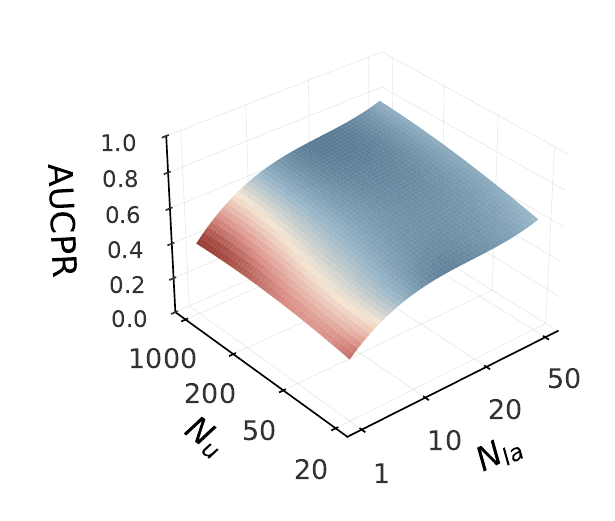}
    \caption{XGBOD}
    \label{fig:3d_unlabe_XGBOD}
  \end{subfigure}
  \hfill
  \begin{subfigure}[b]{0.16\textwidth}
    \centering
    \includegraphics[width=\textwidth, height=0.94\linewidth]{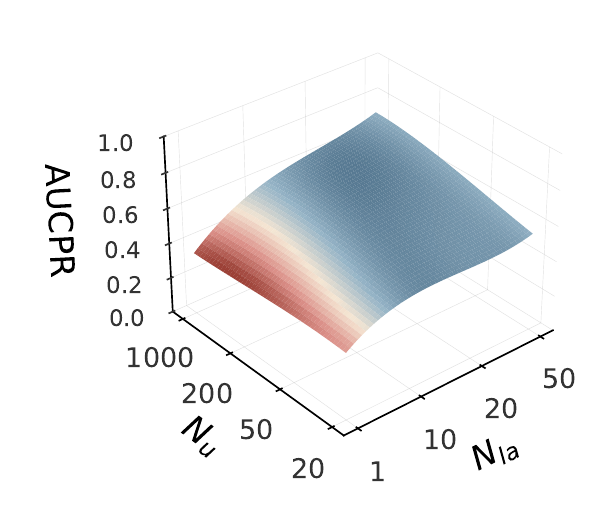}
    \caption{RoSAS}
    \label{fig:3d_unlabe_RoSAS}
  \end{subfigure}

  \vspace{4mm}

  % --- Row 2 ---
  \begin{subfigure}[b]{0.16\textwidth}
    \centering
    \includegraphics[width=\textwidth, height=0.94\linewidth]{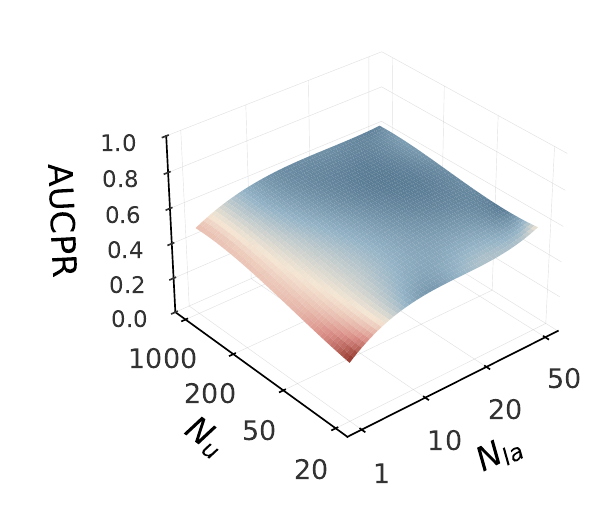}
    \caption{Dual-MGAN}
    \label{fig:3d_unlabe_Dual_MGAN}
  \end{subfigure}
  \hfill
  \begin{subfigure}[b]{0.16\textwidth}
    \centering
    \includegraphics[width=\textwidth, height=0.94\linewidth]{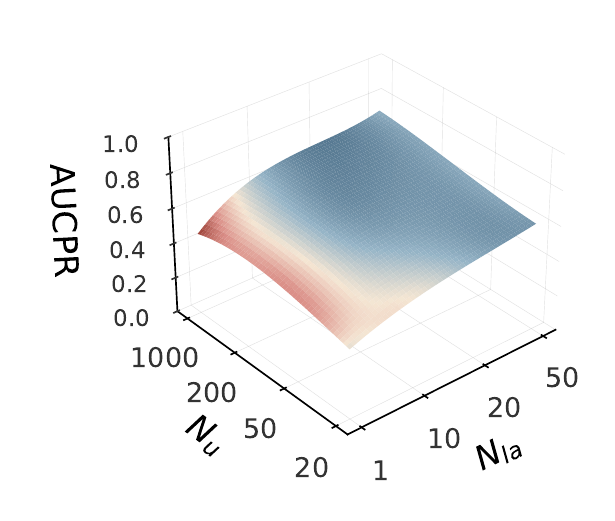}
    \caption{FEAWAD}
    \label{fig:3d_unlabe_FEAWAD}
  \end{subfigure}
  \hfill
  \begin{subfigure}[b]{0.16\textwidth}
    \centering
    \includegraphics[width=\textwidth, height=0.94\linewidth]{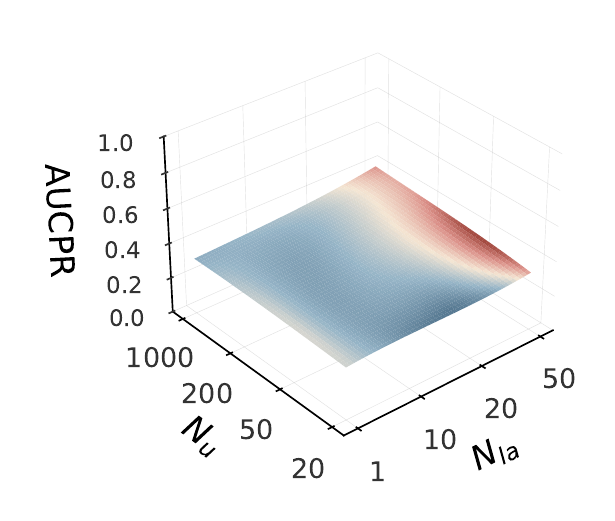}
    \caption{AnoDDAE}
    \label{fig:3d_unlabe_AnoDDAE}
  \end{subfigure}
  \hfill
  \begin{subfigure}[b]{0.16\textwidth}
    \centering
    \includegraphics[width=\textwidth, height=0.94\linewidth]{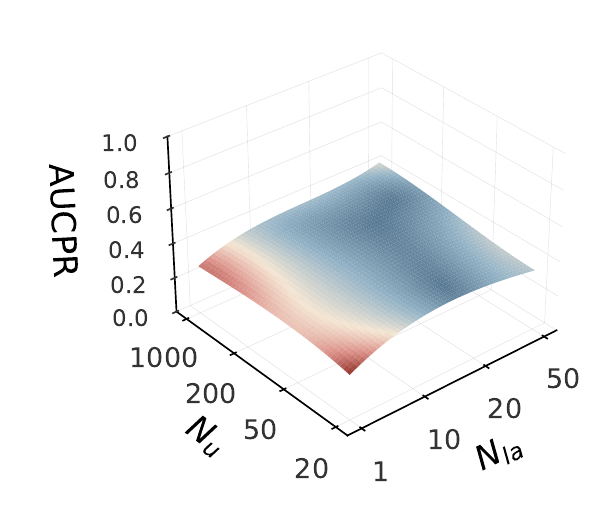}
    \caption{SOEL-NTL}
    \label{fig:3d_unlabe_SOEL_NTL}
  \end{subfigure}
  \hfill
  \begin{subfigure}[b]{0.16\textwidth}
    \centering
    \includegraphics[width=\textwidth, height=0.94\linewidth]{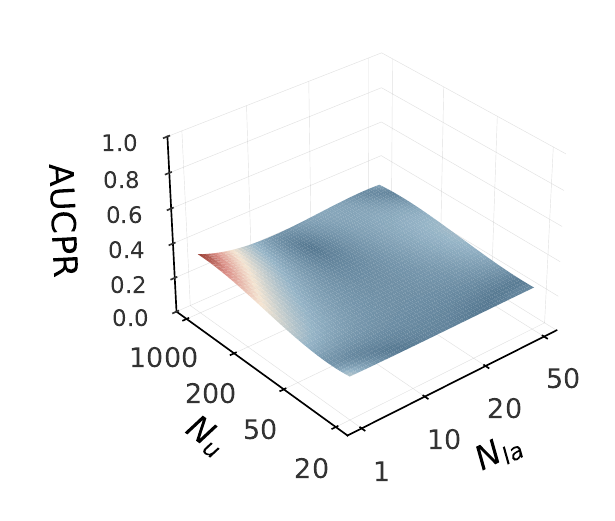}
    \caption{AA-BiGAN}
    \label{fig:3d_unlabe_AA_BiGAN}
  \end{subfigure}
  \hfill
  \begin{subfigure}[b]{0.16\textwidth}
    \centering
    \includegraphics[width=\textwidth, height=0.94\linewidth]{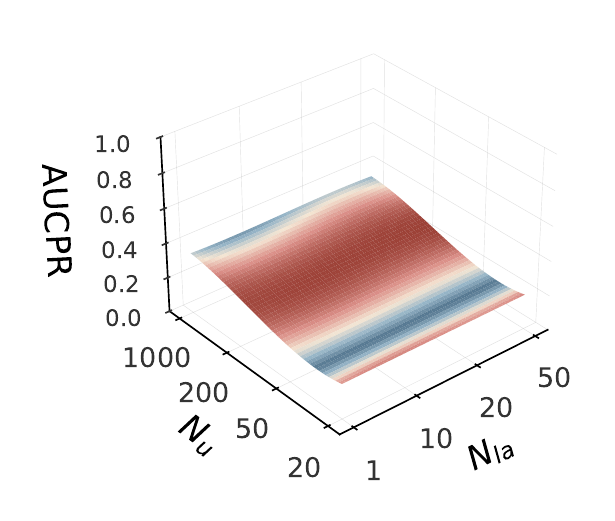}
    \caption{GANomaly}
    \label{fig:3d_unlabe_GANomaly}
  \end{subfigure}

  \vspace{4mm}

  % --- Row 3 ---
  \begin{subfigure}[b]{0.16\textwidth}
    \centering
    \includegraphics[width=\textwidth, height=0.94\linewidth]{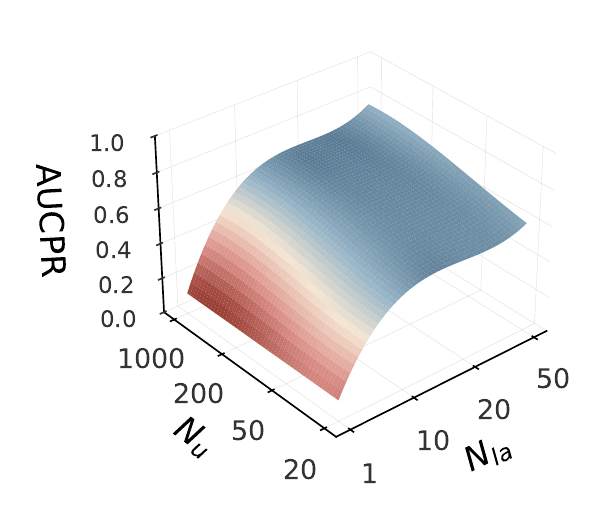}
    \caption{XGBoost}
    \label{fig:3d_unlabe_XGBoost}
  \end{subfigure}
  \hfill
  \begin{subfigure}[b]{0.16\textwidth}
    \centering
    \includegraphics[width=\textwidth, height=0.94\linewidth]{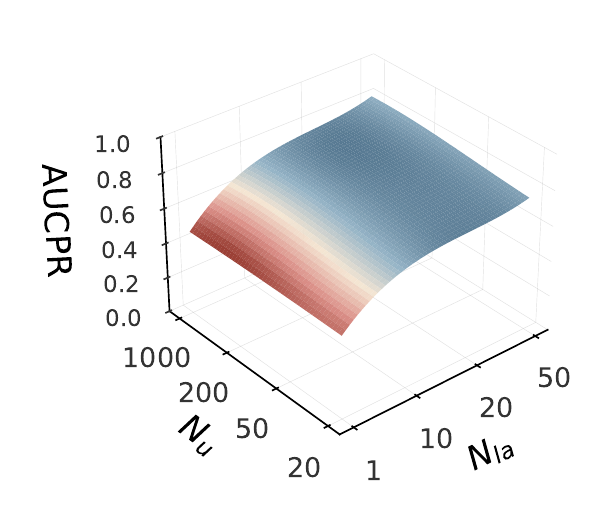}
    \caption{CatBoost}
    \label{fig:3d_unlabe_CatBoost}
  \end{subfigure}
  \hfill
  \begin{subfigure}[b]{0.16\textwidth}
    \centering
    \includegraphics[width=\textwidth, height=0.94\linewidth]{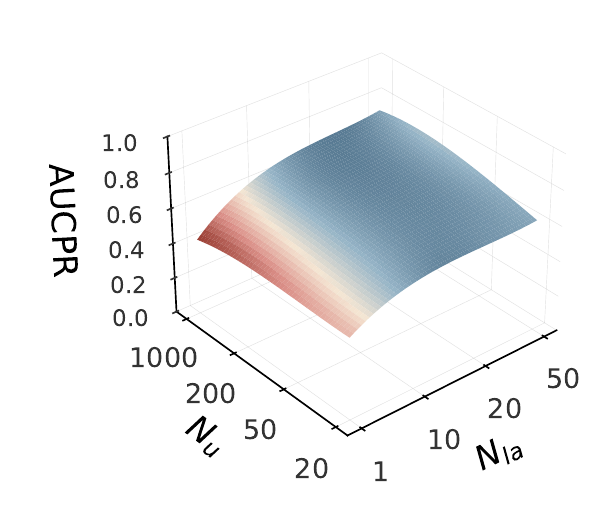}
    \caption{TabM}
    \label{fig:3d_unlabe_TabM}
  \end{subfigure}
  \hfill
  \begin{subfigure}[b]{0.16\textwidth}
    \centering
    \includegraphics[width=\textwidth, height=0.94\linewidth]{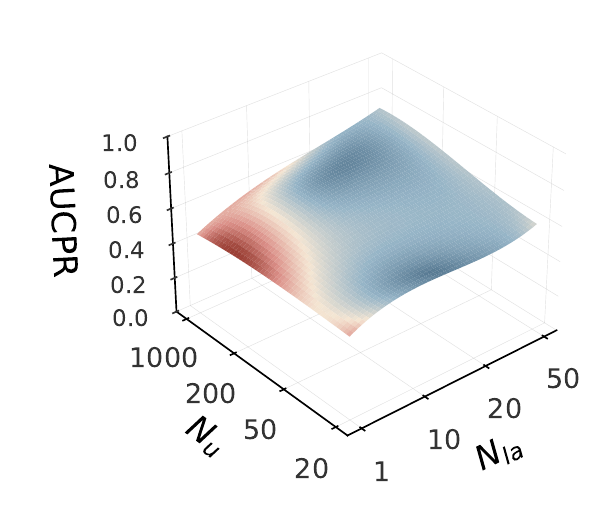}
    \caption{TabR-S}
    \label{fig:3d_unlabe_TabR_S}
  \end{subfigure}
  \hfill
  \begin{subfigure}[b]{0.16\textwidth}
    \centering
    \includegraphics[width=\textwidth, height=0.94\linewidth]{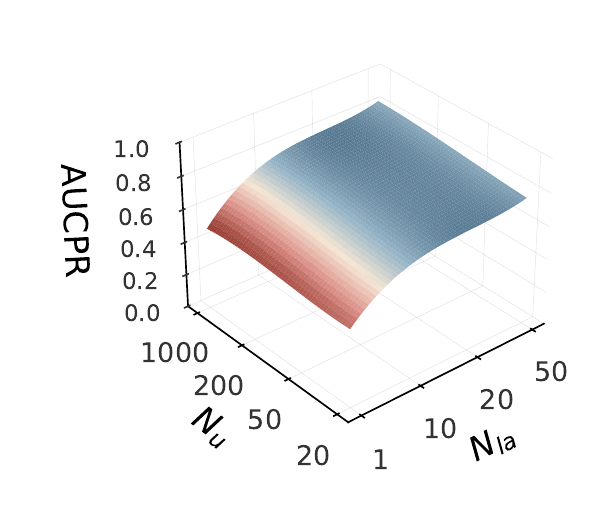}
    \caption{TabPFN}
    \label{fig:3d_unlabe_TabPFN}
  \end{subfigure}
  \hfill
  \begin{subfigure}[b]{0.16\textwidth}
    \centering
    \includegraphics[width=\textwidth, height=0.94\linewidth]{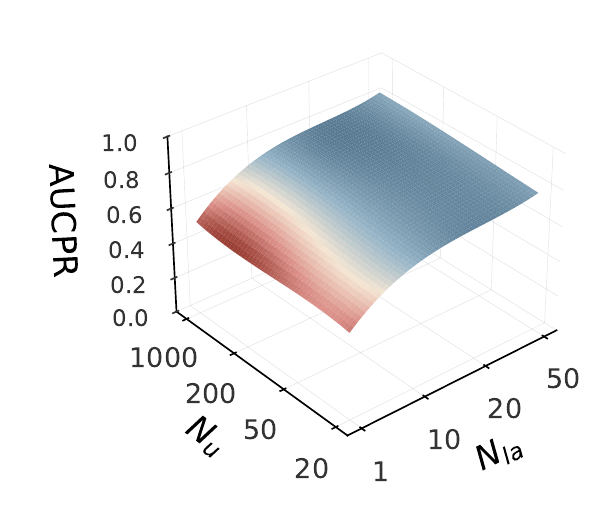}
    \caption{LimiX}
    \label{fig:3d_unlabe_LimiX}
  \end{subfigure}

  % --- Row 4 --- (Centered, extra model FTTransformer - commented out)
  % \begin{subfigure}[b]{0.16\textwidth}
  %   \centering
  %   \includegraphics[width=\textwidth, height=0.94\linewidth]{figures/Unlabel_exp/Classical/appendix_figures_split/FTTransformer.pdf}
  %   \caption{FTTransformer}
  %   \label{fig:3d_unlabe_FTTransformer}
  % \end{subfigure}
  \vspace{-5pt}
  \caption{3D surfaces visualizing AUCPR results of different models on tabular datasets under varying labeled ($\nla$) and unlabeled ($\numu$) data amounts.}
  \vspace{5pt}
  \label{fig:3d_full_unlabe}
\end{figure*}

\begin{figure*}[p]
  \centering

  % --- Row 1 ---
  \begin{subfigure}[b]{0.16\textwidth}
    \centering
    \includegraphics[width=\textwidth, height=0.94\linewidth]{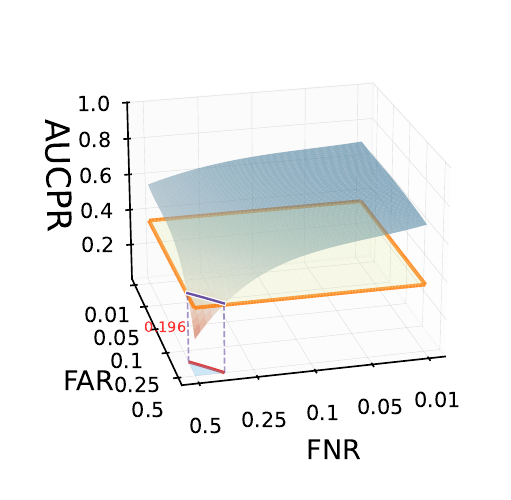}
    \caption{DevNet}
    \label{fig:3d_inacc_DevNet}
  \end{subfigure}
  \hfill
  \begin{subfigure}[b]{0.16\textwidth}
    \centering
    \includegraphics[width=\textwidth, height=0.94\linewidth]{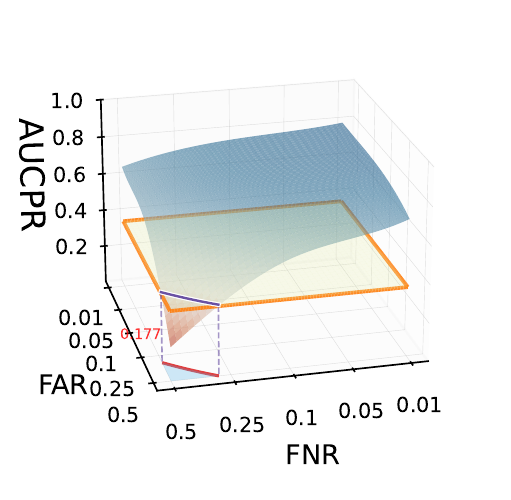}
    \caption{DeepSAD}
    \label{fig:3d_inacc_DeepSAD}
  \end{subfigure}
  \hfill
  \begin{subfigure}[b]{0.16\textwidth}
    \centering
    \includegraphics[width=\textwidth, height=0.94\linewidth]{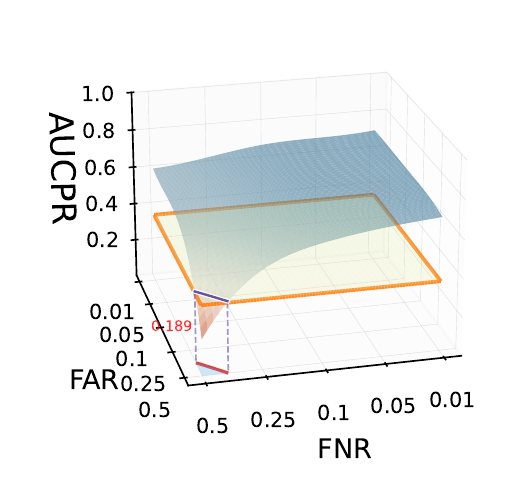}
    \caption{PReNet}
    \label{fig:3d_inacc_PReNet}
  \end{subfigure}
  \hfill
  \begin{subfigure}[b]{0.16\textwidth}
    \centering
    \includegraphics[width=\textwidth, height=0.94\linewidth]{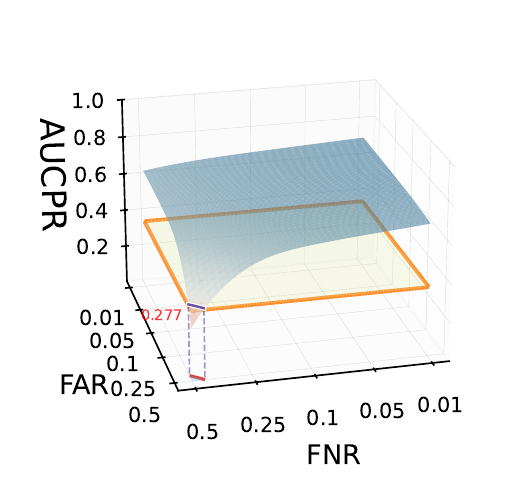}
    \caption{REPEN}
    \label{fig:3d_inacc_REPEN}
  \end{subfigure}
  \hfill
  \begin{subfigure}[b]{0.16\textwidth}
    \centering
    \includegraphics[width=\textwidth, height=0.94\linewidth]{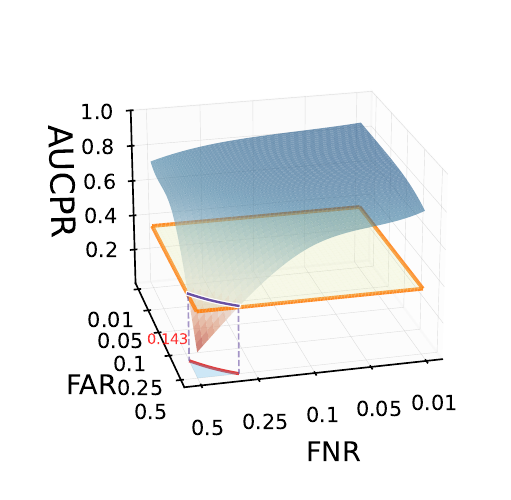}
    \caption{XGBOD}
    \label{fig:3d_inacc_XGBOD}
  \end{subfigure}
  \hfill
  \begin{subfigure}[b]{0.16\textwidth}
    \centering
    \includegraphics[width=\textwidth, height=0.94\linewidth]{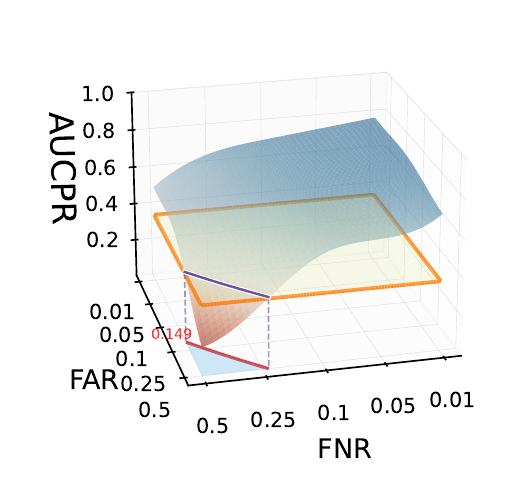}
    \caption{RoSAS}
    \label{fig:3d_inacc_RoSAS}
  \end{subfigure}

  \vspace{4mm}

  % --- Row 2 ---
  \begin{subfigure}[b]{0.16\textwidth}
    \centering
    \includegraphics[width=\textwidth, height=0.94\linewidth]{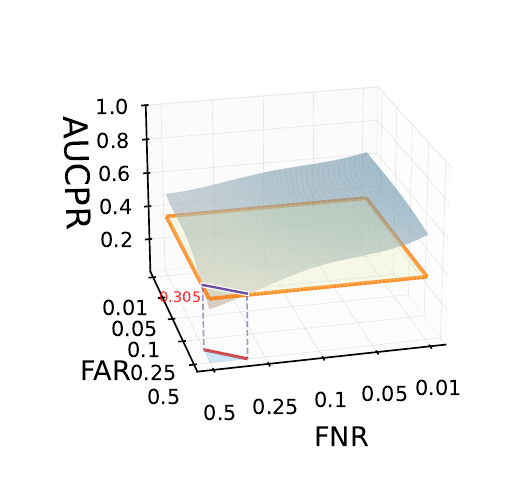}
    \caption{Dual-MGAN}
    \label{fig:3d_inacc_Dual_MGAN}
  \end{subfigure}
  \hfill
  \begin{subfigure}[b]{0.16\textwidth}
    \centering
    \includegraphics[width=\textwidth, height=0.94\linewidth]{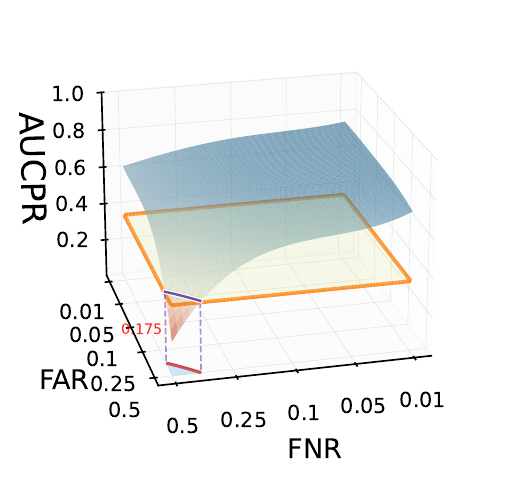}
    \caption{FEAWAD}
    \label{fig:3d_inacc_FEAWAD}
  \end{subfigure}
  \hfill
  \begin{subfigure}[b]{0.16\textwidth}
    \centering
    \includegraphics[width=\textwidth, height=0.94\linewidth]{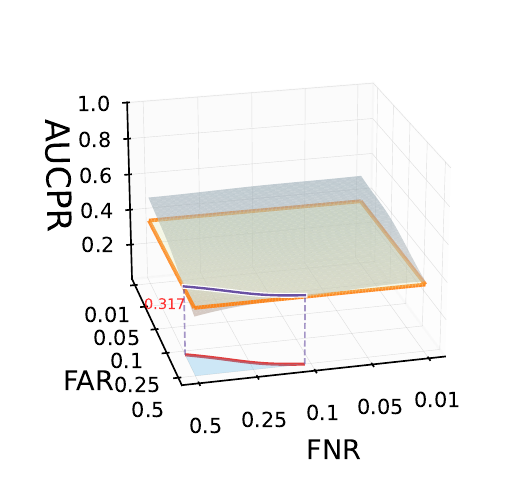}
    \caption{AnoDDAE}
    \label{fig:3d_inacc_AnoDDAE}
  \end{subfigure}
  \hfill
  \begin{subfigure}[b]{0.16\textwidth}
    \centering
    \includegraphics[width=\textwidth, height=0.94\linewidth]{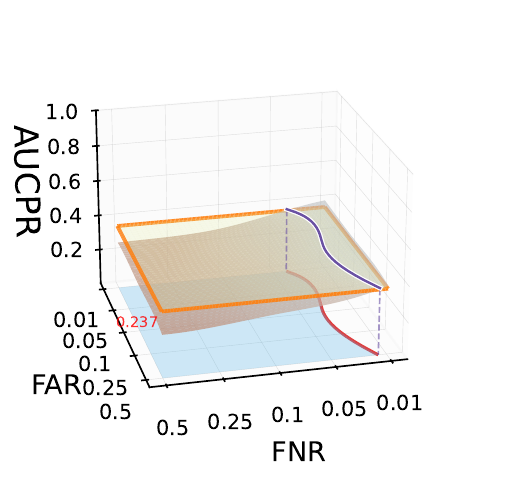}
    \caption{SOEL-NTL}
    \label{fig:3d_inacc_SOEL_NTL}
  \end{subfigure}
  \hfill
  \begin{subfigure}[b]{0.16\textwidth}
    \centering
    \includegraphics[width=\textwidth, height=0.94\linewidth]{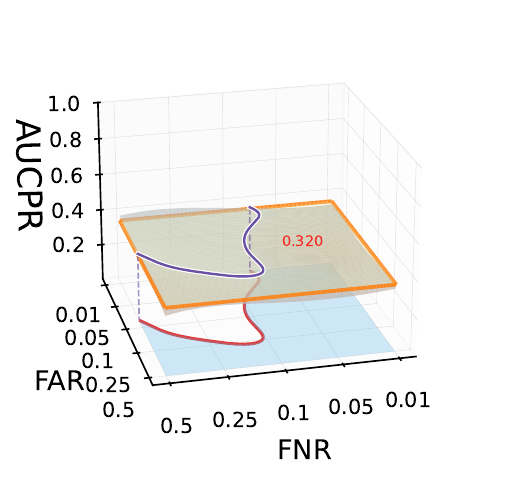}
    \caption{AA-BiGAN}
    \label{fig:3d_inacc_AA_BiGAN}
  \end{subfigure}
  \hfill
  \begin{subfigure}[b]{0.16\textwidth}
    \centering
    \includegraphics[width=\textwidth, height=0.94\linewidth]{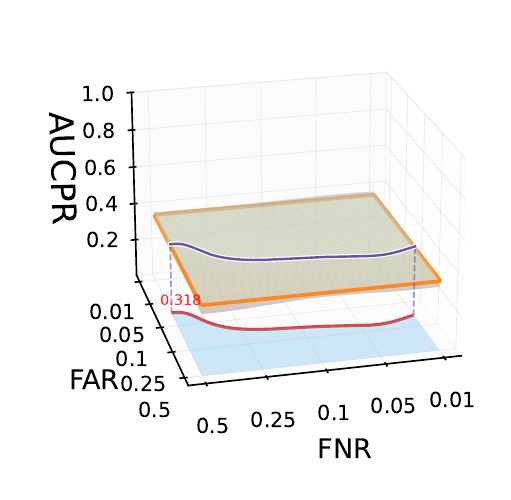}
    \caption{GANomaly}
    \label{fig:3d_inacc_GANomaly}
  \end{subfigure}

  \vspace{4mm}

  % --- Row 3 ---
  \begin{subfigure}[b]{0.16\textwidth}
    \centering
    \includegraphics[width=\textwidth, height=0.94\linewidth]{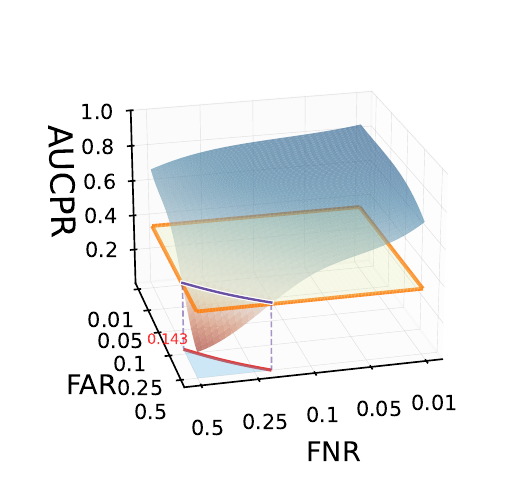}
    \caption{XGBoost}
    \label{fig:3d_inacc_XGBoost}
  \end{subfigure}
  \hfill
  \begin{subfigure}[b]{0.16\textwidth}
    \centering
    \includegraphics[width=\textwidth, height=0.94\linewidth]{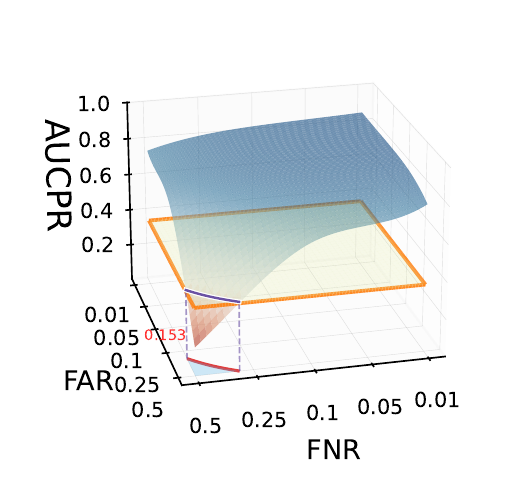}
    \caption{CatBoost}
    \label{fig:3d_inacc_CatBoost}
  \end{subfigure}
  \hfill
  \begin{subfigure}[b]{0.16\textwidth}
    \centering
    \includegraphics[width=\textwidth, height=0.94\linewidth]{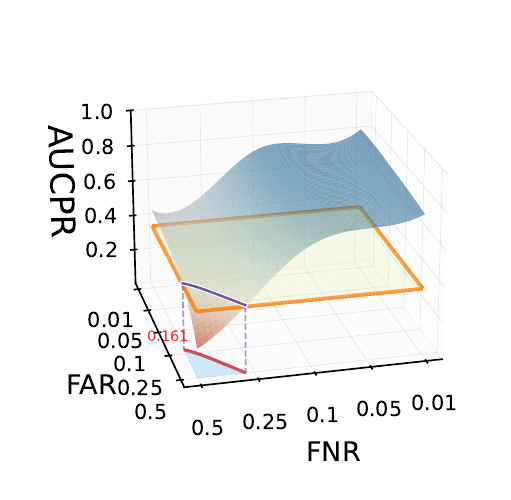}
    \caption{TabM}
    \label{fig:3d_inacc_TabM}
  \end{subfigure}
  \hfill
  \begin{subfigure}[b]{0.16\textwidth}
    \centering
    \includegraphics[width=\textwidth, height=0.94\linewidth]{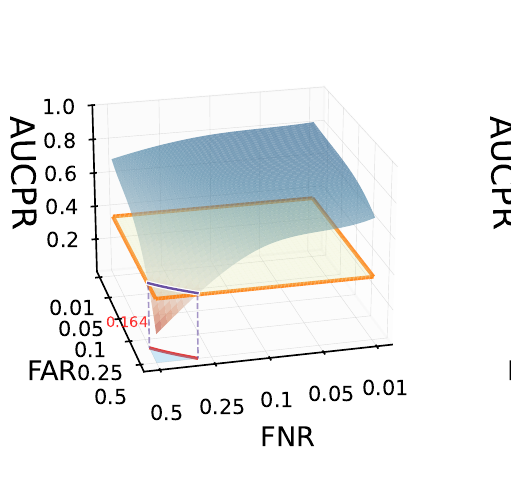}
    \caption{TabR-S}
    \label{fig:3d_inacc_TabR_S}
  \end{subfigure}
  \hfill
  \begin{subfigure}[b]{0.16\textwidth}
    \centering
    \includegraphics[width=\textwidth, height=0.94\linewidth]{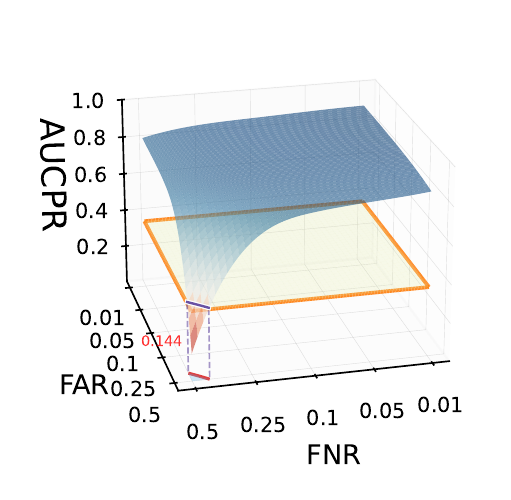}
    \caption{TabPFN}
    \label{fig:3d_inacc_TabPFN}
  \end{subfigure}
  \hfill
  \begin{subfigure}[b]{0.16\textwidth}
    \centering
    \includegraphics[width=\textwidth, height=0.94\linewidth]{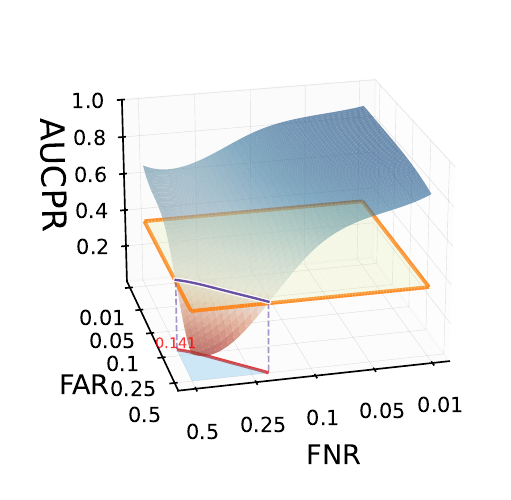}
    \caption{LimiX}
    \label{fig:3d_inacc_LimiX}
  \end{subfigure}

  \vspace{-5pt}
  \caption{3D surfaces visualizing AUCPR degradation of different models under varying flip normal ratios (FNR) and flip abnormal ratios (FAR) on tabular datasets.}
  \vspace{5pt}
  \label{fig:3d_inacc_overall}
\end{figure*}

%tabular Benchmark (rla/nla:AUCPR)
\begin{table*}[p]
  \centering
  \caption{The average $\pm$ standard deviation and ranking of AUCPR under different $\rla$ (=1\%, 5\%, 10\%, 25\%, 50\%, 100\%) settings on tabular datasets.}
  \label{tab:supp_incomp_aucpr_rla_cls}
  \resizebox{\textwidth}{!}{
\begin{tabular}{lllcccccc}
      \toprule
      \multirow{2}{*}{\textbf{Type}} & \multirow{2}{*}{\textbf{Model}} & \multirow{2}{*}{\textbf{Key Mechanism}} & \multicolumn{6}{c}{\textbf{Label Ratio ($\rla$)}} \\
      \cmidrule(lr){4-9}
      & & & 1\% & 5\% & 10\% & 25\% & 50\% & 100\% \\
      \midrule
      % === Semi-supervised Block ===
      \multirow{12}{*}{Semi-supervised}
      & XGBOD & Repr. Learning
      & 0.481 {\tiny $\pm$ 0.257} (9) & 0.623 {\tiny $\pm$ 0.261} (7) & 0.688 {\tiny $\pm$ 0.257} (5) & 0.767 {\tiny $\pm$ 0.248} (4) & 0.816 {\tiny $\pm$ 0.239} (4) & 0.864 {\tiny $\pm$ 0.219} (4) \\
      & DeepSAD & Repr. Learning
      & 0.379 {\tiny $\pm$ 0.251} (13) & 0.497 {\tiny $\pm$ 0.276} (14) & 0.568 {\tiny $\pm$ 0.283} (15) & 0.670 {\tiny $\pm$ 0.278} (14) & 0.752 {\tiny $\pm$ 0.263} (11) & 0.813 {\tiny $\pm$ 0.243} (10) \\
      & REPEN & Repr. Learning
      & 0.408 {\tiny $\pm$ 0.259} (11) & 0.538 {\tiny $\pm$ 0.275} (11) & 0.602 {\tiny $\pm$ 0.276} (12) & 0.686 {\tiny $\pm$ 0.278} (13) & 0.739 {\tiny $\pm$ 0.272} (12) & 0.779 {\tiny $\pm$ 0.265} (12) \\
      & AA-BiGAN & GAN-based
      & 0.308 {\tiny $\pm$ 0.267} (17) & 0.306 {\tiny $\pm$ 0.263} (18) & 0.310 {\tiny $\pm$ 0.269} (18) & 0.306 {\tiny $\pm$ 0.267} (19) & 0.321 {\tiny $\pm$ 0.280} (19) & 0.351 {\tiny $\pm$ 0.286} (19) \\
      & GANomaly & GAN-based
      & 0.314 {\tiny $\pm$ 0.261} (16) & 0.315 {\tiny $\pm$ 0.255} (17) & 0.316 {\tiny $\pm$ 0.260} (17) & 0.319 {\tiny $\pm$ 0.254} (18) & 0.335 {\tiny $\pm$ 0.264} (18) & 0.382 {\tiny $\pm$ 0.291} (18) \\
      & DevNet & Score Learning
      & 0.555 {\tiny $\pm$ 0.314} (2) & 0.635 {\tiny $\pm$ 0.304} (6) & 0.666 {\tiny $\pm$ 0.302} (8) & 0.695 {\tiny $\pm$ 0.293} (10) & 0.713 {\tiny $\pm$ 0.287} (14) & 0.722 {\tiny $\pm$ 0.284} (14) \\
\bottomrule
  \end{tabular}}
\end{table*}

\begin{table*}
  \centering
  \caption{The average $\pm$ standard deviation and ranking of AUCPR under different $\nla$ (=1, 3, 5, 10, 15, 20, 50) settings on tabular datasets.}
  \label{tab:supp_incomp_aucpr_nla_cls}
  \resizebox{\textwidth}{!}{
    \begin{tabular}{llccccccc}
      \toprule
      \multirow{2}{*}{\textbf{Type}} & \multirow{2}{*}{\textbf{Model}} & \multicolumn{7}{c}{\textbf{Label Count ($\nla$)}} \\
      \cmidrule(lr){3-9}
      & & 1 & 3 & 5 & 10 & 15 & 20 & 50 \\
      \midrule
      % === Semi-supervised Block ===
      \multirow{12}{*}{Semi-supervised}
      & XGBOD
      & 0.385 {\tiny $\pm$ 0.244} (11) & 0.518 {\tiny $\pm$ 0.255} (10) & 0.580 {\tiny $\pm$ 0.258} (8) & 0.657 {\tiny $\pm$ 0.260} (5) & 0.697 {\tiny $\pm$ 0.260} (4) & 0.723 {\tiny $\pm$ 0.253} (4) & 0.791 {\tiny $\pm$ 0.241} (4) \\
      & DeepSAD
      & 0.332 {\tiny $\pm$ 0.223} (15) & 0.423 {\tiny $\pm$ 0.255} (14) & 0.474 {\tiny $\pm$ 0.266} (15) & 0.556 {\tiny $\pm$ 0.283} (14) & 0.604 {\tiny $\pm$ 0.290} (14) & 0.637 {\tiny $\pm$ 0.287} (13) & 0.715 {\tiny $\pm$ 0.270} (10) \\
      & REPEN
      & 0.353 {\tiny $\pm$ 0.235} (13) & 0.482 {\tiny $\pm$ 0.261} (12) & 0.537 {\tiny $\pm$ 0.274} (12) & 0.607 {\tiny $\pm$ 0.286} (11) & 0.643 {\tiny $\pm$ 0.291} (10) & 0.667 {\tiny $\pm$ 0.290} (10) & 0.714 {\tiny $\pm$ 0.280} (11) \\
      & AA-BiGAN
      & 0.322 {\tiny $\pm$ 0.274} (16) & 0.284 {\tiny $\pm$ 0.244} (19) & 0.282 {\tiny $\pm$ 0.245} (19) & 0.282 {\tiny $\pm$ 0.249} (19) & 0.299 {\tiny $\pm$ 0.259} (19) & 0.288 {\tiny $\pm$ 0.249} (19) & 0.331 {\tiny $\pm$ 0.274} (19) \\
      & GANomaly
      & 0.310 {\tiny $\pm$ 0.257} (18) & 0.313 {\tiny $\pm$ 0.257} (17) & 0.314 {\tiny $\pm$ 0.258} (17) & 0.319 {\tiny $\pm$ 0.258} (18) & 0.318 {\tiny $\pm$ 0.258} (18) & 0.323 {\tiny $\pm$ 0.265} (18) & 0.341 {\tiny $\pm$ 0.275} (18) \\
      & DevNet
      & \textbf{0.517 {\tiny $\pm$ 0.315} (1)} & 0.576 {\tiny $\pm$ 0.317} (5) & 0.610 {\tiny $\pm$ 0.311} (4) & 0.648 {\tiny $\pm$ 0.304} (7) & 0.666 {\tiny $\pm$ 0.302} (9) & 0.676 {\tiny $\pm$ 0.296} (8) & 0.697 {\tiny $\pm$ 0.291} (14) \\
\bottomrule
  \end{tabular}}
\end{table*}

%CV_by_ViT (rla/nla:AUCPR)
\begin{table*}
  \centering
  \caption{The average $\pm$ standard deviation and ranking of AUCPR under different $\rla$ (=1\%, 5\%, 10\%, 25\%, 50\%, 100\%) settings on image datasets (preprocessed by Vision Transformer).}
  \label{tab:supp_incomp_aucpr_rla_cv}
  \resizebox{\textwidth}{!}{
    \begin{tabular}{lllcccccc}
      \toprule
      \multirow{2}{*}{\textbf{Type}} & \multirow{2}{*}{\textbf{Model}} & \multirow{2}{*}{\textbf{Key Mechanism}} & \multicolumn{6}{c}{\textbf{Label Ratio ($\rla$)}} \\
      \cmidrule(lr){4-9}
      & & & 1\% & 5\% & 10\% & 25\% & 50\% & 100\% \\
      \midrule
      % === Semi-supervised Block ===
      \multirow{12}{*}{Semi-supervised}
      & XGBOD & Repr. Learning
      & 0.388 {\tiny $\pm$ 0.242} (9) & 0.571 {\tiny $\pm$ 0.301} (9) & 0.633 {\tiny $\pm$ 0.307} (9) & 0.711 {\tiny $\pm$ 0.311} (9) & 0.759 {\tiny $\pm$ 0.307} (9) & 0.798 {\tiny $\pm$ 0.301} (10) \\
      & DeepSAD & Repr. Learning
      & 0.268 {\tiny $\pm$ 0.168} (18) & 0.395 {\tiny $\pm$ 0.226} (14) & 0.474 {\tiny $\pm$ 0.249} (14) & 0.606 {\tiny $\pm$ 0.278} (13) & 0.722 {\tiny $\pm$ 0.291} (13) & 0.826 {\tiny $\pm$ 0.280} (3) \\
      & REPEN & Repr. Learning
      & 0.288 {\tiny $\pm$ 0.198} (13) & 0.481 {\tiny $\pm$ 0.264} (11) & 0.592 {\tiny $\pm$ 0.292} (11) & 0.711 {\tiny $\pm$ 0.307} (10) & 0.773 {\tiny $\pm$ 0.305} (6) & 0.816 {\tiny $\pm$ 0.298} (5) \\
      & AA-BiGAN & GAN-based
      & 0.332 {\tiny $\pm$ 0.246} (10) & 0.318 {\tiny $\pm$ 0.220} (16) & 0.320 {\tiny $\pm$ 0.227} (16) & 0.321 {\tiny $\pm$ 0.230} (17) & 0.330 {\tiny $\pm$ 0.236} (17) & 0.366 {\tiny $\pm$ 0.272} (17) \\
      & GANomaly & GAN-based
      & 0.277 {\tiny $\pm$ 0.282} (17) & 0.279 {\tiny $\pm$ 0.282} (18) & 0.279 {\tiny $\pm$ 0.280} (18) & 0.283 {\tiny $\pm$ 0.283} (18) & 0.285 {\tiny $\pm$ 0.285} (18) & 0.311 {\tiny $\pm$ 0.295} (18) \\
      & DevNet & Score Learning
      & 0.459 {\tiny $\pm$ 0.296} (5) & 0.629 {\tiny $\pm$ 0.307} (4) & 0.694 {\tiny $\pm$ 0.312} (2) & 0.756 {\tiny $\pm$ 0.308} (2) & 0.782 {\tiny $\pm$ 0.299} (4) & 0.798 {\tiny $\pm$ 0.291} (9) \\
\bottomrule
  \end{tabular}}
\end{table*}

\begin{table*}
  \centering
  \caption{The average $\pm$ standard deviation and ranking of AUCPR under different $\nla$ (=1, 3, 5, 10, 15, 20, 50) settings on image datasets (preprocessed by Vision Transformer).}
  \label{tab:supp_incomp_aucpr_nla_cv}
  \resizebox{\textwidth}{!}{
    \begin{tabular}{llccccccc}
      \toprule
      \multirow{2}{*}{\textbf{Type}} & \multirow{2}{*}{\textbf{Model}} & \multicolumn{7}{c}{\textbf{Label Count ($\nla$)}} \\
      \cmidrule(lr){3-9}
      & & 1 & 3 & 5 & 10 & 15 & 20 & 50 \\
      \midrule
      % === Semi-supervised Block ===
      \multirow{12}{*}{Semi-supervised}
      & XGBOD
      & 0.300 {\tiny $\pm$ 0.198} (12) & 0.404 {\tiny $\pm$ 0.248} (9) & 0.471 {\tiny $\pm$ 0.271} (9) & 0.557 {\tiny $\pm$ 0.297} (9) & 0.598 {\tiny $\pm$ 0.305} (9) & 0.631 {\tiny $\pm$ 0.308} (8) & 0.710 {\tiny $\pm$ 0.316} (8) \\
      & DeepSAD
      & 0.217 {\tiny $\pm$ 0.147} (16) & 0.269 {\tiny $\pm$ 0.168} (18) & 0.302 {\tiny $\pm$ 0.180} (17) & 0.376 {\tiny $\pm$ 0.213} (14) & 0.423 {\tiny $\pm$ 0.232} (15) & 0.455 {\tiny $\pm$ 0.239} (14) & 0.599 {\tiny $\pm$ 0.279} (13) \\
      & REPEN
      & 0.213 {\tiny $\pm$ 0.175} (17) & 0.298 {\tiny $\pm$ 0.203} (15) & 0.350 {\tiny $\pm$ 0.215} (12) & 0.447 {\tiny $\pm$ 0.243} (12) & 0.517 {\tiny $\pm$ 0.265} (12) & 0.564 {\tiny $\pm$ 0.277} (11) & 0.707 {\tiny $\pm$ 0.308} (10) \\
      & AA-BiGAN
      & nan {\tiny $\pm$ nan} (18) & 0.315 {\tiny $\pm$ 0.228} (13) & 0.327 {\tiny $\pm$ 0.231} (13) & 0.324 {\tiny $\pm$ 0.227} (16) & 0.324 {\tiny $\pm$ 0.217} (17) & 0.322 {\tiny $\pm$ 0.228} (16) & 0.345 {\tiny $\pm$ 0.237} (16) \\
      & GANomaly
      & 0.282 {\tiny $\pm$ 0.284} (14) & 0.275 {\tiny $\pm$ 0.281} (17) & 0.278 {\tiny $\pm$ 0.281} (18) & 0.278 {\tiny $\pm$ 0.282} (18) & 0.284 {\tiny $\pm$ 0.283} (18) & 0.286 {\tiny $\pm$ 0.286} (17) & 0.280 {\tiny $\pm$ 0.281} (17) \\
      & DevNet
      & 0.333 {\tiny $\pm$ 0.249} (7) & 0.467 {\tiny $\pm$ 0.288} (4) & 0.541 {\tiny $\pm$ 0.299} (4) & 0.625 {\tiny $\pm$ 0.309} (3) & \textbf{0.670 {\tiny $\pm$ 0.315} (1)} & 0.694 {\tiny $\pm$ 0.314} (2) & 0.756 {\tiny $\pm$ 0.314} (2) \\
\bottomrule
  \end{tabular}}
\end{table*}

%NLP_by_RoBERTa (rla/nla:AUCPR)
\begin{table*}
  \centering
  \caption{The average $\pm$ standard deviation and ranking of AUCPR under different $\nla$ (=1, 3, 5, 10, 15, 20, 50) settings on text datasets (preprocessed by RoBERTa).}
  \label{tab:supp_incomp_aucpr_nla_nlp}
  \resizebox{\textwidth}{!}{
    \begin{tabular}{llccccccc}
      \toprule
      \multirow{2}{*}{\textbf{Type}} & \multirow{2}{*}{\textbf{Model}} & \multicolumn{7}{c}{\textbf{Label Count ($\nla$)}} \\
      \cmidrule(lr){3-9}
      & & 1 & 3 & 5 & 10 & 15 & 20 & 50 \\
      \midrule
      % === Semi-supervised Block ===
      \multirow{12}{*}{Semi-supervised}
      & XGBOD
      & 0.132 {\tiny $\pm$ 0.073} (9) & 0.180 {\tiny $\pm$ 0.105} (9) & 0.221 {\tiny $\pm$ 0.125} (8) & 0.282 {\tiny $\pm$ 0.147} (9) & 0.334 {\tiny $\pm$ 0.166} (9) & 0.376 {\tiny $\pm$ 0.178} (9) & 0.535 {\tiny $\pm$ 0.183} (9) \\
      & DeepSAD
      & 0.088 {\tiny $\pm$ 0.038} (18) & 0.112 {\tiny $\pm$ 0.064} (17) & 0.130 {\tiny $\pm$ 0.074} (16) & 0.171 {\tiny $\pm$ 0.092} (16) & 0.214 {\tiny $\pm$ 0.128} (14) & 0.247 {\tiny $\pm$ 0.146} (14) & 0.436 {\tiny $\pm$ 0.217} (13) \\
      & REPEN
      & 0.108 {\tiny $\pm$ 0.057} (15) & 0.146 {\tiny $\pm$ 0.080} (13) & 0.172 {\tiny $\pm$ 0.088} (13) & 0.231 {\tiny $\pm$ 0.124} (12) & 0.301 {\tiny $\pm$ 0.158} (11) & 0.353 {\tiny $\pm$ 0.173} (11) & 0.565 {\tiny $\pm$ 0.193} (8) \\
      & AA-BiGAN
      & 0.065 {\tiny $\pm$ 0.025} (19) & 0.077 {\tiny $\pm$ 0.040} (19) & 0.076 {\tiny $\pm$ 0.048} (19) & 0.076 {\tiny $\pm$ 0.051} (19) & 0.085 {\tiny $\pm$ 0.079} (19) & 0.091 {\tiny $\pm$ 0.085} (19) & 0.117 {\tiny $\pm$ 0.093} (18) \\
      & GANomaly
      & 0.109 {\tiny $\pm$ 0.054} (14) & 0.111 {\tiny $\pm$ 0.054} (18) & 0.110 {\tiny $\pm$ 0.052} (18) & 0.111 {\tiny $\pm$ 0.055} (17) & 0.111 {\tiny $\pm$ 0.056} (17) & 0.106 {\tiny $\pm$ 0.051} (17) & 0.107 {\tiny $\pm$ 0.056} (19) \\
      & DevNet
      & 0.171 {\tiny $\pm$ 0.114} (5) & \textbf{0.306 {\tiny $\pm$ 0.184} (1)} & 0.352 {\tiny $\pm$ 0.193} (3) & 0.436 {\tiny $\pm$ 0.201} (2) & 0.501 {\tiny $\pm$ 0.199} (2) & \textbf{0.544 {\tiny $\pm$ 0.193} (1)} & 0.645 {\tiny $\pm$ 0.173} (2) \\
\bottomrule
  \end{tabular}}
\end{table*}

\begin{table*}
  \centering
  \caption{The average $\pm$ standard deviation and ranking of AUCPR under different $\rla$ (=1\%, 5\%, 10\%, 25\%, 50\%, 100\%) settings on text datasets (preprocessed by RoBERTa).}
  \label{tab:supp_incomp_aucpr_rla_nlp}
  \resizebox{\textwidth}{!}{
    \begin{tabular}{lllcccccc}
      \toprule
      \multirow{2}{*}{\textbf{Type}} & \multirow{2}{*}{\textbf{Model}} & \multirow{2}{*}{\textbf{Key Mechanism}} & \multicolumn{6}{c}{\textbf{Label Ratio ($\rla$)}} \\
      \cmidrule(lr){4-9}
      & & & 1\% & 5\% & 10\% & 25\% & 50\% & 100\% \\
      \midrule
      % === Semi-supervised Block ===
      \multirow{12}{*}{Semi-supervised}
      & XGBOD & Repr. Learning
      & 0.166 {\tiny $\pm$ 0.087} (9) & 0.284 {\tiny $\pm$ 0.147} (8) & 0.366 {\tiny $\pm$ 0.172} (8) & 0.485 {\tiny $\pm$ 0.184} (10) & 0.571 {\tiny $\pm$ 0.174} (9) & 0.669 {\tiny $\pm$ 0.165} (11) \\
      & DeepSAD & Repr. Learning
      & 0.105 {\tiny $\pm$ 0.046} (16) & 0.171 {\tiny $\pm$ 0.100} (15) & 0.225 {\tiny $\pm$ 0.125} (14) & 0.344 {\tiny $\pm$ 0.151} (14) & 0.503 {\tiny $\pm$ 0.159} (13) & 0.726 {\tiny $\pm$ 0.146} (4) \\
      & REPEN & Repr. Learning
      & 0.134 {\tiny $\pm$ 0.058} (13) & 0.248 {\tiny $\pm$ 0.132} (11) & 0.344 {\tiny $\pm$ 0.186} (11) & 0.499 {\tiny $\pm$ 0.203} (8) & 0.618 {\tiny $\pm$ 0.187} (7) & 0.741 {\tiny $\pm$ 0.141} (3) \\
      & AA-BiGAN & GAN-based
      & 0.072 {\tiny $\pm$ 0.023} (19) & 0.072 {\tiny $\pm$ 0.021} (19) & 0.070 {\tiny $\pm$ 0.021} (19) & 0.070 {\tiny $\pm$ 0.020} (19) & 0.071 {\tiny $\pm$ 0.021} (19) & 0.073 {\tiny $\pm$ 0.026} (19) \\
      & GANomaly & GAN-based
      & 0.110 {\tiny $\pm$ 0.055} (15) & 0.113 {\tiny $\pm$ 0.056} (16) & 0.107 {\tiny $\pm$ 0.050} (16) & 0.112 {\tiny $\pm$ 0.055} (17) & 0.114 {\tiny $\pm$ 0.056} (18) & 0.136 {\tiny $\pm$ 0.081} (18) \\
      & DevNet & Score Learning
      & 0.272 {\tiny $\pm$ 0.172} (2) & \textbf{0.430 {\tiny $\pm$ 0.219} (1)} & 0.512 {\tiny $\pm$ 0.219} (2) & 0.607 {\tiny $\pm$ 0.190} (3) & 0.661 {\tiny $\pm$ 0.159} (3) & 0.705 {\tiny $\pm$ 0.152} (6) \\
\bottomrule
  \end{tabular}}
\end{table*}

%---------------------------------------------------
%---------------------------------------------------

%tabular Benchmark (rla/nla:AUCROC)

\begin{table*}
  \centering
  \caption{The average $\pm$ standard deviation and ranking of AUCROC under different $\rla$ (=1\%, 5\%, 10\%, 25\%, 50\%, 100\%) settings on tabular datasets.}
  \label{tab:classical_aucroc_refined}
  \resizebox{\textwidth}{!}{
    \begin{tabular}{lllcccccc}
      \toprule
      \multirow{2}{*}{\textbf{Type}} & \multirow{2}{*}{\textbf{Model}} & \multirow{2}{*}{\textbf{Key Mechanism}} & \multicolumn{6}{c}{\textbf{Label Ratio ($\gamma_L$)}} \\
      \cmidrule(lr){4-9}
      & & & 1\% & 5\% & 10\% & 25\% & 50\% & 100\% \\
      \midrule
      % === Semi-supervised Block ===
      \multirow{12}{*}{Semi-supervised}
      & XGBOD & Repr. Learning
      & 0.807 {\tiny $\pm$ 0.154} (5) & 0.869 {\tiny $\pm$ 0.135} (4) & 0.903 {\tiny $\pm$ 0.116} (4) & 0.935 {\tiny $\pm$ 0.096} (4) & 0.954 {\tiny $\pm$ 0.072} (4) & 0.968 {\tiny $\pm$ 0.057} (4) \\
      & DeepSAD & Repr. Learning
      & 0.755 {\tiny $\pm$ 0.159} (10) & 0.818 {\tiny $\pm$ 0.144} (12) & 0.854 {\tiny $\pm$ 0.133} (13) & 0.900 {\tiny $\pm$ 0.117} (9) & 0.928 {\tiny $\pm$ 0.099} (7) & 0.949 {\tiny $\pm$ 0.083} (8) \\
      & REPEN & Repr. Learning
      & 0.755 {\tiny $\pm$ 0.157} (9) & 0.828 {\tiny $\pm$ 0.138} (11) & 0.865 {\tiny $\pm$ 0.131} (10) & 0.898 {\tiny $\pm$ 0.125} (10) & 0.916 {\tiny $\pm$ 0.118} (11) & 0.930 {\tiny $\pm$ 0.109} (12) \\
      & AABiGAN & GAN-based
      & 0.663 {\tiny $\pm$ 0.219} (16) & 0.654 {\tiny $\pm$ 0.229} (19) & 0.662 {\tiny $\pm$ 0.222} (19) & 0.675 {\tiny $\pm$ 0.226} (19) & 0.681 {\tiny $\pm$ 0.215} (19) & 0.698 {\tiny $\pm$ 0.225} (19) \\
      & GANomaly & GAN-based
      & 0.677 {\tiny $\pm$ 0.205} (15) & 0.683 {\tiny $\pm$ 0.199} (17) & 0.682 {\tiny $\pm$ 0.202} (17) & 0.686 {\tiny $\pm$ 0.197} (18) & 0.703 {\tiny $\pm$ 0.194} (18) & 0.729 {\tiny $\pm$ 0.194} (18) \\
      & DevNet & Score Learning
      & 0.802 {\tiny $\pm$ 0.174} (6) & 0.859 {\tiny $\pm$ 0.151} (5) & 0.881 {\tiny $\pm$ 0.138} (6) & 0.898 {\tiny $\pm$ 0.128} (11) & 0.911 {\tiny $\pm$ 0.106} (13) & 0.917 {\tiny $\pm$ 0.103} (14) \\
\bottomrule
  \end{tabular}}
\end{table*}

\begin{table*}
  \centering
  \caption{The average $\pm$ standard deviation and ranking of AUCROC under different $\nla$ (=1, 3, 5, 10, 15, 20, 50) settings on tabular datasets.}
  \label{tab:supp_incomp_aucroc_nla_cls}
  \resizebox{\textwidth}{!}{
    \begin{tabular}{llccccccc}
      \toprule
      \multirow{2}{*}{\textbf{Type}} & \multirow{2}{*}{\textbf{Model}} & \multicolumn{7}{c}{\textbf{Label Count ($\nla$)}} \\
      \cmidrule(lr){3-9}
      & & 1 & 3 & 5 & 10 & 15 & 20 & 50 \\
      \midrule
      % === Semi-supervised Block ===
      \multirow{12}{*}{Semi-supervised}
      & XGBOD
      & 0.737 {\tiny $\pm$ 0.179} (9) & 0.807 {\tiny $\pm$ 0.153} (6) & 0.840 {\tiny $\pm$ 0.143} (4) & 0.880 {\tiny $\pm$ 0.127} (4) & 0.900 {\tiny $\pm$ 0.114} (4) & 0.911 {\tiny $\pm$ 0.107} (4) & 0.939 {\tiny $\pm$ 0.087} (4) \\
      & DeepSAD
      & 0.726 {\tiny $\pm$ 0.161} (12) & 0.770 {\tiny $\pm$ 0.159} (11) & 0.794 {\tiny $\pm$ 0.154} (11) & 0.831 {\tiny $\pm$ 0.151} (12) & 0.849 {\tiny $\pm$ 0.146} (12) & 0.863 {\tiny $\pm$ 0.141} (12) & 0.901 {\tiny $\pm$ 0.120} (10) \\
      & REPEN
      & 0.714 {\tiny $\pm$ 0.164} (13) & 0.778 {\tiny $\pm$ 0.163} (10) & 0.807 {\tiny $\pm$ 0.160} (9) & 0.847 {\tiny $\pm$ 0.153} (9) & 0.868 {\tiny $\pm$ 0.144} (9) & 0.879 {\tiny $\pm$ 0.138} (8) & 0.900 {\tiny $\pm$ 0.129} (11) \\
      & AA-BiGAN
      & 0.690 {\tiny $\pm$ 0.209} (15) & 0.640 {\tiny $\pm$ 0.221} (19) & 0.648 {\tiny $\pm$ 0.216} (19) & 0.634 {\tiny $\pm$ 0.220} (19) & 0.661 {\tiny $\pm$ 0.215} (19) & 0.660 {\tiny $\pm$ 0.211} (19) & 0.680 {\tiny $\pm$ 0.221} (19) \\
      & GANomaly
      & 0.672 {\tiny $\pm$ 0.204} (17) & 0.682 {\tiny $\pm$ 0.200} (16) & 0.680 {\tiny $\pm$ 0.205} (17) & 0.687 {\tiny $\pm$ 0.199} (18) & 0.686 {\tiny $\pm$ 0.201} (18) & 0.682 {\tiny $\pm$ 0.206} (18) & 0.696 {\tiny $\pm$ 0.205} (18) \\
      & DevNet
      & 0.759 {\tiny $\pm$ 0.197} (5) & 0.809 {\tiny $\pm$ 0.176} (5) & 0.836 {\tiny $\pm$ 0.157} (6) & 0.864 {\tiny $\pm$ 0.140} (5) & 0.877 {\tiny $\pm$ 0.133} (6) & 0.885 {\tiny $\pm$ 0.126} (6) & 0.903 {\tiny $\pm$ 0.112} (8) \\
\bottomrule
  \end{tabular}}
\end{table*}

%CV_by_ViT (rla/nla:AUCROC)
\begin{table*}
  \centering
  \caption{The average $\pm$ standard deviation and ranking of AUCROC under different $\rla$ (=1\%, 5\%, 10\%, 25\%, 50\%, 100\%) settings on image datasets (preprocessed by Vision Transformer).}
  \label{tab:supp_incomp_aucroc_rla_cv}
  \resizebox{\textwidth}{!}{
    \begin{tabular}{lllcccccc}
      \toprule
      \multirow{2}{*}{\textbf{Type}} & \multirow{2}{*}{\textbf{Model}} & \multirow{2}{*}{\textbf{Key Mechanism}} & \multicolumn{6}{c}{\textbf{Label Ratio ($\rla$)}} \\
      \cmidrule(lr){4-9}
      & & & \textbf{1\%} & \textbf{5\%} & \textbf{10\%} & \textbf{25\%} & \textbf{50\%} & \textbf{100\%} \\
      \midrule
      % === Semi-supervised Block ===
      \multirow{12}{*}{Semi-supervised}
      & XGBOD & Repr. Learning
      & 0.752 {\tiny $\pm$ 0.168} (6) & 0.844 {\tiny $\pm$ 0.152} (6) & 0.873 {\tiny $\pm$ 0.144} (6) & 0.906 {\tiny $\pm$ 0.130} (7) & 0.925 {\tiny $\pm$ 0.118} (7) & 0.940 {\tiny $\pm$ 0.110} (6) \\
      & DeepSAD & Repr. Learning
      & 0.688 {\tiny $\pm$ 0.139} (16) & 0.750 {\tiny $\pm$ 0.148} (14) & 0.786 {\tiny $\pm$ 0.143} (14) & 0.847 {\tiny $\pm$ 0.135} (14) & 0.900 {\tiny $\pm$ 0.125} (13) & 0.938 {\tiny $\pm$ 0.115} (9) \\
      & REPEN & Repr. Learning
      & 0.695 {\tiny $\pm$ 0.119} (15) & 0.799 {\tiny $\pm$ 0.136} (11) & 0.847 {\tiny $\pm$ 0.138} (11) & 0.895 {\tiny $\pm$ 0.133} (11) & 0.921 {\tiny $\pm$ 0.125} (10) & 0.939 {\tiny $\pm$ 0.116} (7) \\
      & AA-BiGAN & GAN-based
      & 0.745 {\tiny $\pm$ 0.172} (7) & 0.728 {\tiny $\pm$ 0.171} (16) & 0.736 {\tiny $\pm$ 0.169} (16) & 0.743 {\tiny $\pm$ 0.169} (16) & 0.742 {\tiny $\pm$ 0.169} (17) & 0.766 {\tiny $\pm$ 0.171} (17) \\
      & GANomaly & GAN-based
      & 0.718 {\tiny $\pm$ 0.169} (12) & 0.721 {\tiny $\pm$ 0.169} (17) & 0.723 {\tiny $\pm$ 0.166} (17) & 0.724 {\tiny $\pm$ 0.167} (18) & 0.734 {\tiny $\pm$ 0.161} (18) & 0.752 {\tiny $\pm$ 0.161} (18) \\
      & DevNet & Score Learning
      & 0.739 {\tiny $\pm$ 0.165} (9) & 0.846 {\tiny $\pm$ 0.147} (4) & 0.887 {\tiny $\pm$ 0.134} (3) & 0.920 {\tiny $\pm$ 0.120} (2) & 0.934 {\tiny $\pm$ 0.109} (4) & 0.941 {\tiny $\pm$ 0.104} (5) \\
\bottomrule
  \end{tabular}}
\end{table*}

\begin{table*}
  \centering
  \caption{The average $\pm$ standard deviation and ranking of AUCROC under different $\nla$ (=1, 3, 5, 10, 15, 20, 50) settings on image datasets (preprocessed by Vision Transformer).}
  \label{tab:supp_incomp_aucroc_nla_cv}
  \resizebox{\textwidth}{!}{
    \begin{tabular}{llccccccc}
      \toprule
      \multirow{2}{*}{\textbf{Type}} & \multirow{2}{*}{\textbf{Model}} & \multicolumn{7}{c}{\textbf{Label Count ($\nla$)}} \\
      \cmidrule(lr){3-9}
      & & 1 & 3 & 5 & 10 & 15 & 20 & 50 \\
      \midrule
      % === Semi-supervised Block ===
      \multirow{12}{*}{Semi-supervised}
      & XGBOD
      & 0.708 {\tiny $\pm$ 0.130} (8) & 0.767 {\tiny $\pm$ 0.151} (6) & 0.802 {\tiny $\pm$ 0.152} (5) & 0.841 {\tiny $\pm$ 0.152} (6) & 0.861 {\tiny $\pm$ 0.148} (6) & 0.872 {\tiny $\pm$ 0.145} (6) & 0.905 {\tiny $\pm$ 0.136} (7) \\
      & DeepSAD
      & 0.662 {\tiny $\pm$ 0.130} (12) & 0.690 {\tiny $\pm$ 0.135} (17) & 0.709 {\tiny $\pm$ 0.136} (17) & 0.743 {\tiny $\pm$ 0.142} (14) & 0.765 {\tiny $\pm$ 0.141} (14) & 0.782 {\tiny $\pm$ 0.140} (14) & 0.849 {\tiny $\pm$ 0.136} (14) \\
      & REPEN
      & 0.639 {\tiny $\pm$ 0.102} (15) & 0.705 {\tiny $\pm$ 0.117} (14) & 0.738 {\tiny $\pm$ 0.124} (11) & 0.788 {\tiny $\pm$ 0.129} (11) & 0.818 {\tiny $\pm$ 0.136} (11) & 0.838 {\tiny $\pm$ 0.136} (11) & 0.894 {\tiny $\pm$ 0.137} (11) \\
      & AA-BiGAN
      & nan {\tiny $\pm$ nan} (18) & 0.733 {\tiny $\pm$ 0.168} (11) & 0.735 {\tiny $\pm$ 0.167} (13) & 0.733 {\tiny $\pm$ 0.170} (16) & 0.732 {\tiny $\pm$ 0.168} (16) & 0.735 {\tiny $\pm$ 0.168} (16) & 0.752 {\tiny $\pm$ 0.169} (16) \\
      & GANomaly
      & 0.719 {\tiny $\pm$ 0.169} (7) & 0.715 {\tiny $\pm$ 0.171} (12) & 0.718 {\tiny $\pm$ 0.169} (14) & 0.718 {\tiny $\pm$ 0.170} (17) & 0.721 {\tiny $\pm$ 0.173} (17) & 0.717 {\tiny $\pm$ 0.179} (17) & 0.719 {\tiny $\pm$ 0.170} (17) \\
      & DevNet
      & 0.662 {\tiny $\pm$ 0.147} (13) & 0.746 {\tiny $\pm$ 0.160} (9) & 0.795 {\tiny $\pm$ 0.154} (8) & 0.845 {\tiny $\pm$ 0.149} (4) & 0.872 {\tiny $\pm$ 0.144} (3) & 0.888 {\tiny $\pm$ 0.137} (2) & 0.919 {\tiny $\pm$ 0.123} (2) \\
\bottomrule
  \end{tabular}}
\end{table*}

%NLP_by_RoBERTa (rla/nla:AUCROC)
\begin{table*}
  \centering
  \caption{The average $\pm$ standard deviation and ranking of AUCROC under different $\rla$ (=1\%, 5\%, 10\%, 25\%, 50\%, 100\%) settings on text datasets (preprocessed by RoBERTa).}
  \label{tab:supp_incomp_aucroc_rla_nlp}
  \resizebox{\textwidth}{!}{
    \begin{tabular}{lllcccccc}
      \toprule
      \multirow{2}{*}{\textbf{Type}} & \multirow{2}{*}{\textbf{Model}} & \multirow{2}{*}{\textbf{Key Mechanism}} & \multicolumn{6}{c}{\textbf{Label Ratio ($\rla$)}} \\
      \cmidrule(lr){4-9}
      & & & \textbf{1\%} & \textbf{5\%} & \textbf{10\%} & \textbf{25\%} & \textbf{50\%} & \textbf{100\%} \\
      \midrule
      % === Semi-supervised Block ===
      \multirow{11}{*}{Semi-supervised}
      & XGBOD & Repr. Learning
      & 0.698 {\tiny $\pm$ 0.088} (8) & 0.795 {\tiny $\pm$ 0.098} (7) & 0.844 {\tiny $\pm$ 0.086} (7) & 0.892 {\tiny $\pm$ 0.069} (8) & 0.917 {\tiny $\pm$ 0.053} (9) & 0.942 {\tiny $\pm$ 0.040} (9) \\
      & DeepSAD & Repr. Learning
      & 0.611 {\tiny $\pm$ 0.086} (16) & 0.671 {\tiny $\pm$ 0.098} (13) & 0.710 {\tiny $\pm$ 0.101} (14) & 0.785 {\tiny $\pm$ 0.085} (13) & 0.856 {\tiny $\pm$ 0.063} (13) & 0.925 {\tiny $\pm$ 0.051} (12) \\
      & REPEN & Repr. Learning
      & 0.633 {\tiny $\pm$ 0.072} (13) & 0.730 {\tiny $\pm$ 0.091} (12) & 0.785 {\tiny $\pm$ 0.104} (12) & 0.862 {\tiny $\pm$ 0.091} (12) & 0.911 {\tiny $\pm$ 0.066} (11) & 0.944 {\tiny $\pm$ 0.045} (8) \\
      & AA-BiGAN & GAN-based
      & 0.606 {\tiny $\pm$ 0.079} (17) & 0.592 {\tiny $\pm$ 0.078} (17) & 0.584 {\tiny $\pm$ 0.079} (17) & 0.583 {\tiny $\pm$ 0.079} (18) & 0.587 {\tiny $\pm$ 0.082} (18) & 0.588 {\tiny $\pm$ 0.089} (18) \\
      & GANomaly & GAN-based
      & 0.665 {\tiny $\pm$ 0.121} (11) & 0.667 {\tiny $\pm$ 0.126} (14) & 0.662 {\tiny $\pm$ 0.116} (15) & 0.670 {\tiny $\pm$ 0.122} (15) & 0.674 {\tiny $\pm$ 0.112} (16) & 0.700 {\tiny $\pm$ 0.119} (16) \\
      & DevNet & Score Learning
      & 0.745 {\tiny $\pm$ 0.122} (5) & 0.851 {\tiny $\pm$ 0.110} (4) & 0.891 {\tiny $\pm$ 0.089} (3) & \textbf{0.932 {\tiny $\pm$ 0.050} (1)} & 0.945 {\tiny $\pm$ 0.038} (2) & 0.956 {\tiny $\pm$ 0.030} (3) \\
\bottomrule
  \end{tabular}}
\end{table*}

\begin{table*}
  \centering
  \caption{The average $\pm$ standard deviation and ranking of AUCROC under different $\nla$ (=1, 3, 5, 10, 15, 20, 50) settings on text datasets (preprocessed by RoBERTa).}
  \label{tab:supp_incomp_aucroc_nla_nlp}
  \resizebox{\textwidth}{!}{
    \begin{tabular}{llccccccc}
      \toprule
      \multirow{2}{*}{\textbf{Type}} & \multirow{2}{*}{\textbf{Model}} & \multicolumn{7}{c}{\textbf{Label Count ($\nla$)}} \\
      \cmidrule(lr){3-9}
      & & 1 & 3 & 5 & 10 & 15 & 20 & 50 \\
      \midrule
      % === Semi-supervised Block ===
      \multirow{12}{*}{Semi-supervised}
      & XGBOD
      & 0.651 {\tiny $\pm$ 0.085} (11) & 0.719 {\tiny $\pm$ 0.079} (9) & 0.755 {\tiny $\pm$ 0.088} (8) & 0.802 {\tiny $\pm$ 0.074} (8) & 0.833 {\tiny $\pm$ 0.073} (9) & 0.854 {\tiny $\pm$ 0.070} (9) & 0.909 {\tiny $\pm$ 0.054} (8) \\
      & DeepSAD
      & 0.582 {\tiny $\pm$ 0.074} (18) & 0.612 {\tiny $\pm$ 0.078} (16) & 0.636 {\tiny $\pm$ 0.084} (16) & 0.682 {\tiny $\pm$ 0.083} (15) & 0.708 {\tiny $\pm$ 0.089} (15) & 0.738 {\tiny $\pm$ 0.090} (13) & 0.823 {\tiny $\pm$ 0.093} (13) \\
      & REPEN
      & 0.597 {\tiny $\pm$ 0.073} (17) & 0.640 {\tiny $\pm$ 0.069} (15) & 0.675 {\tiny $\pm$ 0.067} (13) & 0.730 {\tiny $\pm$ 0.076} (12) & 0.769 {\tiny $\pm$ 0.077} (12) & 0.801 {\tiny $\pm$ 0.080} (12) & 0.891 {\tiny $\pm$ 0.063} (11) \\
      & AA-BiGAN
      & 0.537 {\tiny $\pm$ 0.120} (19) & 0.577 {\tiny $\pm$ 0.099} (19) & 0.544 {\tiny $\pm$ 0.103} (18) & 0.530 {\tiny $\pm$ 0.118} (18) & 0.539 {\tiny $\pm$ 0.121} (18) & 0.548 {\tiny $\pm$ 0.125} (17) & 0.610 {\tiny $\pm$ 0.110} (17) \\
      & GANomaly
      & 0.665 {\tiny $\pm$ 0.126} (9) & 0.668 {\tiny $\pm$ 0.121} (14) & 0.666 {\tiny $\pm$ 0.124} (14) & 0.663 {\tiny $\pm$ 0.128} (16) & 0.667 {\tiny $\pm$ 0.128} (16) & 0.658 {\tiny $\pm$ 0.123} (16) & 0.656 {\tiny $\pm$ 0.127} (16) \\
      & DevNet
      & 0.655 {\tiny $\pm$ 0.094} (10) & 0.770 {\tiny $\pm$ 0.106} (5) & 0.819 {\tiny $\pm$ 0.093} (3) & \textbf{0.874 {\tiny $\pm$ 0.072} (1)} & \textbf{0.901 {\tiny $\pm$ 0.058} (1)} & \textbf{0.915 {\tiny $\pm$ 0.052} (1)} & \textbf{0.943 {\tiny $\pm$ 0.038} (1)} \\
\bottomrule
  \end{tabular}}
\end{table*}

\begin{table*}
  \centering
  \caption{The average AUCPR performance on individual datasets under the $\rla=1\%$ setting.}
  \label{tab:supp_incomp_alldata_rla_1}
  \resizebox{\textwidth}{!}{ % The average AUCPR performance on individual datasets under the $\rla=1\%$ setting.
    \begin{tabular}{lccccccccccccccccc}
      \toprule
      Dataset & XGBOD & DeepSAD & REPEN & AA-BiGAN & GANomaly & DevNet & FEAWAD & PReNet & RoSAS & Dual-MGAN & SOEL-NTL & XGBoost & CatBoost & TabM & TabR-S & TabPFN & LimiX \\
      \midrule
      10\_cover & 0.467 (7) & 0.211 (12) & 0.269 (10) & 0.091 (15) & 0.009 (17) & 0.644 (4) & 0.653 (3) & 0.661 (2) & 0.132 (14) & \textbf{0.727 (1)} & 0.184 (13) & 0.010 (16) & 0.409 (8) & 0.492 (5) & 0.231 (11) & 0.338 (9) & 0.478 (6) \\
      11\_donors & 0.542 (14) & 0.713 (11) & 0.860 (6) & 0.116 (16) & 0.097 (17) & \textbf{0.990 (1)} & 0.989 (2) & 0.969 (3) & 0.795 (8) & 0.750 (10) & 0.487 (15) & 0.569 (13) & 0.753 (9) & 0.905 (4) & 0.676 (12) & 0.884 (5) & 0.855 (7) \\
      12\_fault & 0.467 (12) & 0.510 (3) & 0.469 (10) & 0.373 (17) & 0.493 (7) & 0.506 (4) & 0.469 (11) & 0.489 (8) & 0.393 (16) & 0.476 (9) & \textbf{0.536 (1)} & 0.427 (15) & 0.500 (5) & 0.461 (13) & 0.429 (14) & 0.499 (6) & 0.515 (2) \\
      13\_fraud & 0.414 (7) & 0.213 (15) & 0.269 (13) & 0.402 (8) & \textbf{0.556 (1)} & 0.530 (4) & 0.537 (3) & 0.530 (5) & 0.235 (14) & 0.539 (2) & 0.380 (11) & 0.002 (17) & 0.393 (10) & 0.372 (12) & 0.194 (16) & 0.400 (9) & 0.478 (6) \\
      14\_glass & 0.393 (7) & 0.233 (12) & 0.307 (10) & 0.073 (16) & 0.171 (15) & 0.209 (13) & 0.611 (2) & 0.311 (9) & 0.567 (3) & 0.473 (5) & 0.288 (11) & 0.041 (17) & 0.380 (8) & \textbf{0.636 (1)} & 0.196 (14) & 0.534 (4) & 0.398 (6) \\
      15\_Hepatitis & 0.454 (8) & 0.359 (13) & 0.396 (12) & 0.295 (15) & 0.240 (16) & 0.527 (5) & 0.442 (10) & 0.480 (7) & 0.342 (14) & 0.448 (9) & 0.398 (11) & 0.160 (17) & 0.542 (4) & 0.543 (3) & \textbf{0.560 (1)} & 0.512 (6) & 0.556 (2) \\
      16\_http & 0.837 (9) & 0.757 (10) & 0.933 (6) & 0.280 (15) & 0.450 (14) & \textbf{1.000 (1)} & 1.000 (2) & 1.000 (3) & 0.918 (7) & 0.554 (13) & 0.041 (16) & 0.004 (17) & 0.860 (8) & 1.000 (4) & 0.575 (12) & 0.720 (11) & 0.947 (5) \\
\bottomrule
    \end{tabular}
  }
\end{table*}

% The average AUCPR performance on individual datasets under the $\rla=5\%$ setting.
\begin{table*}
  \centering
  \caption{The average AUCPR performance on individual datasets under the $\rla=5\%$ setting.}
  \label{tab:supp_incomp_alldata_rla_5}
  \resizebox{\textwidth}{!}{
    \begin{tabular}{lcccccccccccccccccc}
      \toprule
      Dataset & XGBOD & DeepSAD & REPEN & AA-BiGAN & GANomaly & DevNet & FEAWAD & PReNet & RoSAS & Dual-MGAN & SOEL-NTL & DDAE & XGBoost & CatBoost & TabM & TabR-S & TabPFN & LimiX \\
      \midrule
      10\_cover & 0.765 (10) & 0.678 (12) & 0.794 (8) & 0.310 (16) & 0.010 (18) & 0.892 (4) & 0.887 (5) & 0.900 (3) & 0.531 (13) & 0.917 (2) & 0.358 (15) & 0.100 (17) & 0.771 (9) & 0.686 (11) & 0.851 (7) & 0.461 (14) & 0.874 (6) & \textbf{0.932 (1)} \\
      11\_donors & 0.933 (11) & 0.979 (6) & 0.984 (4) & 0.057 (18) & 0.121 (15) & 0.995 (2) & \textbf{0.998 (1)} & 0.977 (7) & 0.946 (9) & 0.830 (13) & 0.068 (17) & 0.096 (16) & 0.774 (14) & 0.901 (12) & 0.989 (3) & 0.943 (10) & 0.982 (5) & 0.970 (8) \\
      12\_fault & 0.544 (5) & 0.522 (10) & 0.510 (13) & 0.439 (18) & 0.497 (14) & 0.542 (6) & 0.518 (11) & 0.557 (4) & 0.465 (17) & 0.537 (7) & 0.486 (15) & 0.536 (8) & 0.515 (12) & 0.562 (3) & 0.523 (9) & 0.477 (16) & \textbf{0.580 (1)} & 0.569 (2) \\
      13\_fraud & 0.414 (8) & 0.213 (16) & 0.269 (14) & 0.402 (9) & \textbf{0.556 (1)} & 0.530 (4) & 0.537 (3) & 0.530 (5) & 0.235 (15) & 0.539 (2) & 0.380 (12) & 0.491 (6) & 0.002 (18) & 0.393 (11) & 0.372 (13) & 0.194 (17) & 0.400 (10) & 0.478 (7) \\
      14\_glass & 0.706 (7) & 0.277 (14) & 0.460 (11) & 0.113 (17) & 0.172 (15) & 0.285 (13) & 0.724 (6) & 0.541 (9) & 0.752 (4) & 0.480 (10) & 0.337 (12) & 0.156 (16) & 0.041 (18) & 0.762 (3) & \textbf{0.827 (1)} & 0.553 (8) & 0.769 (2) & 0.728 (5) \\
      15\_Hepatitis & 0.709 (8) & 0.504 (14) & 0.739 (6) & 0.280 (16) & 0.247 (18) & 0.741 (5) & 0.695 (10) & 0.717 (7) & 0.547 (13) & 0.489 (15) & 0.549 (12) & 0.264 (17) & 0.744 (4) & 0.755 (3) & 0.701 (9) & 0.625 (11) & \textbf{0.772 (1)} & 0.755 (2) \\
\bottomrule
    \end{tabular}
  }
\end{table*}

\begin{table*}
  \centering
  \caption{The average AUCPR performance on individual datasets under the $\rla=10\%$ setting.
  }
  \label{tab:rank_aucpr_0.1}
  \resizebox{\textwidth}{!}{
    \begin{tabular}{lcccccccccccccccccc}
      \toprule
      Dataset & XGBOD & DeepSAD & REPEN & AA-BiGAN & GANomaly & DevNet & FEAWAD & PReNet & RoSAS & Dual-MGAN & SOEL-NTL & DDAE & XGBoost & CatBoost & TabM & TabR-S & TabPFN & LimiX \\
      \midrule
      10\_cover & 0.832 (9) & 0.794 (12) & 0.866 (8) & 0.190 (16) & 0.008 (18) & 0.929 (4) & 0.921 (5) & 0.938 (3) & 0.650 (13) & 0.941 (2) & 0.389 (15) & 0.095 (17) & 0.803 (11) & 0.811 (10) & 0.879 (7) & 0.614 (14) & 0.908 (6) & \textbf{0.967 (1)} \\
      11\_donors & 0.975 (11) & 0.998 (3) & 0.997 (5) & 0.059 (18) & 0.105 (16) & 0.999 (2) & \textbf{0.999 (1)} & 0.984 (10) & 0.993 (8) & 0.849 (14) & 0.136 (15) & 0.100 (17) & 0.886 (13) & 0.954 (12) & 0.996 (6) & 0.992 (9) & 0.997 (4) & 0.994 (7) \\
      12\_fault & 0.583 (4) & 0.557 (10) & 0.565 (8) & 0.448 (18) & 0.500 (16) & 0.560 (9) & 0.547 (11) & 0.573 (5) & 0.506 (15) & 0.571 (6) & 0.496 (17) & 0.547 (12) & 0.543 (13) & 0.595 (3) & 0.566 (7) & 0.515 (14) & 0.624 (2) & \textbf{0.639 (1)} \\
      13\_fraud & 0.391 (15) & 0.321 (16) & 0.437 (11) & 0.434 (12) & 0.556 (7) & 0.639 (2) & 0.626 (3) & \textbf{0.644 (1)} & 0.406 (14) & 0.572 (6) & 0.428 (13) & 0.484 (10) & 0.002 (18) & 0.604 (4) & 0.595 (5) & 0.303 (17) & 0.524 (9) & 0.537 (8) \\
      14\_glass & 0.765 (4) & 0.314 (14) & 0.478 (10) & 0.130 (18) & 0.172 (16) & 0.207 (15) & 0.741 (6) & 0.402 (12) & 0.731 (7) & 0.439 (11) & 0.392 (13) & 0.159 (17) & 0.637 (9) & \textbf{0.862 (1)} & 0.815 (3) & 0.665 (8) & 0.841 (2) & 0.764 (5) \\
      15\_Hepatitis & 0.875 (5) & 0.676 (15) & 0.862 (8) & 0.376 (16) & 0.249 (18) & 0.840 (9) & 0.864 (6) & 0.825 (11) & 0.775 (12) & 0.704 (14) & 0.732 (13) & 0.296 (17) & 0.892 (4) & 0.901 (3) & 0.862 (7) & 0.826 (10) & \textbf{0.909 (1)} & 0.907 (2) \\
\bottomrule
    \end{tabular}
  }
\end{table*}

\begin{table*}
  \centering
  \caption{The average AUCPR performance on individual datasets under the $\rla=25\%$ setting.}
  \label{tab:rank_aucpr_0.25}
  \resizebox{\textwidth}{!}{
    \begin{tabular}{lcccccccccccccccccc}
      \toprule
      Dataset & XGBOD & DeepSAD & REPEN & AA-BiGAN & GANomaly & DevNet & FEAWAD & PReNet & RoSAS & Dual-MGAN & SOEL-NTL & DDAE & XGBoost & CatBoost & TabM & TabR-S & TabPFN & LimiX \\
      \midrule
      10\_cover & 0.890 (13) & 0.935 (7) & 0.950 (5) & 0.159 (15) & 0.009 (18) & 0.961 (2) & 0.955 (3) & 0.954 (4) & 0.898 (12) & 0.949 (6) & 0.108 (16) & 0.098 (17) & 0.901 (10) & 0.899 (11) & 0.933 (9) & 0.839 (14) & 0.934 (8) & \textbf{0.975 (1)} \\
      11\_donors & 0.988 (10) & \textbf{1.000 (1)} & 0.993 (8) & 0.201 (15) & 0.137 (16) & 0.999 (4) & 0.999 (3) & 0.983 (12) & 0.998 (7) & 0.858 (14) & 0.098 (18) & 0.105 (17) & 0.963 (13) & 0.988 (11) & 0.998 (6) & 0.992 (9) & 1.000 (2) & 0.998 (5) \\
      12\_fault & 0.617 (4) & 0.584 (12) & 0.606 (6) & 0.437 (18) & 0.508 (16) & 0.558 (15) & 0.600 (8) & 0.597 (9) & 0.590 (11) & 0.559 (14) & 0.494 (17) & 0.561 (13) & 0.609 (5) & 0.652 (3) & 0.601 (7) & 0.590 (10) & 0.689 (2) & \textbf{0.728 (1)} \\
      13\_fraud & 0.633 (10) & 0.495 (14) & 0.674 (7) & 0.434 (17) & 0.559 (12) & \textbf{0.712 (1)} & 0.711 (2) & 0.658 (9) & 0.534 (13) & 0.679 (6) & 0.391 (18) & 0.494 (15) & 0.602 (11) & 0.683 (5) & 0.658 (8) & 0.447 (16) & 0.704 (3) & 0.695 (4) \\
      14\_glass & 0.864 (6) & 0.412 (12) & 0.623 (10) & 0.102 (18) & 0.188 (16) & 0.286 (14) & 0.858 (8) & 0.410 (13) & 0.903 (5) & 0.482 (11) & 0.263 (15) & 0.180 (17) & 0.859 (7) & \textbf{0.952 (1)} & 0.919 (4) & 0.836 (9) & 0.941 (2) & 0.921 (3) \\
      15\_Hepatitis & 0.983 (5) & 0.909 (12) & 0.965 (10) & 0.273 (17) & 0.264 (18) & 0.882 (13) & 0.974 (7) & 0.919 (11) & 0.965 (9) & 0.792 (14) & 0.364 (16) & 0.461 (15) & 0.988 (2) & 0.987 (4) & 0.978 (6) & 0.973 (8) & \textbf{0.990 (1)} & 0.988 (3) \\
\bottomrule
    \end{tabular}
  }
\end{table*}

\begin{table*}
  \centering
  \caption{The average AUCPR performance on individual datasets under the $\rla=50\%$ setting.
  }
  \label{tab:rank_aucpr_0.5}
  \resizebox{\textwidth}{!}{
    \begin{tabular}{lcccccccccccccccccc}
      \toprule
      Dataset & XGBOD & DeepSAD & REPEN & AA-BiGAN & GANomaly & DevNet & FEAWAD & PReNet & RoSAS & Dual-MGAN & SOEL-NTL & DDAE & XGBoost & CatBoost & TabM & TabR-S & TabPFN & LimiX \\
      \midrule
      10\_cover & 0.911 (14) & 0.956 (7) & 0.967 (2) & 0.258 (16) & 0.010 (18) & 0.960 (5) & 0.957 (6) & 0.954 (8) & 0.944 (10) & 0.951 (9) & 0.267 (15) & 0.120 (17) & 0.934 (11) & 0.923 (13) & 0.963 (4) & 0.924 (12) & 0.966 (3) & \textbf{0.985 (1)} \\
      11\_donors & 0.997 (11) & 1.000 (4) & 1.000 (2) & 0.135 (15) & 0.076 (18) & 0.999 (8) & 1.000 (6) & 0.984 (13) & 1.000 (5) & 0.838 (14) & 0.089 (17) & 0.133 (16) & 0.993 (12) & 0.997 (10) & 1.000 (7) & 0.999 (9) & 1.000 (3) & \textbf{1.000 (1)} \\
      12\_fault & 0.679 (4) & 0.634 (10) & 0.667 (6) & 0.479 (17) & 0.522 (16) & 0.588 (14) & 0.649 (8) & 0.628 (12) & 0.647 (9) & 0.617 (13) & 0.477 (18) & 0.584 (15) & 0.669 (5) & 0.729 (3) & 0.633 (11) & 0.657 (7) & 0.774 (2) & \textbf{0.787 (1)} \\
      13\_fraud & 0.670 (9) & 0.575 (14) & 0.680 (6) & 0.373 (18) & 0.567 (15) & 0.714 (2) & 0.707 (3) & 0.672 (8) & 0.622 (12) & 0.689 (4) & 0.451 (17) & 0.495 (16) & 0.673 (7) & \textbf{0.760 (1)} & 0.634 (10) & 0.618 (13) & 0.688 (5) & 0.625 (11) \\
      14\_glass & 0.969 (5) & 0.554 (12) & 0.693 (10) & 0.098 (18) & 0.191 (17) & 0.286 (15) & 0.909 (9) & 0.325 (14) & 0.973 (4) & 0.614 (11) & 0.354 (13) & 0.200 (16) & 0.958 (7) & \textbf{0.985 (1)} & 0.967 (6) & 0.917 (8) & 0.984 (3) & 0.985 (2) \\
      15\_Hepatitis & 0.994 (10) & 0.992 (11) & 1.000 (3) & 0.361 (17) & 0.331 (18) & 0.881 (13) & 1.000 (9) & 0.926 (12) & 1.000 (4) & 0.786 (14) & 0.405 (16) & 0.677 (15) & 1.000 (8) & \textbf{1.000 (1)} & 1.000 (5) & 1.000 (7) & 1.000 (6) & 1.000 (2) \\
\bottomrule
    \end{tabular}
  }
\end{table*}

\begin{table*}  % The average AUCROC performance on individual datasets under the $\rla=100\%$ setting.
  \centering
  \caption{The average AUCPR performance on individual datasets under the $\rla=100\%$ setting.}
  \label{tab:supp_incomp_alldata_rla_100}
  \resizebox{\textwidth}{!}{
    \begin{tabular}{lcccccccccccccccccc}
      \toprule
      Dataset & XGBOD & DeepSAD & REPEN & AA- BiGAN & GANomaly & DevNet & FEAWAD & PReNet & RoSAS & Dual-MGAN & SOEL-NTL & DDAE & XGBoost & CatBoost & TabM & TabR-S & TabPFN & LimiX \\
      \midrule
      10\_cover & 0.974 (10) & 0.984 (6) & 0.983 (7) & 0.133 (17) & 0.009 (18) & 0.966 (13) & 0.969 (12) & 0.972 (11) & 0.979 (9) & 0.959 (14) & 0.266 (15) & 0.145 (16) & 0.980 (8) & 0.985 (5) & 0.988 (4) & 0.991 (3) & 0.996 (2) & \textbf{0.997 (1)} \\
      11\_donors & 1.000 (10) & 1.000 (8) & 1.000 (3) & 0.064 (18) & 0.089 (17) & 0.999 (12) & 1.000 (9) & 0.989 (13) & 1.000 (4) & 0.942 (14) & 0.172 (16) & 0.545 (15) & 1.000 (11) & \textbf{1.000 (1)} & 1.000 (5) & 1.000 (7) & 1.000 (6) & 1.000 (2) \\
      12\_fault & 0.773 (6) & 0.729 (10) & 0.730 (9) & 0.518 (18) & 0.553 (16) & 0.622 (15) & 0.709 (11) & 0.664 (12) & 0.756 (7) & 0.645 (13) & 0.553 (17) & 0.637 (14) & 0.823 (4) & 0.834 (3) & 0.736 (8) & 0.781 (5) & 0.895 (2) & \textbf{0.895 (1)} \\
      13\_fraud & 0.722 (4) & 0.690 (10) & 0.688 (11) & 0.467 (18) & 0.572 (14) & 0.718 (6) & 0.721 (5) & 0.712 (7) & 0.542 (15) & 0.677 (12) & 0.508 (16) & 0.501 (17) & 0.697 (8) & 0.741 (2) & 0.737 (3) & 0.670 (13) & \textbf{0.742 (1)} & 0.693 (9) \\
      14\_glass & 0.990 (8) & 0.708 (10) & 0.666 (11) & 0.077 (18) & 0.194 (17) & 0.306 (15) & 0.875 (9) & 0.340 (14) & 1.000 (3) & 0.541 (12) & 0.440 (13) & 0.237 (16) & 1.000 (7) & \textbf{1.000 (1)} & 1.000 (4) & 1.000 (6) & 1.000 (5) & 1.000 (2) \\
      15\_Hepatitis & 0.996 (10) & 1.000 (2) & 1.000 (4) & 0.465 (17) & 0.423 (18) & 0.870 (14) & 0.996 (11) & 0.949 (13) & 1.000 (5) & 0.866 (15) & 0.609 (16) & 0.982 (12) & 1.000 (9) & \textbf{1.000 (1)} & 1.000 (6) & 1.000 (8) & 1.000 (7) & 1.000 (3) \\
\bottomrule
    \end{tabular}
  }
\end{table*}

%AUCROC
\begin{table*}  % % The average AUCROC performance on individual datasets under the $\rla=1\%$ setting.
  \centering
  \caption{The average AUCROC performance on individual datasets under the $\rla=1\%$ setting.}
  \label{tab:rank_aucroc_0.01}
  \resizebox{\textwidth}{!}{
    \begin{tabular}{lcccccccccccccccccc}
      \toprule
      Dataset & XGBOD & DeepSAD & REPEN & AA-BiGAN & GANomaly & DevNet & FEAWAD & PReNet & RoSAS & Dual-MGAN & SOEL-NTL & DDAE & XGBoost & CatBoost & TabM & TabR-S & TabPFN & LimiX \\
      \midrule
      10\_cover & 0.942 (8) & 0.835 (13) & 0.788 (15) & 0.829 (14) & 0.479 (18) & 0.985 (4) & 0.985 (3) & 0.986 (2) & 0.532 (16) & \textbf{0.993 (1)} & 0.865 (11) & 0.922 (10) & 0.500 (17) & 0.969 (6) & 0.930 (9) & 0.853 (12) & 0.982 (5) & 0.951 (7) \\
      11\_donors & 0.945 (11) & 0.963 (9) & 0.964 (8) & 0.591 (17) & 0.573 (18) & \textbf{0.999 (1)} & 0.999 (2) & 0.997 (3) & 0.893 (13) & 0.990 (6) & 0.890 (14) & 0.691 (16) & 0.938 (12) & 0.976 (7) & 0.953 (10) & 0.885 (15) & 0.995 (4) & 0.991 (5) \\
      12\_fault & 0.588 (12) & 0.685 (2) & 0.614 (9) & 0.478 (17) & 0.642 (5) & 0.624 (7) & 0.563 (15) & 0.603 (10) & 0.469 (18) & 0.586 (13) & 0.669 (3) & \textbf{0.705 (1)} & 0.542 (16) & 0.618 (8) & 0.602 (11) & 0.566 (14) & 0.651 (4) & 0.641 (6) \\
      13\_fraud & 0.924 (7) & 0.895 (11) & 0.840 (14) & 0.806 (15) & \textbf{0.964 (1)} & 0.910 (9) & 0.891 (12) & 0.928 (6) & 0.654 (17) & 0.933 (4) & 0.920 (8) & 0.947 (2) & 0.500 (18) & 0.943 (3) & 0.848 (13) & 0.740 (16) & 0.904 (10) & 0.930 (5) \\
      14\_glass & 0.822 (12) & 0.801 (13) & 0.873 (8) & 0.404 (18) & 0.719 (16) & 0.860 (10) & 0.872 (9) & 0.907 (6) & 0.744 (15) & 0.930 (3) & 0.911 (5) & 0.745 (14) & 0.500 (17) & 0.918 (4) & 0.905 (7) & 0.823 (11) & \textbf{0.950 (1)} & 0.934 (2) \\
      15\_Hepatitis & 0.688 (9) & 0.657 (12) & 0.654 (13) & 0.696 (7) & 0.605 (15) & 0.715 (6) & 0.578 (16) & 0.686 (10) & 0.532 (17) & 0.679 (11) & 0.689 (8) & 0.635 (14) & 0.500 (18) & \textbf{0.841 (1)} & 0.783 (4) & 0.824 (3) & 0.776 (5) & 0.838 (2) \\
\bottomrule
    \end{tabular}
  }
\end{table*}

\begin{table*}
  \centering
  \caption{The average AUCROC performance on individual datasets under the $\rla=5\%$ setting.
  }
  \label{tab:rank_aucroc_0.05}
  \resizebox{\textwidth}{!}{
    \begin{tabular}{lcccccccccccccccccc}
      \toprule
      Dataset & XGBOD & DeepSAD & REPEN & AA-BiGAN & GANomaly & DevNet & FEAWAD & PReNet & RoSAS & Dual-MGAN & SOEL-NTL & DDAE & XGBoost & CatBoost & TabM & TabR-S & TabPFN & LimiX \\
      \midrule
      10\_cover & 0.965 (10) & 0.934 (12) & 0.956 (11) & 0.917 (14) & 0.489 (18) & 0.996 (7) & 0.997 (4) & 0.996 (6) & 0.781 (17) & 0.998 (3) & 0.903 (15) & 0.923 (13) & 0.995 (8) & 0.980 (9) & 0.996 (5) & 0.881 (16) & 0.999 (2) & \textbf{0.999 (1)} \\
      11\_donors & 0.996 (8) & 0.999 (6) & 0.999 (5) & 0.388 (18) & 0.608 (16) & 1.000 (2) & \textbf{1.000 (1)} & 0.999 (4) & 0.957 (14) & 0.994 (10) & 0.492 (17) & 0.696 (15) & 0.977 (13) & 0.994 (9) & 0.993 (11) & 0.983 (12) & 0.999 (3) & 0.998 (7) \\
      12\_fault & 0.706 (4) & 0.702 (5) & 0.677 (9) & 0.576 (17) & 0.644 (14) & 0.689 (7) & 0.634 (15) & 0.678 (8) & 0.551 (18) & 0.668 (12) & 0.655 (13) & 0.699 (6) & 0.675 (11) & 0.720 (3) & 0.677 (10) & 0.618 (16) & 0.733 (2) & \textbf{0.734 (1)} \\
      13\_fraud & 0.924 (7) & 0.895 (11) & 0.840 (14) & 0.806 (15) & \textbf{0.964 (1)} & 0.910 (9) & 0.891 (12) & 0.928 (6) & 0.654 (17) & 0.933 (4) & 0.920 (8) & 0.947 (2) & 0.500 (18) & 0.943 (3) & 0.848 (13) & 0.740 (16) & 0.904 (10) & 0.930 (5) \\
      14\_glass & 0.957 (5) & 0.877 (13) & 0.913 (9) & 0.484 (18) & 0.721 (16) & 0.884 (11) & 0.878 (12) & 0.954 (6) & 0.820 (14) & 0.936 (8) & 0.911 (10) & 0.748 (15) & 0.500 (17) & 0.973 (2) & 0.963 (4) & 0.944 (7) & \textbf{0.976 (1)} & 0.971 (3) \\
      15\_Hepatitis & 0.836 (9) & 0.763 (13) & 0.848 (7) & 0.724 (14) & 0.614 (18) & 0.893 (3) & 0.787 (11) & 0.851 (6) & 0.659 (17) & 0.697 (15) & 0.779 (12) & 0.666 (16) & 0.859 (5) & \textbf{0.912 (1)} & 0.843 (8) & 0.794 (10) & 0.898 (2) & 0.887 (4) \\
\bottomrule
    \end{tabular}
  }
\end{table*}

\begin{table*}
  \centering
  \caption{The average AUCROC performance on individual datasets under the $\rla=10\%$ setting.
  }
  \label{tab:rank_aucroc_0.1}
  \resizebox{\textwidth}{!}{
    \begin{tabular}{lcccccccccccccccccc}
      \toprule
      Dataset & XGBOD & DeepSAD & REPEN & AA-BiGAN & GANomaly & DevNet & FEAWAD & PReNet & RoSAS & Dual-MGAN & SOEL-NTL & DDAE & XGBoost & CatBoost & TabM & TabR-S & TabPFN & LimiX \\
      \midrule
      10\_cover & 0.990 (9) & 0.964 (12) & 0.978 (11) & 0.887 (16) & 0.392 (18) & 0.999 (4) & 0.999 (6) & 0.999 (3) & 0.841 (17) & 0.999 (5) & 0.897 (15) & 0.921 (14) & 0.996 (8) & 0.990 (10) & 0.998 (7) & 0.924 (13) & 0.999 (2) & \textbf{1.000 (1)} \\
      11\_donors & 0.999 (8) & 1.000 (3) & 1.000 (6) & 0.456 (18) & 0.567 (17) & \textbf{1.000 (1)} & 1.000 (2) & 0.999 (7) & 0.994 (13) & 0.995 (12) & 0.607 (16) & 0.711 (15) & 0.991 (14) & 0.998 (9) & 0.997 (10) & 0.997 (11) & 1.000 (4) & 1.000 (5) \\
      12\_fault & 0.738 (4) & 0.729 (5) & 0.723 (6) & 0.586 (18) & 0.646 (16) & 0.713 (8) & 0.688 (13) & 0.710 (10) & 0.612 (17) & 0.708 (11) & 0.655 (15) & 0.711 (9) & 0.698 (12) & 0.741 (3) & 0.716 (7) & 0.662 (14) & 0.765 (2) & \textbf{0.770 (1)} \\
      13\_fraud & 0.903 (12) & 0.906 (10) & 0.899 (13) & 0.839 (16) & \textbf{0.964 (1)} & 0.913 (8) & 0.904 (11) & 0.918 (6) & 0.696 (17) & 0.933 (4) & 0.906 (9) & 0.947 (2) & 0.500 (18) & 0.945 (3) & 0.846 (15) & 0.856 (14) & 0.917 (7) & 0.926 (5) \\
      14\_glass & 0.974 (4) & 0.899 (13) & 0.928 (8) & 0.596 (18) & 0.721 (17) & 0.862 (14) & 0.905 (12) & 0.923 (9) & 0.780 (15) & 0.909 (11) & 0.929 (7) & 0.745 (16) & 0.917 (10) & \textbf{0.989 (1)} & 0.975 (3) & 0.959 (6) & 0.979 (2) & 0.972 (5) \\
      15\_Hepatitis & 0.937 (6) & 0.854 (14) & 0.937 (7) & 0.763 (16) & 0.615 (18) & 0.953 (3) & 0.885 (12) & 0.937 (8) & 0.798 (15) & 0.888 (11) & 0.855 (13) & 0.671 (17) & 0.949 (4) & \textbf{0.957 (1)} & 0.928 (9) & 0.910 (10) & 0.956 (2) & 0.946 (5) \\
\bottomrule
    \end{tabular}
  }
\end{table*}

\begin{table*}
  \centering
  \caption{The average AUCROC performance on individual datasets under the $\rla=25\%$ setting.
  }
  \label{tab:rank_aucroc_0.25}
  \resizebox{\textwidth}{!}{
    \begin{tabular}{lcccccccccccccccccc}
      \toprule
      Dataset & XGBOD & DeepSAD & REPEN & AA-BiGAN & GANomaly & DevNet & FEAWAD & PReNet & RoSAS & Dual-MGAN & SOEL-NTL & DDAE & XGBoost & CatBoost & TabM & TabR-S & TabPFN & LimiX \\
      \midrule
      10\_cover & 0.992 (12) & 0.994 (11) & 0.999 (9) & 0.863 (16) & 0.457 (18) & 0.999 (3) & 0.999 (6) & 0.999 (4) & 0.985 (14) & 0.999 (7) & 0.759 (17) & 0.920 (15) & 0.999 (8) & 0.999 (10) & 0.999 (5) & 0.991 (13) & 1.000 (2) & \textbf{1.000 (1)} \\
      11\_donors & 0.999 (7) & \textbf{1.000 (1)} & 1.000 (6) & 0.655 (16) & 0.591 (18) & 1.000 (5) & 1.000 (3) & 0.999 (9) & 0.998 (12) & 0.995 (14) & 0.642 (17) & 0.727 (15) & 0.998 (13) & 0.999 (8) & 0.998 (10) & 0.998 (11) & 1.000 (2) & 1.000 (4) \\
      12\_fault & 0.765 (4) & 0.753 (6) & 0.755 (5) & 0.595 (18) & 0.649 (16) & 0.726 (12) & 0.739 (10) & 0.745 (9) & 0.719 (13) & 0.700 (15) & 0.631 (17) & 0.716 (14) & 0.748 (8) & 0.787 (3) & 0.753 (7) & 0.732 (11) & 0.823 (2) & \textbf{0.839 (1)} \\
      13\_fraud & 0.943 (5) & 0.940 (7) & 0.912 (15) & 0.909 (18) & \textbf{0.964 (1)} & 0.932 (8) & 0.911 (17) & 0.932 (9) & 0.911 (16) & 0.951 (2) & 0.919 (14) & 0.948 (3) & 0.928 (12) & 0.947 (4) & 0.928 (13) & 0.931 (11) & 0.941 (6) & 0.931 (10) \\
      14\_glass & 0.990 (4) & 0.938 (11) & 0.968 (10) & 0.516 (18) & 0.723 (17) & 0.889 (14) & 0.975 (9) & 0.921 (13) & 0.983 (7) & 0.931 (12) & 0.858 (15) & 0.746 (16) & 0.983 (8) & \textbf{0.997 (1)} & 0.987 (6) & 0.989 (5) & 0.995 (2) & 0.993 (3) \\
      15\_Hepatitis & 0.991 (6) & 0.969 (13) & 0.982 (9) & 0.731 (16) & 0.634 (18) & 0.974 (12) & 0.986 (8) & 0.976 (11) & 0.980 (10) & 0.953 (14) & 0.648 (17) & 0.772 (15) & 0.995 (3) & 0.996 (2) & 0.991 (5) & 0.989 (7) & \textbf{0.996 (1)} & 0.994 (4) \\
\bottomrule
    \end{tabular}
  }
\end{table*}

\begin{table*}
  \centering
  \caption{The average AUCROC performance on individual datasets under the $\rla=50\%$ setting.
  }
  \label{tab:rank_aucroc_0.5}
  \resizebox{\textwidth}{!}{
    \begin{tabular}{lcccccccccccccccccc}
      \toprule
      Dataset & XGBOD & DeepSAD & REPEN & AA-BiGAN & GANomaly & DevNet & FEAWAD & PReNet & RoSAS & Dual-MGAN & SOEL-NTL & DDAE & XGBoost & CatBoost & TabM & TabR-S & TabPFN & LimiX \\
      \midrule
      10\_cover & 0.998 (12) & 0.999 (11) & 1.000 (4) & 0.933 (16) & 0.484 (18) & 0.999 (6) & 1.000 (5) & 0.999 (7) & 0.994 (14) & 0.999 (10) & 0.922 (17) & 0.933 (15) & 0.999 (9) & 0.999 (8) & 1.000 (3) & 0.998 (13) & 1.000 (2) & \textbf{1.000 (1)} \\
      11\_donors & 1.000 (11) & 1.000 (4) & 1.000 (2) & 0.562 (16) & 0.539 (18) & 1.000 (8) & 1.000 (6) & 0.999 (13) & 1.000 (5) & 0.994 (14) & 0.555 (17) & 0.788 (15) & 1.000 (12) & 1.000 (10) & 1.000 (7) & 1.000 (9) & 1.000 (3) & \textbf{1.000 (1)} \\
      12\_fault & 0.809 (4) & 0.788 (8) & 0.791 (7) & 0.617 (18) & 0.659 (16) & 0.739 (13) & 0.772 (10) & 0.757 (12) & 0.765 (11) & 0.735 (15) & 0.650 (17) & 0.735 (14) & 0.803 (5) & 0.836 (3) & 0.776 (9) & 0.793 (6) & 0.870 (2) & \textbf{0.883 (1)} \\
      13\_fraud & 0.948 (6) & 0.936 (12) & 0.947 (7) & 0.840 (18) & 0.965 (2) & 0.941 (9) & 0.892 (17) & 0.964 (3) & 0.894 (15) & \textbf{0.967 (1)} & 0.939 (11) & 0.949 (5) & 0.933 (14) & 0.962 (4) & 0.939 (10) & 0.894 (16) & 0.935 (13) & 0.941 (8) \\
      14\_glass & 0.998 (4) & 0.969 (11) & 0.979 (10) & 0.543 (18) & 0.727 (17) & 0.887 (15) & 0.988 (8) & 0.907 (13) & 0.986 (9) & 0.957 (12) & 0.895 (14) & 0.759 (16) & 0.996 (7) & \textbf{0.999 (1)} & 0.997 (5) & 0.997 (6) & 0.999 (3) & 0.999 (2) \\
      15\_Hepatitis & 0.999 (10) & 0.998 (11) & 1.000 (3) & 0.811 (16) & 0.700 (17) & 0.982 (13) & 1.000 (9) & 0.990 (12) & 1.000 (4) & 0.957 (14) & 0.685 (18) & 0.881 (15) & 1.000 (8) & \textbf{1.000 (1)} & 1.000 (5) & 1.000 (7) & 1.000 (6) & 1.000 (2) \\
\bottomrule
    \end{tabular}
  }
\end{table*}

\begin{table*}
  \centering
  \caption{The average AUCROC performance on individual datasets under the $\rla=100\%$ setting.
  }
  \label{tab:rank_aucroc_1.0}
  \resizebox{\textwidth}{!}{
    \begin{tabular}{lcccccccccccccccccc}
      \toprule
      Dataset & XGBOD & DeepSAD & REPEN & AA-BiGAN & GANomaly & DevNet & FEAWAD & PReNet & RoSAS & Dual-MGAN & SOEL-NTL & DDAE & XGBoost & CatBoost & TabM & TabR-S & TabPFN & LimiX \\
      \midrule
      10\_cover & 1.000 (10) & 1.000 (8) & 1.000 (7) & 0.862 (17) & 0.448 (18) & 1.000 (13) & 1.000 (11) & 1.000 (12) & 1.000 (4) & 0.999 (14) & 0.890 (16) & 0.940 (15) & 1.000 (9) & 1.000 (6) & 1.000 (5) & 1.000 (3) & 1.000 (2) & \textbf{1.000 (1)} \\
      11\_donors & 1.000 (10) & 1.000 (8) & 1.000 (3) & 0.535 (18) & 0.594 (17) & 1.000 (12) & 1.000 (9) & 0.999 (13) & 1.000 (4) & 0.998 (14) & 0.775 (16) & 0.974 (15) & 1.000 (11) & \textbf{1.000 (1)} & 1.000 (5) & 1.000 (7) & 1.000 (6) & 1.000 (2) \\
      12\_fault & 0.856 (6) & 0.831 (8) & 0.823 (10) & 0.667 (18) & 0.681 (17) & 0.759 (15) & 0.806 (11) & 0.775 (12) & 0.827 (9) & 0.767 (13) & 0.700 (16) & 0.764 (14) & 0.891 (4) & 0.897 (3) & 0.835 (7) & 0.860 (5) & 0.936 (2) & \textbf{0.937 (1)} \\
      13\_fraud & 0.974 (2) & 0.939 (12) & 0.962 (6) & 0.846 (18) & 0.964 (5) & 0.961 (7) & 0.931 (14) & 0.969 (3) & 0.932 (13) & 0.941 (11) & 0.931 (16) & 0.949 (10) & \textbf{0.980 (1)} & 0.967 (4) & 0.952 (8) & 0.912 (17) & 0.931 (15) & 0.950 (9) \\
      14\_glass & 1.000 (8) & 0.987 (10) & 0.977 (11) & 0.567 (18) & 0.732 (17) & 0.891 (15) & 0.988 (9) & 0.904 (13) & 1.000 (3) & 0.936 (12) & 0.904 (14) & 0.765 (16) & 1.000 (7) & \textbf{1.000 (1)} & 1.000 (4) & 1.000 (6) & 1.000 (5) & 1.000 (2) \\
      15\_Hepatitis & 0.999 (11) & 1.000 (2) & 1.000 (4) & 0.832 (16) & 0.777 (18) & 0.978 (14) & 0.999 (10) & 0.993 (13) & 1.000 (5) & 0.974 (15) & 0.800 (17) & 0.996 (12) & 1.000 (9) & \textbf{1.000 (1)} & 1.000 (6) & 1.000 (8) & 1.000 (7) & 1.000 (3) \\
\bottomrule
    \end{tabular}
  }
\end{table*}

\begin{table*}  % The average AUCPR performance on individual datasets under the $\nla=1$ setting.
  \centering
  \caption{The average AUCPR performance on individual datasets under the $\nla=1$ setting.}
  \label{tab:rank_aucpr_1}
  \resizebox{\textwidth}{!}{
    \begin{tabular}{lcccccccccccccccccc}
      \toprule
      Dataset & XGBOD & DeepSAD & REPEN & AA-BiGAN & GANomaly & DevNet & FEAWAD & PReNet & RoSAS & Dual-MGAN & SOEL-NTL & DDAE & XGBoost & CatBoost & TabM & TabR-S & TabPFN & LimiX \\
      \midrule
      10\_cover & 0.467 (7) & 0.211 (13) & 0.269 (11) & 0.091 (17) & 0.009 (18) & 0.642 (4) & 0.653 (3) & 0.661 (2) & 0.132 (14) & \textbf{0.727 (1)} & 0.093 (16) & 0.095 (15) & 0.533 (5) & 0.409 (9) & 0.461 (8) & 0.231 (12) & 0.338 (10) & 0.490 (6) \\
      11\_donors & 0.199 (13) & 0.324 (12) & 0.512 (7) & 0.140 (14) & 0.080 (17) & \textbf{0.872 (1)} & 0.822 (3) & 0.850 (2) & 0.496 (8) & 0.645 (4) & 0.084 (16) & 0.100 (15) & 0.408 (11) & 0.442 (10) & 0.578 (6) & 0.460 (9) & 0.062 (18) & 0.602 (5) \\
      12\_fault & 0.407 (14) & 0.496 (4) & 0.439 (9) & 0.436 (10) & 0.492 (5) & 0.425 (13) & 0.443 (8) & 0.430 (12) & 0.397 (16) & 0.434 (11) & 0.499 (3) & \textbf{0.545 (1)} & 0.399 (15) & 0.484 (7) & 0.382 (17) & 0.368 (18) & 0.526 (2) & 0.486 (6) \\
      13\_fraud & 0.414 (9) & 0.213 (17) & 0.269 (15) & 0.400 (10) & \textbf{0.556 (1)} & 0.530 (4) & 0.537 (3) & 0.530 (5) & 0.235 (16) & 0.539 (2) & 0.321 (14) & 0.491 (6) & 0.419 (8) & 0.393 (13) & 0.393 (12) & 0.194 (18) & 0.400 (11) & 0.473 (7) \\
      14\_glass & 0.393 (7) & 0.233 (13) & 0.307 (11) & 0.073 (18) & 0.171 (17) & 0.209 (14) & 0.611 (2) & 0.311 (10) & 0.567 (3) & 0.473 (5) & 0.263 (12) & 0.190 (16) & 0.357 (9) & 0.380 (8) & \textbf{0.636 (1)} & 0.196 (15) & 0.534 (4) & 0.399 (6) \\
      15\_Hepatitis & 0.406 (9) & 0.327 (15) & 0.436 (6) & 0.351 (13) & 0.241 (17) & 0.469 (5) & 0.360 (12) & 0.423 (7) & 0.328 (14) & 0.410 (8) & 0.218 (18) & 0.249 (16) & 0.393 (11) & 0.481 (4) & 0.492 (2) & \textbf{0.518 (1)} & 0.406 (10) & 0.484 (3) \\
\bottomrule
    \end{tabular}
  }
\end{table*}

\begin{table*}
  \centering
  \caption{The average of AUCPR performance on individual datasets under $N_{la}=3$ setting}
  \label{tab:rank_aucpr_3}
  \resizebox{\textwidth}{!}{
    \begin{tabular}{lcccccccccccccccccc}
      \toprule
      Dataset & XGBOD & DeepSAD & REPEN & AABiGAN & GANomaly & DevNet & FEAWAD & PReNet & RoSAS & DualMGAN & SOEL-NTL & AnoDDAE & XGB & CatB & TabMCls & TabR-S & TabPFN & LimiX \\
      \midrule
      10\_cover & 0.760 (8) & 0.621 (11) & 0.725 (10) & 0.192 (15) & 0.010 (18) & 0.888 (4) & 0.880 (5) & \textbf{0.909 (1)} & 0.435 (13) & 0.902 (2) & 0.099 (17) & 0.104 (16) & 0.753 (9) & 0.591 (12) & 0.822 (6) & 0.371 (14) & 0.815 (7) & 0.894 (3) \\
      11\_donors & 0.454 (14) & 0.567 (13) & 0.797 (6) & 0.088 (17) & 0.103 (15) & \textbf{0.972 (1)} & 0.945 (3) & 0.966 (2) & 0.662 (11) & 0.715 (7) & 0.054 (18) & 0.094 (16) & 0.622 (12) & 0.697 (10) & 0.829 (4) & 0.702 (9) & 0.711 (8) & 0.808 (5) \\
      12\_fault & 0.442 (12) & 0.503 (2) & 0.457 (9) & 0.364 (17) & 0.492 (6) & 0.449 (10) & 0.441 (13) & 0.448 (11) & 0.400 (16) & 0.472 (8) & 0.496 (5) & \textbf{0.534 (1)} & 0.418 (14) & 0.500 (4) & 0.405 (15) & 0.359 (18) & 0.501 (3) & 0.489 (7) \\
      13\_fraud & 0.479 (13) & 0.440 (16) & 0.612 (8) & 0.370 (18) & 0.560 (10) & 0.674 (4) & 0.697 (2) & 0.644 (6) & 0.479 (14) & 0.664 (5) & 0.437 (17) & 0.492 (12) & 0.518 (11) & 0.560 (9) & 0.624 (7) & 0.448 (15) & 0.695 (3) & \textbf{0.735 (1)} \\
      14\_glass & 0.767 (4) & 0.301 (13) & 0.508 (11) & 0.104 (18) & 0.172 (16) & 0.207 (15) & 0.730 (7) & 0.406 (12) & 0.737 (6) & 0.536 (10) & 0.258 (14) & 0.167 (17) & 0.627 (9) & \textbf{0.863 (1)} & 0.817 (3) & 0.667 (8) & 0.838 (2) & 0.753 (5) \\
      15\_Hepatitis & 0.498 (12) & 0.398 (14) & 0.564 (10) & 0.305 (15) & 0.242 (18) & 0.640 (3) & 0.579 (8) & 0.598 (7) & 0.404 (13) & 0.519 (11) & 0.293 (16) & 0.255 (17) & 0.579 (9) & 0.649 (2) & 0.618 (6) & 0.634 (4) & 0.630 (5) & \textbf{0.666 (1)} \\
\bottomrule
    \end{tabular}
  }
\end{table*}

\begin{table*}
  \centering
  \caption{The average of AUCPR performance on individual datasets under $N_{la}=5$ setting}
  \label{tab:rank_aucpr_5}
  \resizebox{\textwidth}{!}{
    \begin{tabular}{lcccccccccccccccccc}
      \toprule
      Dataset & XGBOD & DeepSAD & REPEN & AABiGAN & GANomaly & DevNet & FEAWAD & PReNet & RoSAS & DualMGAN & SOEL-NTL & AnoDDAE & XGB & CatB & TabMCls & TabR-S & TabPFN & LimiX \\
      \midrule
      10\_cover & 0.826 (9) & 0.738 (12) & 0.863 (8) & 0.289 (15) & 0.009 (18) & 0.909 (4) & 0.904 (5) & 0.903 (6) & 0.605 (13) & 0.919 (3) & 0.095 (17) & 0.097 (16) & 0.793 (10) & 0.740 (11) & 0.894 (7) & 0.576 (14) & 0.922 (2) & \textbf{0.954 (1)} \\
      11\_donors & 0.551 (14) & 0.753 (9) & 0.880 (6) & 0.113 (15) & 0.099 (16) & 0.986 (2) & \textbf{0.988 (1)} & 0.967 (3) & 0.810 (8) & 0.749 (11) & 0.076 (18) & 0.090 (17) & 0.661 (13) & 0.751 (10) & 0.909 (4) & 0.693 (12) & 0.894 (5) & 0.863 (7) \\
      12\_fault & 0.467 (13) & 0.510 (3) & 0.469 (11) & 0.373 (18) & 0.493 (8) & 0.506 (4) & 0.469 (12) & 0.489 (9) & 0.393 (17) & 0.476 (10) & 0.497 (7) & \textbf{0.542 (1)} & 0.441 (15) & 0.500 (5) & 0.461 (14) & 0.429 (16) & 0.499 (6) & 0.513 (2) \\
      13\_fraud & 0.634 (11) & 0.565 (14) & 0.679 (7) & 0.443 (17) & 0.561 (15) & \textbf{0.722 (1)} & 0.709 (2) & 0.676 (8) & 0.622 (12) & 0.685 (6) & 0.370 (18) & 0.493 (16) & 0.697 (4) & 0.708 (3) & 0.666 (9) & 0.590 (13) & 0.690 (5) & 0.645 (10) \\
      14\_glass & 0.811 (5) & 0.342 (13) & 0.518 (10) & 0.136 (18) & 0.188 (16) & 0.237 (15) & 0.722 (9) & 0.383 (11) & 0.810 (6) & 0.361 (12) & 0.255 (14) & 0.166 (17) & 0.750 (8) & \textbf{0.916 (1)} & 0.824 (3) & 0.797 (7) & 0.885 (2) & 0.822 (4) \\
      15\_Hepatitis & 0.667 (7) & 0.466 (14) & 0.665 (8) & 0.252 (17) & 0.246 (18) & \textbf{0.749 (1)} & 0.657 (10) & 0.702 (6) & 0.540 (12) & 0.489 (13) & 0.286 (15) & 0.276 (16) & 0.706 (5) & 0.728 (2) & 0.662 (9) & 0.605 (11) & 0.719 (3) & 0.712 (4) \\
\bottomrule
    \end{tabular}
  }
\end{table*}

\begin{table*}
  \centering
  \caption{The average of AUCPR performance on individual datasets under $N_{la}=10$ setting}
  \label{tab:rank_aucpr_10}
  \resizebox{\textwidth}{!}{
    \begin{tabular}{lcccccccccccccccccc}
      \toprule
      Dataset & XGBOD & DeepSAD & REPEN & AABiGAN & GANomaly & DevNet & FEAWAD & PReNet & RoSAS & DualMGAN & SOEL-NTL & AnoDDAE & XGB & CatB & TabMCls & TabR-S & TabPFN & LimiX \\
      \midrule
      10\_cover & 0.847 (10) & 0.847 (11) & 0.882 (7) & 0.170 (16) & 0.009 (18) & 0.943 (3) & 0.928 (5) & \textbf{0.950 (1)} & 0.753 (13) & 0.943 (2) & 0.253 (15) & 0.115 (17) & 0.866 (9) & 0.842 (12) & 0.868 (8) & 0.653 (14) & 0.903 (6) & 0.941 (4) \\
      11\_donors & 0.857 (10) & 0.882 (8) & 0.928 (5) & 0.115 (15) & 0.111 (16) & 0.958 (3) & \textbf{0.993 (1)} & 0.966 (2) & 0.876 (9) & 0.809 (13) & 0.076 (18) & 0.095 (17) & 0.677 (14) & 0.832 (12) & 0.947 (4) & 0.837 (11) & 0.923 (6) & 0.895 (7) \\
      12\_fault & 0.487 (12) & 0.507 (7) & 0.489 (11) & 0.390 (18) & 0.493 (10) & 0.497 (9) & 0.480 (13) & 0.506 (8) & 0.392 (17) & 0.519 (5) & 0.508 (6) & \textbf{0.536 (1)} & 0.463 (15) & 0.522 (4) & 0.476 (14) & 0.426 (16) & 0.531 (2) & 0.531 (3) \\
      13\_fraud & \textbf{0.736 (1)} & 0.682 (9) & 0.677 (11) & 0.384 (18) & 0.567 (13) & 0.722 (5) & 0.720 (6) & 0.723 (4) & 0.509 (14) & 0.680 (10) & 0.488 (17) & 0.497 (15) & 0.707 (7) & 0.727 (3) & 0.731 (2) & 0.494 (16) & 0.695 (8) & 0.650 (12) \\
      14\_glass & 0.896 (7) & 0.455 (11) & 0.635 (10) & 0.096 (18) & 0.188 (17) & 0.318 (15) & 0.851 (9) & 0.424 (12) & \textbf{0.966 (1)} & 0.392 (13) & 0.375 (14) & 0.199 (16) & 0.905 (6) & 0.962 (2) & 0.943 (5) & 0.861 (8) & 0.959 (3) & 0.950 (4) \\
      15\_Hepatitis & 0.851 (4) & 0.624 (13) & 0.809 (8) & 0.293 (17) & 0.248 (18) & 0.784 (10) & 0.826 (6) & 0.797 (9) & 0.747 (12) & 0.612 (14) & 0.332 (15) & 0.320 (16) & 0.844 (5) & 0.856 (3) & 0.822 (7) & 0.768 (11) & \textbf{0.880 (1)} & 0.867 (2) \\
\bottomrule
    \end{tabular}
  }
\end{table*}

\begin{table*}
  \centering  % The average AUCROC performance on individual datasets under the $\nla=1$ setting.
  \caption{The average AUCROC performance on individual datasets under the $\nla=1$ setting.
  }
  \label{tab:rank_aucroc_1}
  \resizebox{\textwidth}{!}{
    \begin{tabular}{lcccccccccccccccccc}
      \toprule
      Dataset & XGBOD & DeepSAD & REPEN & AA-BiGAN & GANomaly & DevNet & FEAWAD & PReNet & RoSAS & Dual-MGAN & SOEL-NTL & DDAE & XGBoost & CatBoost & TabM & TabR-S & TabPFN & LimiX \\
      \midrule
      10\_cover & 0.942 (9) & 0.835 (13) & 0.788 (15) & 0.829 (14) & 0.479 (18) & 0.984 (4) & 0.985 (3) & 0.986 (2) & 0.532 (17) & \textbf{0.993 (1)} & 0.759 (16) & 0.922 (10) & 0.972 (6) & 0.969 (7) & 0.919 (11) & 0.853 (12) & 0.982 (5) & 0.952 (8) \\
      11\_donors & 0.788 (12) & 0.846 (10) & 0.848 (9) & 0.779 (13) & 0.594 (16) & \textbf{0.988 (1)} & 0.920 (6) & 0.983 (2) & 0.697 (15) & 0.948 (4) & 0.580 (17) & 0.708 (14) & 0.863 (7) & 0.941 (5) & 0.843 (11) & 0.855 (8) & 0.534 (18) & 0.956 (3) \\
      12\_fault & 0.530 (14) & 0.673 (3) & 0.584 (9) & 0.590 (8) & 0.641 (5) & 0.530 (13) & 0.547 (10) & 0.535 (12) & 0.482 (17) & 0.543 (11) & 0.663 (4) & \textbf{0.705 (1)} & 0.515 (15) & 0.615 (7) & 0.482 (16) & 0.469 (18) & 0.690 (2) & 0.621 (6) \\
      13\_fraud & 0.924 (8) & 0.895 (11) & 0.840 (15) & 0.839 (16) & \textbf{0.964 (1)} & 0.910 (9) & 0.891 (12) & 0.928 (7) & 0.654 (18) & 0.933 (5) & 0.928 (6) & 0.947 (2) & 0.871 (13) & 0.943 (3) & 0.849 (14) & 0.740 (17) & 0.904 (10) & 0.940 (4) \\
      14\_glass & 0.822 (13) & 0.801 (14) & 0.873 (8) & 0.404 (18) & 0.719 (17) & 0.860 (11) & 0.872 (9) & 0.907 (5) & 0.744 (16) & 0.930 (3) & 0.863 (10) & 0.745 (15) & 0.876 (7) & 0.918 (4) & 0.905 (6) & 0.823 (12) & \textbf{0.950 (1)} & 0.934 (2) \\
      15\_Hepatitis & 0.664 (10) & 0.637 (13) & 0.677 (7) & 0.719 (5) & 0.604 (15) & 0.674 (8) & 0.560 (17) & 0.632 (14) & 0.568 (16) & 0.671 (9) & 0.534 (18) & 0.640 (12) & 0.655 (11) & \textbf{0.813 (1)} & 0.782 (3) & 0.786 (2) & 0.682 (6) & 0.778 (4) \\
\bottomrule
    \end{tabular}
  }
\end{table*}

\begin{table*}
  \centering
  \caption{The average of AUCROC performance on individual datasets under $N_{la}=3$ setting}
  \label{tab:rank_aucroc_3}
  \resizebox{\textwidth}{!}{
    \begin{tabular}{lcccccccccccccccccc}
      \toprule
      Dataset & XGBOD & DeepSAD & REPEN & AABiGAN & GANomaly & DevNet & FEAWAD & PReNet & RoSAS & DualMGAN & SOEL-NTL & AnoDDAE & XGB & CatB & TabMCls & TabR-S & TabPFN & LimiX \\
      \midrule
      10\_cover & 0.973 (8) & 0.930 (11) & 0.914 (13) & 0.896 (14) & 0.491 (18) & 0.996 (6) & 0.997 (4) & 0.996 (5) & 0.705 (17) & 0.997 (3) & 0.774 (16) & 0.926 (12) & 0.947 (10) & 0.972 (9) & 0.992 (7) & 0.895 (15) & 0.998 (2) & \textbf{0.999 (1)} \\
      11\_donors & 0.933 (11) & 0.938 (10) & 0.956 (9) & 0.647 (16) & 0.552 (17) & \textbf{0.998 (1)} & 0.977 (6) & 0.997 (2) & 0.783 (14) & 0.966 (7) & 0.464 (18) & 0.690 (15) & 0.961 (8) & 0.978 (5) & 0.932 (12) & 0.910 (13) & 0.982 (4) & 0.989 (3) \\
      12\_fault & 0.562 (12) & 0.680 (2) & 0.613 (7) & 0.467 (18) & 0.642 (4) & 0.576 (10) & 0.544 (13) & 0.566 (11) & 0.484 (16) & 0.585 (9) & 0.654 (3) & \textbf{0.699 (1)} & 0.525 (15) & 0.611 (8) & 0.526 (14) & 0.475 (17) & 0.639 (5) & 0.621 (6) \\
      13\_fraud & 0.940 (7) & 0.938 (8) & 0.911 (16) & 0.804 (18) & \textbf{0.964 (1)} & 0.931 (10) & 0.911 (17) & 0.932 (9) & 0.917 (14) & 0.951 (2) & 0.921 (13) & 0.948 (3) & 0.947 (4) & 0.946 (5) & 0.916 (15) & 0.924 (12) & 0.941 (6) & 0.929 (11) \\
      14\_glass & 0.974 (4) & 0.897 (11) & 0.934 (8) & 0.600 (18) & 0.721 (17) & 0.863 (13) & 0.880 (12) & 0.924 (10) & 0.792 (15) & 0.933 (9) & 0.837 (14) & 0.747 (16) & 0.938 (7) & \textbf{0.989 (1)} & 0.975 (3) & 0.962 (6) & 0.979 (2) & 0.971 (5) \\
      15\_Hepatitis & 0.726 (12) & 0.690 (14) & 0.764 (9) & 0.738 (11) & 0.606 (16) & 0.816 (5) & 0.691 (13) & 0.768 (8) & 0.491 (18) & 0.748 (10) & 0.556 (17) & 0.641 (15) & 0.790 (7) & \textbf{0.882 (1)} & 0.802 (6) & 0.817 (4) & 0.820 (3) & 0.875 (2) \\
\bottomrule
    \end{tabular}
  }
\end{table*}

\begin{table*}
  \centering
  \caption{The average AUCROC performance on individual datasets under $N_{la}=5$ setting}
  \label{tab:rank_aucroc_5}
  \resizebox{\textwidth}{!}{
    \begin{tabular}{lcccccccccccccccccc}
      \toprule
      Dataset & XGBOD & DeepSAD & REPEN & AABiGAN & GANomaly & DevNet & FEAWAD & PReNet & RoSAS & DualMGAN & SOEL-NTL & AnoDDAE & XGB & CatB & TabMCls & TabR-S & TabPFN & LimiX \\
      \midrule
      10\_cover & 0.972 (10) & 0.941 (12) & 0.980 (8) & 0.903 (15) & 0.411 (18) & 0.997 (6) & 0.997 (4) & 0.997 (5) & 0.795 (17) & 0.998 (3) & 0.843 (16) & 0.919 (13) & 0.954 (11) & 0.977 (9) & 0.997 (7) & 0.912 (14) & 0.999 (2) & \textbf{0.999 (1)} \\
      11\_donors & 0.941 (12) & 0.969 (8) & 0.968 (9) & 0.552 (18) & 0.570 (17) & 0.999 (2) & \textbf{0.999 (1)} & 0.997 (3) & 0.898 (13) & 0.990 (6) & 0.574 (16) & 0.678 (15) & 0.955 (11) & 0.976 (7) & 0.958 (10) & 0.887 (14) & 0.996 (4) & 0.991 (5) \\
      12\_fault & 0.588 (12) & 0.685 (2) & 0.614 (9) & 0.478 (17) & 0.642 (5) & 0.624 (7) & 0.563 (15) & 0.603 (11) & 0.469 (18) & 0.586 (13) & 0.662 (3) & \textbf{0.705 (1)} & 0.552 (16) & 0.618 (8) & 0.603 (10) & 0.566 (14) & 0.651 (4) & 0.641 (6) \\
      13\_fraud & 0.929 (12) & 0.919 (13) & 0.962 (3) & 0.894 (16) & 0.965 (2) & 0.934 (11) & 0.892 (17) & 0.942 (6) & 0.904 (15) & \textbf{0.970 (1)} & 0.935 (10) & 0.949 (5) & 0.911 (14) & 0.959 (4) & 0.940 (8) & 0.878 (18) & 0.939 (9) & 0.940 (7) \\
      14\_glass & 0.977 (4) & 0.916 (9) & 0.948 (8) & 0.566 (18) & 0.723 (17) & 0.876 (13) & 0.882 (12) & 0.916 (10) & 0.875 (14) & 0.905 (11) & 0.866 (15) & 0.747 (16) & 0.956 (7) & \textbf{0.994 (1)} & 0.971 (5) & 0.971 (6) & 0.986 (2) & 0.978 (3) \\
      15\_Hepatitis & 0.821 (7) & 0.732 (12) & 0.773 (10) & 0.704 (14) & 0.612 (17) & 0.866 (4) & 0.755 (11) & 0.835 (6) & 0.678 (15) & 0.711 (13) & 0.573 (18) & 0.662 (16) & 0.862 (5) & \textbf{0.909 (1)} & 0.820 (8) & 0.812 (9) & 0.883 (2) & 0.881 (3) \\
\bottomrule
    \end{tabular}
  }
\end{table*}

\begin{table*}
  \centering
  \caption{The average of AUCROC performance on individual datasets under $N_{la}=10$ setting}
  \label{tab:rank_aucroc_10}
  \resizebox{\textwidth}{!}{
    \begin{tabular}{lcccccccccccccccccc}
      \toprule
      Dataset & XGBOD & DeepSAD & REPEN & AABiGAN & GANomaly & DevNet & FEAWAD & PReNet & RoSAS & DualMGAN & SOEL-NTL & AnoDDAE & XGB & CatB & TabMCls & TabR-S & TabPFN & LimiX \\
      \midrule
      10\_cover & 0.997 (8) & 0.968 (12) & 0.992 (10) & 0.847 (17) & 0.431 (18) & 0.999 (5) & 0.999 (6) & 0.999 (3) & 0.900 (15) & 0.999 (4) & 0.872 (16) & 0.929 (14) & 0.991 (11) & 0.995 (9) & 0.998 (7) & 0.961 (13) & 0.999 (2) & \textbf{1.000 (1)} \\
      11\_donors & 0.988 (8) & 0.991 (7) & 0.982 (10) & 0.609 (16) & 0.534 (18) & 0.999 (3) & \textbf{0.999 (1)} & 0.999 (2) & 0.937 (13) & 0.993 (6) & 0.578 (17) & 0.696 (15) & 0.934 (14) & 0.985 (9) & 0.969 (11) & 0.952 (12) & 0.997 (4) & 0.994 (5) \\
      12\_fault & 0.650 (7) & 0.688 (4) & 0.647 (8) & 0.487 (17) & 0.642 (10) & 0.631 (12) & 0.585 (15) & 0.630 (13) & 0.470 (18) & 0.645 (9) & 0.662 (6) & \textbf{0.700 (1)} & 0.595 (14) & 0.684 (5) & 0.633 (11) & 0.548 (16) & 0.694 (2) & 0.689 (3) \\
      13\_fraud & 0.962 (4) & 0.945 (10) & 0.927 (14) & 0.735 (18) & 0.964 (3) & 0.956 (5) & 0.928 (13) & 0.948 (8) & 0.946 (9) & 0.954 (6) & 0.923 (16) & 0.951 (7) & \textbf{0.968 (1)} & 0.965 (2) & 0.927 (15) & 0.838 (17) & 0.929 (12) & 0.943 (11) \\
      14\_glass & 0.995 (4) & 0.947 (11) & 0.968 (9) & 0.658 (18) & 0.723 (17) & 0.891 (15) & 0.961 (10) & 0.918 (12) & 0.994 (5) & 0.912 (13) & 0.899 (14) & 0.743 (16) & 0.988 (7) & \textbf{0.998 (1)} & 0.992 (6) & 0.987 (8) & 0.996 (2) & 0.995 (3) \\
      15\_Hepatitis & 0.929 (3) & 0.831 (12) & 0.889 (9) & 0.731 (15) & 0.617 (17) & 0.921 (6) & 0.870 (10) & 0.910 (7) & 0.799 (14) & 0.812 (13) & 0.596 (18) & 0.682 (16) & 0.923 (5) & \textbf{0.944 (1)} & 0.909 (8) & 0.857 (11) & 0.943 (2) & 0.928 (4) \\
\bottomrule
    \end{tabular}
  }
\end{table*}

\newcommand{\res}[4]{%
  \ifx&#4&%
  #1
  \else%
  \textbf{#1}
  \fi%
  \hspace{1pt}%
  {\tiny $\pm$#2}
  \hspace{2pt}%
  {\scriptsize (#3)}
}

\begin{table*}
  \centering
  \caption{\textbf{AUCPR performance on tabular datasets.} Performance (mean $\pm$ std, rank) under varying label ratios ($\rla$) and label counts ($\nla$).}
  \label{tab:supp_incomp_baseline_cls}
  \resizebox{\textwidth}{!}{
    \begin{tabular}{lllcccccc}
      \toprule
      \multirow{2}{*}{\textbf{Type}} & \multirow{2}{*}{\textbf{Model}} & \multirow{2}{*}{\textbf{Key Mechanism}} & \multicolumn{3}{c}{\textbf{Label Ratio ($\rla$)}} & \multicolumn{3}{c}{\textbf{Label Count ($\nla$)}} \\
      \cmidrule(lr){4-6} \cmidrule(lr){7-9}
      & & & 1\% & 5\% & 10\% & 1 & 5 & 10 \\
      \midrule

      % === Semi-supervised Block ===
      \multirow{11}{*}{Semi-supervised}
      & XGBOD & Repr. Learning
      & \res{0.481}{0.257}{9}{} & \res{0.623}{0.261}{7}{} & \res{0.688}{0.257}{5}{} & \res{0.385}{0.244}{11}{} & \res{0.580}{0.258}{8}{} & \res{0.657}{0.260}{5}{} \\
      & DeepSAD & Repr. Learning
      & \res{0.379}{0.251}{13}{} & \res{0.497}{0.276}{14}{} & \res{0.568}{0.283}{15}{} & \res{0.332}{0.223}{16}{} & \res{0.474}{0.266}{15}{} & \res{0.556}{0.283}{14}{} \\
      & REPEN & Repr. Learning
      & \res{0.408}{0.259}{11}{} & \res{0.538}{0.275}{11}{} & \res{0.602}{0.276}{12}{} & \res{0.353}{0.235}{13}{} & \res{0.537}{0.274}{12}{} & \res{0.607}{0.286}{11}{} \\
      \addlinespace[0.2em]
      & AA-BiGAN & GAN-based
      & \res{0.308}{0.267}{17}{} & \res{0.306}{0.263}{18}{} & \res{0.310}{0.269}{18}{} & \res{0.322}{0.274}{17}{} & \res{0.282}{0.245}{18}{} & \res{0.282}{0.249}{18}{} \\
      & GANomaly & GAN-based
      & \res{0.314}{0.261}{16}{} & \res{0.315}{0.255}{17}{} & \res{0.316}{0.260}{17}{} & \res{0.310}{0.257}{18}{} & \res{0.314}{0.258}{17}{} & \res{0.319}{0.258}{17}{} \\
      \addlinespace[0.2em]
      & DevNet &  Score Learning
      & \res{0.555}{0.314}{2}{} & \res{0.635}{0.304}{6}{} & \res{0.666}{0.302}{8}{} & \res{0.517}{0.315}{1}{b} & \res{0.610}{0.311}{4}{} & \res{0.648}{0.304}{7}{} \\
\bottomrule
  \end{tabular}}
\end{table*}

\begin{table*}
  \centering
  \caption{\textbf{AUCPR performance on image datasets (ViT features).} Performance (mean $\pm$ std, rank) under varying $\rla$ and $\nla$.}
  \label{tab:supp_incomp_baseline_cv}
  \resizebox{\textwidth}{!}{
    \begin{tabular}{lllcccccc}
      \toprule
      \multirow{2}{*}{\textbf{Type}} & \multirow{2}{*}{\textbf{Model}} & \multirow{2}{*}{\textbf{Key Mechanism}} & \multicolumn{3}{c}{\textbf{Label Ratio ($\gamma_a$)}} & \multicolumn{3}{c}{\textbf{Label Count ($\nla$)}} \\
      \cmidrule(lr){4-6} \cmidrule(lr){7-9}
      & & & 1\% & 5\% & 10\% & 1 & 5 & 10 \\
      \midrule

      % === Semi-supervised Block ===
      \multirow{11}{*}{Semi-supervised}
      & XGBOD & Repr. Learning
      & \res{0.388}{0.242}{9}{} & \res{0.571}{0.301}{9}{} & \res{0.633}{0.307}{9}{} & \res{0.300}{0.198}{11}{} & \res{0.471}{0.271}{9}{} & \res{0.557}{0.297}{9}{} \\
      & DeepSAD & Repr. Learning
      & \res{0.268}{0.168}{17}{} & \res{0.395}{0.226}{14}{} & \res{0.474}{0.249}{14}{} & \res{0.217}{0.147}{16}{} & \res{0.302}{0.180}{16}{} & \res{0.376}{0.213}{14}{} \\
      & REPEN & Repr. Learning
      & \res{0.288}{0.198}{12}{} & \res{0.481}{0.264}{11}{} & \res{0.592}{0.292}{11}{} & \res{0.213}{0.175}{17}{} & \res{0.350}{0.215}{11}{} & \res{0.447}{0.243}{11}{} \\
      & AA-BiGAN & GAN-based
      & \res{0.332}{0.246}{10}{} & \res{0.318}{0.220}{15}{} & \res{0.320}{0.227}{15}{} & \res{0.314}{0.253}{10}{} & \res{0.327}{0.231}{13}{} & \res{0.324}{0.227}{16}{} \\
      & GANomaly & GAN-based
      & \res{0.277}{0.282}{16}{} & \res{0.279}{0.282}{17}{} & \res{0.279}{0.280}{17}{} & \res{0.282}{0.284}{13}{} & \res{0.278}{0.281}{17}{} & \res{0.278}{0.282}{17}{} \\
      & DevNet & Score Learning
      & \res{0.459}{0.296}{5}{} & \res{0.629}{0.307}{4}{} & \res{0.694}{0.312}{2}{} & \res{0.333}{0.249}{6}{} & \res{0.541}{0.299}{4}{} & \res{0.625}{0.309}{3}{} \\
\bottomrule
  \end{tabular}}
\end{table*}

\begin{table*}
  \centering
  \caption{\textbf{AUCPR performance on text datasets (RoBERTa features).} Performance (mean $\pm$ std, rank) under varying $\rla$ and $\nla$.}
  \label{tab:supp_incomp_baseline_nlp}
  \resizebox{\textwidth}{!}{
    \begin{tabular}{lllcccccc}
      \toprule
      \multirow{2}{*}{\textbf{Type}} & \multirow{2}{*}{\textbf{Model}} & \multirow{2}{*}{\textbf{Key Mechanism}} & \multicolumn{3}{c}{\textbf{Label Ratio ($\rla$)}} & \multicolumn{3}{c}{\textbf{Label Count ($\nla$)}} \\
      \cmidrule(lr){4-6} \cmidrule(lr){7-9}
      & & & 1\% & 5\% & 10\% & 1 & 5 & 10 \\
      \midrule

      % === Semi-supervised Block ===
      \multirow{11}{*}{Semi-supervised}
      & XGBOD & Repr. Learning
      & \res{0.166}{0.087}{8}{} & \res{0.284}{0.147}{8}{} & \res{0.366}{0.172}{8}{} & \res{0.132}{0.073}{8}{} & \res{0.221}{0.125}{8}{} & \res{0.282}{0.147}{9}{} \\
      & DeepSAD & Repr. Learning
      & \res{0.105}{0.046}{14}{} & \res{0.171}{0.100}{13}{} & \res{0.225}{0.125}{13}{} & \res{0.088}{0.038}{16}{} & \res{0.130}{0.074}{14}{} & \res{0.171}{0.092}{14}{} \\
      & REPEN & Repr. Learning
      & \res{0.134}{0.058}{12}{} & \res{0.248}{0.132}{11}{} & \res{0.344}{0.186}{11}{} & \res{0.108}{0.057}{13}{} & \res{0.172}{0.088}{12}{} & \res{0.231}{0.124}{12}{} \\
      & AA-BiGAN & GAN-based
      & \res{0.067}{0.025}{17}{} & \res{0.072}{0.029}{17}{} & \res{0.078}{0.050}{17}{} & \res{0.065}{0.025}{17}{} & \res{0.076}{0.048}{17}{} & \res{0.076}{0.051}{17}{} \\
      & GANomaly & GAN-based
      & \res{0.110}{0.055}{13}{} & \res{0.113}{0.056}{14}{} & \res{0.107}{0.050}{14}{} & \res{0.109}{0.054}{12}{} & \res{0.110}{0.052}{16}{} & \res{0.111}{0.055}{15}{} \\
      & DevNet & Score Learning
      & \res{0.272}{0.172}{2}{} & \res{0.430}{0.219}{1}{b} & \res{0.512}{0.219}{2}{} & \res{0.171}{0.114}{4}{} & \res{0.352}{0.193}{3}{} & \res{0.436}{0.201}{2}{} \\
\bottomrule
  \end{tabular}}
\end{table*}

\begin{table*}[htbp]
\centering
\caption{Performance comparison of video anomaly detection models}
\label{tab:inexcat_vad_tab_pretrain}
\renewcommand{\arraystretch}{1.2}
\resizebox{\textwidth}{!}{%
\begin{tabular}{lllcccccccccc}
\toprule
\multicolumn{1}{l}{\textbf{Pretrain}} & \multicolumn{1}{l}{\textbf{Model}} & \multicolumn{1}{l}{\textbf{Category}} & \multicolumn{5}{c}{\textbf{AUCROC}} & \multicolumn{5}{c}{\textbf{AUCPR}} \\
\cmidrule(lr){4-8} \cmidrule(lr){9-13}
& & & \textbf{TAD} & \textbf{UCF-Crime} & \textbf{XD-Violence} & \textbf{ShanghaiTech} & \textbf{Mean} & \textbf{TAD} & \textbf{UCF-Crime} & \textbf{XD-Violence} & \textbf{ShanghaiTech} & \textbf{Mean} \\
\midrule
\multirow{12}{*}{i3d} 
& AR-Net & Dynamic MIL & 0.858 {\tiny $\pm$ 0.001} (5) & \textbf{0.806 {\tiny $\pm$ 0.001} (1)} & 0.871 {\tiny $\pm$ 0.000} (2) & 0.898 {\tiny $\pm$ 0.003} (6) & 0.858 (3) & 0.298 {\tiny $\pm$ 0.001} (4) & 0.207 {\tiny $\pm$ 0.000} (4) & 0.658 {\tiny $\pm$ 0.001} (2) & \textbf{0.557 {\tiny $\pm$ 0.002} (1)} & \textbf{0.430 (1)} \\
& MGFN & Magnitude MIL & 0.849 {\tiny $\pm$ 0.042} (7) & 0.795 {\tiny $\pm$ 0.006} (4) & 0.779 {\tiny $\pm$ 0.021} (11) & 0.688 {\tiny $\pm$ 0.032} (11) & 0.778 (11) & 0.262 {\tiny $\pm$ 0.060} (9) & 0.171 {\tiny $\pm$ 0.003} (10) & 0.446 {\tiny $\pm$ 0.039} (11) & 0.154 {\tiny $\pm$ 0.028} (11) & 0.258 (11) \\
& RTFM & Magnitude MIL & 0.808 {\tiny $\pm$ 0.017} (10) & 0.779 {\tiny $\pm$ 0.014} (8) & 0.837 {\tiny $\pm$ 0.007} (6) & 0.845 {\tiny $\pm$ 0.010} (8) & 0.817 (8) & 0.268 {\tiny $\pm$ 0.013} (8) & 0.209 {\tiny $\pm$ 0.024} (3) & 0.606 {\tiny $\pm$ 0.012} (5) & 0.460 {\tiny $\pm$ 0.014} (6) & 0.386 (6) \\
& Sultani & Vanilla MIL & 0.834 {\tiny $\pm$ 0.003} (9) & \textbf{0.806 {\tiny $\pm$ 0.001} (1)} & 0.797 {\tiny $\pm$ 0.003} (10) & 0.851 {\tiny $\pm$ 0.001} (7) & 0.822 (7) & 0.285 {\tiny $\pm$ 0.002} (7) & \textbf{0.216 {\tiny $\pm$ 0.002} (1)} & 0.590 {\tiny $\pm$ 0.007} (7) & 0.477 {\tiny $\pm$ 0.004} (4) & 0.392 (5) \\
\bottomrule
\end{tabular}
}
\end{table*}

% vad

\begin{table*}[htbp]
\centering
\caption{Comparison of AUCROC results on different datasets.}
\label{tab:inexact_vad_auc}
\resizebox{\textwidth}{!}{
\begin{tabular}{lllccccccccccc}
\toprule
\multirow{2}{*}{\textbf{Pretrain}} & \multirow{2}{*}{\textbf{Model}} & \multirow{2}{*}{\textbf{Loss Type}} & \multicolumn{5}{c}{\textbf{32 Segments}} & \multicolumn{5}{c}{\textbf{200 Segments}} \\
\cmidrule(lr){4-8} \cmidrule(lr){9-13}
 &  &  & \textbf{TAD} & \textbf{UCF-Crime} & \textbf{XD-Violence} & \textbf{ShanghaiTech} & \textbf{Mean} & \textbf{TAD} & \textbf{UCF-Crime} & \textbf{XD-Violence} & \textbf{ShanghaiTech} & \textbf{Mean} \\
\midrule

% ==================== i3d BLOCK ====================
\multirow{7}{*}{i3d} 
& AR-Net & Dynamic MIL 
& 0.858 {\tiny $\pm$ 0.001} (2) & \textbf{0.806 {\tiny $\pm$ 0.001} (1)} & \textbf{0.871 {\tiny $\pm$ 0.000} (1)} & 0.898 {\tiny $\pm$ 0.003} (2) & \textbf{0.858 (1)} 
& 0.858 {\tiny $\pm$ 0.001} (3) & \textbf{0.806 {\tiny $\pm$ 0.001} (1)} & \textbf{0.871 {\tiny $\pm$ 0.000} (1)} & \textbf{0.898 {\tiny $\pm$ 0.003} (1)} & \textbf{0.858 (1)} \\

& MGFN & Magnitude MIL 
& 0.849 {\tiny $\pm$ 0.042} (3) & 0.795 {\tiny $\pm$ 0.006} (3) & 0.779 {\tiny $\pm$ 0.021} (7) & 0.688 {\tiny $\pm$ 0.032} (7) & 0.778 (7) 
& \textbf{0.893 {\tiny $\pm$ 0.018} (1)} & 0.795 {\tiny $\pm$ 0.009} (3) & 0.792 {\tiny $\pm$ 0.034} (6) & 0.744 {\tiny $\pm$ 0.070} (6) & 0.806 (6) \\

& RTFM & Magnitude MIL 
& 0.808 {\tiny $\pm$ 0.017} (7) & 0.779 {\tiny $\pm$ 0.014} (5) & 0.837 {\tiny $\pm$ 0.007} (3) & 0.845 {\tiny $\pm$ 0.010} (4) & 0.817 (4) 
& 0.809 {\tiny $\pm$ 0.022} (7) & 0.777 {\tiny $\pm$ 0.014} (5) & 0.843 {\tiny $\pm$ 0.005} (3) & 0.844 {\tiny $\pm$ 0.021} (4) & 0.818 (4) \\

& Sultani & Vanilla MIL 
& 0.834 {\tiny $\pm$ 0.003} (5) & \textbf{0.806 {\tiny $\pm$ 0.001} (1) }& 0.797 {\tiny $\pm$ 0.003} (6) & 0.851 {\tiny $\pm$ 0.001} (3) & 0.822 (3) 
& 0.824 {\tiny $\pm$ 0.007} (5) & 0.801 {\tiny $\pm$ 0.001} (2) & 0.787 {\tiny $\pm$ 0.003} (7) & 0.829 {\tiny $\pm$ 0.003} (5) & 0.810 (5) \\
\bottomrule
\end{tabular}
}
\end{table*}

\begin{table*}[htbp]
\centering
\caption{Comparison of AUCPR results on different datasets.}
\label{tab:inexact_vad_pr}
\resizebox{\textwidth}{!}{
\begin{tabular}{lllcccccccccc}
\toprule
\multirow{2}{*}{\textbf{Pretrain}} & \multirow{2}{*}{\textbf{Model}} & \multirow{2}{*}{\textbf{Loss Type}} & \multicolumn{5}{c}{\textbf{32 Segments}} & \multicolumn{5}{c}{\textbf{200 Segments}} \\
\cmidrule(lr){4-8} \cmidrule(lr){9-13}
 &  &  & \textbf{TAD} & \textbf{UCF-Crime} & \textbf{XD-Violence} & \textbf{ShanghaiTech} & \textbf{Mean} & \textbf{TAD} & \textbf{UCF-Crime} & \textbf{XD-Violence} & \textbf{ShanghaiTech} & \textbf{Mean} \\
\midrule
% ==================== i3d BLOCK ====================
\multirow{7}{*}{i3d} 
& AR-Net & Dynamic MIL 
& \textbf{0.298 {\tiny $\pm$ 0.001} (1)} & 0.207 {\tiny $\pm$ 0.000} (3) & \textbf{0.658 {\tiny $\pm$ 0.001} (1)} & \textbf{0.557 {\tiny $\pm$ 0.002} (1)} & \textbf{0.430 (1)} 
& 0.298 {\tiny $\pm$ 0.001} (2) & \textbf{0.208 {\tiny $\pm$ 0.000} (1)} & \textbf{0.658 {\tiny $\pm$ 0.001} (1)} & \textbf{0.554 {\tiny $\pm$ 0.002} (1)} & \textbf{0.429 (1)} \\

& MGFN & Magnitude MIL 
& 0.262 {\tiny $\pm$ 0.060} (5) & 0.171 {\tiny $\pm$ 0.003} (6) & 0.446 {\tiny $\pm$ 0.039} (7) & 0.154 {\tiny $\pm$ 0.028} (7) & 0.258 (7) 
& \textbf{0.329 {\tiny $\pm$ 0.023} (1)} & 0.170 {\tiny $\pm$ 0.006} (6) & 0.471 {\tiny $\pm$ 0.077} (7) & 0.219 {\tiny $\pm$ 0.074} (6) & 0.297 (7) \\

& RTFM & Magnitude MIL 
& 0.268 {\tiny $\pm$ 0.013} (4) & 0.209 {\tiny $\pm$ 0.024} (2) & 0.606 {\tiny $\pm$ 0.012} (3) & 0.460 {\tiny $\pm$ 0.014} (3) & 0.386 (3) 
& 0.257 {\tiny $\pm$ 0.007} (5) & 0.204 {\tiny $\pm$ 0.019} (3) & 0.596 {\tiny $\pm$ 0.023} (4) & 0.460 {\tiny $\pm$ 0.020} (2) & 0.380 (2) \\

& Sultani & Vanilla MIL 
& 0.285 {\tiny $\pm$ 0.002} (3) & \textbf{0.216 {\tiny $\pm$ 0.002} (1)} & 0.590 {\tiny $\pm$ 0.007} (5) & 0.477 {\tiny $\pm$ 0.004} (2) & 0.392 (2) 
& 0.277 {\tiny $\pm$ 0.008} (4) & 0.205 {\tiny $\pm$ 0.003} (2) & 0.576 {\tiny $\pm$ 0.004} (5) & 0.449 {\tiny $\pm$ 0.012} (3) & 0.377 (3) \\
\bottomrule
\end{tabular}
}
\end{table*}

\fi
\end{document}